\documentclass[a4paper, 11pt]{article}
\usepackage[top=3cm, bottom=3cm, left=2cm, right=2cm]{geometry}
\usepackage{url}
\usepackage{hyperref}  
\usepackage{graphicx} 
\usepackage{subcaption}
\usepackage{float}
\usepackage[export]{adjustbox}
\usepackage{comment}
\usepackage{appendix}
\usepackage{cite}
\usepackage{enumitem}
\usepackage{amssymb, amsmath}
\usepackage{tikz}
\usepackage{ifthen}
\usepackage{xstring}
\usetikzlibrary{shapes.geometric}
\usepackage{multirow}
\usepackage{longtable}
\usepackage{array}
\usepackage{arydshln}
\usepackage{caption}
\usepackage{makecell}
\usepackage{hhline}
\usepackage{booktabs}
\newcommand{\tabitem}{~~\llap{\textbullet}~~}

\usepackage{mwe} 
\usepackage{lmodern}  

\usepackage{xcolor}
\definecolor{blue}{HTML}{636EFA}
\definecolor{red}{HTML}{EF553B}
\definecolor{green}{HTML}{00CC96}
\definecolor{purple}{HTML}{AB63FA}
\definecolor{oranje}{HTML}{FFA15A}
\definecolor{cyan}{HTML}{19D3F3}
\definecolor{cherry}{HTML}{FF6692}
\definecolor{pomme}{HTML}{B6E880}
\definecolor{pink}{HTML}{FF97FF}
\definecolor{gold}{HTML}{fecb52}
\definecolor{grey}{HTML}{808080}

\usepackage{pifont}
\newcommand{\cmark}{\ding{51}}%
\newcommand{\xmark}{\ding{55}}%

\makeatletter
\newcommand\notsotiny{\@setfontsize\notsotiny\@vipt\@viipt}
\makeatother

\newcommand{\checkcolor}[1]{
    \IfStrEq{#1}{black}{
        \def\fillcolor{white} 
    }{
        \def\fillcolor{black} 
    }
}

\newcommand{\cblacksquare}[2][0.3]{
\begin{tikzpicture}[scale=#1]
    \filldraw[fill=#2, draw=black] (0,0) -- (1,0) -- (1,1) -- (0,1) -- (0,0);
\end{tikzpicture}
}

\newcommand{\cwhitesquare}[2][0.3]{
\begin{tikzpicture}[scale=#1]
 \draw[#2] (0,0) -- (1,0) -- (1,1) -- (0,1) -- (0,0);
\end{tikzpicture}
}

\newcommand{\cblacksquaredot}[2][0.3]{
\begin{tikzpicture}[scale=#1]
\checkcolor{#2}
 \filldraw[fill=#2, draw=black] (0,0) -- (1,0) -- (1,1) -- (0,1) -- (0,0);
 \fill[\fillcolor] (0.5,0.5) circle (4pt);
\end{tikzpicture}
}

\newcommand{\cwhitesquaredot}[2][0.3]{
\begin{tikzpicture}[scale=#1]
 \draw[#2] (0,0) rectangle (1,1);
 \filldraw[#2] (0.5,0.5) circle (4pt);
\end{tikzpicture}
}

\newcommand{\cblacktriangleup}[2][0.3]{
\begin{tikzpicture}[scale=#1]
    \filldraw[fill=#2, draw=black] (0,0) -- (1,0) -- (0.5, 1) -- cycle;
\end{tikzpicture}
}

\newcommand{\cwhitetriangleup}[2][0.3]{
\begin{tikzpicture}[scale=#1]
    \draw[#2] (0,0) -- (1,0) -- (0.5, 1) -- cycle;
\end{tikzpicture}
}

\newcommand{\cblacktriangleleft}[2][0.3]{
\begin{tikzpicture}[scale=#1, rotate=90]
    \filldraw[fill=#2, draw=black] (0,0) -- (1,0) -- (0.5, 1) -- cycle;
\end{tikzpicture}
}

\newcommand{\cwhitetriangleleft}[2][0.3]{
\begin{tikzpicture}[scale=#1, rotate=90]
    \draw[#2] (0,0) -- (1,0) -- (0.5, 1) -- cycle;
\end{tikzpicture}
}

\newcommand{\cwhitetriangleright}[2][0.3]{
\begin{tikzpicture}[scale=#1, rotate=-90]
    \draw[#2] (0,0) -- (1,0) -- (0.5, 1) -- cycle;
\end{tikzpicture}
}

\newcommand{\cblacktriangledown}[2][0.3]{
\begin{tikzpicture}[scale=#1, rotate=180]
    \filldraw[fill=#2, draw=black] (0,0) -- (1,0) -- (0.5, 1) -- cycle;
\end{tikzpicture}
}

\newcommand{\cwhitetriangledown}[2][0.3]{
\begin{tikzpicture}[scale=#1, rotate=180]
    \draw[#2] (0,0) -- (1,0) -- (0.5, 1) -- cycle;
\end{tikzpicture}
}

\newcommand{\cwhitesquarex}[2][0.3]{
\begin{tikzpicture}[scale=#1]
 \draw[#2] (0,0) -- (1,0) -- (1,1) -- (0,1) -- (0,0);
 \draw[#2] (1,0) -- (0,1);
 \draw[#2] (0,0) -- (1,1);
\end{tikzpicture}
}

\newcommand{\cblacksquarex}[2][0.3]{
\begin{tikzpicture}[scale=#1]
\checkcolor{#2}
 \filldraw[fill=#2, draw=black] (0,0) -- (1,0) -- (1,1) -- (0,1) -- (0,0);
 \draw[\fillcolor, line width=0.4mm] (1,0) -- (0,1);
 \draw[\fillcolor, line width=0.4mm] (0,0) -- (1,1);
\end{tikzpicture}
}

\newcommand{\cwhitesquarecross}[2][0.3]{
\begin{tikzpicture}[scale=#1]
 \draw[#2] (0,0) -- (1,0) -- (1,1) -- (0,1) -- (0,0);
 \draw[#2] (1,0.5) -- (0,0.5);
 \draw[#2] (0.5,0) -- (0.5,1);
\end{tikzpicture}
}

\newcommand{\cblacksquarecross}[2][0.3]{
\begin{tikzpicture}[scale=#1]
\checkcolor{#2}
 \filldraw[fill=#2, draw=black] (0,0) -- (1,0) -- (1,1) -- (0,1) -- (0,0);
 \draw[\fillcolor, line width=0.4mm] (1,0.5) -- (0,0.5);
 \draw[\fillcolor, line width=0.4mm] (0.5,0) -- (0.5,1);
\end{tikzpicture}
}

\newcommand{\cwhitediamondcross}[2][0.3]{
\begin{tikzpicture}[scale=#1]
 \draw[#2] (0,0.5) -- (0.5, 1) -- (1, 0.5) -- (0.5, 0) -- (0,0.5);
 \draw[#2] (1,0.5) -- (0,0.5);
 \draw[#2] (0.5,0) -- (0.5,1);
\end{tikzpicture}
}

\newcommand{\cblackdiamondcross}[2][0.3]{
\begin{tikzpicture}[scale=#1]
\checkcolor{#2}
 \filldraw[fill=#2, draw=black] (0,0.5) -- (0.5, 1) -- (1, 0.5) -- (0.5, 0) -- (0,0.5);
 \draw[\fillcolor, line width=0.4mm] (1,0.5) -- (0,0.5);
 \draw[\fillcolor, line width=0.4mm] (0.5,0) -- (0.5,1);
\end{tikzpicture}
}

\newcommand{\cwhitediamondx}[2][0.3]{
\begin{tikzpicture}[scale=#1]
 \draw[#2] (0,0.5) -- (0.5, 1) -- (1, 0.5) -- (0.5, 0) -- (0,0.5);
 \draw[#2] (0.25,0.75) -- (0.75,0.25);
 \draw[#2] (0.25,0.25) -- (0.75,0.75);
\end{tikzpicture}
}

\newcommand{\cblackdiamondx}[2][0.3]{
\begin{tikzpicture}[scale=#1]
\checkcolor{#2}
 \filldraw[fill=#2, draw=black] (0,0.5) -- (0.5, 1) -- (1, 0.5) -- (0.5, 0) -- (0,0.5);
 \draw[\fillcolor, line width=0.4mm] (0.25,0.75) -- (0.75,0.25);
 \draw[\fillcolor, line width=0.4mm] (0.25,0.25) -- (0.75,0.75);
\end{tikzpicture}
}

\newcommand{\cblackcircledot}[2][0.3]{
\begin{tikzpicture}[scale=#1]
\checkcolor{#2}
 \filldraw[fill=#2] (0.5,0.5) circle (14pt);
 \filldraw[\fillcolor] (0.5, 0.5) circle (4pt);
\end{tikzpicture}
}

\newcommand{\cwhitecircledot}[2][0.3]{
\begin{tikzpicture}[scale=#1]s
 \draw[#2] (0.5,0.5) circle (14pt);
 \filldraw[#2] (0.5, 0.5) circle (4pt);
\end{tikzpicture}
}

\newcommand{\cwhitecircle}[2][0.3]{
\begin{tikzpicture}[scale=#1]
 \draw[#2] (0.5,0.5) circle (14pt);
\end{tikzpicture}
}

\newcommand{\cwhitestartriangleup}[2][0.3]{
\begin{tikzpicture}[scale=#1]
    \draw[#2] (0,0) to[out=20, in=160] (1,0) to[out=160, in=90] (0.5, 0.9) to[out=90, in=20] cycle;
\end{tikzpicture}
}

\newcommand{\cwhitestartriangleupdot}[2][0.3]{
\begin{tikzpicture}[scale=#1]
    \draw[#2] (0,0) to[out=20, in=160] (1,0) to[out=160, in=90] (0.5, 0.9) to[out=90, in=20] cycle;
    \filldraw[#2] (0.5,0.4) circle (3pt);
\end{tikzpicture}
}

\newcommand{\cwhitestartriangledown}[2][0.3]{
\begin{tikzpicture}[scale=#1]
    \draw[#2] (0,1) to[out=-20, in=-160] (1,1) to[out=-160, in=65] (0.5, 0) to[out=115, in=-20] cycle;
\end{tikzpicture}
}

\newcommand{\cblackstartriangledown}[2][0.3]{
\begin{tikzpicture}[scale=#1]
    \filldraw[fill=#2] (0,1) to[out=-20, in=-160] (1,1) to[out=-160, in=65] (0.5, 0) to[out=115, in=-20] cycle;
\end{tikzpicture}
}

\newcommand{\cwhitestartriangledowndot}[2][0.3]{
\begin{tikzpicture}[scale=#1]
    \draw[#2] (0,1) to[out=-20, in=-160] (1,1) to[out=-160, in=65] (0.5, 0) to[out=115, in=-20] cycle;
    \fill[fill=#2] (0.5,0.6) circle (3pt);
\end{tikzpicture}
}

\newcommand{\cblackstartriangledowndot}[2][0.3]{
\begin{tikzpicture}[scale=#1]
\checkcolor{#2}
    \filldraw[fill=#2] (0,1) to[out=-20, in=-160] (1,1) to[out=-160, in=65] (0.5, 0) to[out=115, in=-20] cycle;
    \filldraw[\fillcolor] (0.5,0.6) circle (3pt);
\end{tikzpicture}
}

\newcommand{\cblackstar}[2][0.7]{
\begin{tikzpicture}
    \node[star, star points=5, star point ratio=2.25, fill=#2, draw=black, scale=#1] at (0,0) {};
\end{tikzpicture}
}

\newcommand{\cwhitestar}[2][0.7]{
\begin{tikzpicture}
    \node[star, star points=5, star point ratio=2.25, draw=#2, scale=#1] at (0,0) {};
\end{tikzpicture}
}

\title{Zero-shot capability of SAM-family models \\ for bone segmentation in CT scans}
\author{Caroline Magg\textsuperscript{1,2},
Hoel Kervadec\textsuperscript{1,2},
Clara I. S\'anchez\textsuperscript{1,2}}
\date{\normalsize \textsuperscript{1} University of Amsterdam, The Netherlands  \\ \textsuperscript{2} Amsterdam UMC location University of Amsterdam, The Netherlands}

\begin{document}

\maketitle

\begin{abstract}
    \noindent 
    The Segment Anything Model (\textsc{Sam}) and similar models build a family of promptable foundation models (FMs) for image and video segmentation. The object of interest is identified using prompts, such as bounding boxes or points.
    With these FMs becoming part of medical image segmentation, extensive evaluation studies are required to assess their strengths and weaknesses in clinical setting. Since the performance is highly dependent on the chosen prompting strategy, it is important to investigate different prompting techniques to define optimal guidelines that ensure effective use in medical image segmentation.
    Currently, no dedicated evaluation studies exist specifically for bone segmentation in CT scans, leaving a gap in understanding the performance for this task. Thus, we use non-iterative, ``optimal'' prompting strategies composed of bounding box, points and combinations to test the zero-shot capability of \textsc{Sam}-family models for bone CT segmentation on three different skeletal regions. 
    Our results show that the best settings depend on the model type and size, dataset characteristics and objective to optimize. Overall, \textsc{Sam} and \textsc{Sam2} prompted with a bounding box in combination with the center point for all the components of an object yield the best results across all tested settings. As the results depend on multiple factors, we provide a guideline for informed decision-making in 2D prompting with non-interactive, ``optimal'' prompts.
\end{abstract}



\section{Introduction}

The release of Segment Anything Model (\textit{\textsc{Sam}}) \cite{kirillov2023sam} started a family of promptable foundation models (FMs) for segmentation. Spatial or textual information in form of bounding box, points inside and outside the object or descriptive text are used as prompts to identify the object of interest. FMs are trained on huge datasets and their design allows them to generalize to unseen tasks and data. As data scarcity and domain shifts are common problems in medical image segmentation, FMs enter medical image segmentation as new alternative to fully supervised, specialized models trained on annotated data.

Since \textit{\textsc{Sam}} and its extension \textit{\textsc{Sam2}} (for video segmentation) are trained on natural image materials, there remains a gap in application to medical data due to the domain shift (natural images vs. medical images) and modality differences (2D vs. 3D). Efforts to address this gap have focused on fine-tuning and modifying the \textit{\textsc{Sam}} architecture to improve its suitability for medical imaging, resulting in versions such as \textit{Med-\textsc{Sam}} \cite{jun2024medsam}, \textit{\textsc{Sam}-Med2d} \cite{cheng2023sammed2d}, \textit{\textsc{Sam}-Med3d} \cite{wang2024sammed3d}, \textit{Med-\textsc{Sam2}} \cite{zhu2024medsam2}. Beyond model adaptations, thorough evaluation studies are essential to understand the current performance behavior, to identify potential weaknesses, risks and limitations in clinical settings and to formulate application guidelines for specific medical use cases. 

\begin{figure}[h!]
    \centering
    \includegraphics[width=0.99\linewidth, trim=0 200 0 100, clip]{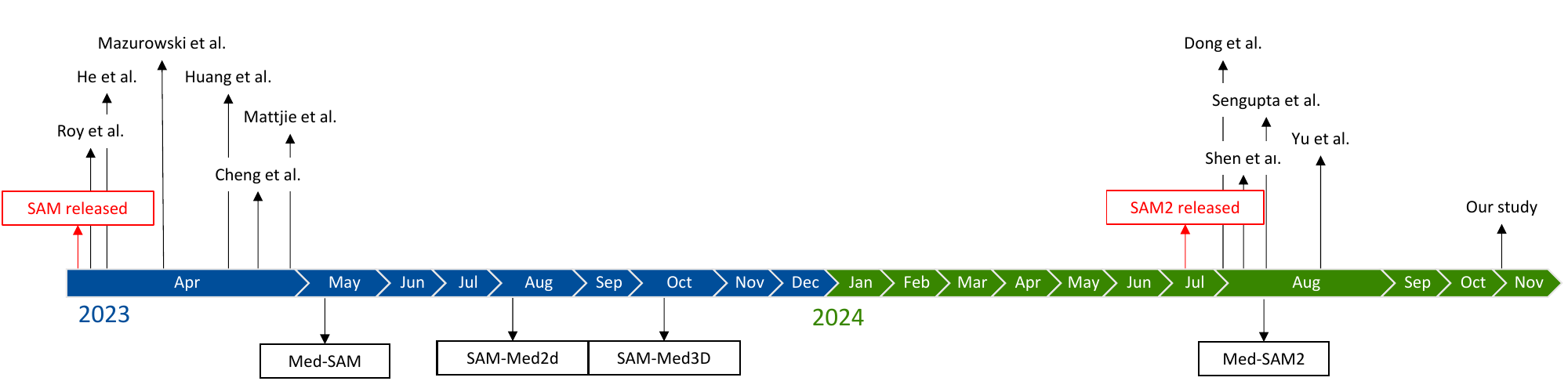}
    \caption{Timeline overview of relevant publications: models used for this study (highlighted with a bounding box) and related evaluation studies as mentioned in Table \ref{tab:related_work}. The first timestamp of pre-prints (not accepted peer-reviewed publications) are used for the visualization.}
    \label{fig:time_line}
\end{figure}

Table \ref{tab:related_work} collects a selection of evaluation studies closely related to our work, especially those conducted with a variety of medical image datasets. The conclusion of several evaluation studies \cite{MAZUROWSKI2023eval, he2023eval, mattjie2023eval, cheng2023eval, HUANG2024eval, dong2024eval} is the unstable performance across different datasets and task. The models tend to struggle with small, irregular structures with low-contrast or fuzzy boundaries, leading to unsatisfying results. In contrast, they show promising results on larger structures with clear, sharp boundaries. Given that bone appears in CT scans with high-intensity values and well-defined boundaries, we hypothesize that \textsc{Sam}-family models are well-suited to achieve promising results for this task. While we do not claim that the provided list in Table \ref{tab:related_work} is exhaustive, a gap in current evaluation studies is recognizable: There is no dedicated study focused on CT scans for bone segmentation. Existing studies primarily only evaluate the performance of \textsc{Sam} and \textsc{Sam2} across general medical imaging tasks, which allows them to use publicly available datasets. In contrast, our study utilizes a private dataset because many fine-tuned \textsc{Sam} models have already been trained on public CT datasets with bone segmentation labels, which prevents a fair comparison.
Fig. \ref{fig:time_line} provides a timeline overview of relevant publications (i.e., FMs and evaluation studies from Table \ref{tab:related_work}).

\begin{table}[h!]
    \centering
    \footnotesize
    \renewcommand{\arraystretch}{1.1}
    \caption{Overview of evaluation studies similar to our work. The prompting strategies marked with \textsuperscript{\textdagger} are applied for the largest and multiple disconnected components of one object. Unknown model sizes are denoted as N/A. The last column indicates whether CT scans with bone segmentation are included in the dataset. \xmark* corresponds to testing on X-Rays of the hip \cite{Gut2021}.}
    \label{tab:related_work}
    \begin{tabular}{llclc}
        Authors & Prompting strategies & Models & Dataset & Bone CT \\
        \hhline{=====}
        Roy et al. \cite{roy2023sammd} &
        \makecell[l]{\tabitem pos. random points \\
        \tabitem perturbed bounding box}
        & \makecell[c]{\textsc{Sam} \\ (size N/A)} & 1 public & \xmark \\
        \hdashline
        He et al. \cite{he2023eval} &
        \makecell[l]{\tabitem mass center \\
        \tabitem dilated bounding box}
        & \makecell[c]{\textsc{Sam} \\ (size N/A)} & 12 public & \xmark\\
        \hdashline
        Mazurowski et al. \cite{MAZUROWSKI2023eval} &
        \makecell[l]{\tabitem center\textsuperscript{\textdagger} \\
        \tabitem bounding box\textsuperscript{\textdagger} \\
        \tabitem simulated interactive 2D prompting}
        & \makecell[c]{\textsc{Sam} \\ (size N/A)} & 19 public & \xmark* \\
        \hdashline
        Huang et al. \cite{HUANG2024eval} &
        \makecell[l]{\tabitem center of mass/positive point \\
        \tabitem 5 pos. random points \\
        \tabitem 5 pos. and neg. points \\
        \tabitem bounding box \\
        \tabitem bounding box + 1 positive point}
        & \makecell[c]{\textsc{Sam} \\ (all sizes)} & 53 public & \cmark \\
        \hdashline
        Cheng et al. \cite{cheng2023eval} &
        \makecell[l]{\tabitem bounding box \\
        \tabitem center point of bounding box}
        & \makecell[c]{\textsc{Sam} \\ (size N/A)} & 12 public & \xmark \\
        \hdashline
        Mattjie et al. \cite{mattjie2023eval} & 
        \makecell[l]{\tabitem center \\
        \tabitem (distributed) pos. random points \\ 
        \tabitem (perturbed) bounding box}
        & \makecell[c]{\textsc{Sam} \\ (all sizes)} & 6 public & \xmark * \\
        \hdashline
        Dong et al. \cite{dong2024eval} &
        \makecell[l]{\tabitem center\textsuperscript{\textdagger} \\
        \tabitem bounding box\textsuperscript{\textdagger}} 
        & \makecell[c]{\textsc{Sam2} \\ (size N/A)} & 18 public & \xmark *\\
        \hdashline
        Shen et al. \cite{shen2024eval} &
        \makecell[l]{\tabitem pos. and neg. points \\
        \tabitem bounding box} & 
        \makecell[c]{\textsc{Sam}, \textsc{Sam2}, \\ Med-\textsc{Sam}, \textsc{Sam}-Med2 \\ (sizes N/A)} & 1 public & \xmark \\
        \hdashline
        Sengupta et al. \cite{sengupta2024eval} &
        \makecell[l]{\tabitem pos. random points \\
        \tabitem 1 pos. and 2 neg. points}
        & \makecell[c]{\textsc{Sam} \& \textsc{Sam2} \\ (all sizes)} & 11 public & \xmark \\
        \hdashline
        Yu et al. \cite{yu2024eval} & 
        \makecell[l]{\tabitem 1 point (undefined) \\
        \tabitem bounding box} & 
        \makecell[c]{\textsc{Sam} \& \textsc{Sam2} \\ (size N/A)} & 2 public & \xmark \\
        \hline
        Our study & 
        \makecell[l]{\tabitem center\textsuperscript{\textdagger} \\ \tabitem centroid\textsuperscript{\textdagger} \\
        \tabitem bounding box\textsuperscript{\textdagger} \\
        \tabitem pos random points\textsuperscript{\textdagger} \\
        \tabitem bounding box + center\textsuperscript{\textdagger} \\
        \tabitem bounding box + pos/neg. points\textsuperscript{\textdagger} \\
        \tabitem center + 1,5 neg. points\textsuperscript{\textdagger} \\
        \tabitem 1,5 pos. + neg. points\textsuperscript{\textdagger}} 
        & \makecell[c]{\textsc{Sam}, \textsc{Sam2}, \\ Med-\textsc{Sam}, \textsc{Sam}-Med2d \\ (all sizes)} & 3 private & \cmark \\
        \hline
    \end{tabular}
\end{table}

As a first step, the aim of this study is to investigate different non-iterative prompting strategies for \textsc{Sam}-family models on bone segmentation in CT scans under ``optimal'' 2D settings, i.e., prompts are based on reference masks without manipulation or human error. We test $9$ \textsc{Sam}-family models with $32$ prompting strategies on three different skeletal regions containing different bone structures. 
Based on our extensive analysis considering segmentation performance and model inference time, we introduce a preliminary guideline for choosing a 2D prompting strategy and model for a use case.

\section{SAM-family}

We refer to \textsc{Sam} and its related FMs as the \textsc{Sam}-family, each model used in our evaluation study described in the following section. 

\subsection{SAM}

The Segment Anything Model (\textit{\textsc{Sam}}) \cite{kirillov2023sam} was introduced in April 2023 by \textit{Meta} as promptable ``foundation model for image segmentation''. \textsc{Sam} supports sparse prompts, i.e., bounding box and clicks (positive and negative), and dense prompts, i.e., masks. The architecture consists of three parts (Fig. \ref{fig:sam_architecture}): First, the image encoder, a Masked Autoencoder (MAE) pre-trained Vision Transformer (ViT), is run once per image to create image embeddings of the 2D image input. Then, the prompt encoder creates prompt embeddings, i.e., the sum of positional encodings and learned embeddings, for each prompt type. Finally, the lightweight mask decoder combines both embeddings and an output token and predicts the final segmentation mask. Since prompts can be ambiguous, the model produces by default three masks with confidence scores that are based on estimated IoU. We employ the option for a single output mask.
The model was trained in three stages employing a model-assisted annotation workflow to annotate the final training dataset SA-1B, which consists of 11M images and over 1B masks.
The model is available in three different sizes: base (B), huge (H) and large (L), which depends on the ViT encoder.

\subsection{SAM2}

\begin{figure}[t]
    \centering
    \begin{subfigure}[b]{.75\textwidth}
        \centering
        \includegraphics[width=0.85\textwidth, trim=0 150 0 0, clip]{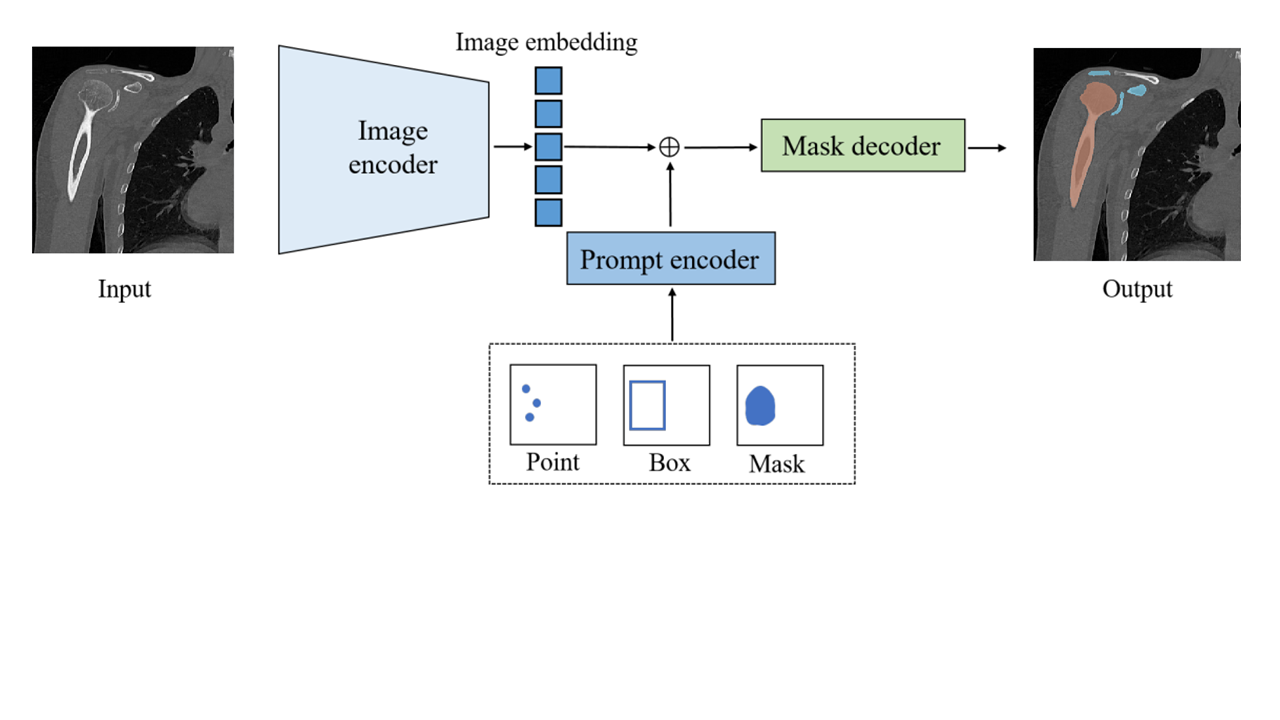}
        \caption{\textsc{Sam} architecture with image \& prompt encoder and mask decoder; Image adapted from \cite{zhang2024evalfirst}.}
        \label{fig:sam_architecture}
    \end{subfigure}
    \\
    \begin{subfigure}[b]{.75\textwidth}
    \centering
        \includegraphics[width=0.85\textwidth, trim=0 150 0 0, clip]{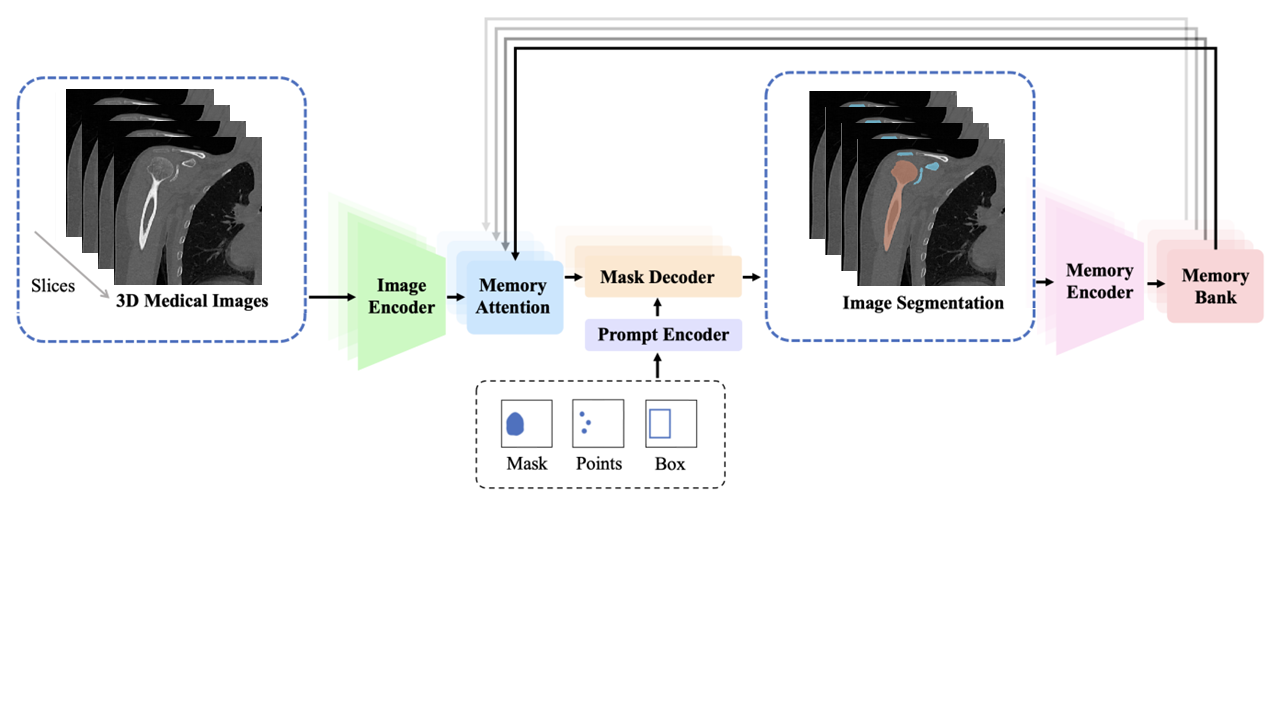}
        \caption{\textsc{Sam2} architecture with image \& prompt encoder, mask decoder and memory mechanics; Image adapted from \cite{zhang2024evalsecond}.}
        \label{fig:sam2_architecture}
    \end{subfigure}
    \caption{Comparison of the \textsc{Sam} and \textsc{Sam2} architectures.}
\end{figure}

\textit{\textsc{Sam2}} \cite{ravi2024sam2} was introduced in July 2024 as extension to \textsc{Sam} with the capability of image and video segmentation. This is realized by changes in the architecture (Fig. \ref{fig:sam2_architecture}): The ViT encoder is replaced by a MAE pre-trained Hiera image encoder, making use of multi-scale features. The image encoder is run once per interaction to produce unconditioned embeddings for each frame. The memory attention block fuses the unconditioned embeddings of a frame with past frame features and predictions. The conditioned embeddings are combined with new available prompt embeddings. Since not every frame in a video might contain the object of interest due to occlusion, the mask decoder is enhanced by an additional occlusion head to allow for an empty prediction. The generated segmentation masks are downsampled and combined with the unconditioned frame embedding to build a spatial feature map. This created memory is fed to a \textsc{Fifo} queue in the memory bank, where information of up to $N$ recent frames and $K$ prompted frames are stored. The model was trained on SA-V, the Segment Anything Video dataset with 35.5M masks in 50.9K videos, in multiple training iterations. Due to different Hiera sizes, four different \textit{\textsc{Sam2}} sizes are available: base plus (B+), tiny (T), small (S)  and large (L).

\subsection{Med-SAM}

Ma et al. introduced \textit{Med-\textsc{Sam}} \cite{jun2024medsam} as ``a foundation model for promptable medical image segmentation". Without any adaptions to the \textsc{Sam} architecture, they fine-tuned \textit{\textsc{Sam} B} on a medical image dataset. While keeping the prompt encoder frozen, all trainable parameters in the image encoder and the mask decoder are trained for 150 epochs. The training dataset with focus on cancer types contains more than one million image-mask pairs including 10 imaging modalities. \textit{Med-\textsc{Sam}} was only retrained for bounding box prompts, as the authors argue that for that task point-based prompts introduce a more ambiguous spatial context than bounding boxes.

\subsection{SAM-Med2d}

Cheng et al. developed \textit{\textsc{Sam}-Med2d} \cite{cheng2023sammed2d} by fine-tuning \textit{\textsc{Sam} B} with an adapter technique. The frozen image encoder is equipped with learnable adapter layers which are updated together with the prompt encoder and the mask decoder during 9 training iterations with 12 epochs. The model keeps the functionality of both sparse prompts (bounding box and point), while using also dense prompts (masks) during training. For fine-tuning, the dataset SA-Med2D-20M \cite{ye2023sammed2d_dataset} was curated from public and private datasets, where 20M corresponds to the number of masks. 

\section{Dataset}

\begin{figure}[h]
   \centering
   \includegraphics[width=1.0\textwidth]{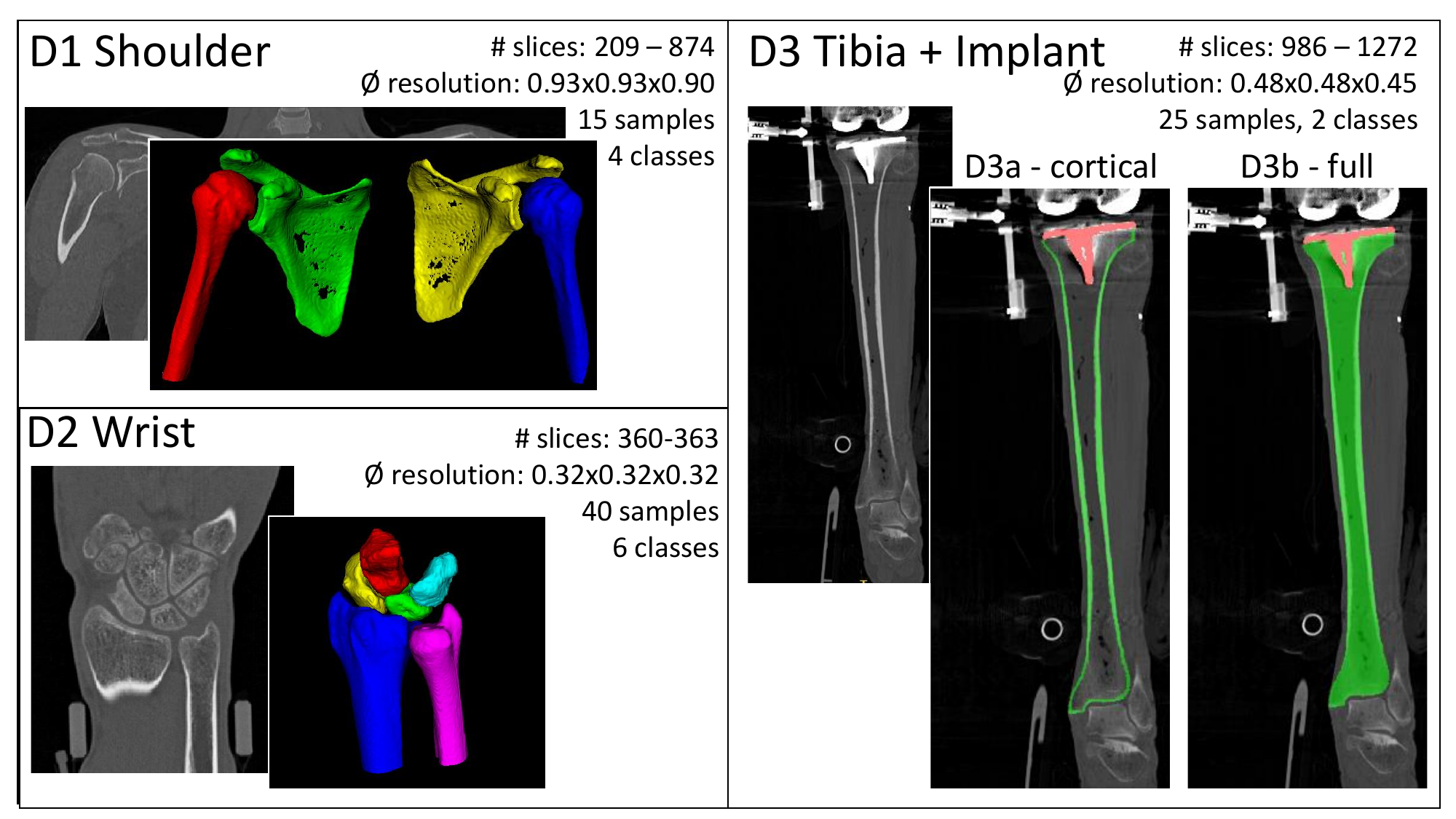}
  \caption{Dataset Overview: 3 private datasets containing 80 CT scans from three skeletal regions, i.e., shoulder (D1), wrist (D2) and knee (D3) are used. The knee dataset comes with two different sets for the tibia segmentation, i.e., cortical (D3a) and full (D3b) tibia segmentation.}
  \label{fig:datasets}
\end{figure}

As medical \textsc{Sam} versions (e.g., \textit{Med-\textsc{Sam}} \cite{jun2024medsam}, \textit{\textsc{Sam}-Med2D} \cite{cheng2023sammed2d}, \textit{\textsc{Sam}-Med3d} \cite{wang2024sammed3d}) are fine-tuned on publicly available datasets containing bone segmentation on CT scans (i.e., TotalSegmentator \cite{wasserthal2023totalsegmentator}, CTPelvic1K \cite{liu2021ctpelvic1k}, VerSe2020 \cite{verse2021}), an independent dataset is required to ensure a fair comparison across all models. Thus, we compiled a private (from the department of Orthopedic Surgery and Sports Medicine of the Amsterdam UMC) dataset of 80 CT Scans with three different skeletal regions (Fig. \ref{fig:datasets}):
\begin{itemize}
    \item D1 Shoulder: 15 bilateral scans with 4 labels for left and right scapula and humerus.
    \item D2 Wrist: 40 unilateral scans with 6 labels for capitate, lunate, radius, scaphoid, triquetrum, and ulna.
    \item D3 Knee: 25 unilateral scans with 2 labels for tibia bone and tibia implant. There are two different labeling protocols for the tibia bone: cortical bone (D3a) and full bone (D3b).
\end{itemize}
All scans were acquired with a Brilliance 64-channel CT Scanner (Philips Healthcare, Best, The Netherlands) or a Siemens SOMATOM Force with $160$ mAs, $120$ kV. The isotropic voxel spacing is $0.93$ mm, $0.32$ mm, and $0.48$ mm, for D1, D2, and D3, respectively. The annotations were generated using an in-house annotation software \cite{dobbe2014} and/or ITK-Snap \cite{itk_snap}.

\section{Experimental Setup}

\subsection{Prompting Strategies}

We use non-iterative prompts, which are automatically extracted from the reference masks. A prompt consists of at least one prompt primitive (i.e., \texttt{bounding box}, \texttt{center}, \texttt{centroid}, \texttt{positive},  \texttt{negative} points) and one component selection criteria (i.e., largest (\texttt{1C}) or up to 5 components (\texttt{5C})).

\subsubsection{Primitives}
\begin{figure}[h!]
     \centering
     \begin{subfigure}[b]{0.19\textwidth}
   \centering
   \includegraphics[width=0.8\textwidth]{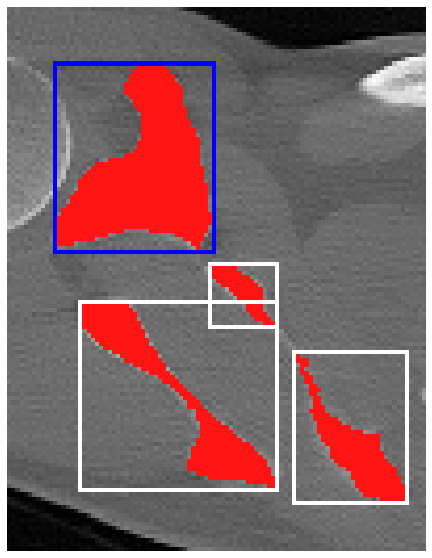}
   \caption{\texttt{bbox}}
   \label{fig:prompts_box}
     \end{subfigure}
     \begin{subfigure}[b]{0.19\textwidth}
   \centering
   \includegraphics[width=0.8\textwidth]{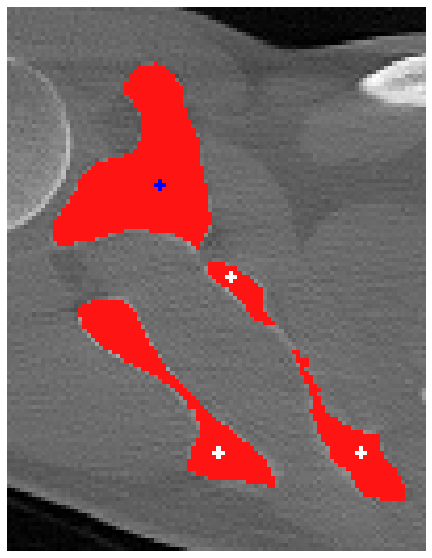}
   \caption{\texttt{center}}
   \label{fig:prompts_center}
     \end{subfigure}
     \begin{subfigure}[b]{0.19\textwidth}
   \centering
   \includegraphics[width=0.8\textwidth]{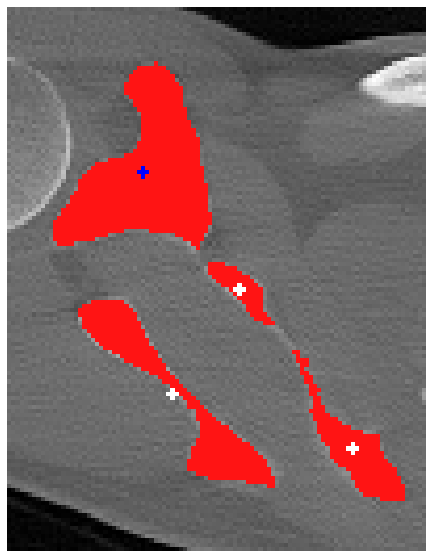}
   \caption{\texttt{centroid}}
   \label{fig:prompts_centroid}
    \end{subfigure}
    \begin{subfigure}[b]{0.19\textwidth}
   \centering
   \includegraphics[width=0.8\textwidth]{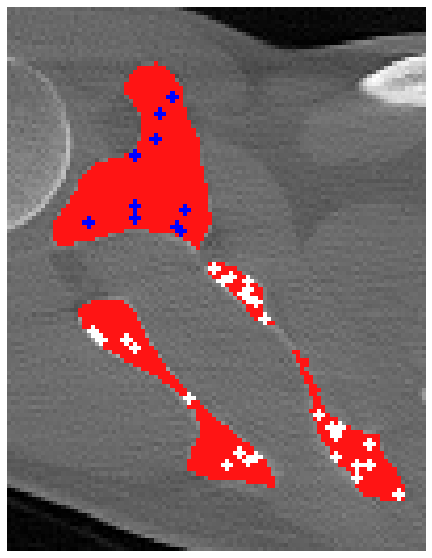}
   \caption{\texttt{positive}}
   \label{fig:prompts_random}
     \end{subfigure}
     \begin{subfigure}[b]{0.19\textwidth}
   \centering
   \includegraphics[width=0.8\textwidth]{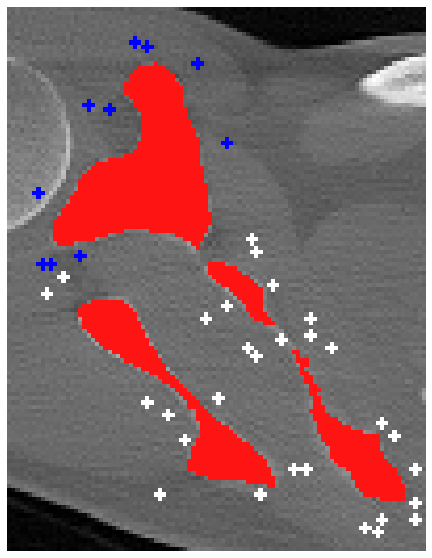}
   \caption{\texttt{negative}}
   \label{fig:prompts_negative}
    \end{subfigure}
  \caption{Prompt primitives: (a) bounding box, (b) (EDT) center, (c) centroid, (d) positive random points inside the object, (e) negative random points outside the object. The largest component's prompt is blue (i.e., one component (\texttt{1C})), while the others are white, resulting in the setting with up to 5 components (\texttt{5C}), when all prompts are used.}
  \label{fig:prompts}
\end{figure}

There are 5 primitives which are the building blocks for a prompt (Fig. \ref{fig:prompts}):
\begin{enumerate}[label=(\alph*)]
    \item \texttt{bounding box}: Enclosing tight rectangle, i.e., no extra space between the object boundary and the box.
    \item \texttt{(EDT) center}: Maximum location of the Euclidean distance transform (EDT), i.e., the point the most further away from the object boundary. In case of equality (two points or more are as far as possible from the boundary), a random point is selected. For shortness, we refer to it simply as \texttt{center} from now on.
    \item \texttt{centroid}: Center of mass with homogeneous density. Note that there is no guarantee that the centroid is inside the object. Despite this shortcoming, we include it as other existing work \cite{he2023eval} used it.
    \item \texttt{positive} points: Random points within the region. To avoid random points on the border, the reference mask is eroded by a $3\times3$ ellipsoidal kernel before the random points are extracted.
    \item \texttt{negative} points: Random points outside the region close to the border. To ensure that the points remain near the object and prevent scattering across distant image regions, a margin through mask dilation is created for point extraction. The mask is dilated in two steps: first the mask is dilated with a $5\times5$ ellipsoidal kernel and then with a $15\times 15$ ellipsoidal kernel. The difference between these two dilated regions is used for point extraction.
\end{enumerate}

\begin{figure}[h!]
    \centering
    \includegraphics[width=0.4\linewidth,trim=0 150 250 0,clip]{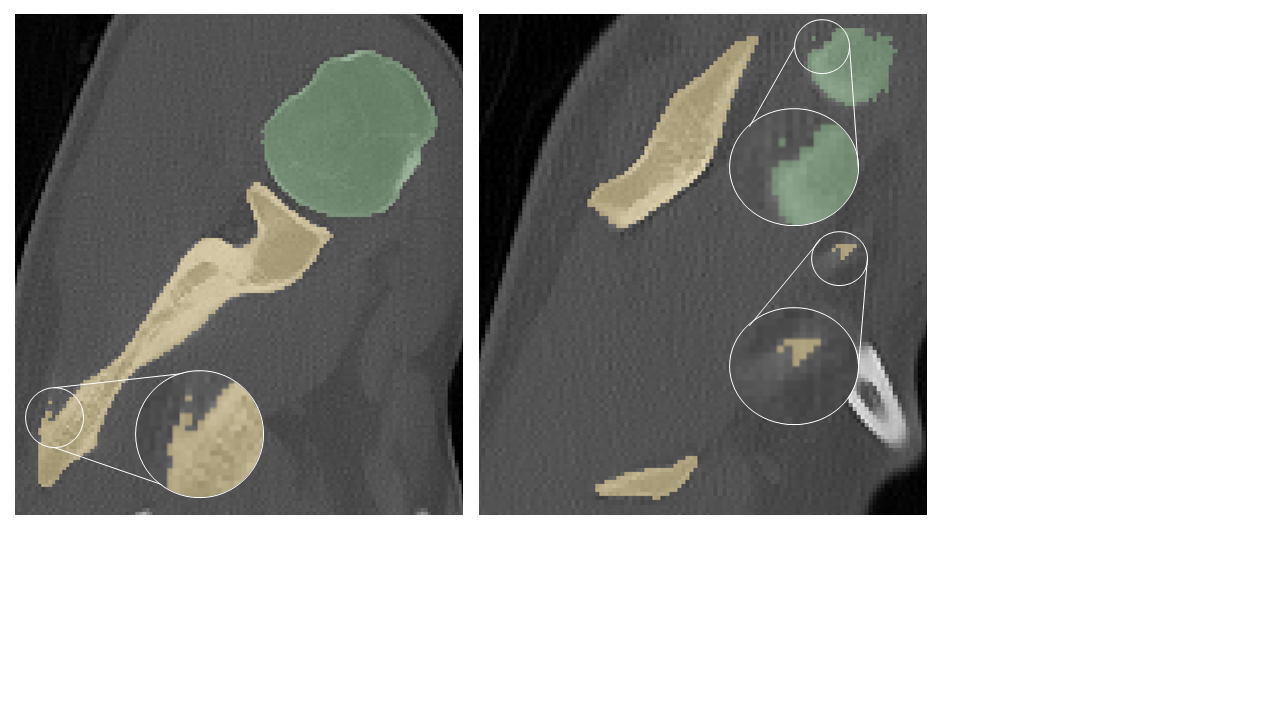}
    \caption{Examples for small components in reference masks that are ignored during prompt primitive extraction. The zoomed-in windows show small reference masks that do not meet the criteria of being at least $15$ pixels in size or covering more than $5\%$ of the total component in the slice.}
    \label{fig:small_reference_mask}
\end{figure}

The prompt primitives are extracted for each component of the reference mask larger than 15 pixels or larger than 5\% of the entire component in a slice. Fig. \ref{fig:small_reference_mask} illustrates that components that are smaller as the defined criteria are unrealistic to be annotated: bounding boxes are collapsing and there are not enough pixels to extract 10 random points inside the object. The thresholds were chosen empirically after dataset inspection.

\subsubsection{Component selection criteria}
As shown in Fig. \ref{fig:prompts}, anatomical structures can consist of multiple disconnected components in a 2D medical image slice. In our datasets, the number of disconnected components does not exceed 6, which only occurs for less than 10 slices in our entire dataset. Thus, for a prompt, the primitives of either the largest component (\texttt{1C}, denoted as open symbols) or up to 5 components (\texttt{5C}, denoted as closed symbols) are used. As those foundational models claim to perform very well in zero-shot scenarios, we expect them to generalize well to the other connected components, even if they are not directly selected with a point or bounding box.

\subsubsection{2D Prompts}
A 2D prompting strategy is defined by one or more prompt primitives and one component selection criteria. 2D strategies can be divided into three categories based on the involved primitives:

\begin{itemize}
    \item One-type prompts (OT prompts): bounding box (\cwhitesquare{black}), center (\cwhitecircledot{black}), centroid (\cwhitecircle{black}), 1, 3, 5 or 10 positive random points (\cwhitetriangleup{black}, \cwhitetriangleright{black}, \cwhitetriangledown{black}, \cwhitetriangleleft{black}).
    \item bounding box + point combination prompts (BPC prompts): bounding box with center (\cwhitesquaredot{black}), with 1 or 5 positive random points (\cwhitesquarecross{black}, \cwhitediamondcross{black}), with 1 or 5 negative random points (\cwhitesquarex{black}, \cwhitediamondx{black}).
    \item Point based combination prompts (PB prompts): center with 1 or 5 negative points (\cwhitestartriangleupdot{black}, \cwhitestartriangledowndot{black}), 1 or 5 positive and negative random points (\cwhitestartriangleup{black}, \cwhitestartriangledown{black}).
\end{itemize}

Certain redundant combinations have been omitted to reduce the number of prompting strategies: The centroid is an unreliable primitive since it may lie outside the object depending on the object's shape, so it is not used for combinations.
Combinations with random points are done with one point as comparison to the center point and with five points to asses the effect of an increased number of points. In total, there are $32$ settings per model. An exception is \textit{Med-\textsc{Sam}} with $2$ settings, as the model only supports bounding boxes. 

\subsection{Implementation details}

The official github repositories (i.e., \textit{\textsc{Sam}}\footnote{\url{https://github.com/facebookresearch/segment-anything}, commit \texttt{6fdee8f}}, \textit{\textsc{Sam2}}\footnote{\url{https://github.com/facebookresearch/sam2}, commits \texttt{0e78a11} \& \texttt{29267c8}}, \textit{Med-\textsc{Sam}}\footnote{\url{https://github.com/bowang-lab/MedSAM}, commit \texttt{2b7c64c}}, \textit{\textsc{Sam}-Med2d}\footnote{\url{https://github.com/OpenGVLab/SAM-Med2D}, commit \texttt{bfd2b93}}) are used for all models. Data preprocessing and weight download is performed as instructed. For \textit{\textsc{Sam2}}, the weights from July 29, 2024 are used. 
The evaluation is performed on GPUs NVIDIA Geforce RTX 2080 Ti 12GB and an Intel Core Xeon Gold 6128 3.40GHz CPU, which are embedded in a server accessible to multiple users. Evaluation code is adapted from \cite{isensee2021nnunet, jia2024seg}. Visualizations are performed with 3D Slicer (\url{https://www.slicer.org/}) and plotly (\url{https://plotly.com/}).

\subsection{Evaluation}

Two common segmentation performance metrics, i.e, Dice Similarity Coefficient (DSC) and 95\%-percentile Hausdorff Distance (HD95), were used to compare all prediction masks with the reference labels. In addition to performance metrics, the inference time for each model prediction was measured. The inference time includes the recommended image and prompt preprocessing and each prediction call, but excludes image and prompt loading. For models with multiple prediction calls due to multi-class segmentation (i.e., each class requires a separate prediction call as binary segmentation masks are returned), the image embedding is done once and reused for all class predictions.\newline
To compare FMs with fully supervised models, 2D and 3D full resolution nnUNets \cite{isensee2021nnunet} were trained for each subset. The training details are reported in Appendix \ref{sec:nnunet_details}. The results are denoted as \cwhitestar[0.4]{black}.

\section{Results}

\subsection{Segmentation performances}

\subsubsection{Overview}

\begin{figure}[h!]
    \centering
    \includegraphics[width=1.0\linewidth]{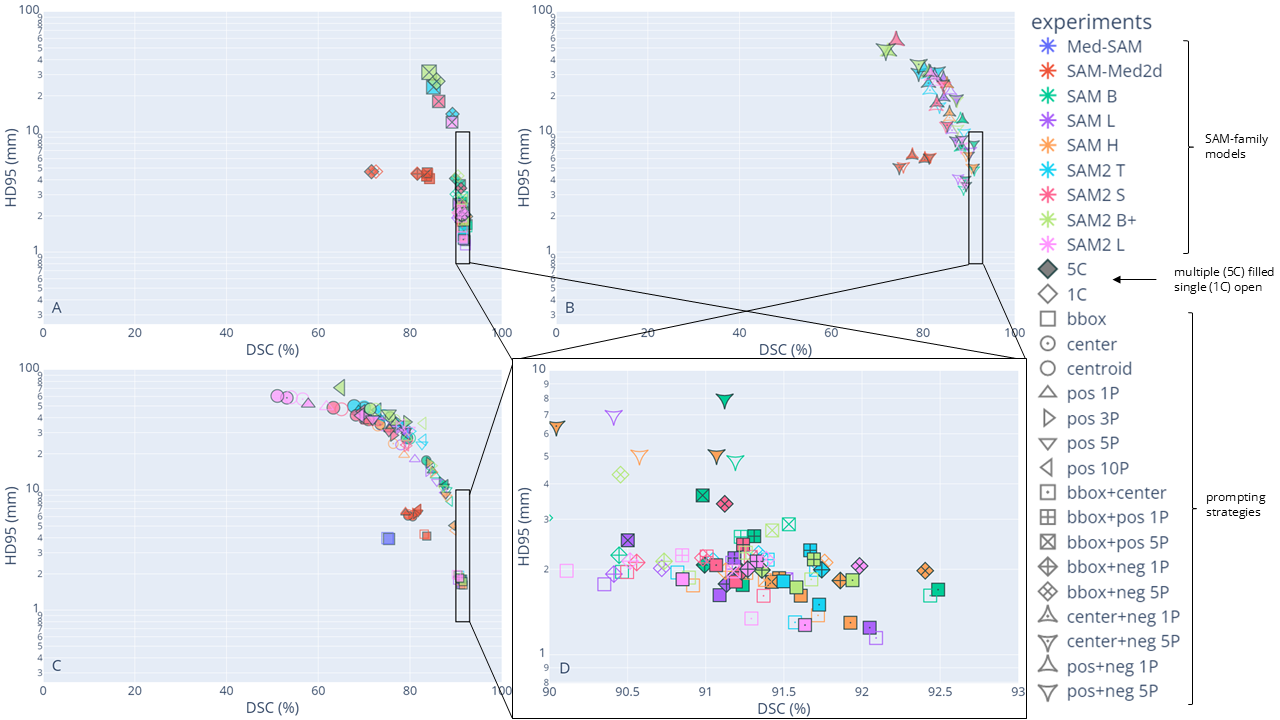}
    \caption{Performance of 2D prompting strategies: Scatterplots of BPC prompts (A), PC prompts (B), OT prompts (C) and zoomed-in view to the lower right corner (D). The symbol size in (A)-(C) is determined by the DSC standard deviation (std), i.e., bigger symbol means higher DSC std. The lower right corner, i.e., high DSC and low HD95 metrics, contains the performance strongest settings.}
    \label{fig:scatter_plot_all}
\end{figure}

\begin{figure}[h!]
\centering
\begin{tabular}{|l|cc|cc|cc|}
    \hline
    Setting & \multicolumn{2}{c|}{DSC $\downarrow$} & \multicolumn{2}{c|}{DSC median} & \multicolumn{2}{c|}{DSC $\uparrow$} \\ 
    \hline
    \multirow{2}{*}{\cblacksquare[0.6]{blue}} & \includegraphics[height=2cm, trim=600 380 100 330, clip]{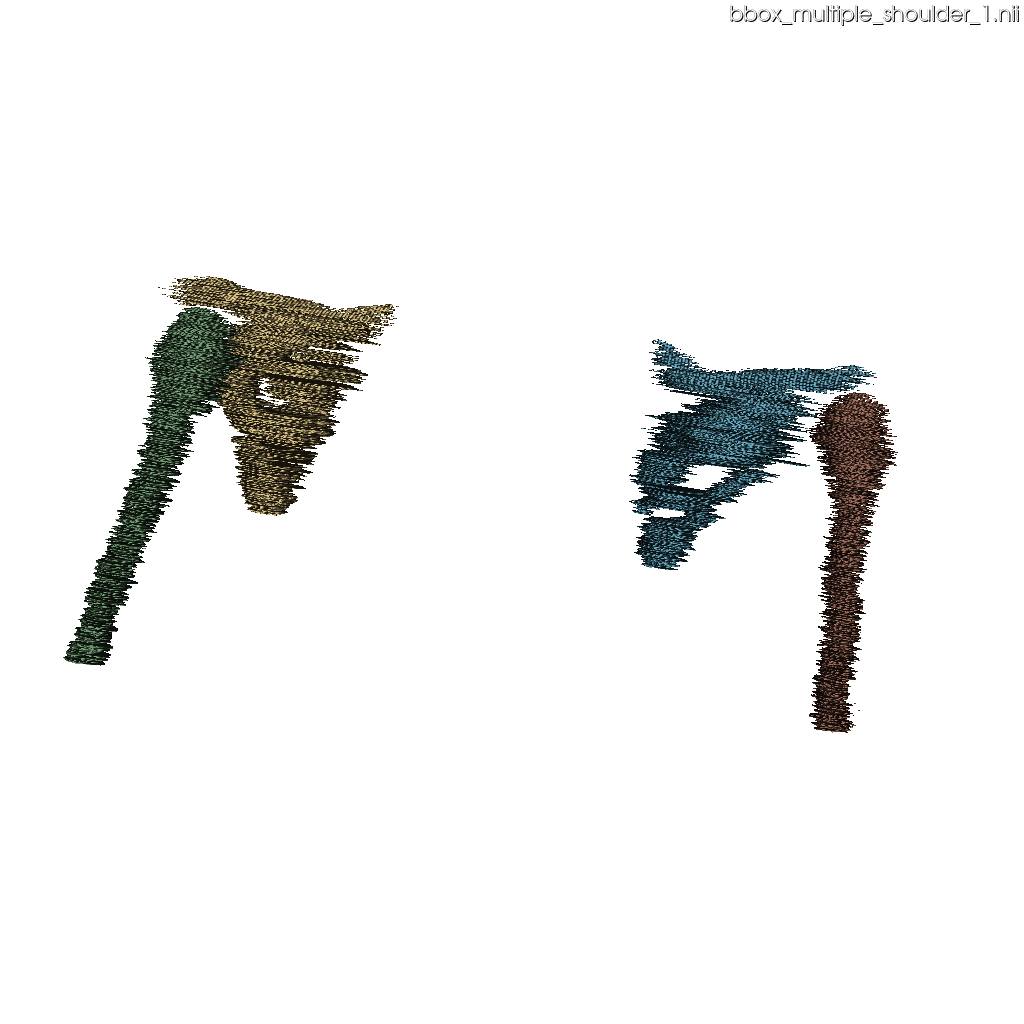} & \raisebox{-0.55\height}[0pt][0pt]{\includegraphics[height=4cm, trim=250 150 300 50, clip]{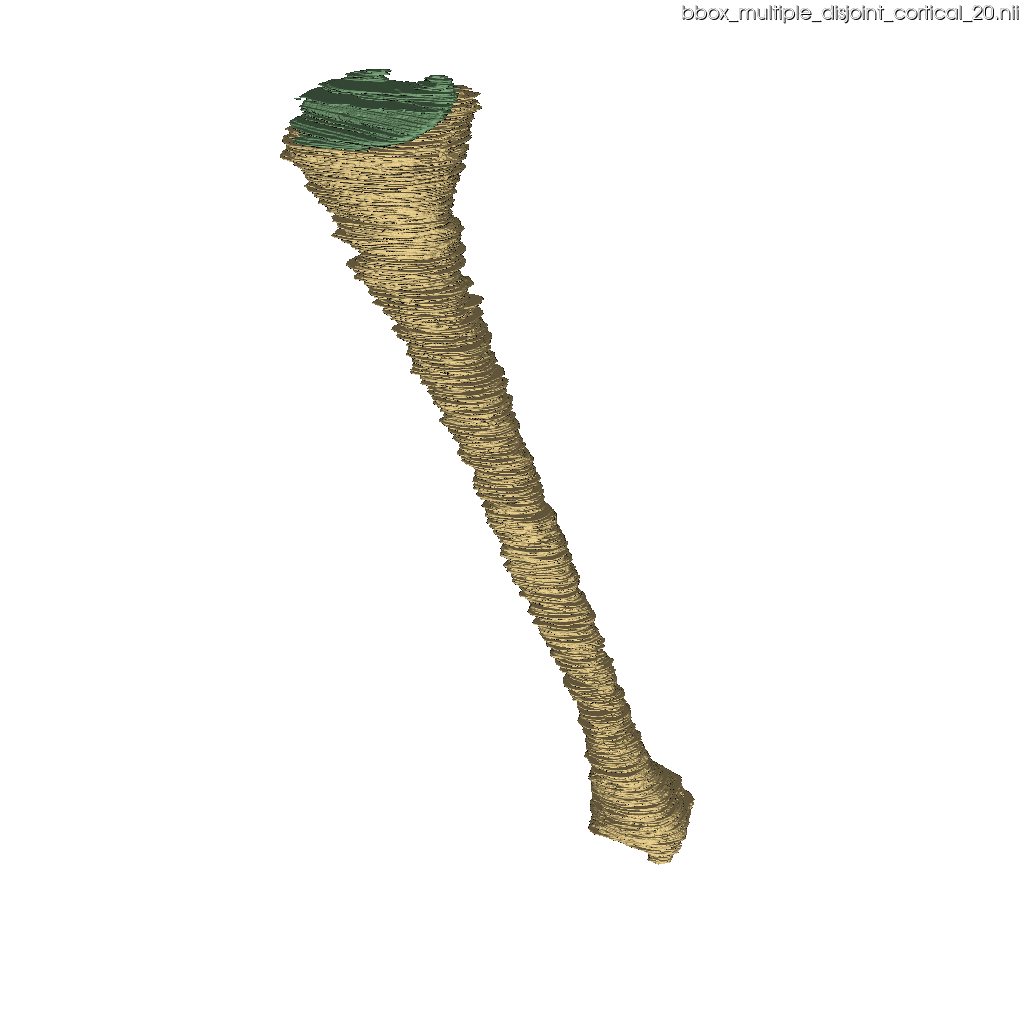}} &
    \includegraphics[height=2cm, trim=560 380 130 360, clip]{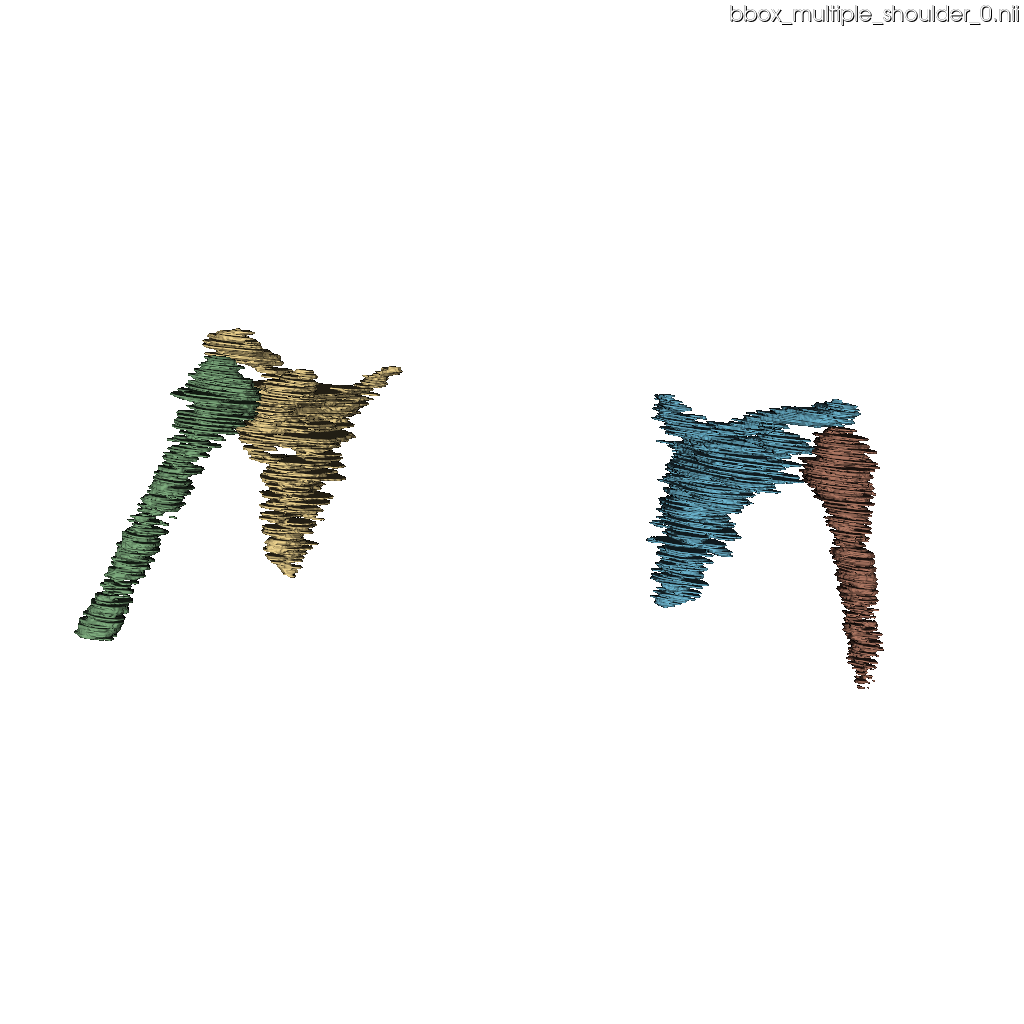} & \raisebox{-0.55\height}[0pt][0pt]{\includegraphics[height=4cm, trim=380 200 400 100, clip]{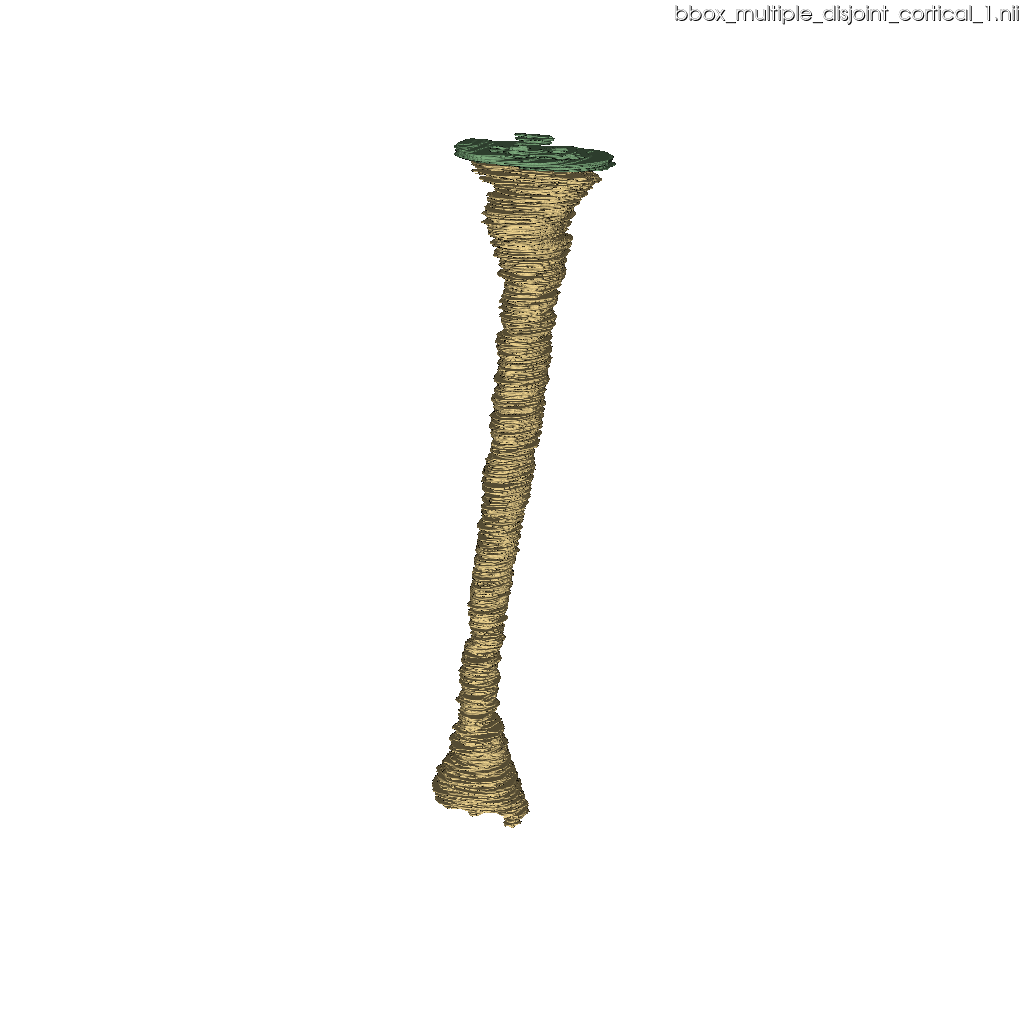}} &
    \includegraphics[height=2cm, trim=560 380 190 330, clip]{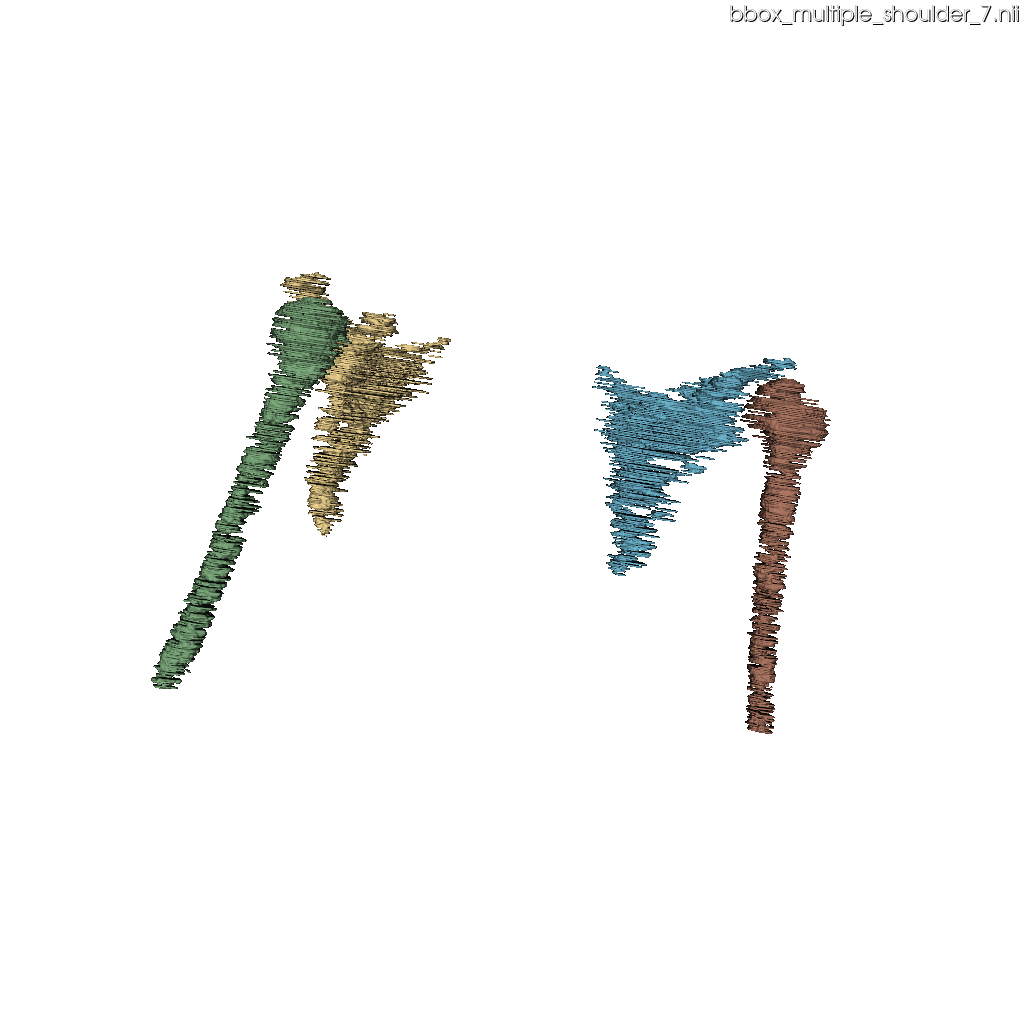} & \raisebox{-0.55\height}[0pt][0pt]{\includegraphics[height=4cm, trim=400 150 400 100, clip]{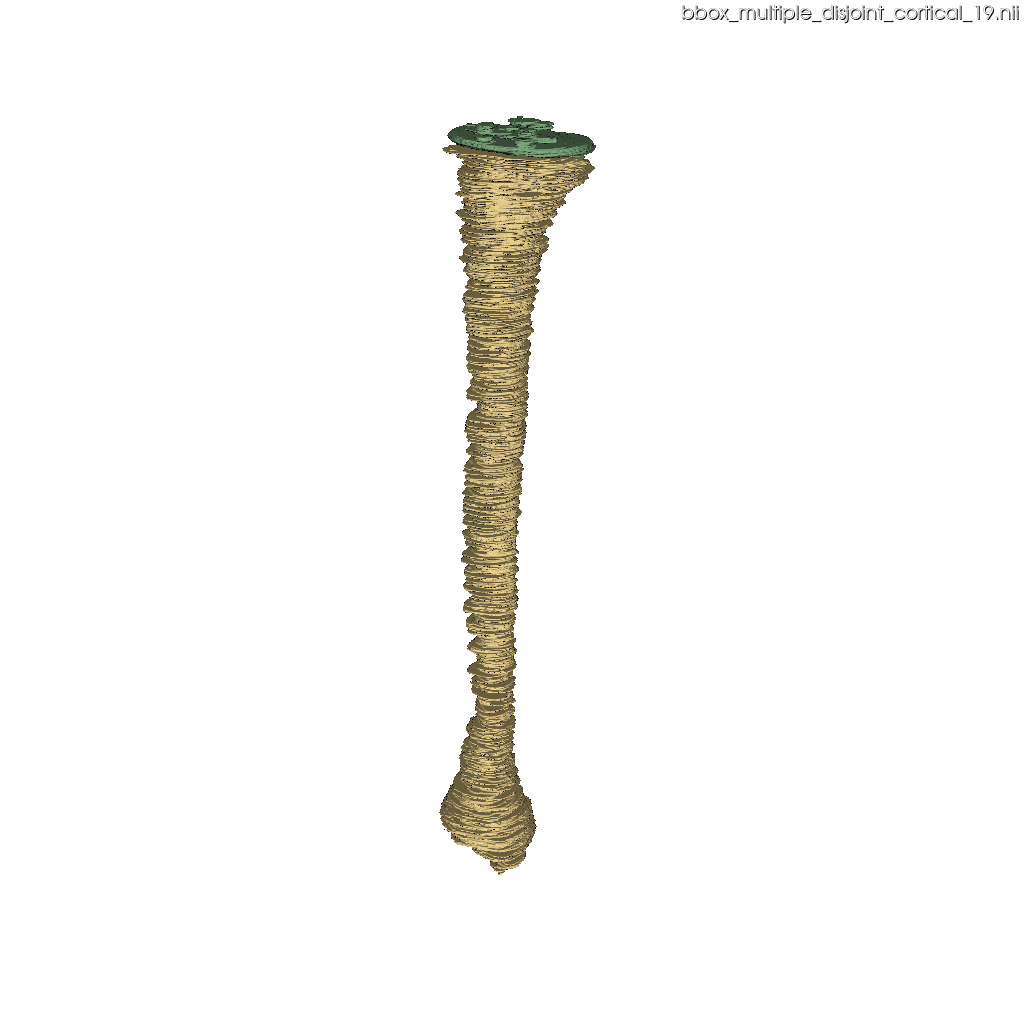}} 
    \\
    & \includegraphics[height=2cm, trim=370 290 370 280, clip]{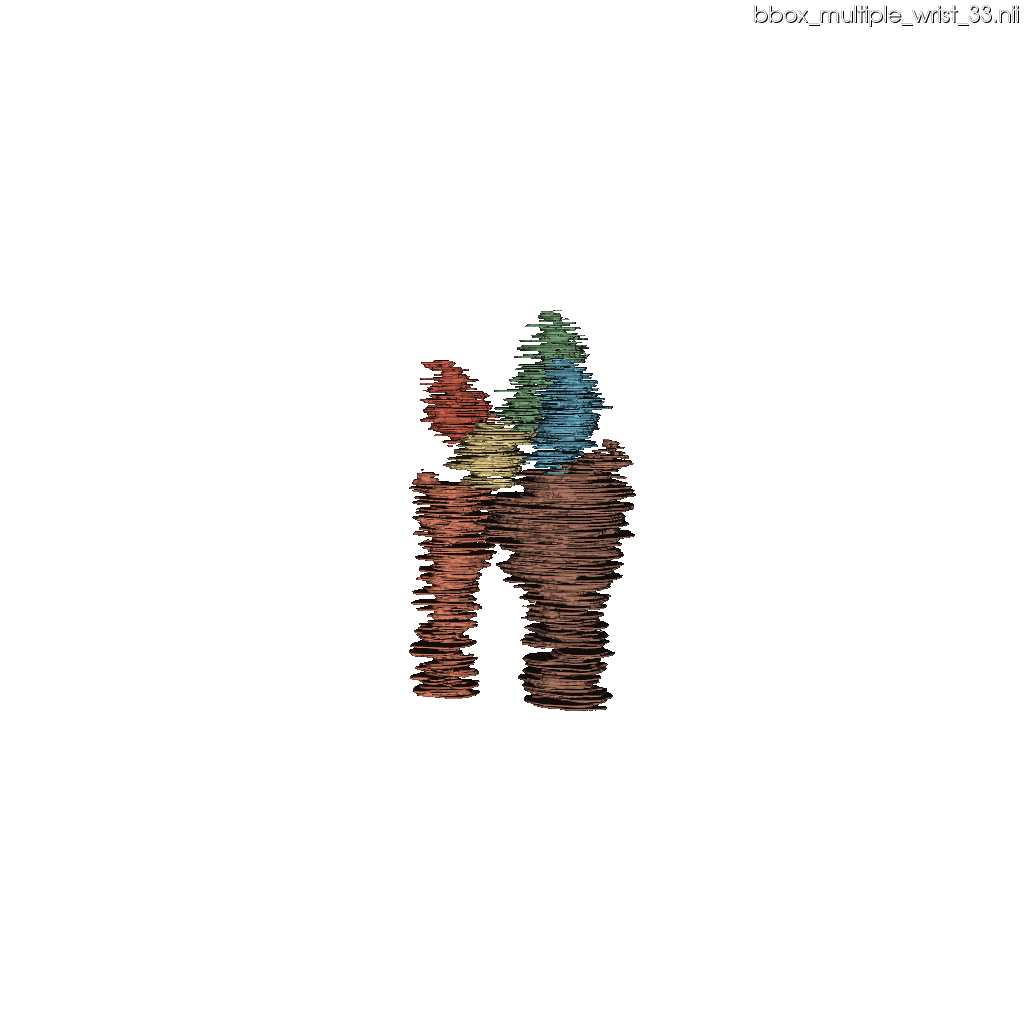} & &
    \includegraphics[height=2cm, trim=370 290 370 280, clip]{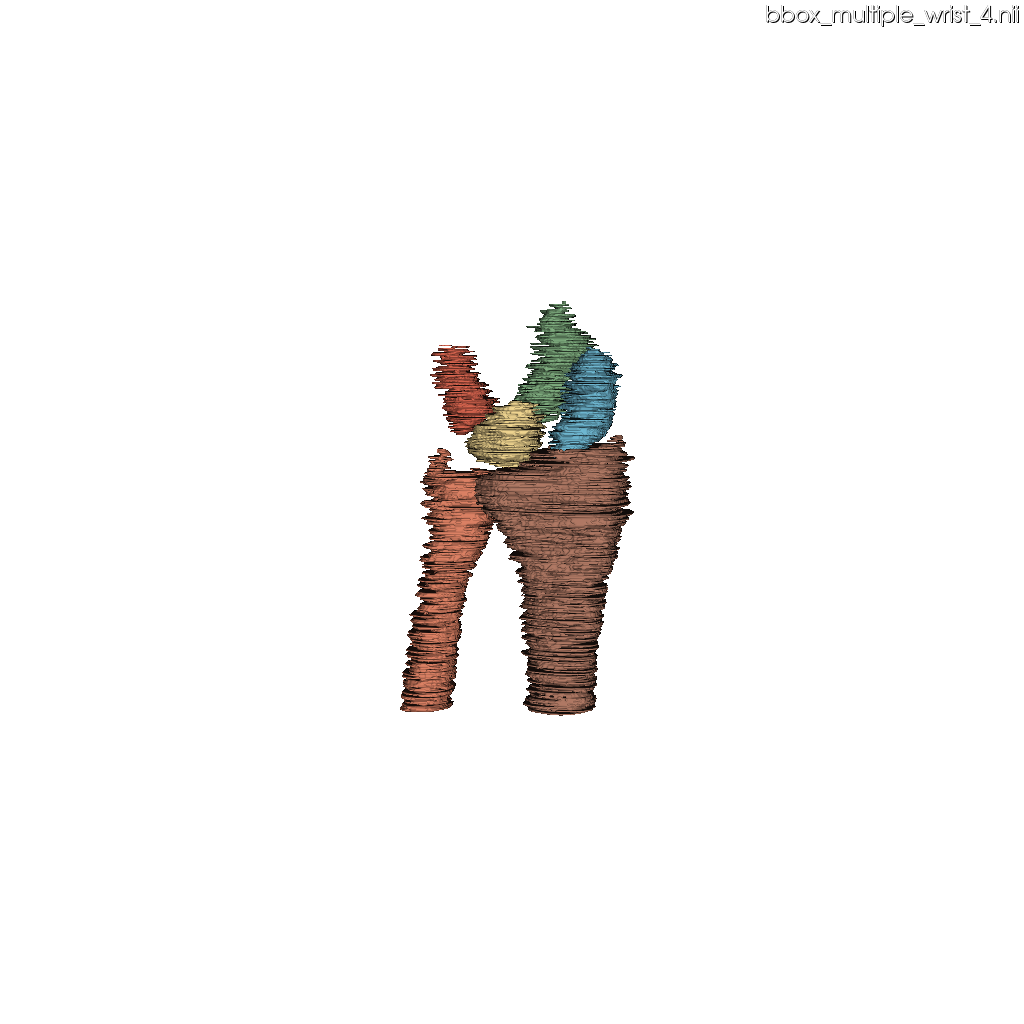} & &
    \includegraphics[height=2cm, trim=370 290 370 280, clip]{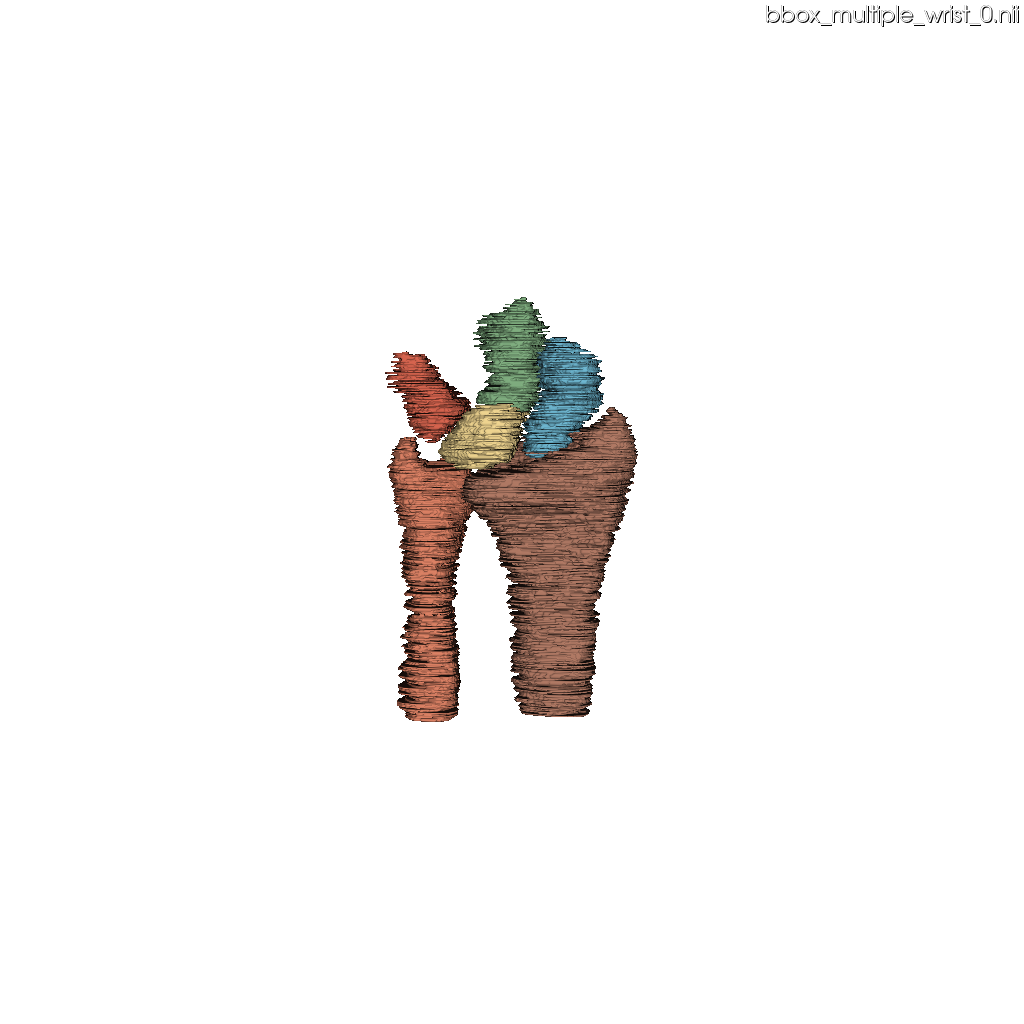} & \\ 
    \hline
    \multirow{2}{*}{\cblacksquare[0.6]{green}} & \includegraphics[height=2cm, trim=580 380 100 320, clip]{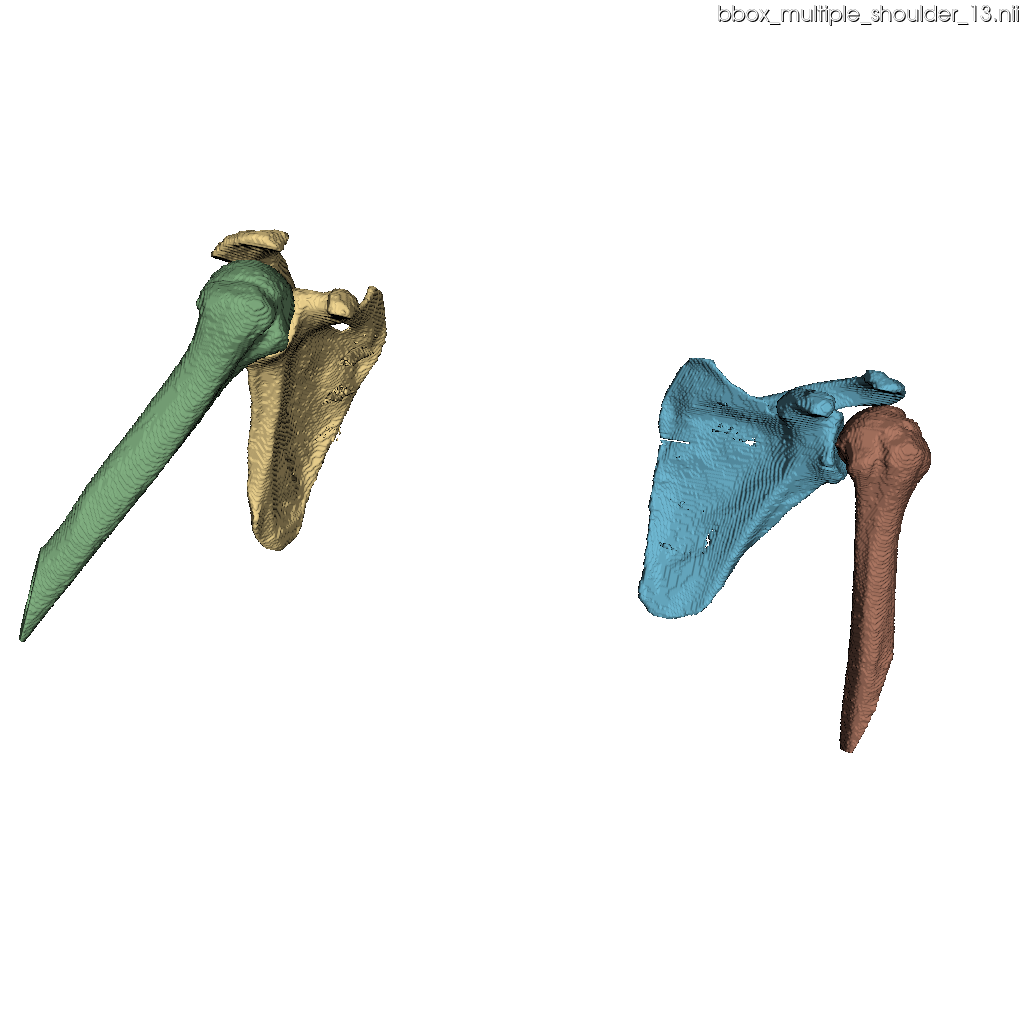} & \raisebox{-0.55\height}[0pt][0pt]{\includegraphics[height=4cm, trim=400 200 430 150, clip]{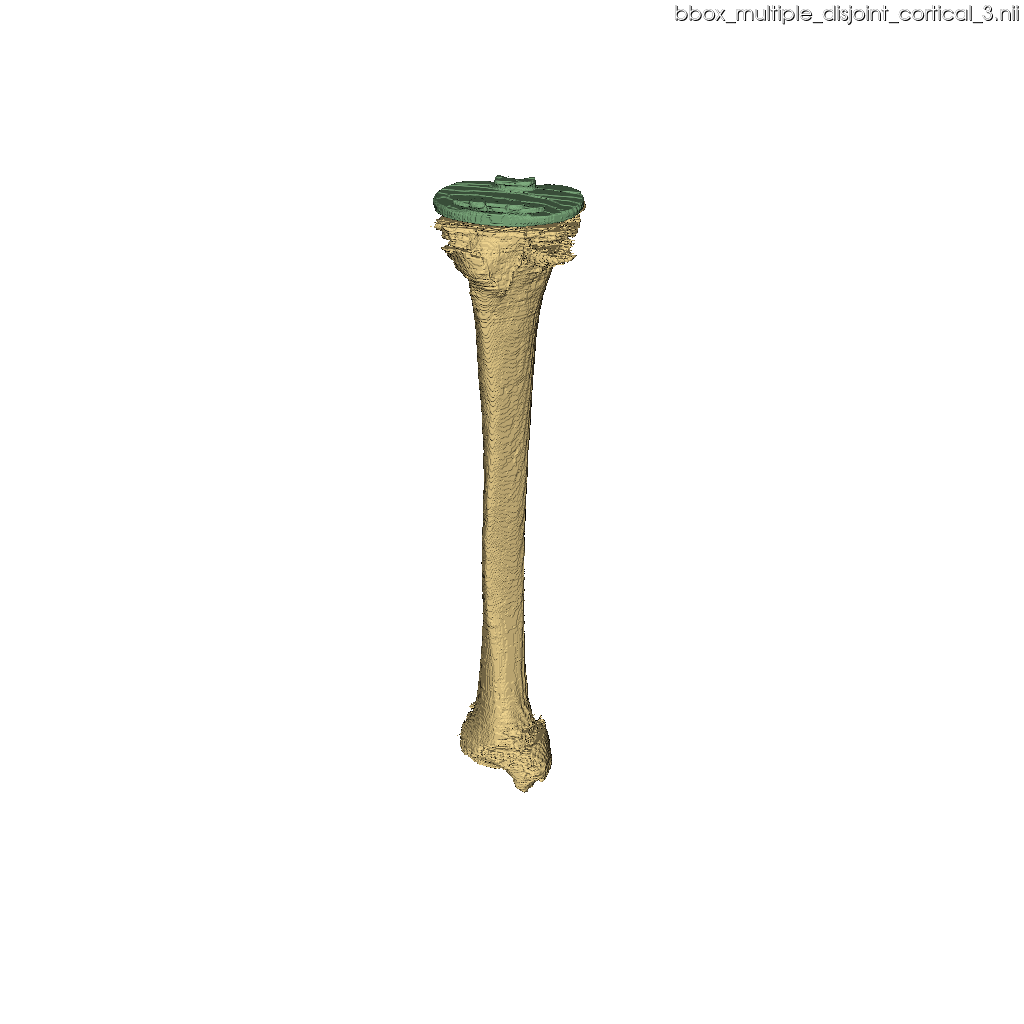}} &
    \includegraphics[height=2cm, trim=560 380 190 330, clip]{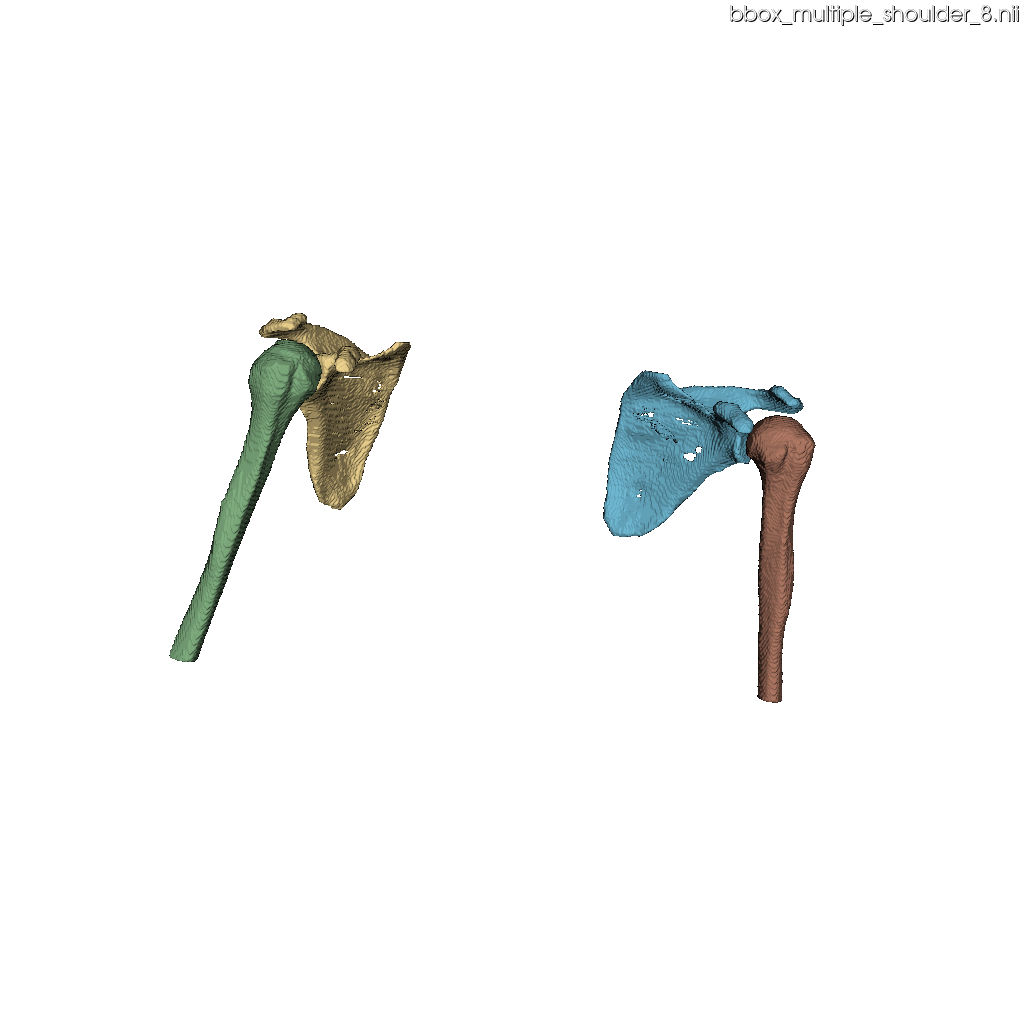} & \raisebox{-0.55\height}[0pt][0pt]{\includegraphics[height=4cm, trim=400 200 410 120, clip]{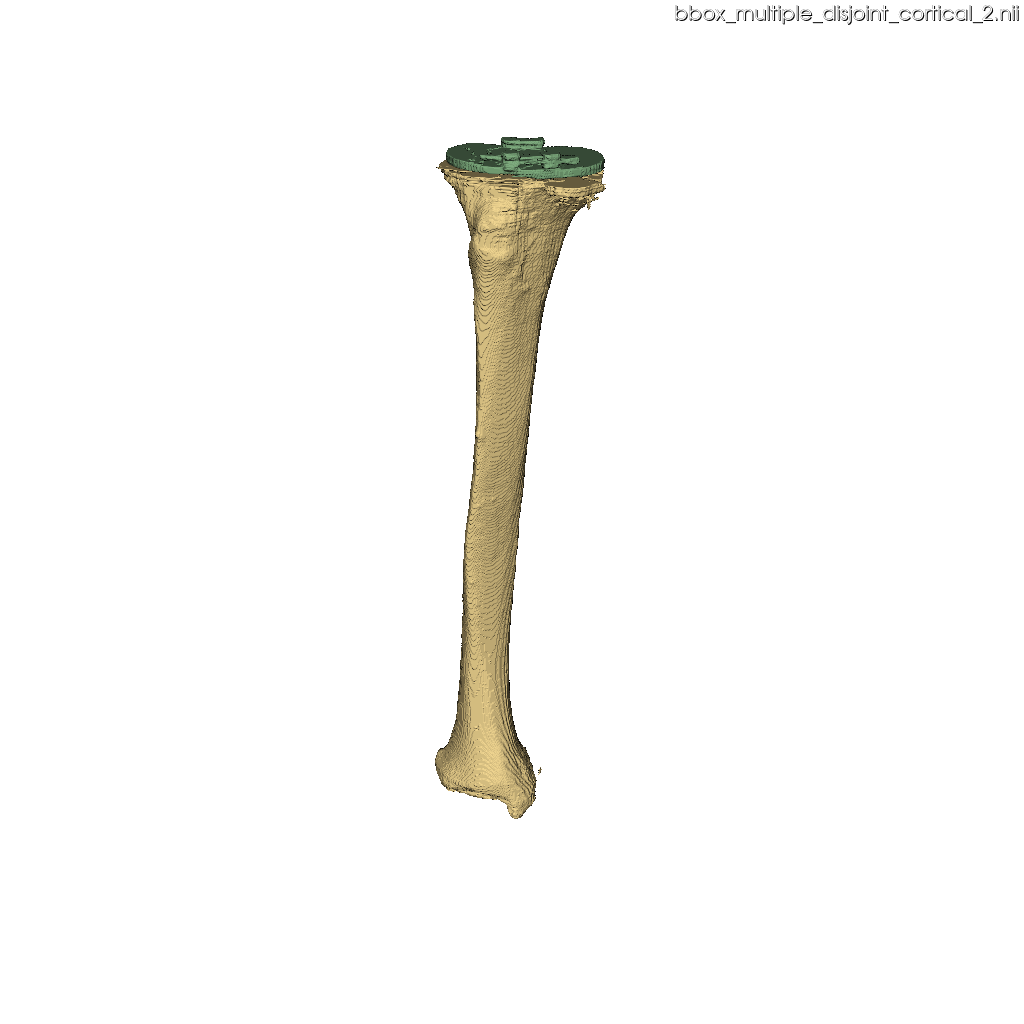}} &
    \includegraphics[height=2cm, trim=560 380 130 320, clip]{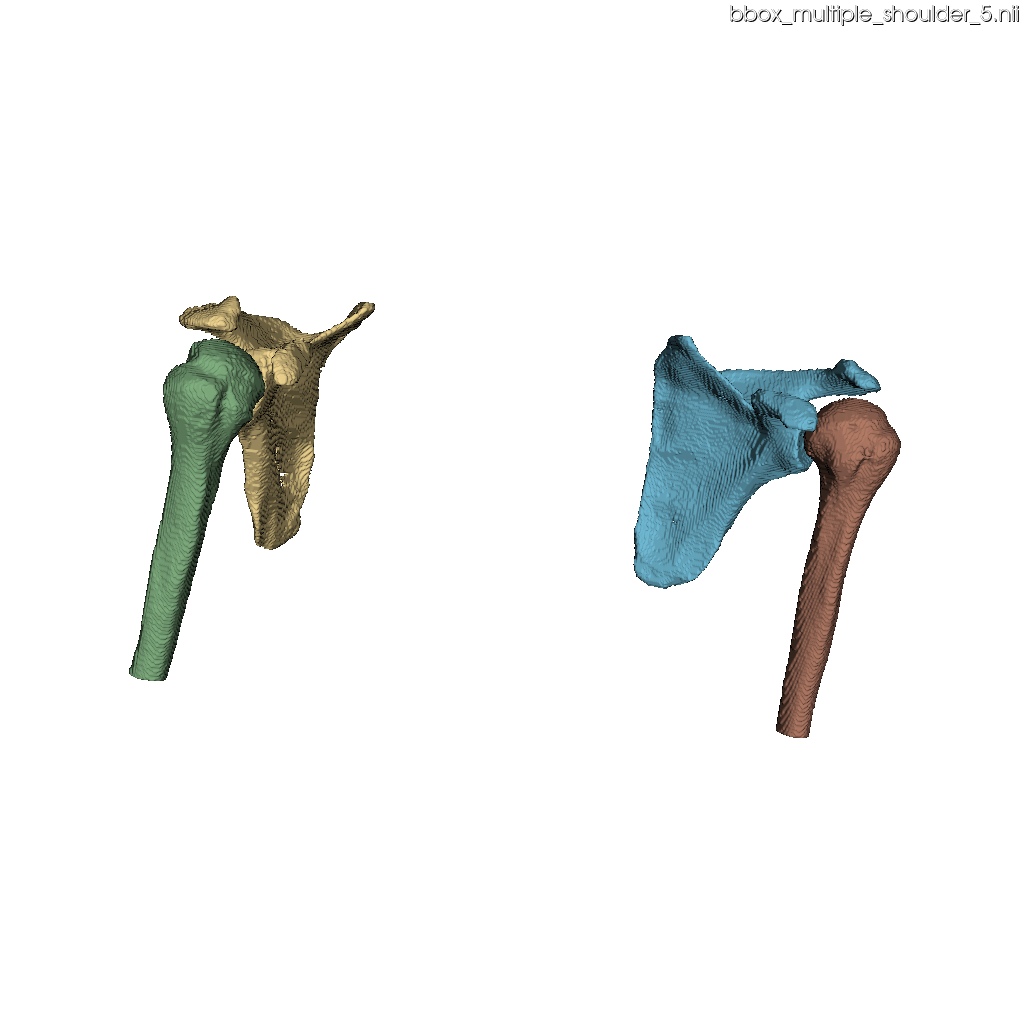} & \raisebox{-0.55\height}[0pt][0pt]{\includegraphics[height=4cm, trim=400 170 410 120, clip]{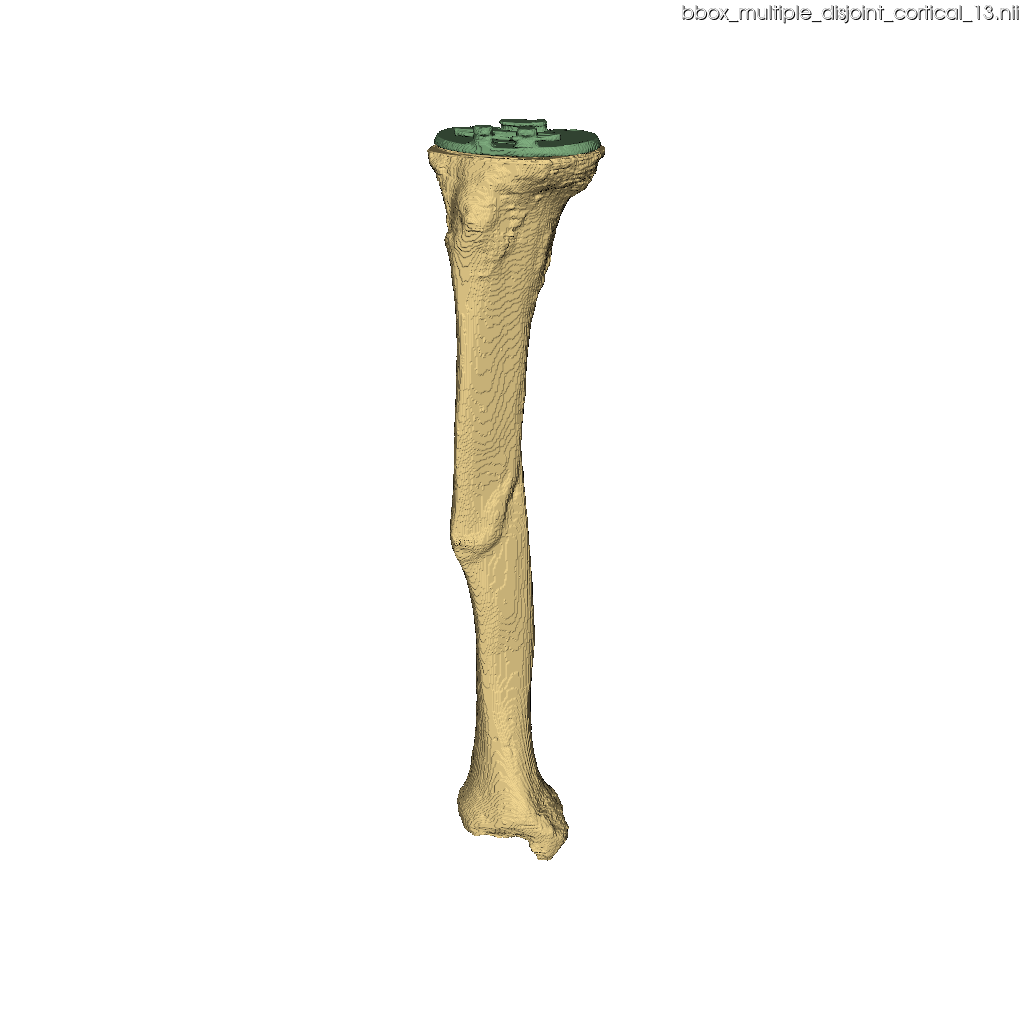}} 
    \\
    & \includegraphics[height=2cm, trim=370 290 370 290, clip]{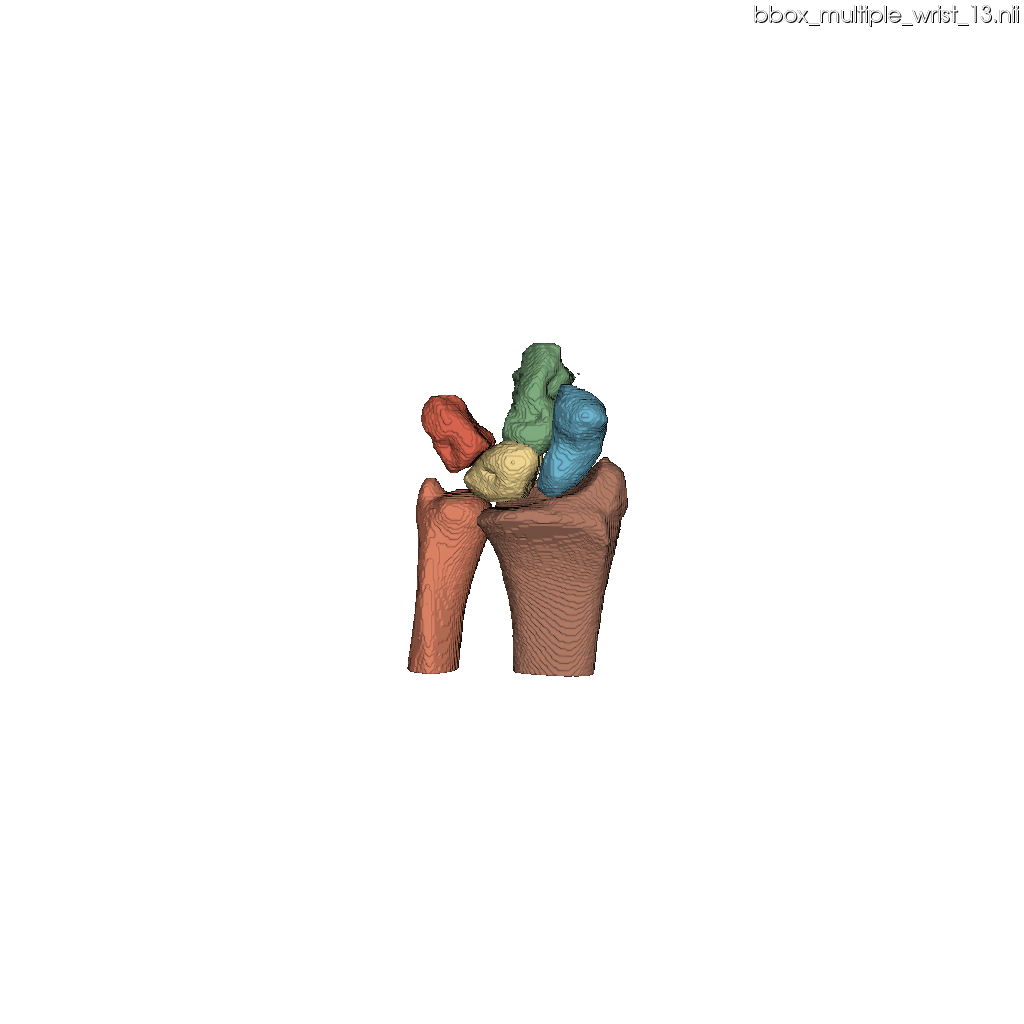} & &
    \includegraphics[height=2cm, trim=370 310 370 250, clip]{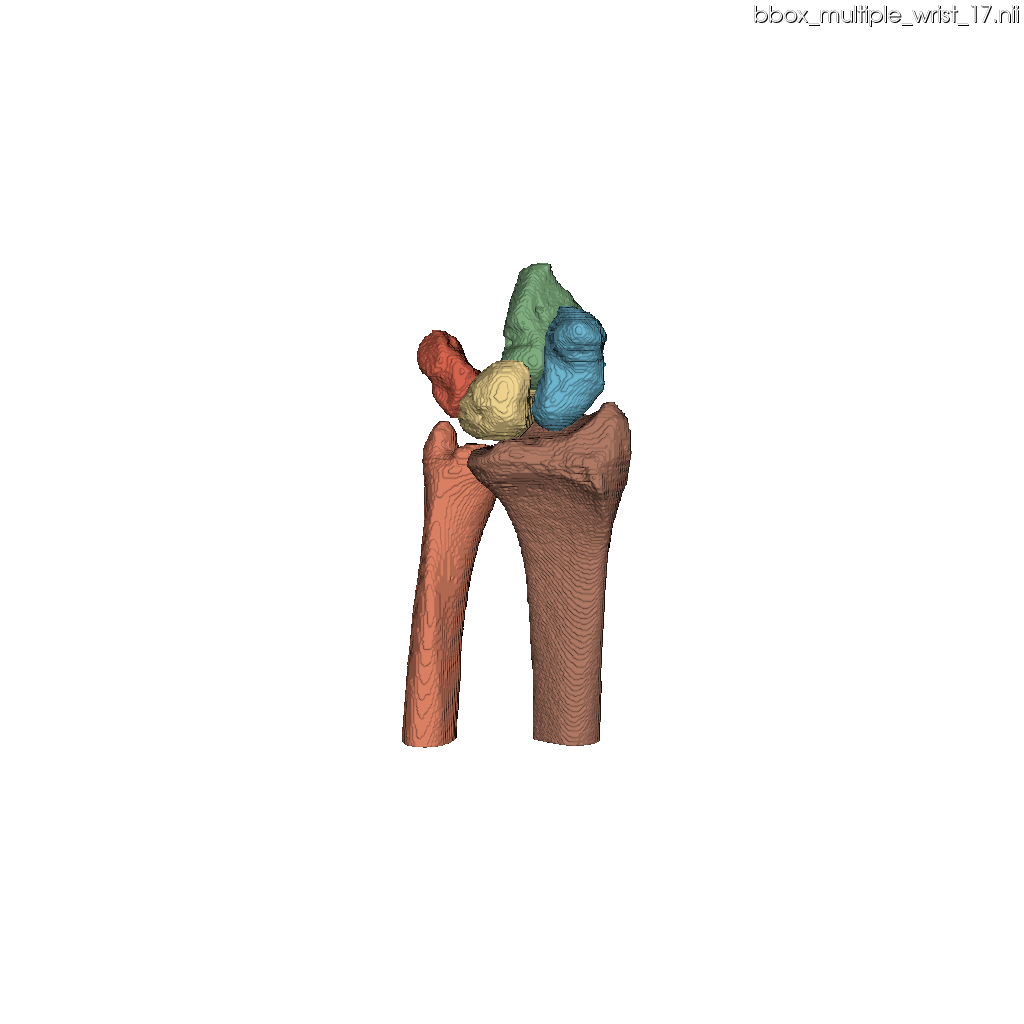} & &
    \includegraphics[height=2cm, trim=370 290 370 280, clip]{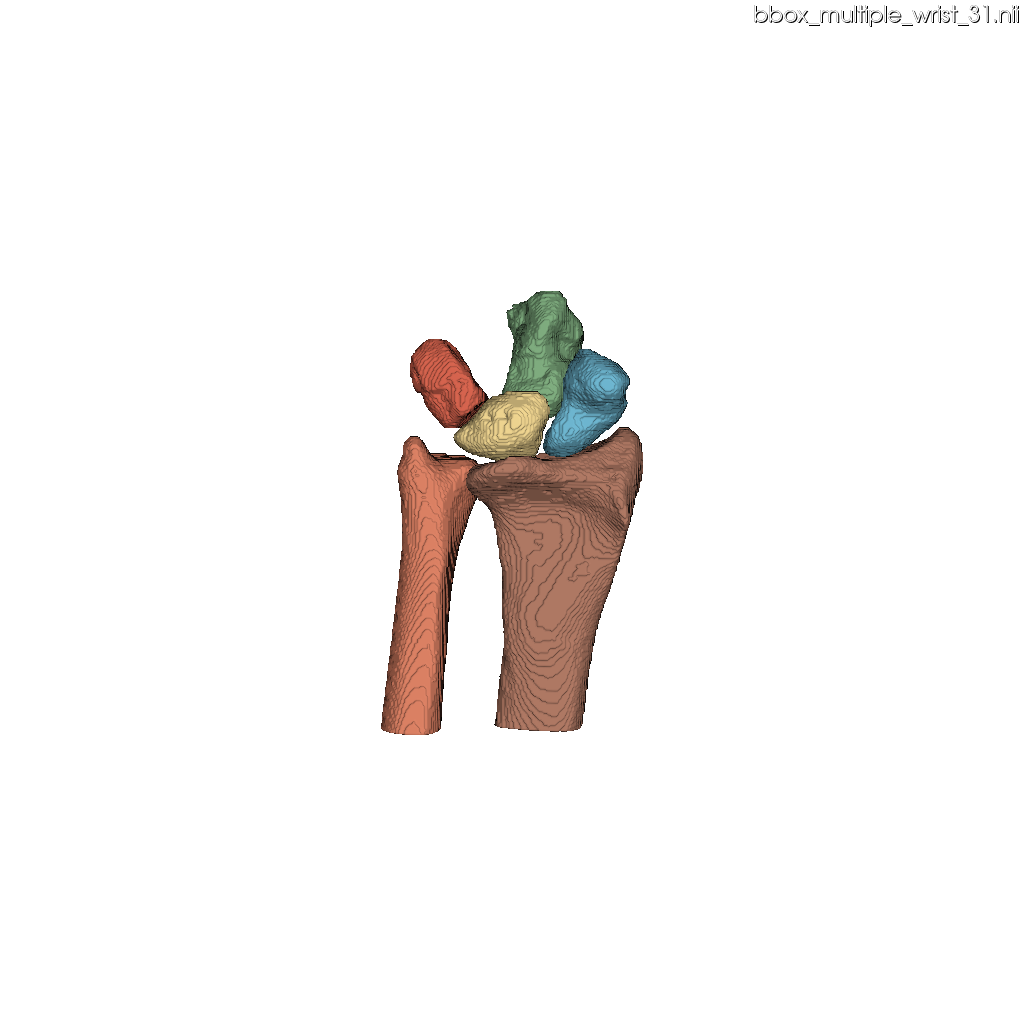} & \\ 
    \hline
    \end{tabular}
\caption{Selected visual examples for \texttt{bounding box 5C} for \textit{Med-\textsc{Sam}} \cblacksquare[0.3]{blue} and \textit{\textsc{Sam} B }\cblacksquare[0.3]{green} with low, medium and high DSC (\%).}
\label{fig:visual_examples}
\end{figure}

Considering pure performance metrics (i.e., maximizing DSC or HD95 or both), the bottom right corner of Fig. \ref{fig:scatter_plot_all} shows an overview of all methods with high DSC and low HD95 averaged over all datasets.
\textit{\textsc{Sam} B} with \texttt{bounding box + center 5C} (\cblacksquaredot{black}) has the highest DSC of $92.5\%$ ($1.7$ mm HD95), whereas \textit{\textsc{Sam} L} with \texttt{bounding box + center 1C} (\cwhitesquaredot{black}) has the lowest HD95 with $1.1$ mm ($92.1\%$ DSC). $59$ out of $258$ methods achieve a DSC higher than $91\%$: $28$ \textit{\textsc{Sam}} and $31$ \textit{\textsc{Sam2}} methods; $43$ methods achieve a HD95 lower than $2$ mm: $23$ \textit{\textsc{Sam}} and $20$ \textit{\textsc{Sam2}} methods. The five highest DSC values ($>92\%$) are achieved by \textit{\textsc{Sam}} methods.
A similar visualization for each dataset individually can be found in Appendix \ref{sec:appendix_scatter_plots}, Fig. \ref{fig:scatter_plot_details}. The highest DSC reached \textit{\textsc{Sam} H} with \texttt{center + 5 negative points 5C} for D1 ($92.8$\%), \textit{\textsc{Sam2} L} with \texttt{bounding box + 5 positive points 1C} for D2 ($97.2$\%), \textit{\textsc{Sam2} B} with \texttt{bounding box + 5 negative points 1C} for D3a ($77.3$\%), and \textit{\textsc{Sam} B} with \texttt{bounding box + center 1C} for D3b ($92.2$\%). The lowest HD95 have \textit{\textsc{Sam} H} with \texttt{bounding box + center 5C} for D1 ($1.0$ mm), \textit{\textsc{Sam2} T} with \texttt{bounding box + center 5C} for D2 ($0.4$ mm),  \textit{\textsc{Sam} L} with \texttt{bounding box 5C} for D3a ($2.7$ mm), and \textit{\textsc{Sam} B} with \texttt{bounding box + center 1C} for D3b ($1.9$ mm). Visual examples are shown in Fig. \ref{fig:visual_examples}, and Appendix \ref{sec:appendx_example_model}, Fig. \ref{fig:example_medsam} - \ref{fig:example_sam2bplus}.

\subsubsection{Best performing prompts}

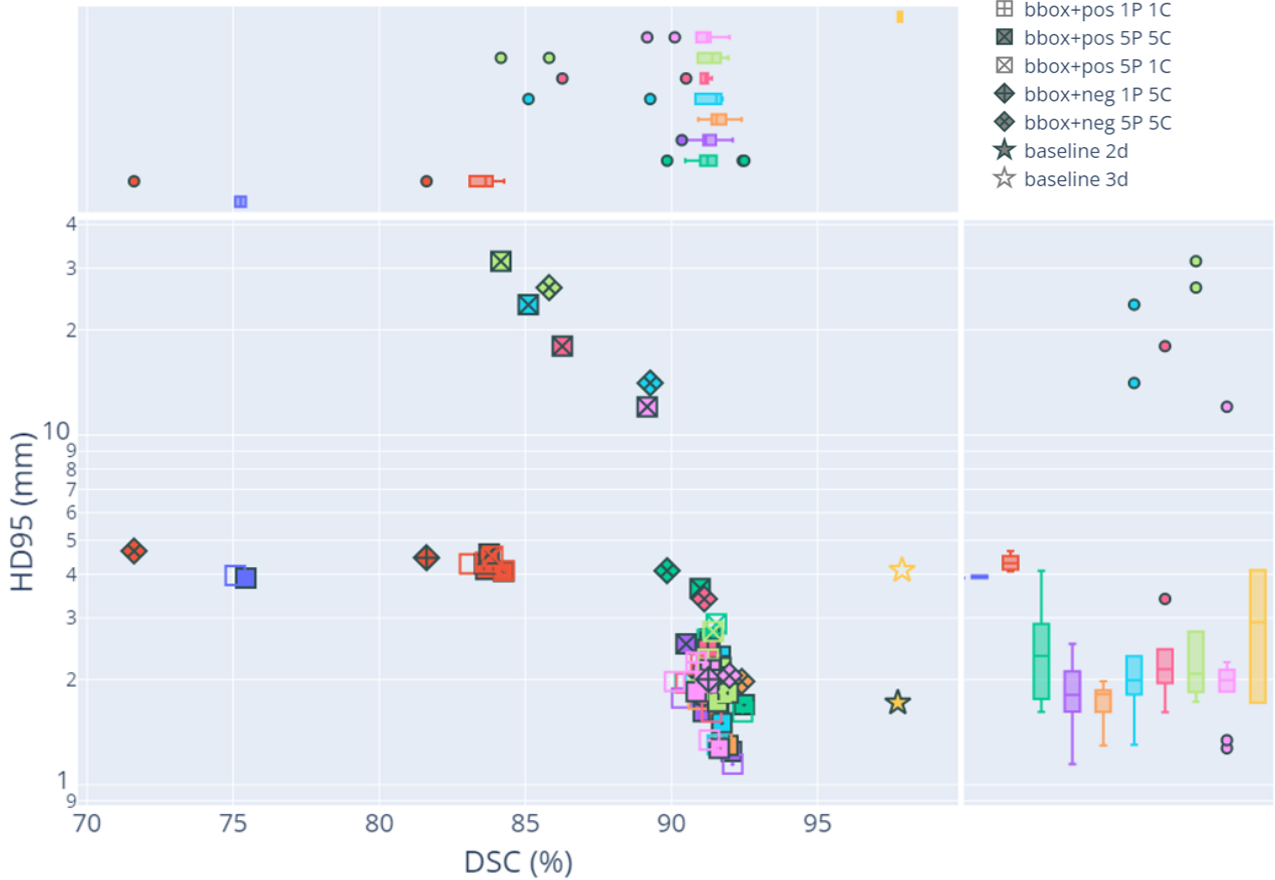
\begin{figure}[h!]
    \centering
    \sbox0{\begin{subfigure}[c]{.75\textwidth}
    \centering
    \setlength{\tabcolsep}{2pt}
    \begin{tabular}{|c|ccc|}
    \hline
    Prompt &  avg Ranking & \makecell{DSC (\%) \\ avg (std)} & \makecell{HD95 (\%) \\ avg (std)}\\
    \hline
    \cblacksquaredot[0.3]{grey}    &  2.38 &  90.89 (10.0) &    1.87 (2.7) \\
    \cwhitesquaredot[0.3]{grey}    &  2.88 &  90.80 (10.0) &    1.79 (2.1) \\
    \cwhitesquarex[0.3]{grey}      &  4.38 &  90.49 (10.2) &    2.50 (2.8) \\
    \cblacksquarecross[0.3]{grey}  &  4.50 &   90.44 (9.6) &    2.50 (2.6) \\
    \cblacksquare[0.3]{grey}       &  5.33 &  88.68 (10.4) &    2.25 (2.2) \\
    \cblackdiamondcross[0.3]{grey} &  6.25 &   90.15 (9.0) &    2.26 (2.3) \\
    \cwhitesquarecross[0.3]{grey}  &  7.38 &   90.22 (9.7) &    2.49 (2.4) \\
    \cwhitesquare[0.3]{grey}       &  9.67 &  88.03 (10.7) &    2.39 (2.2) \\
    \cblackdiamondx[0.3]{grey}     & 10.38 &  87.91 (10.2) &   7.32 (16.1) \\
    \cblacksquarex[0.3]{grey}      & 10.75 &  87.67 (14.1) &  12.20 (28.1) \\
    \hline
    \end{tabular}
    \end{subfigure}}
    \hspace{1em}
    \sbox1{\begin{subfigure}[c]{.2\textwidth}
    \centering
    \adjincludegraphics[trim={{0.82\width} {0.08\height} 0 {0.01\height}}, clip, scale=0.5]{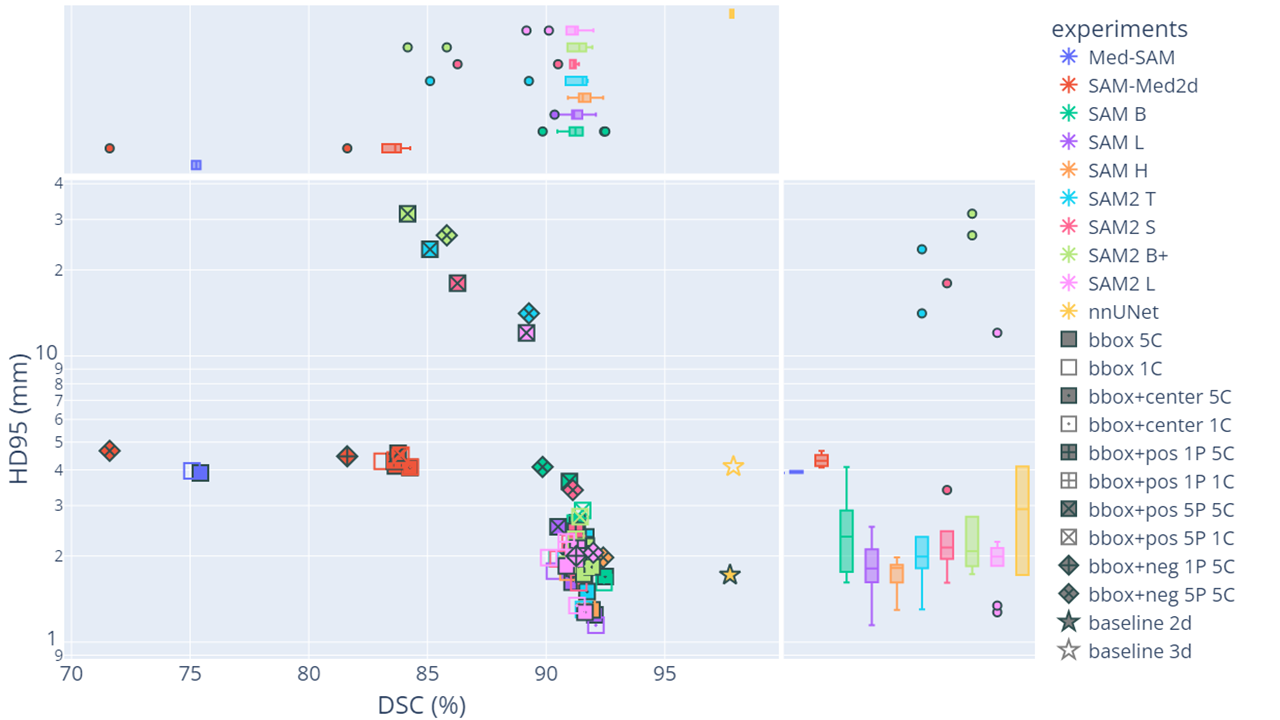}
    \end{subfigure}}

    \sbox2{\adjincludegraphics[trim={0 0 {0.18\width} 0}, clip, width=1.0\linewidth]{scatter_plots/sam_scatter_plot_all_marginal.PNG}}
    
    \begin{tikzpicture}
        \node[inner sep=0pt] at (-1.7,9) {\usebox0};
        \node[inner sep=0pt] at (0,0) {\usebox2};
        \node[inner sep=0pt] at (6,7.65) {\usebox1};
    \end{tikzpicture} 
    \caption{10 best performing 2D prompting strategies across models: The prompt ranking is determined per model by means of the average DSC over all datasets (i.e., highest DSC corresponds to rank 1) and then averaged over all models. The visualization shows the scatter plot with the performance distribution per model across different prompting strategies. Note that the 10 best performing prompts are a subset of the prompts visualized in Fig. \ref{fig:scatter_plot_all} D.}
    \label{fig:scatter_plot_all_best}
\end{figure}

Fig. \ref{fig:scatter_plot_all_best} shows the 10 best performing 2D prompting strategies across different models: \texttt{bounding box 5C and 1C}, \texttt{bounding box + center 5C and 1C}, \texttt{bounding box + 1 positive point 5C and 1C}, \texttt{bounding box + 5 positive points 1C}, \texttt{bounding box + 1 negative points 5C and 1C}, ~and \texttt{bounding box + 5 negative points 5C} (\cblacksquare[0.3]{grey}, \cwhitesquare[0.3]{grey}, \cblacksquaredot[0.3]{grey}, \cwhitesquaredot[0.3]{grey}, \cblacksquarecross[0.3]{grey}, \cwhitesquarecross[0.3]{grey}, \cblacksquarex[0.3]{grey}, \cwhitesquarex[0.3]{grey}, \cblackdiamondcross[0.3]{grey}, \cblackdiamondx[0.3]{grey}). They are determined by ranking the prompts from 1 to 32 (1 being the best) for each model based on their average DSC over all datasets. For the top 10 strategies, \textit{\textsc{Sam} H} has the highest DSC with 91.6\% and the lowest HD95 with 1.73 mm, followed by \textit{\textsc{Sam} L} and \textit{\textsc{Sam} B} with DSC around $91\%$ and HD95 between $1.8$ and $2.5$ mm. \textit{\textsc{Sam2}} models achieve DSC around $90\%$ and HD95 of $2.8$, $3.7$, $5.2$ and $7.4$ mm for \textit{\textsc{Sam2} L, S, T, B+}, respectively. \textit{\textsc{Sam}-Med2d} has $82.3$\% DSC and $4.3$ mm HD95. \textit{Med-\textsc{Sam}} has $75.3$\% DSC and $3.9$ mm HD95. 
As an example for a fully supervised model, the performance of the individually dedicated nnUNets is averaged: $97.7\%$ DSC and $1.7$ mm HD95 for 3D full resolution nnUNet; $97.9\%$ DSC and $4.1$ mm HD95 for 2D nnUNet. \newline
A similar ranking for each dataset individually is reported in Appendix \ref{sec:appendix_results_detailed}, Table \ref{tab:best_settings_detail}.
New strategies in the top 10 are \texttt{center + 5 negative points 1C} for D1, \texttt{5 positive + negative points 1C} for D2, \texttt{bounding box + 5 positive points 1C} and \texttt{bounding box + 5 negative points 1C} for D3a, \texttt{bounding box + 5 positive points 1C} and \texttt{5 positive + negative points 1C} for D3b.

\subsubsection{Different number of points}

\begin{figure}[h!]
\centering
    \begin{subfigure}[t]{1\textwidth}
    \centering
        \includegraphics[width=0.7\linewidth]{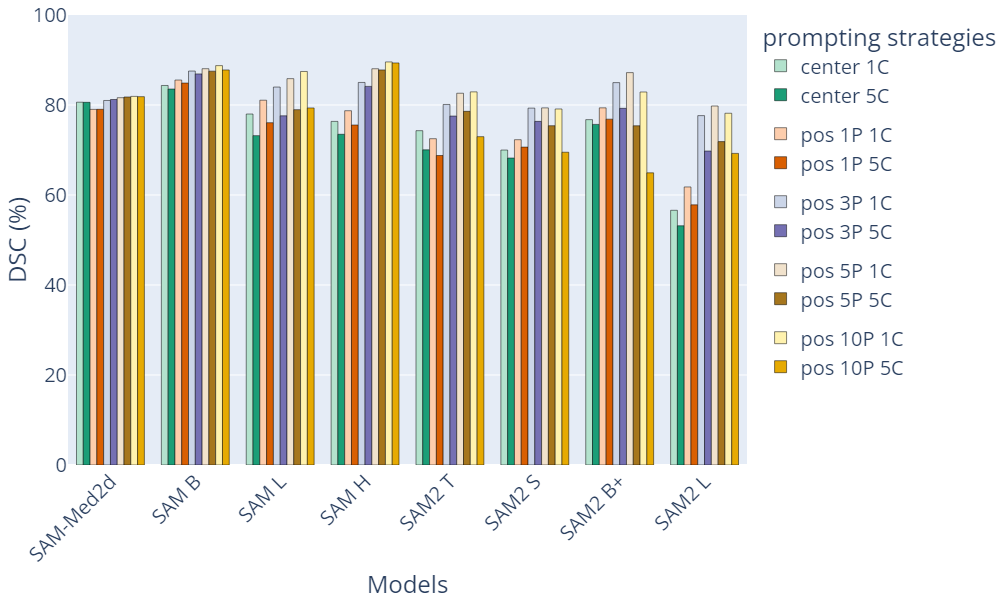}
        \caption{Center and positive random points}
        \label{fig:performance_points_all-random}
    \end{subfigure}
    \begin{subfigure}[t]{1\textwidth}
    \centering
        \includegraphics[width=0.7\linewidth]{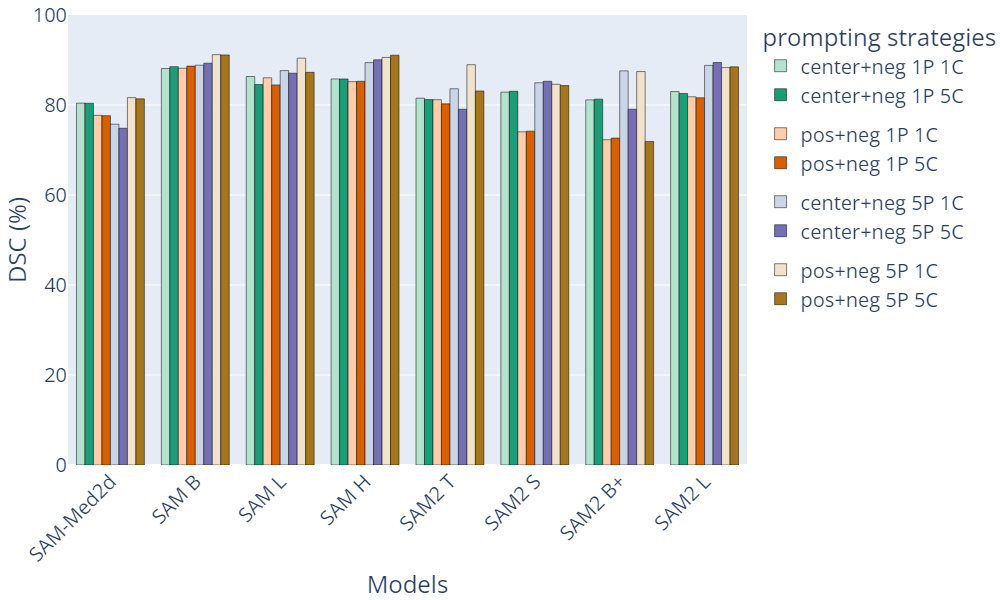}
        \caption{Point combinations}
        \label{fig:performance_points_all-point}
    \end{subfigure}
    \caption{DSC (\%) performance for different number of points per model: (a) center point and 1,3,5,10 random positive points; (b) point combinations of center, 1 or 5 positive and negative points.}
    \label{fig:performance_points_all}
\end{figure}

In Fig. \ref{fig:performance_points_all} prompting strategies with different number of points are compared. For the point prompts, the best DSC achieves \textit{\textsc{Sam} H} \texttt{10 random points 1C} with $89.6$\%, followed by \textit{\textsc{Sam} L} \texttt{10 random points 1C} with $87.4$\% and \textit{\textsc{Sam2} B+} \texttt{5 random points 1C} with $87.2$\%. For \textit{\textsc{Sam}-Med2d} and \textit{\textsc{Sam2} T} the best setting is \texttt{10 random points 1C}, same as for all \textit{\textsc{Sam}} models. The remaining \textit{\textsc{Sam2}} models perform best with \texttt{5 random points 1C}. For the point combination prompts, \texttt{5 positive + negative points} performs best for \textit{\textsc{Sam} H} (\texttt{5C}, $91.1\%$), \textit{\textsc{Sam} B} (\texttt{1C}, $91.2\%$) and \textit{\textsc{Sam} L} (\texttt{1C}, $90\%$). It is also the best setting for \textit{\textsc{Sam}-Med2d} and \textit{\textsc{Sam2} T}. The remaining \textit{\textsc{Sam2}} models work best with \texttt{center + 5 negative points}. The visualizations for each dataset individually are shown in Appendix \ref{sec:appendix_barplot}, Fig. \ref{fig:performance_points_detail}. Visual examples are shown in Appendix \ref{sec:examples_points}, Fig. \ref{fig:examples_points}. 

\subsubsection{Different labeling protocols}

\begin{figure}[h!]
    \centering
    \begin{subfigure}[t]{1\textwidth}
        
    \end{subfigure}
    \includegraphics[width=.7\textwidth]{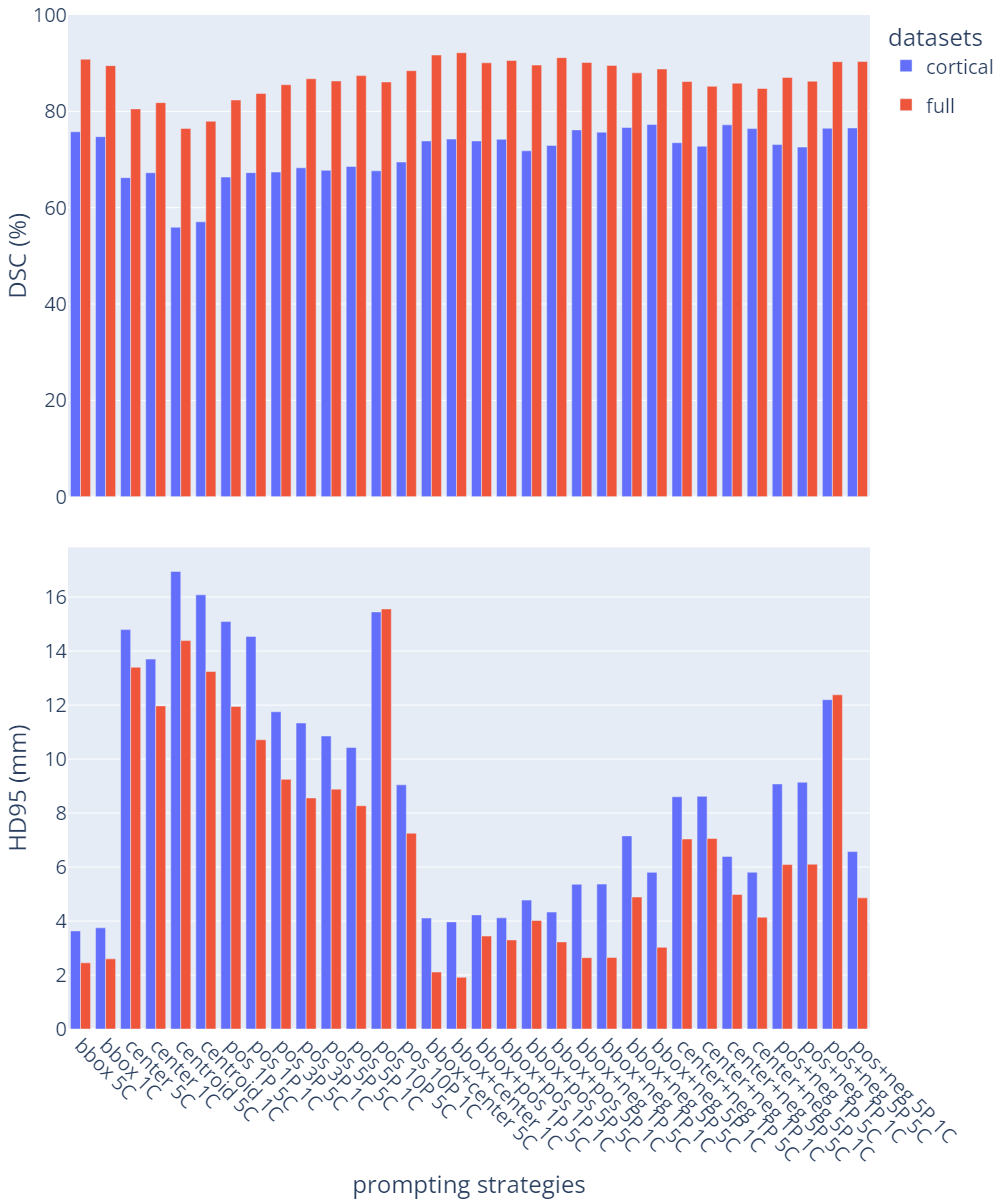}
    \caption{\textit{\textsc{Sam} B} performance (top - DSC (\%); bottom - HD95 (mm)) for different labeling protocols in the knee dataset (D3), i.e., cortical tibia bone (D3a) and full tibia bone (D3b).}
    \label{fig:performance_knee_comparison}
\end{figure}

For both subsets of the knee dataset (D3), the highest DSC is achieved with \textit{\textsc{Sam} B} with \texttt{bounding box 5C} ($74.2$\% DSC and $2.7$ mm HD95) for D3a and with \texttt{bounding box + center 1C} ($92.2$\% DSC and $1.9$ mm HD95) for D3b. Fig. \ref{fig:performance_knee_comparison} shows the performance of both subsets for \textit{\textsc{Sam} B} with all 2D prompting strategies. The full bone labeling protocol achieves higher DSC and lower HD95 for each prompting strategy. Visual examples are shown in Appendix \ref{sec:appendix_example_knee}, Fig. \ref{fig:examples_knee}.

\subsection{Inference Time}

\begin{figure}[h!]   
\captionof{table}{Average prediction time per slice (sec.): The table on the left sorts the inference time averaged over all prompting strategies in ascending order. The line plot on the right shows the time per slice (sec.) for the different prompting strategies for each model.}
\begin{subfigure}[c]{.59\textwidth}
    \centering
    \setlength{\tabcolsep}{2pt}
    \begin{tabular}{|l|ccc|}
    \hline
    Model & \makecell{Avg. time per \\ slice (s)} & \makecell{\# Model \\ Parameter} & Image Size \\
    \hline
    SAM-Med2d &                 0.052 &       271 &    256x256 \\
       SAM2 T &                 0.068 &        38 &  1024x1024 \\
       SAM2 S &                 0.087 &        46 &  1024x1024 \\
      SAM2 B+ &                 0.107 &        80 &  1024x1024 \\
        SAM B &                 0.150 &        93 &  1024x1024 \\
       SAM2 L &                 0.218 &       224 &  1024x1024 \\
        SAM L &                 0.332 &       312 &  1024x1024 \\
        SAM H &                 0.588 &       641 &  1024x1024 \\
      Med-SAM &                 1.658 &        93 &  1024x1024 \\
      \hline
    \end{tabular}
\end{subfigure}
\begin{subfigure}[c]{.39\textwidth}
    \centering
    \includegraphics[width=1.0\linewidth]{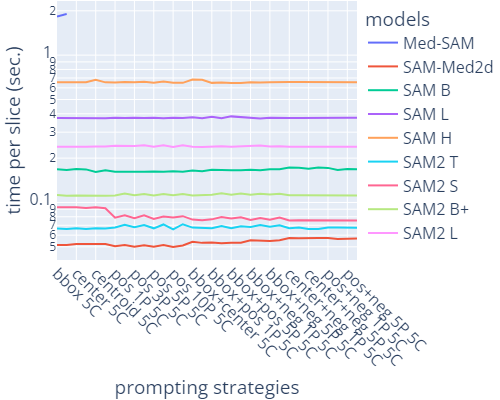}
\end{subfigure}
\label{tab:results_inference_time}
\end{figure}

As inference time per slice (sec.) might be related to number of model parameters, image sizes and prompting strategies, they are all reported together in Table \ref{tab:results_inference_time}. The fastest prediction time has \textit{\textsc{Sam}-Med2D} with $0.052$ sec per slice, followed by \textit{\textsc{Sam2}} and \textit{\textsc{Sam}} versions.

\section{Discussion}

\subsection{Segmentation performance} \label{sec:discussion_segmentation_performance}

\subsubsection{Overview}
\textit{Med-\textsc{Sam}} and \textit{\textsc{Sam}-Med2d}, which are both fine-tuned on a medical image dataset containing CT scans, are outperformed by \textit{\textsc{Sam}} and \textit{\textsc{Sam2}}, trained on natural images, with most of the \texttt{bbox} based prompts. We attribute this to the unlearning of general representations caused by fine-tuning on a large variety of medical domains.  \newline
Fig. \ref{fig:scatter_plot_all} shows a cluster of red symbols in (A)-(C) corresponding to the \textit{\textsc{Sam}-Med2d} experiments. All prompt strategies have the same magnitude of HD95 ($4$-$7$ mm) and DSC between $71$ \% and $85$ \%. We conclude that the prompting strategy has little impact on the performance of \textit{\textsc{Sam}-Med2d}. This is in contrast to \textit{\textsc{Sam}} and \textit{\textsc{Sam2}} experiments. The associated symbols form an arc from high DSC and low HD95 (``good'' performance, i.e., lower right corner) to low DSC and high HD95 (``bad'' performance, i.e., upper left quadrant) with increasing DSC standard deviation (i.e., increasing symbol size). The absence of obvious color and symbol clusters in the described arc shows that the performance is highly dependent on the prompting strategy and no overall best strategy can be determined. However, the zoomed-in lower right corner in Fig. \ref{fig:scatter_plot_all} (D) shows three groups of prompting strategies: bounding box-only prompts and bounding box + point combinations in the lower half and point-based combination with 5 random points (i.e., \texttt{center + 5 negative points} and \texttt{5 positive + negative points}) in the upper left quadrant.

\subsubsection{Best performing prompts}
The top $10$ best performing prompting strategies across models with respect to DSC are bounding box-based, i.e., either bounding on its own or combined with point prompts (Fig. \ref{fig:scatter_plot_all_best}). For each dataset, the majority of the overall $10$ best performing prompting strategies can also be found in the individual top $10$, although the ranking position differs (Fig. \ref{fig:scatter_plot_details}, Table \ref{tab:best_settings_detail}). \newline
A comparison between FMs -- such as \textit{\textsc{Sam}}-based models -- and nnUNet -- a fully supervised and dedicated trained model -- is not completely fair, as FMs do not have any prior knowledge about specific examples and nnUNet is explicitly trained for the given task. However, it gives us valuable insight about the capability and competitiveness of \textsc{Sam}-family models with current, established state-of-art-methods. Both nnUNet versions achieve a higher DSC. However, some \textsc{Sam}-based experiments achieve a lower HD95, especially for the knee dataset (Table \ref{tab:best_settings_detail}). This is due to the oversegmentation of the femur as tibia for nnUNet models (Fig. \ref{fig:examples_knee02}).

\subsubsection{Different number of points}
Comparing prompt strategies with different number of points shows that more points are not always better (Fig. \ref{fig:performance_points_all}). For \textit{\textsc{Sam}-Med2D}, \textit{\textsc{Sam2} T} and \textit{\textsc{Sam}} models, more random points result in better DSC, whereas for \textit{\textsc{Sam2}} models five positive points achieve the highest DSC (Fig. \ref{fig:performance_points_all-random}). Except for \textit{\textsc{Sam}-Med2d} and \textit{\textsc{Sam2} T}, using the center point does not work better than one random positive point. Considering human-created annotations, this is promising: a human annotator may not place the point precisely at the center due to intensity differences in the object, and even when aiming for the center, the prompt might be slightly offset, resembling a random point placement inside the object. Comparing the point combinations shows that prompts with the center point perform better than the random point combination if there is only one positive point for \textit{\textsc{Sam2}}, while with five positive points, the random point combinations demonstrates stronger performance for \textit{\textsc{Sam}} (Fig. \ref{fig:performance_points_all-point}). For all models, more negative points are better. We conclude that \textit{\textsc{Sam2}} versions (except \textit{\textsc{Sam2} T}) do not need as many prompts as \textit{\textsc{Sam}} to reach their best performance. However, while \textit{\textsc{Sam}} models achieve their highest DSC with more points, also the DSC with the lower point equivalent is higher.\newline
For some models, such as \textit{\textsc{Sam} L} and \textit{\textsc{Sam2}} models, there is a noticeable difference between the two component selection criteria, where taking the prompt from the largest component (\texttt{1C}) performs better than taking prompts from up to 5 components (\texttt{5C}). The latter results in oversegmentation of other bone structures (Fig. \ref{fig:examples_points}). This observations is particularly obvious for D1 (Fig. \ref{fig:performance_points_shoulder}), a dataset that contains an object (i.e., scapula) that is split into multiple components in axial slices. 

\subsubsection{Different labeling protocols}

For the knee subset (D3), there are two different label sets available for the tibia bone, i.e., the cortical part of the bone (D3a) or the full bone (D3b). 
Independent of the prompting strategy, the cortical subset performs worse  (Fig. \ref{fig:performance_knee_comparison}). Inspecting visual examples reveals that despite prompt extraction from the cortical bone, the prediction frequently includes the whole bone (Fig. \ref{fig:examples_knee}). We hypothesize that the prompts -- both bounding box and point prompts -- may be too ambiguous for the \textsc{Sam}-family models, leading the models to default to segmenting the full bone rather than individual components. Our objective in this work was to test generic prompting strategies without relying on dataset or task specific knowledge. However, to improve the segmentation of cortical bone, incorporating prior knowledge of the dataset and task could help mitigate ambiguities in the prompts. For example, consistently adding negative points in the center of the bone might guide the models more effectively towards the intended cortical bone structure. In future work, we intend to look further into the knee dataset to test this hypothesis and refine the prompting strategies on this specific task.

\subsection{Inference time}

As the evaluation was performed on a server that is available for multiple users, different server utilization can influence the exact inference time. Despite this limitation, a clear trend can be observed across different datasets, which were performed in separate evaluation runs. The inference time is highly influences by the number of model parameters but not by the different prompting strategies (Table \ref{tab:results_inference_time}). The exceptions are the medically fine-tuned \textsc{Sam}-version. \textit{Med-\textsc{Sam}} has the slowest inference time, which we attribute to code inefficiency, as the number of parameters and image size match with \textit{\textsc{Sam} B}. \textit{\textsc{Sam}-Med2d} has the fastest average inference time due to the smaller image size, as it is the only model using $256\times256$. For all \textit{\textsc{Sam}} and \textit{\textsc{Sam2}} versions, the model size determines the inference time, i.e., the more model parameter, the longer the inference time. 

\subsection{Overall performance}

\subsubsection{Model sizes}
Taking both segmentation performance and inference time into consideration, large models such as \textit{\textsc{Sam} H} and \textit{\textsc{Sam} L} are less favorable due to their longer average time per slice and \textit{Med-\textsc{Sam}} and \textit{\textsc{Sam}-Med2d} due to their sub-par performance. \textit{\textsc{Sam} B} (\textit{\textsc{Sam}} default model size), \textit{\textsc{Sam2} T} and \textit{\textsc{Sam2} B+} (\textit{\textsc{Sam2}} default model size) provide a good trade-off between performance and runtime.

\subsubsection{Prompting Strategies}
Since the inference time is independent of the number of prompts, the general observations in Section \ref{sec:discussion_segmentation_performance} remain valid without the need for reconsideration:
\begin{itemize}
    \item The overall best performing prompt is \texttt{bbox + center}.
    \item All top prompts across different models and datasets are bounding box-based: either bounding box-only or bounding box + point combinations.
    \item The best point combinations are \texttt{center + 5 negative points} for \textit{\textsc{Sam2}} (except \textit{\textsc{Sam2} T}) and \texttt{5 positive + negative points} for \textit{\textsc{Sam}}.
    \item The best one-type point prompt are \texttt{5 positive points} for \textit{\textsc{Sam2}} (except \textit{\textsc{Sam2} T}) and \texttt{10 positive points} for \textit{\textsc{Sam}}.
    \item The worst performing prompt is the \texttt{centroid point}.
\end{itemize}

\subsection{Preliminary guidelines for non-iterative 2D prompting}
\begin{figure}[h!]
    \centering
    \includegraphics[width=0.95\linewidth]{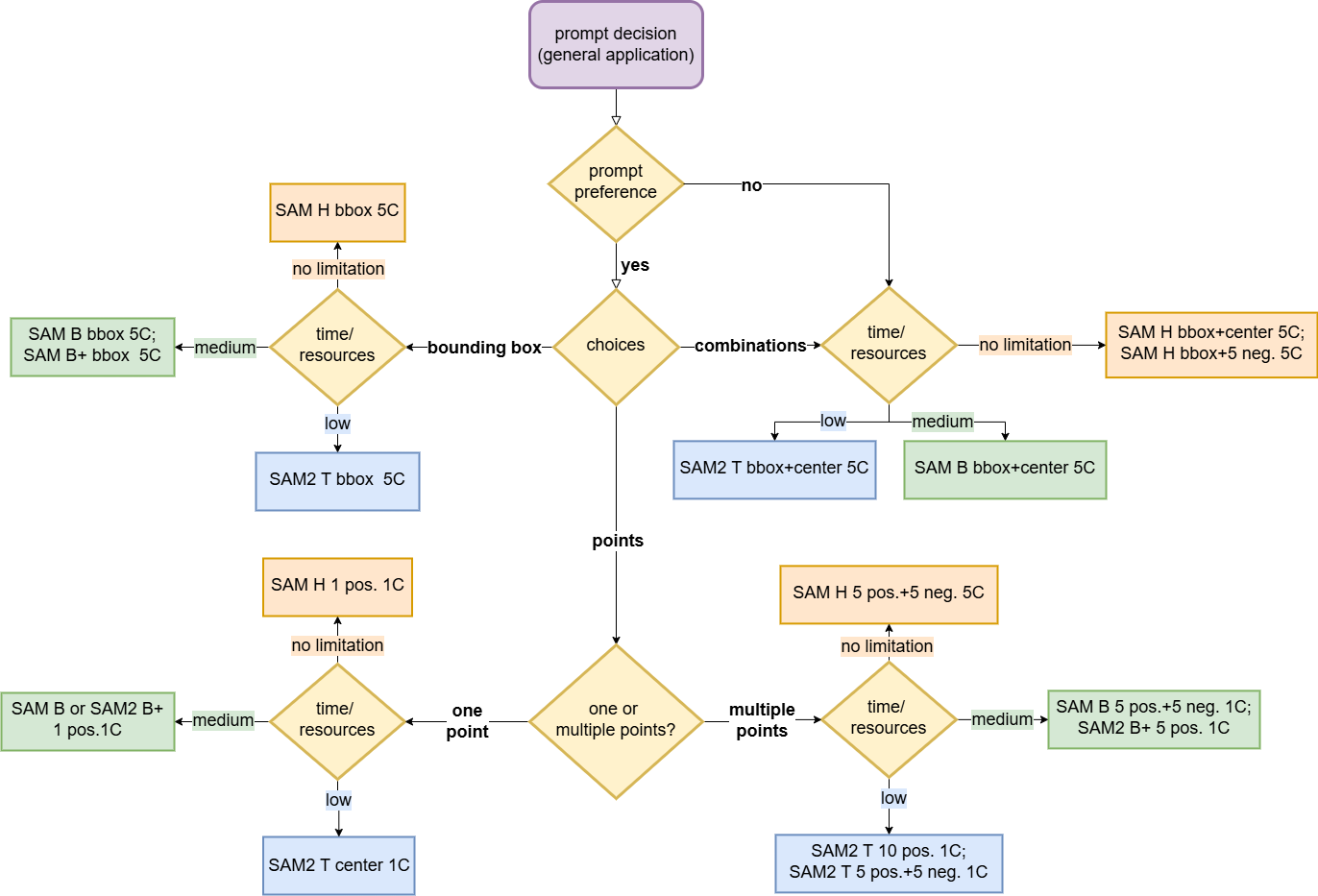}
    \caption{Preliminary guidelines for non-iterative 2D prompting based on our results.}
    \label{fig:guidelines}
\end{figure}

Optimizing over different objectives -- such as inference time of the model, DSC as overlap metric, HD95 as surface-based metric -- for different dataset characteristics leads to a high-dimensional search space. As we can already see in Fig. \ref{fig:scatter_plot_all}, there is no experiment with highest DSC and lowest HD95. Even for the individual datasets, only dataset D3b has a setting that maximizes both metrics: \textit{\textsc{Sam} B} with \texttt{bbox+center 1C}. 
For the rest, a trade-off always remains on which metric to ultimately maximize. Our results can help to identify the ``Pareto front'' -- i.e., the limit where we cannot improve one metric without decreasing the other -- but deciding where to stand on that border remains application specific; highly dependent on the downstream tasks for clinical use and impact of errors.
Nevertheless, based on our current results, we are already able to draw some preliminary guidelines that can help to narrow-down the optimal setting for non-interactive 2D prompting for bone segmentation in CT scans, as shown in Figure \ref{fig:guidelines}.

\subsection{Limitations \& Future Work}
Our private dataset with 80 CT scans is smaller as comparable publicly available datasets (e.g., TotalSegmentator \cite{wasserthal2023totalsegmentator}). As FMs tend to leverage all publicly available datasets, performing an independent and fair comparison is increasingly challenging. We intend to keep our dataset private to prevent its use in FM training and we plan to expand it to include additional skeletal regions (e.g., hip, ankle). We believe it is essential for the research community to work collaboratively toward establishing fair and consistent benchmarks for balanced and transparent model evaluations.\newline
Another limitation of our study is that we did not repeat experiments with varied random point positions or apply jitter to the center point or bounding box extraction which mimics human annotations more. Since prompt positioning can influence the performance, both the number and precise placement of prompt primitives play a role \cite{cheng2023eval}. Since this study aims to evaluate performance under an ``optimal'' setting with ideal prompts (i.e., without human error or bias), we refrain from any ``artificial'' manipulations.
However, in order to develop guidelines for robust and efficient prompting in a real-world usage scenario, additional aspects of the performance need to be considered, such as annotation time and robustness to human errors.
Both are very critical factors for the performance analysis in an interactive setup with a human user. For example, we hypothesize that the annotation time for a prompt combination is longer than for a single prompt primitive and selecting multiple components (\texttt{5C}) takes longer than just one component (\texttt{1C}). Additionally, some models might be more robust to human error, e.g., when the bounding box is not a very tight enclosing rectangle. These considerations might make alternative prompt strategies more favorable considering the trade-off between segmentation performance, inference time, annotation time and robustness to human error. Future research needs to investigate human generated and interactive prompting.\newline
A next step of our evaluation study will focus on 3D prompting for models with 3D capability, such as \textsc{Sam2} \cite{ravi2024sam2}, \textsc{Sam}-Med3d \cite{wang2024sammed3d}, or Med-\textsc{Sam2} \cite{zhu2024medsam2}. We anticipate more contributions from the research community in developing and fine-tuning FMs specifically for 3D medical image data after \textit{\textsc{Sam2}}'s release. The possibilities with 3D prompting are more comprehensive, but also more complex as more components are defining for a prompt (e.g., prompt primitive, component selection, slice selection) \cite{dong2024eval}. Insights gained from this current study on 2D prompting will provide valuable guidance for informed decision-making as we design a study to test 3D models. 

\section{Conclusion}

In this evaluation study, we evaluated four 2D \textsc{Sam}-family models with different model sizes for bone segmentation in CT scans. We tested $32$ non-iterative prompting strategies containing one-type (bounding box, center, centroid, positive random points), bounding box + point and point-based combination prompts. Aside from an extensive comparison of models and prompt strategies, we provide preliminary guidelines for non-iterative 2D prompting.
This evaluation study is the first step towards an extensive testing of \textsc{Sam}-family models to be used in a clinical setting for bone segmentation in CT scans.

\section*{Acknowledgments}
We thank in alphabetical order Leendert Blankevoort, George S. Buijs, Johannes G.G. Dobbe, Arthur J. Kievit, Matthias U. Schafroth, Geert J. Streekstra, Stela Topalova, Lukas P.E. Verweij, and Annemiek ter Wee for their work, support and guidance in data acquisition and curation.

\bibliographystyle{abbrv} 
\bibliography{bibliography}

\newpage
\begin{appendices}

\section*{Appendix}

\section{nnUNet training details} \label{sec:nnunet_details}
A 2D and a 3D full resolution nnUNet \cite{isensee2021nnunet} were trained on each of the datasets individually. The default training settings have been retained, except for the data augmentation for D1 and D3 and the division into training and validation folds. For D1, the mirroring on the vertical axes is removed since bilateral scans contain right and left labels. For D3, the mirroring on the horizontal axes is removed since a horizontally flipped femoral bone and implant show some similarity with the tibial counterparts. The models for D1 and D2 are trained and evaluated on a 5-fold, for D3 on a 4-fold patient-based cross-validation split.

\section{Extra qualitative results}

\subsection{Knee dataset - cortical vs. full bone segmentation} \label{sec:appendix_example_knee}

\begin{figure}[H]
\begin{tabular}{c|c|cccccc}
    \hline
    Reference & \cblackstar[0.5]{yellow} & \cblacksquarecross[0.4]{green} & \cblacksquare[0.4]{green} & \cblackdiamondcross[0.4]{green} & \cblacksquaredot[0.4]{green} & \cblackstartriangledowndot[0.4]{green} & \cblackcircledot[0.4]{green} \\ \hline
    \includegraphics[width=0.1\textwidth, trim=330 0 460 0, clip]{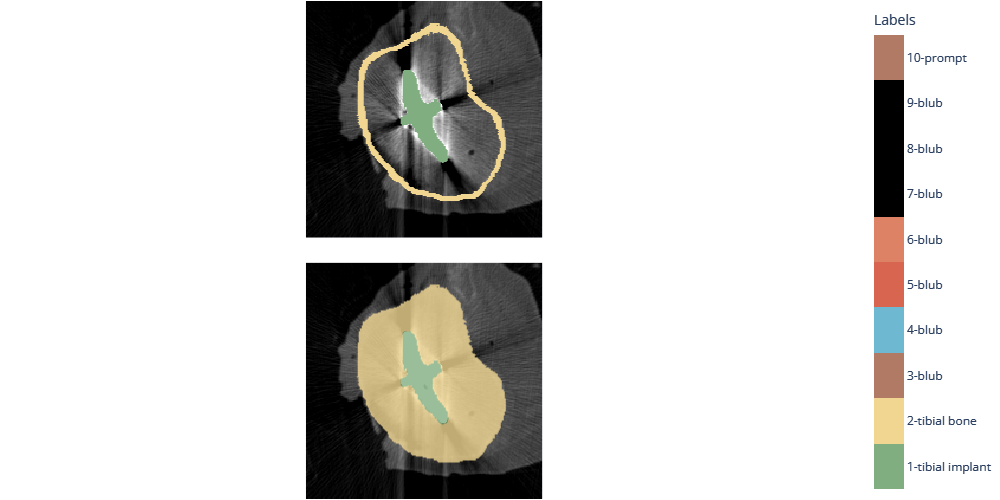} & \includegraphics[width=0.1\textwidth, trim=330 0 460 0, clip]{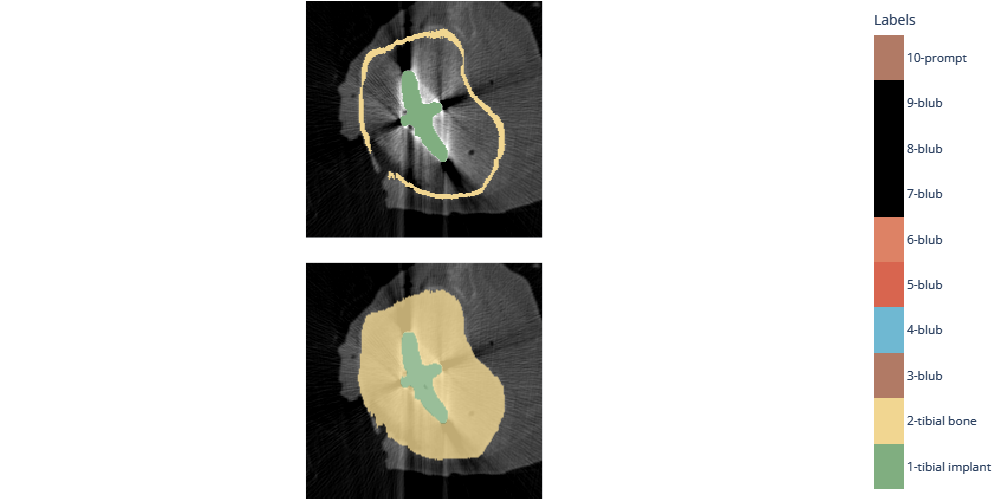} & \includegraphics[width=0.1\textwidth, trim=330 0 460 0, clip]{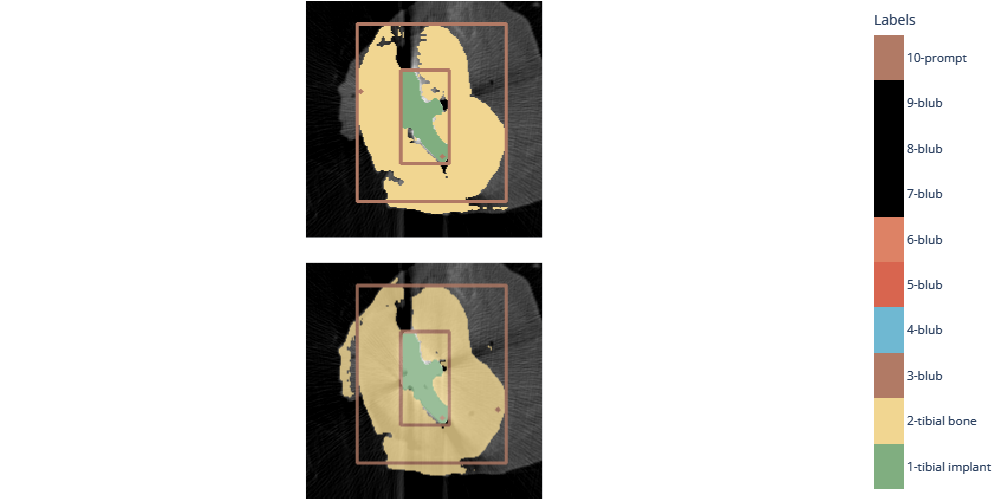} & \includegraphics[width=0.1\textwidth, trim=330 0 460 0, clip]{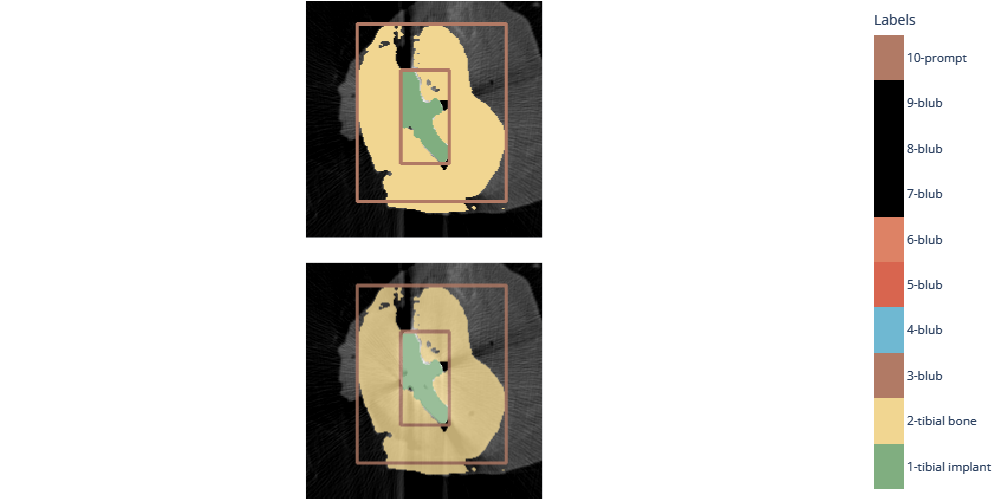} & \includegraphics[width=0.1\textwidth, trim=330 0 460 0, clip]{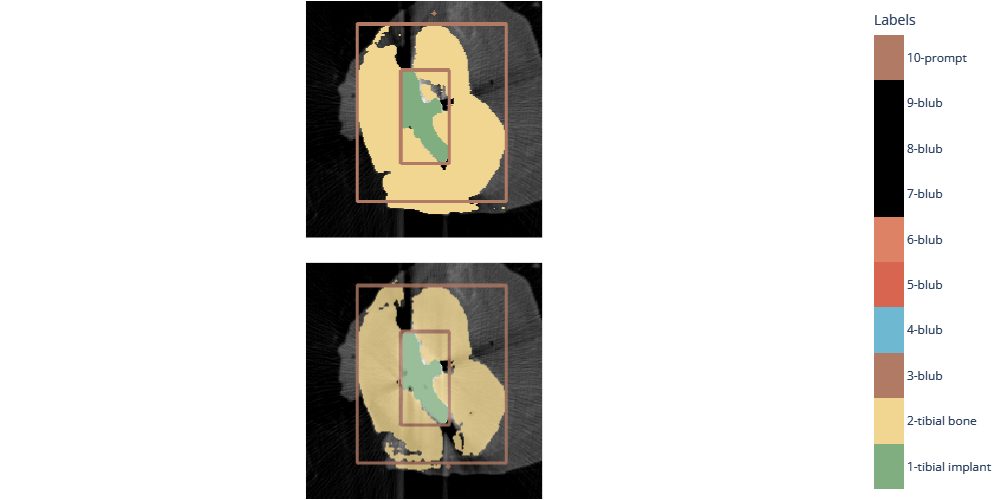} & \includegraphics[width=0.1\textwidth, trim=330 0 460 0, clip]{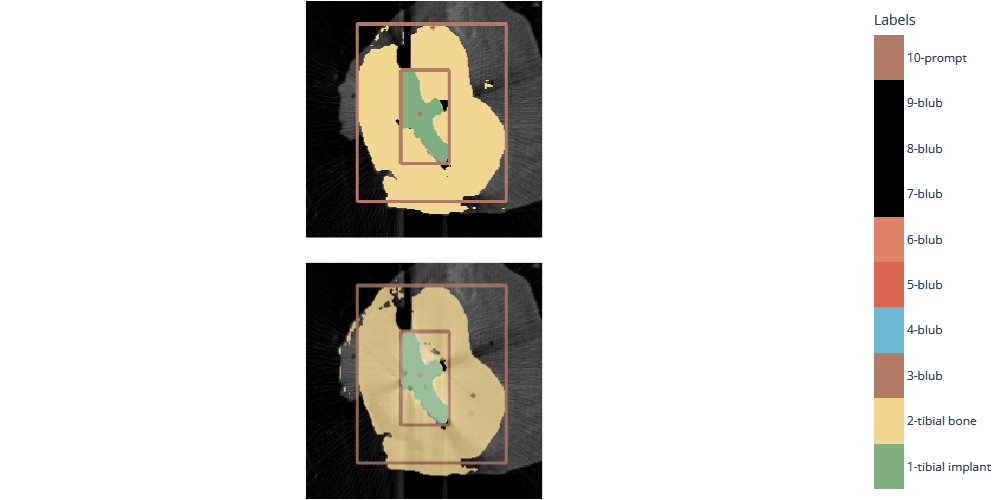} & \includegraphics[width=0.1\textwidth, trim=330 0 460 0, clip]{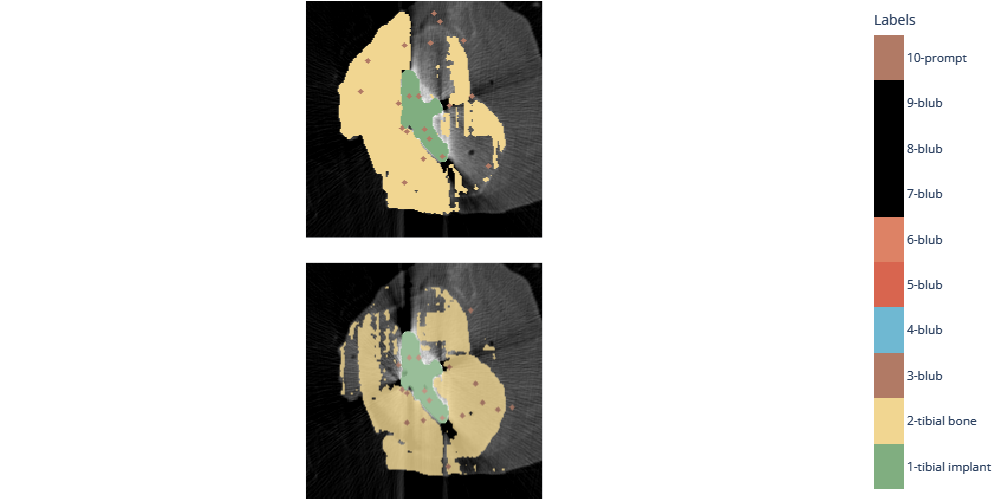} & \includegraphics[width=0.1\textwidth, trim=330 0 460 0, clip]{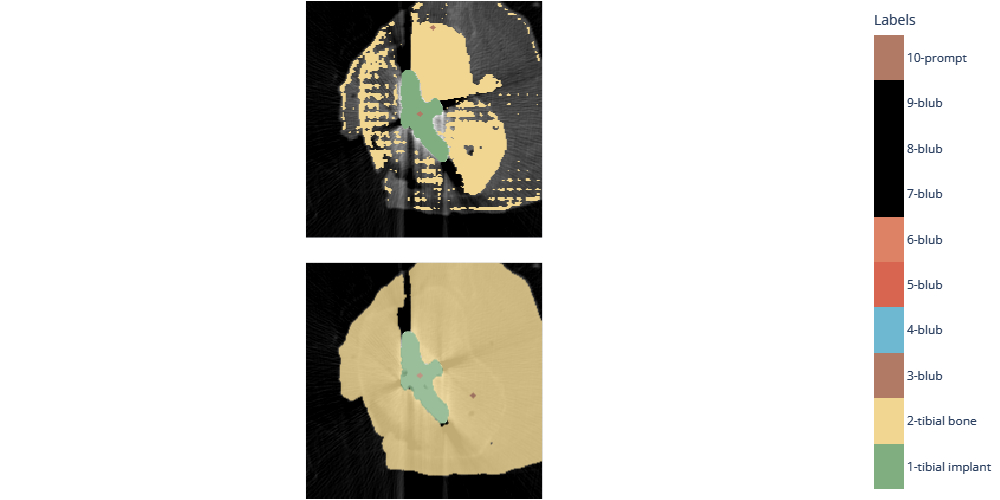} \\ \hdashline
    
    \includegraphics[width=0.1\textwidth, trim=330 0 460 0, clip]{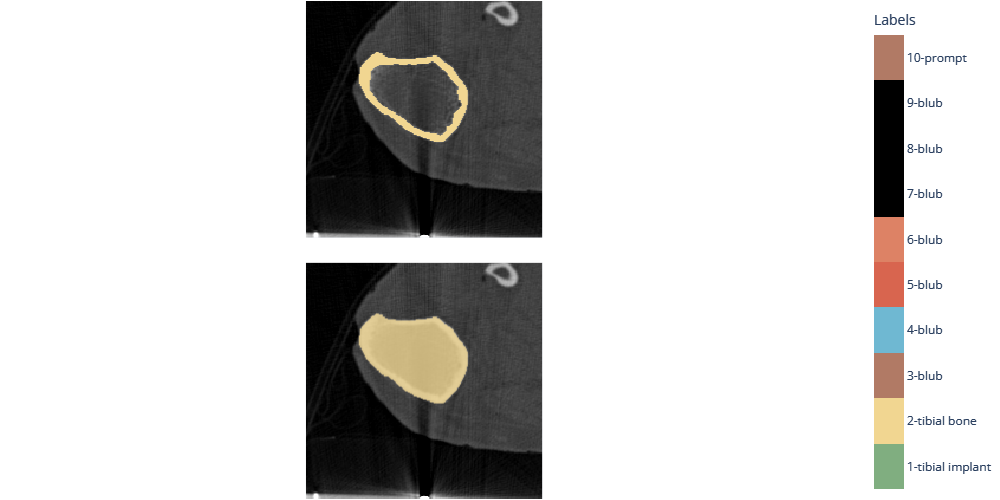} & \includegraphics[width=0.1\textwidth, trim=330 0 460 0, clip]{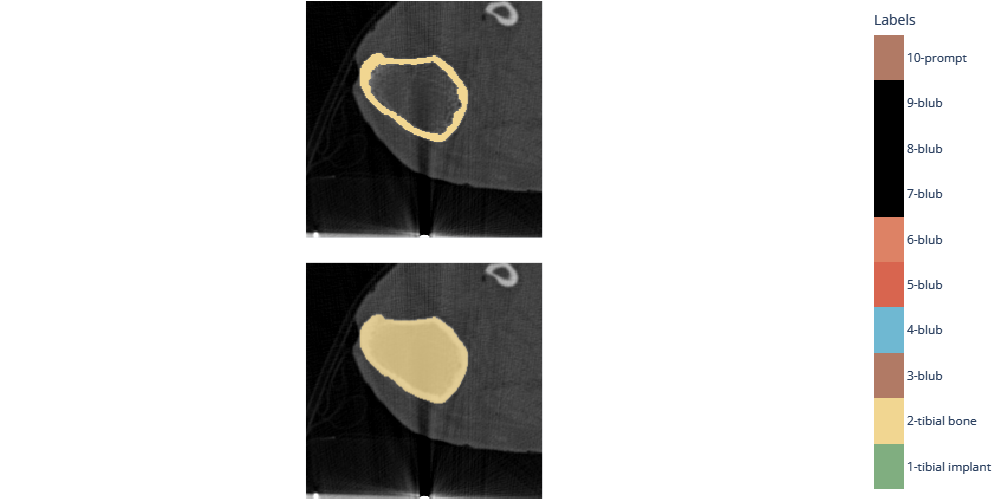} & \includegraphics[width=0.1\textwidth, trim=330 0 460 0, clip]{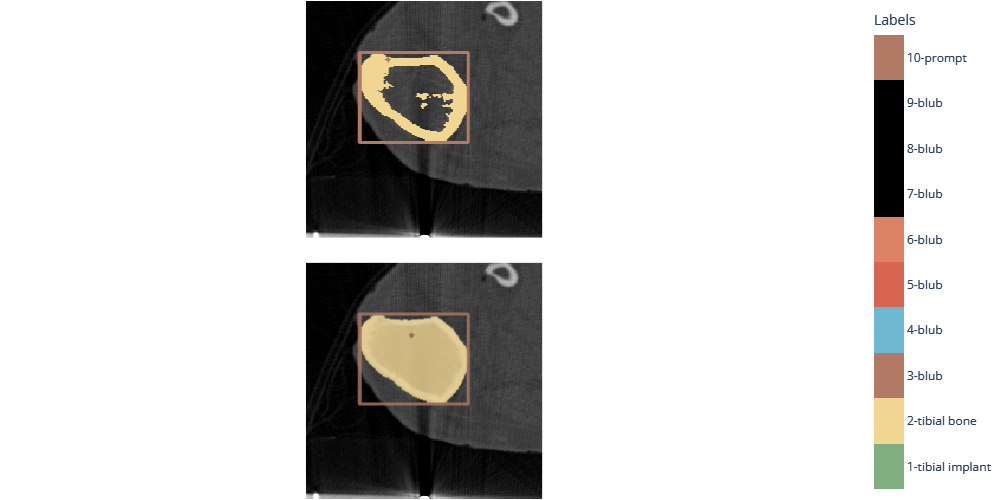} & \includegraphics[width=0.1\textwidth, trim=330 0 460 0, clip]{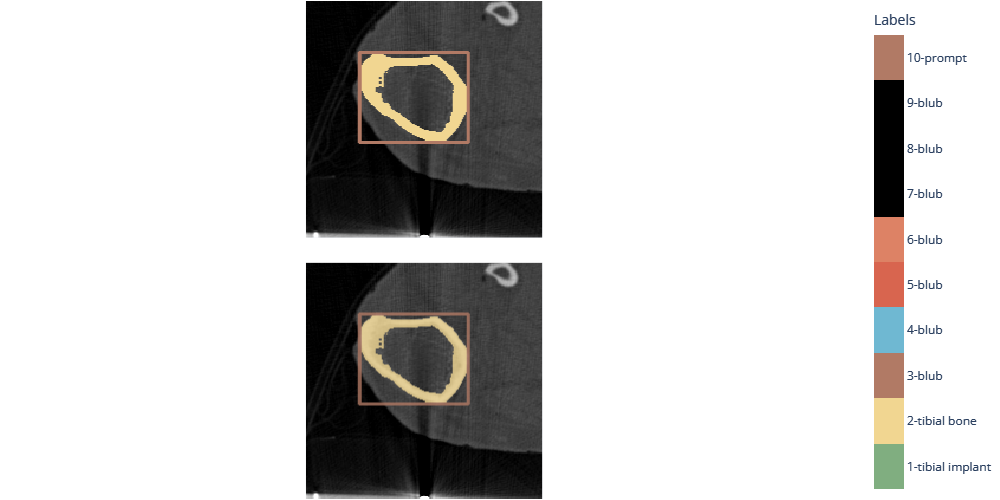} & \includegraphics[width=0.1\textwidth, trim=330 0 460 0, clip]{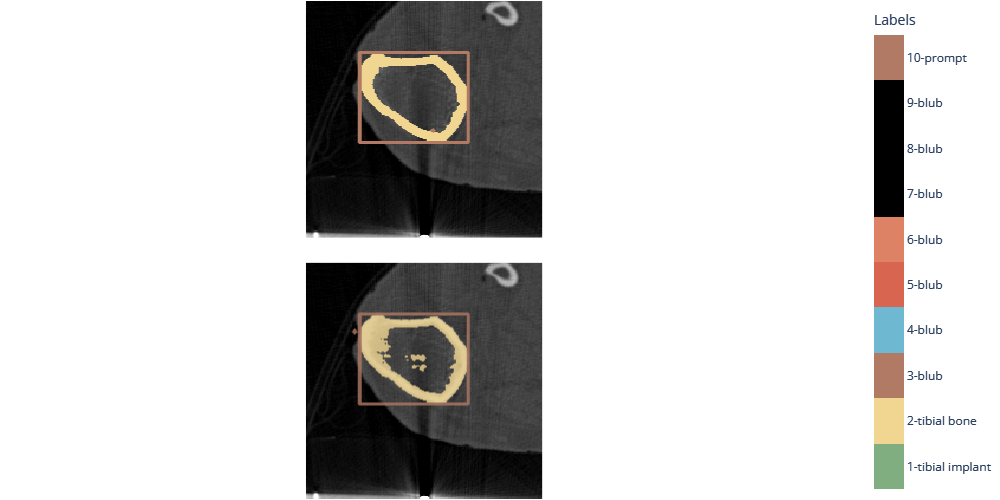} & \includegraphics[width=0.1\textwidth, trim=330 0 460 0, clip]{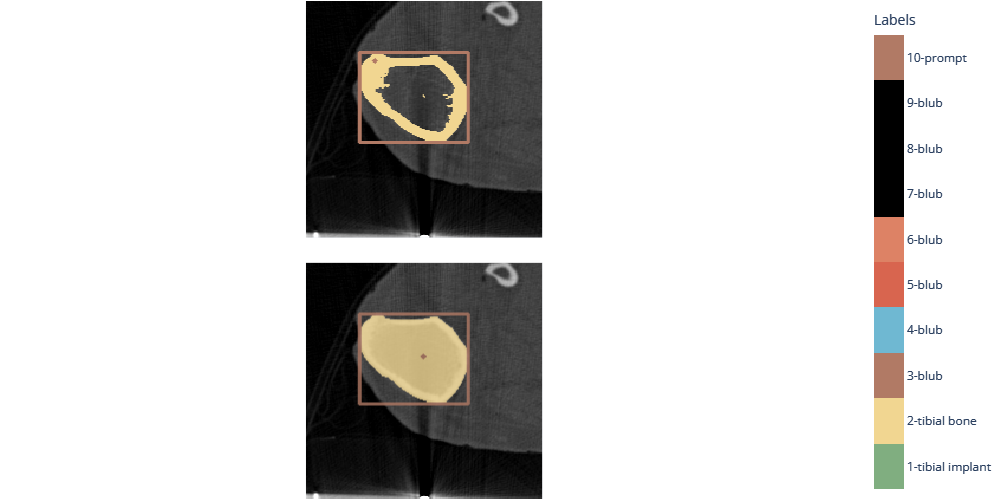} & \includegraphics[width=0.1\textwidth, trim=330 0 460 0, clip]{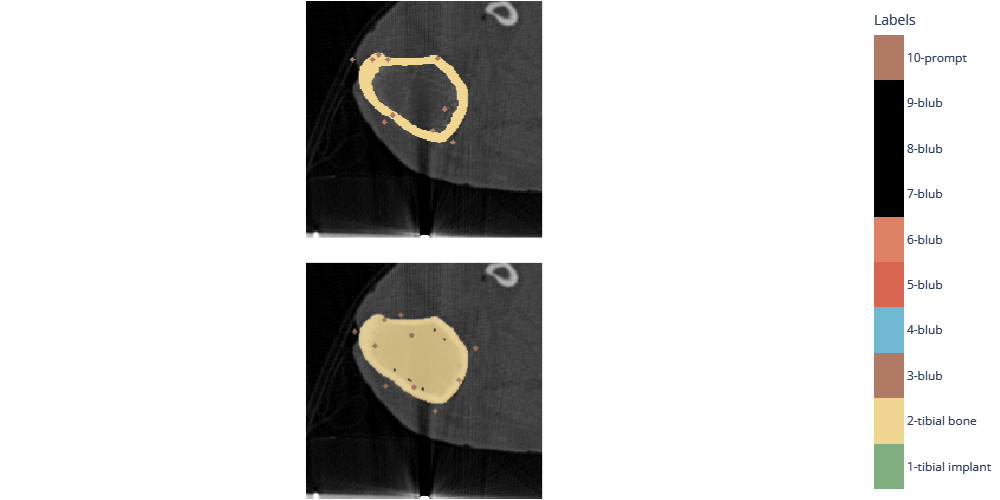} & \includegraphics[width=0.1\textwidth, trim=330 0 460 0, clip]{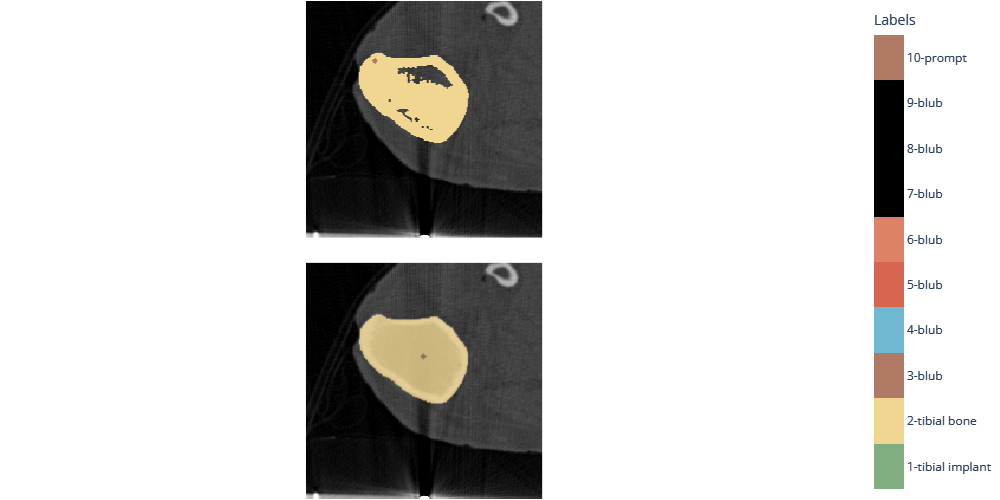} \\ \hdashline
    
    \includegraphics[width=0.1\textwidth, trim=330 0 460 0, clip]{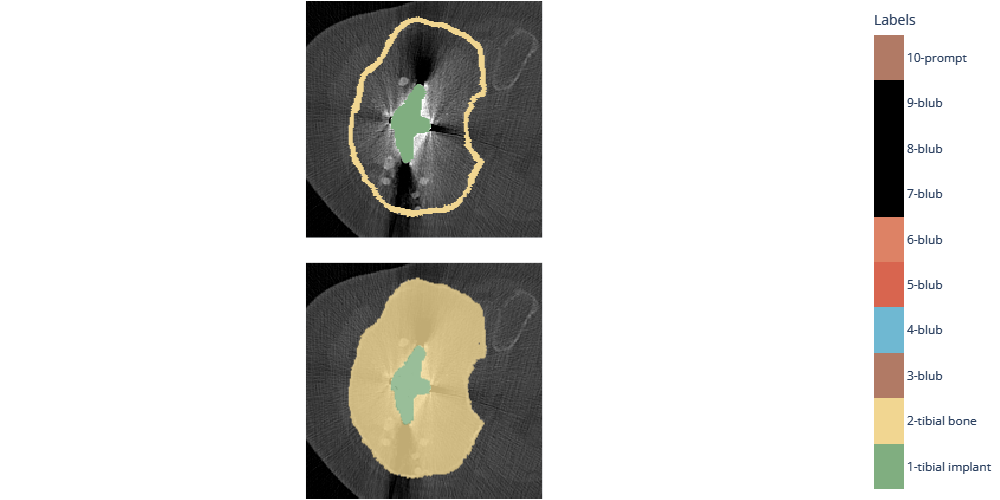} & \includegraphics[width=0.1\textwidth, trim=330 0 460 0, clip]{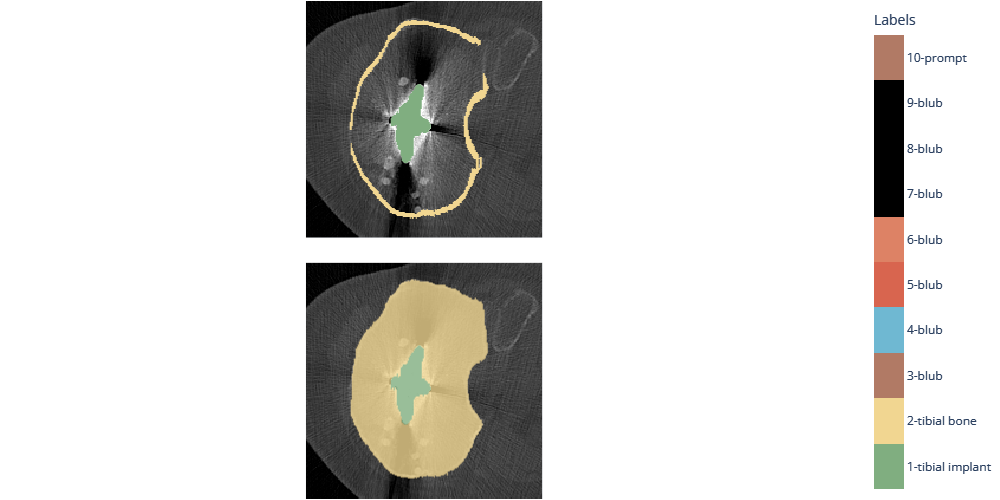} & \includegraphics[width=0.1\textwidth, trim=330 0 460 0, clip]{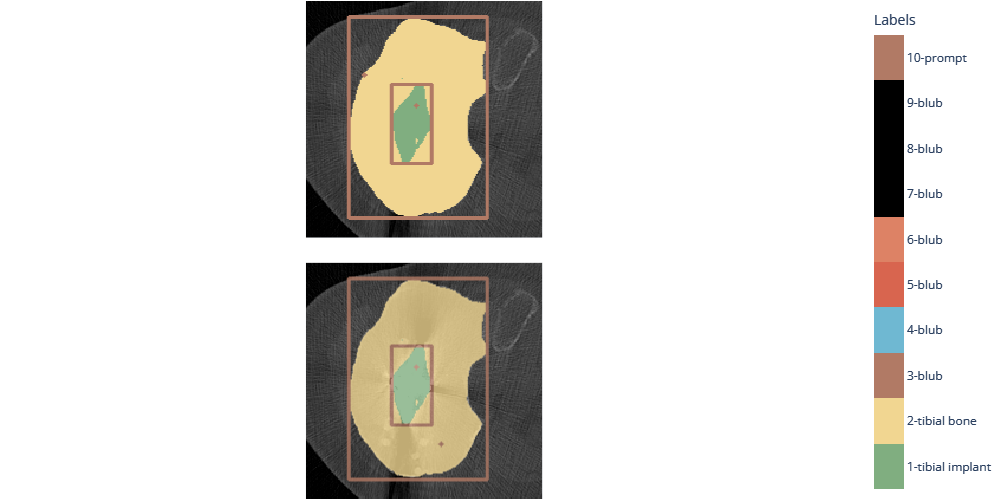} & \includegraphics[width=0.1\textwidth, trim=330 0 460 0, clip]{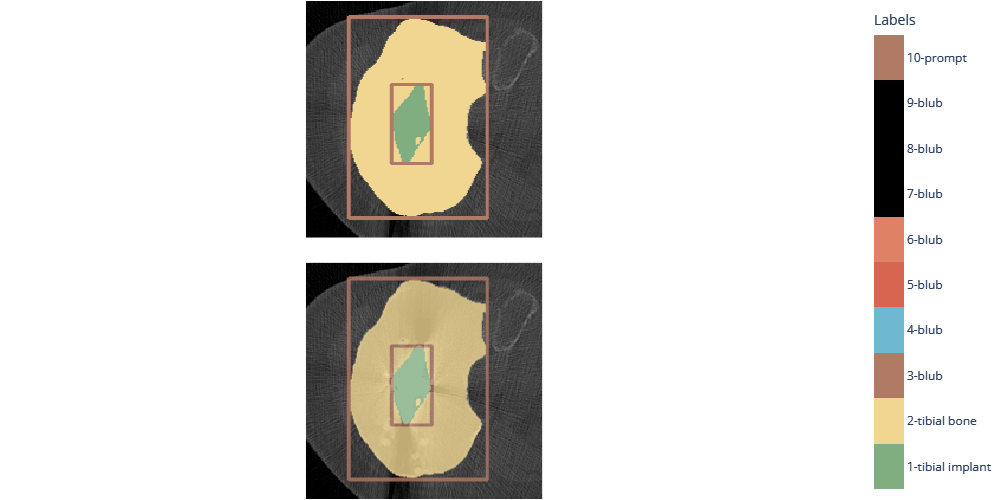} & \includegraphics[width=0.1\textwidth, trim=330 0 460 0, clip]{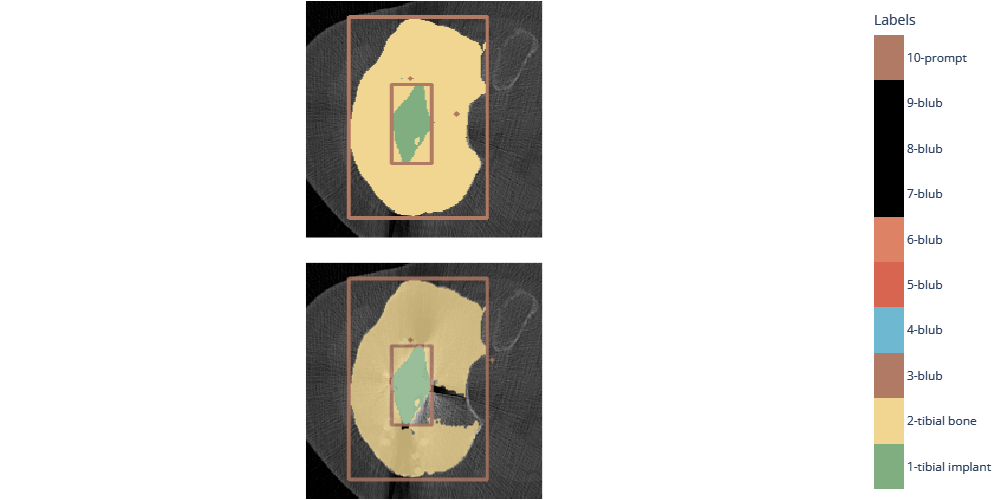} & \includegraphics[width=0.1\textwidth, trim=330 0 460 0, clip]{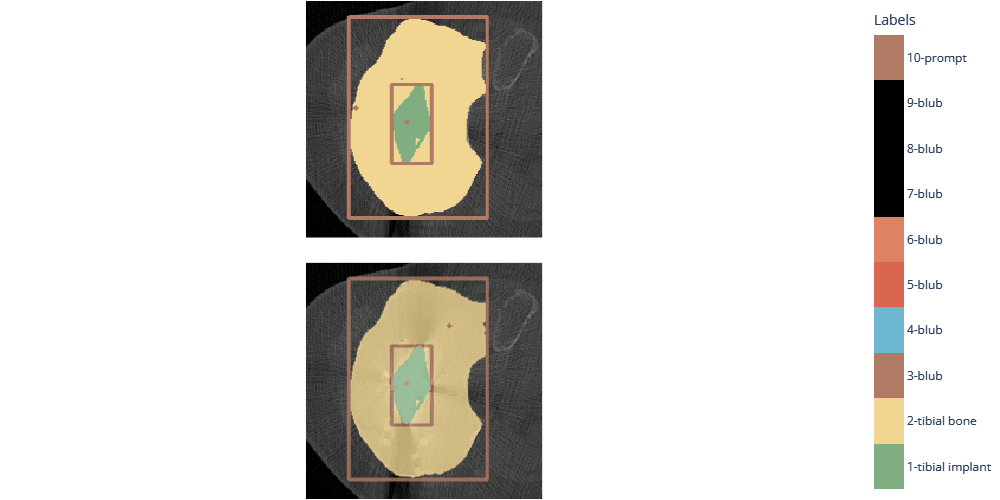} & \includegraphics[width=0.1\textwidth, trim=330 0 460 0, clip]{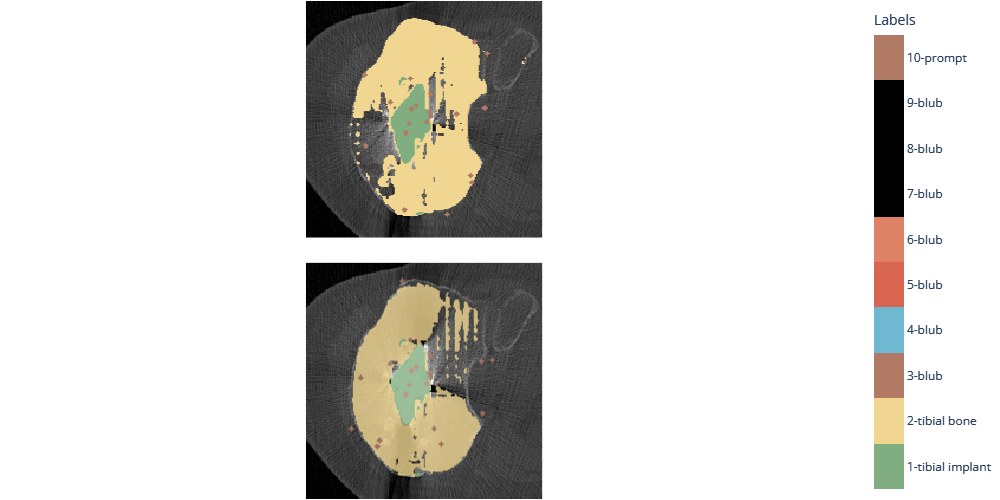} & \includegraphics[width=0.1\textwidth, trim=330 0 460 0, clip]{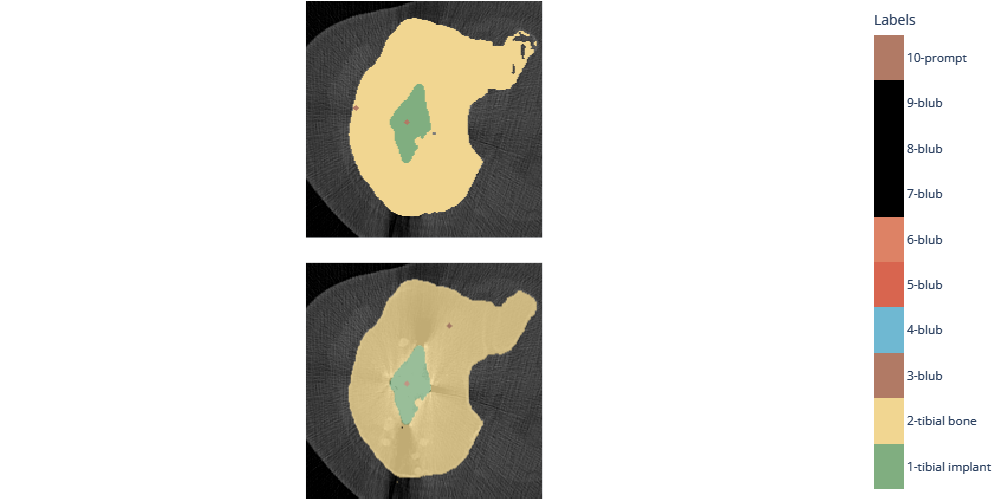} \\ \hline
\end{tabular}
\caption{Examples of knee dataset in axial view for reference labels, 2D nnUNet and \textit{\textsc{Sam} B} with different prompt strategies. In each row, the same axial slice is displayed with cortical (top) and full (bottom) tibia bone segmentation. The labels/prompts are color-encoded: yellow - tibia; green - tibia implant; brown - prompts for  \textit{\textsc{Sam} B} inference.}
\label{fig:examples_knee}
\end{figure}

\begin{figure}[H]
\centering
\begin{tabular}{c|cc|cccc}
    \hline
    Reference &  \cblackstar[0.5]{yellow} & \cwhitestar[0.5]{yellow} &  \cblacksquare[0.4]{green} & \cblacksquaredot[0.4]{green} & \cblackstartriangledowndot[0.4]{green} & \cblackcircledot[0.4]{green} \\
    \hline
    \includegraphics[width=0.1\textwidth, trim=370 650 420 100, clip]{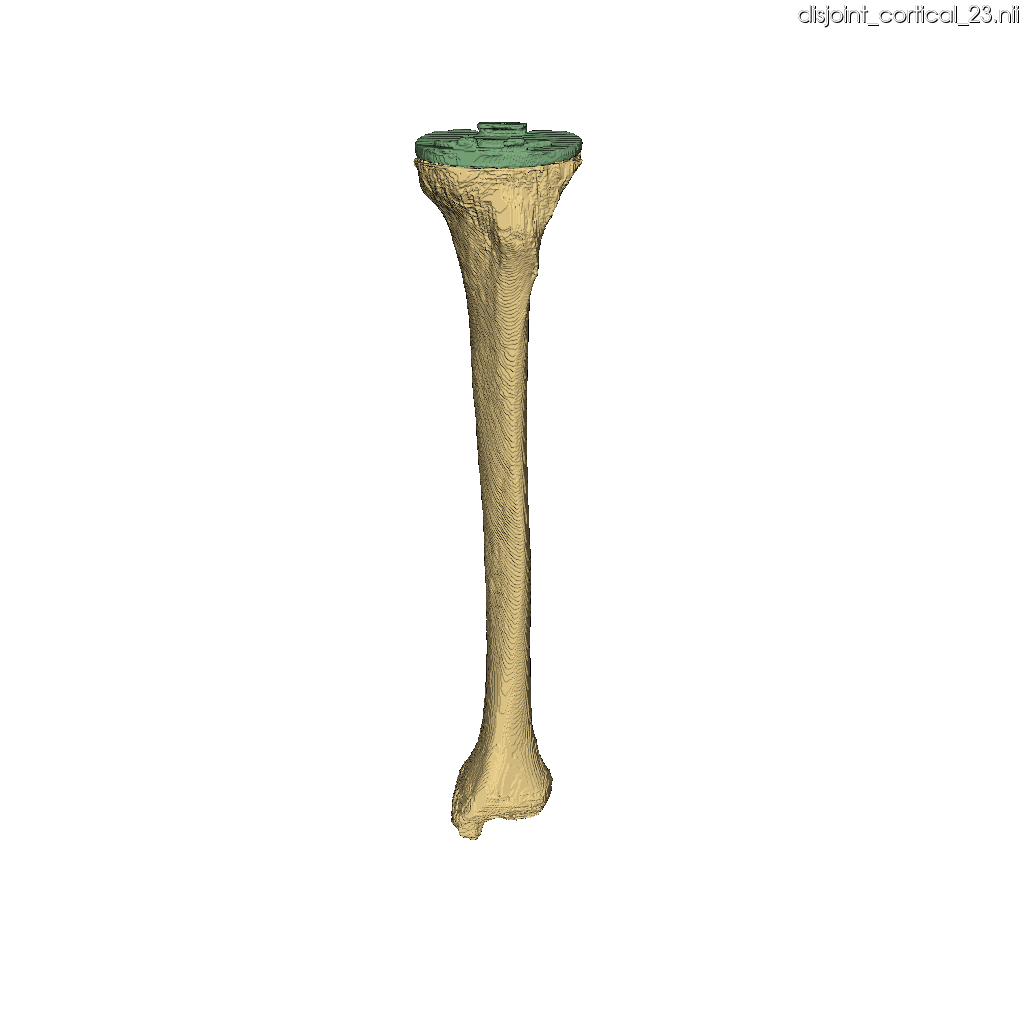} & \includegraphics[height=2cm, trim=370 500 420 100, clip]{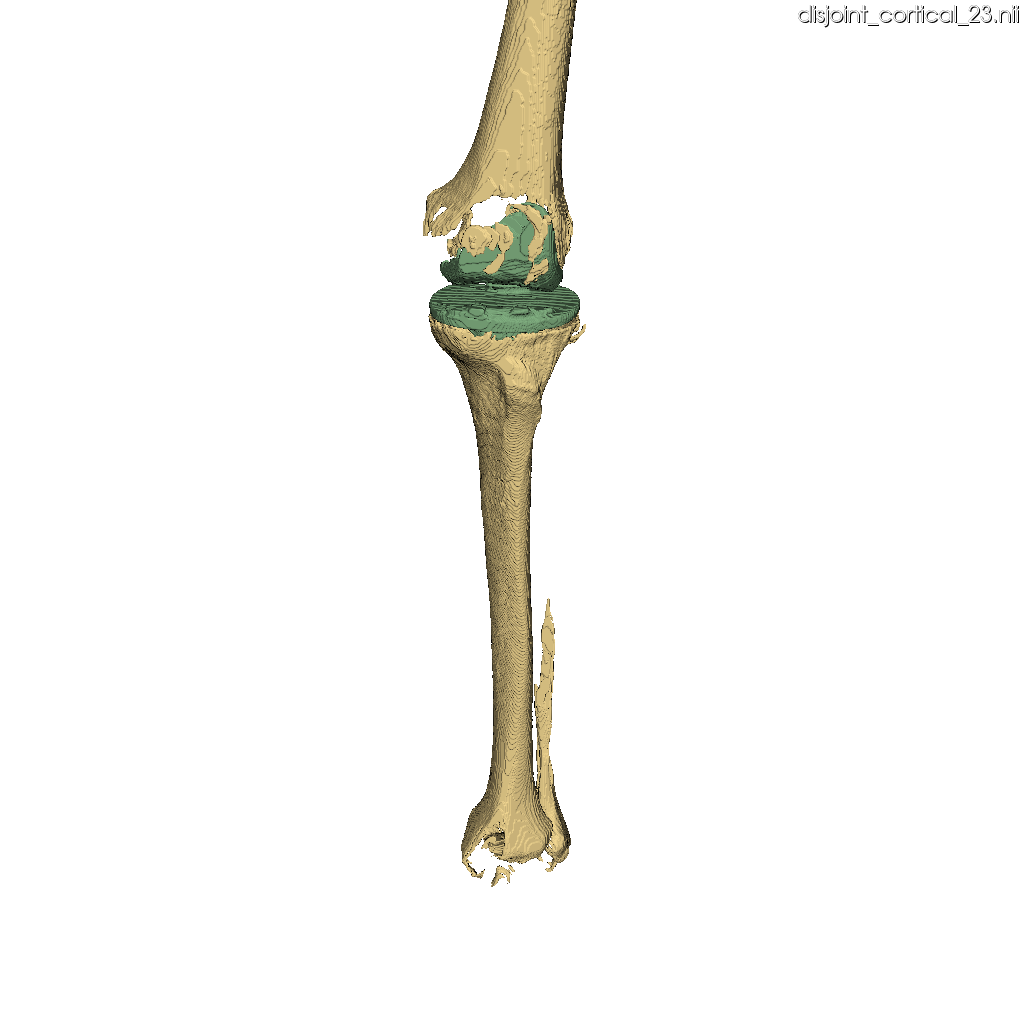} & \includegraphics[height=2cm, trim=370 500 420 100, clip]{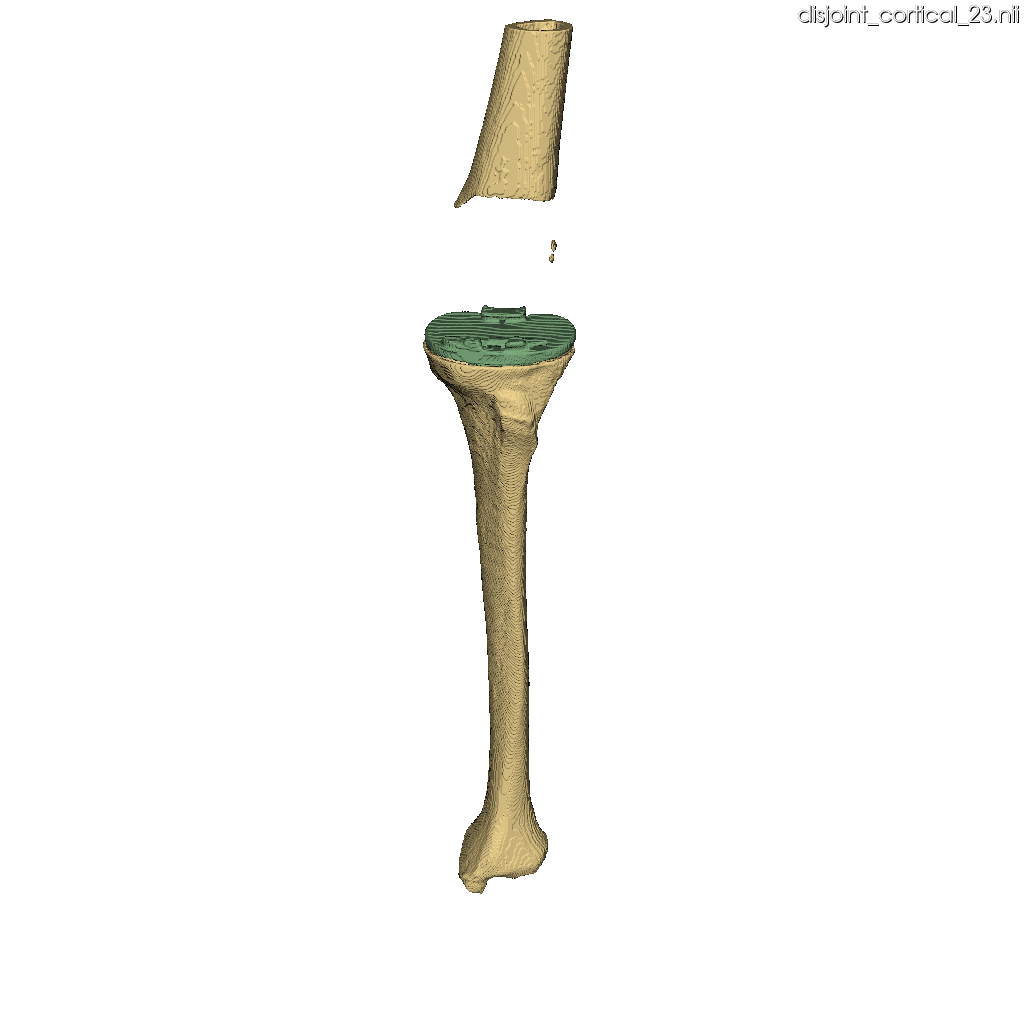} & \includegraphics[height=2cm, trim=370 650 420 100, clip]{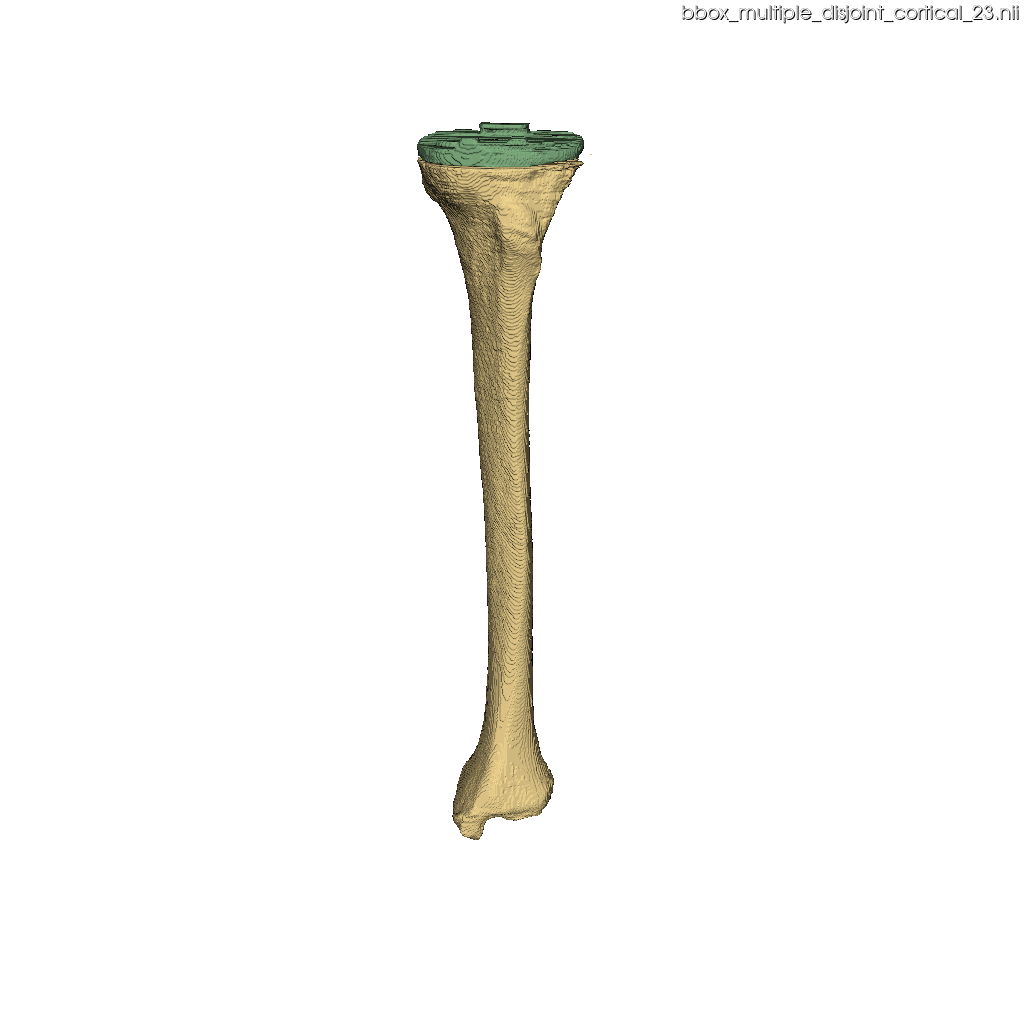} & \includegraphics[height=2cm,trim=370 650 420 100, clip]{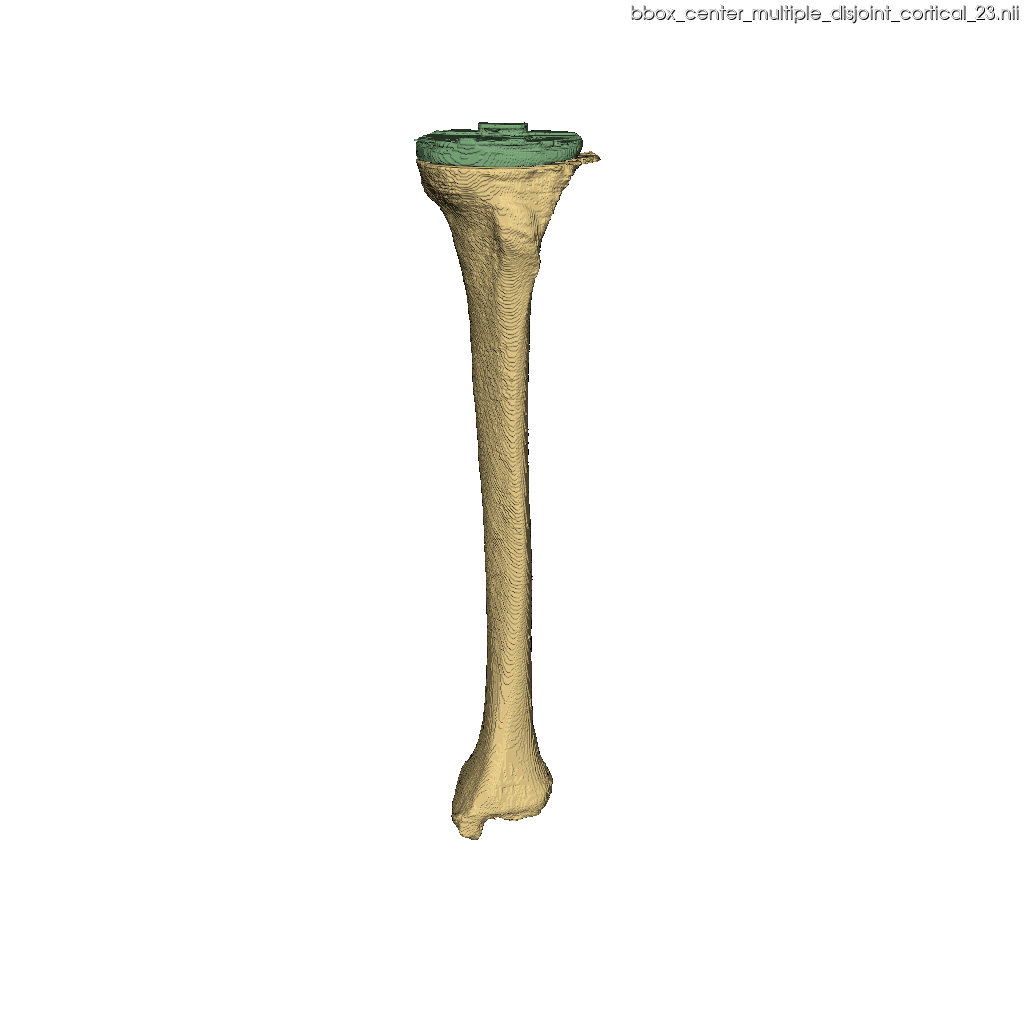} & \includegraphics[height=2cm, trim=370 650 350 100, clip]{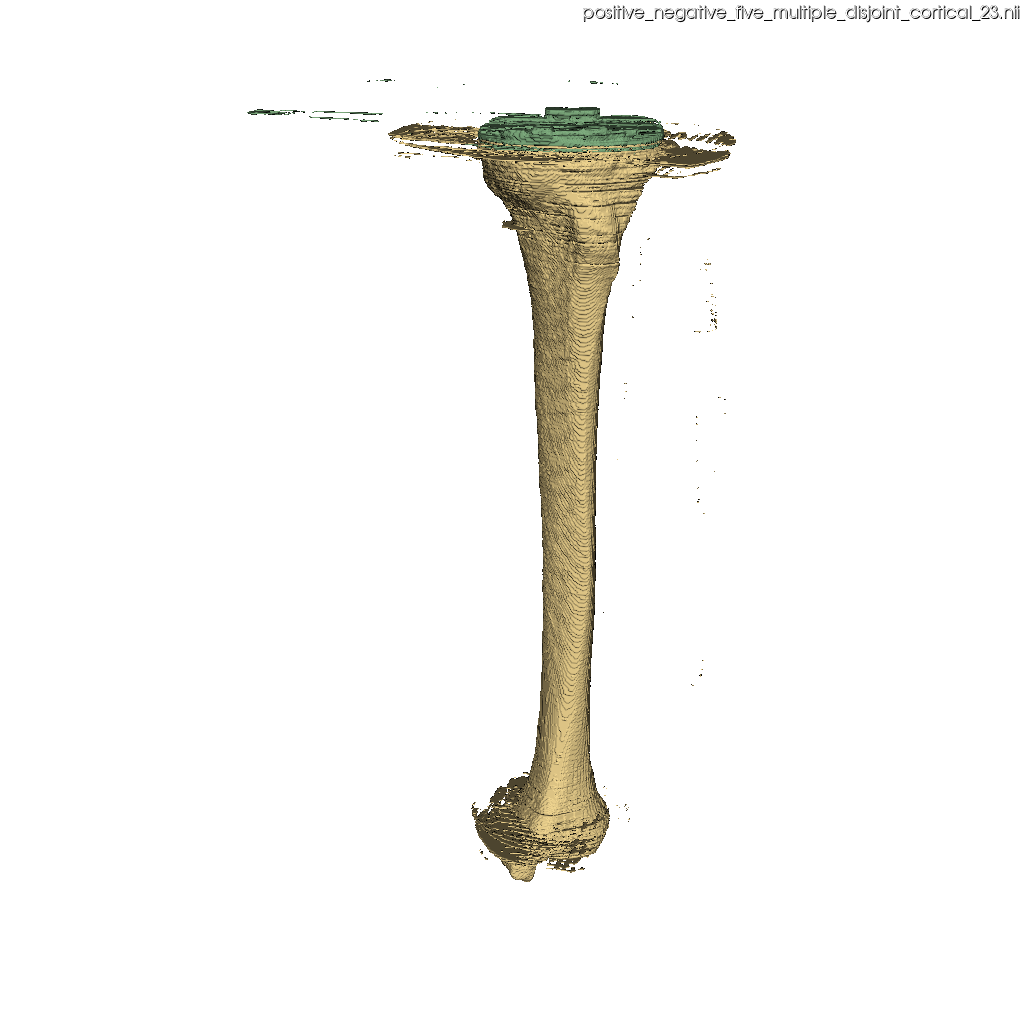} & \includegraphics[height=2cm,trim=370 650 420 100, clip]{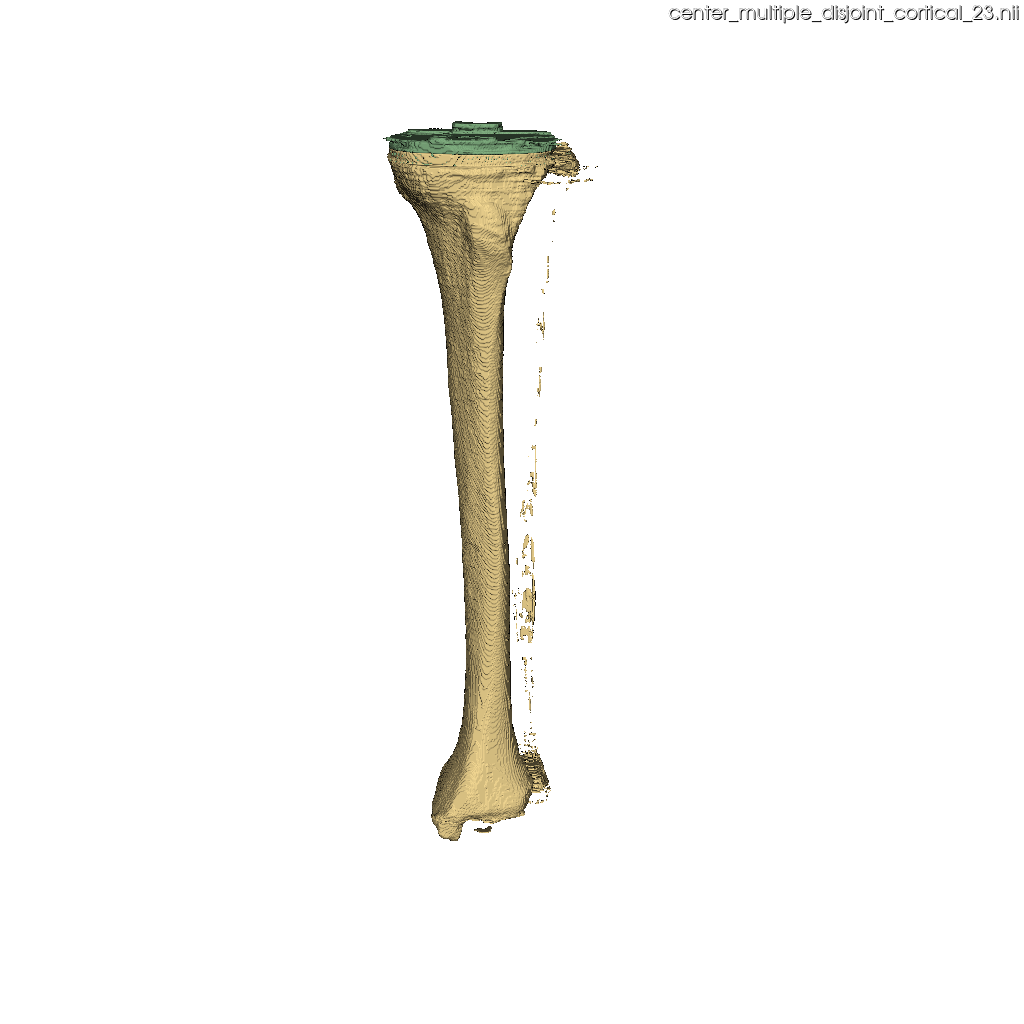} \\
    \hline
\end{tabular}
\caption{3D model example for D3a for reference labels, 2D and 3D nnUNet, and \textit{\textsc{Sam} B} with different prompt strategies. The labels are color-encoded: yellow - tibia; green - tibia implant. }
\label{fig:examples_knee02}
\end{figure}

\subsection{Point-based prompting} \label{sec:examples_points}

\begin{figure}[H]
\centering
\renewcommand{\arraystretch}{1.4}
\begin{tabular}{c|cccc}
    \hline
    Reference & \cblackcircledot[0.4]{grey} & \cwhitecircledot[0.4]{grey} & \cblacktriangleup[0.4]{grey} & \cwhitetriangleup[0.4]{grey} \\
    \includegraphics[width=0.15\textwidth, trim=280 0 440 0, clip]{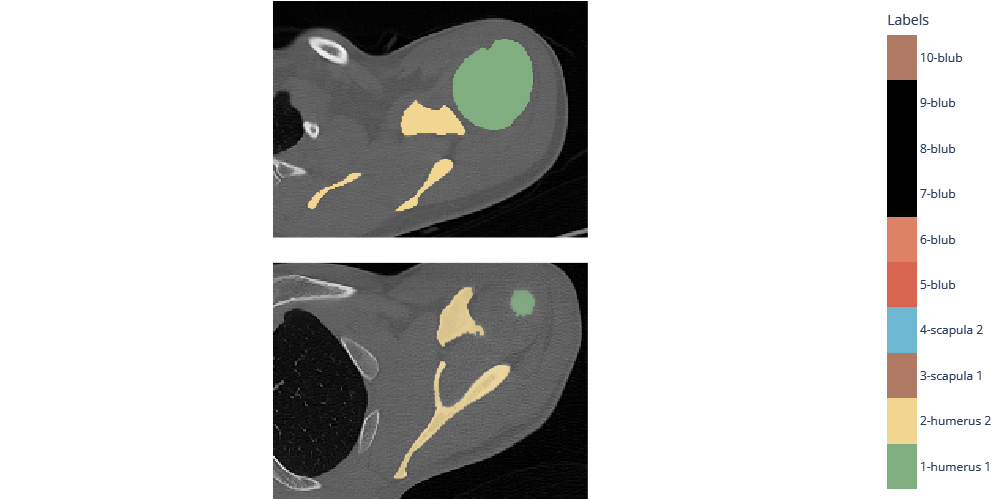} & \includegraphics[width=0.15\textwidth, trim=280 0 440 0, clip]{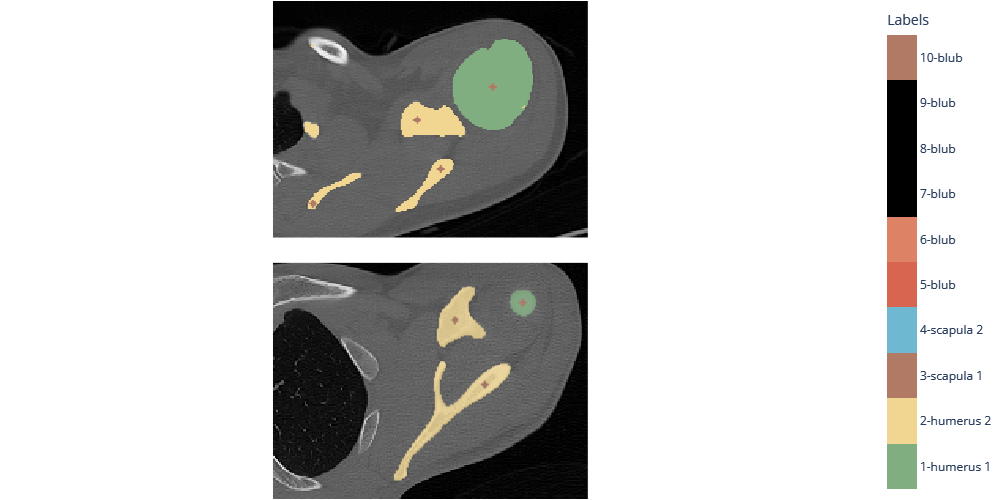} & \includegraphics[width=0.15\textwidth, trim=280 0 440 0, clip]{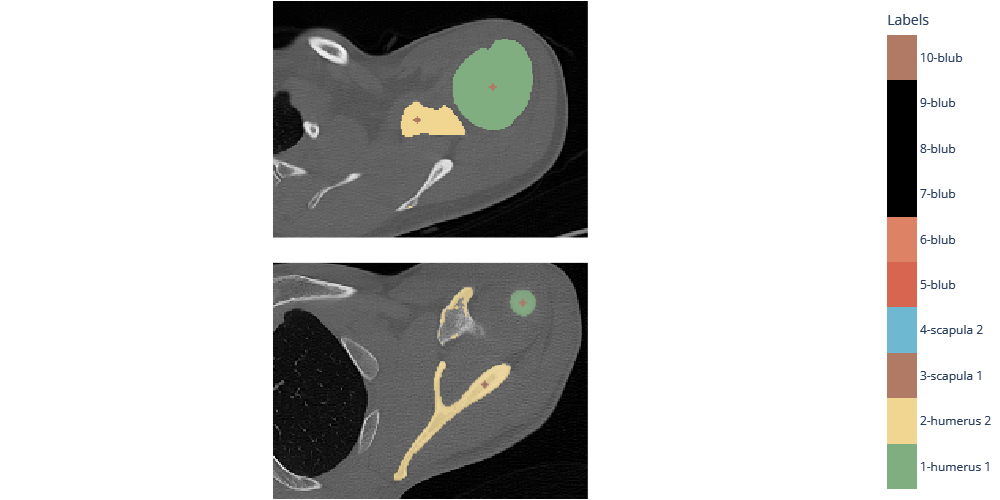} & \includegraphics[width=0.15\textwidth, trim=280 0 440 0, clip]{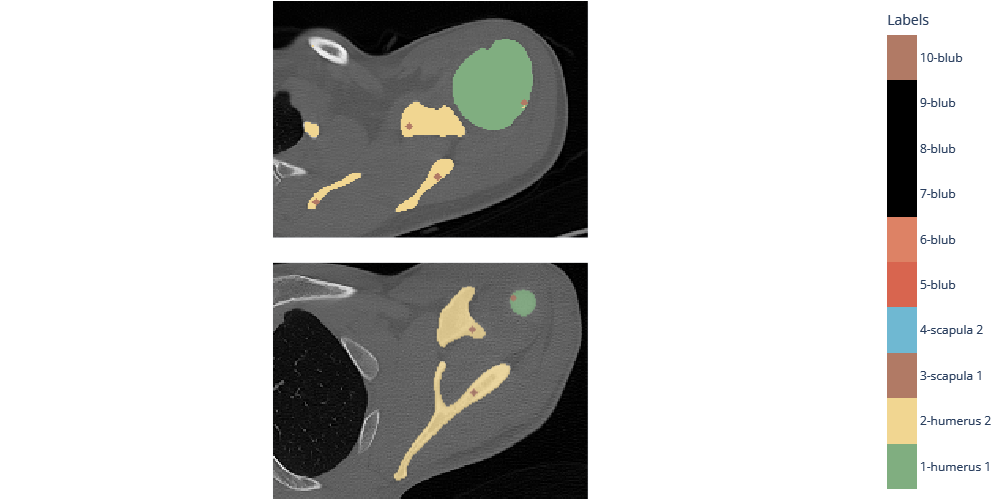} & \includegraphics[width=0.15\textwidth, trim=280 0 440 0, clip]{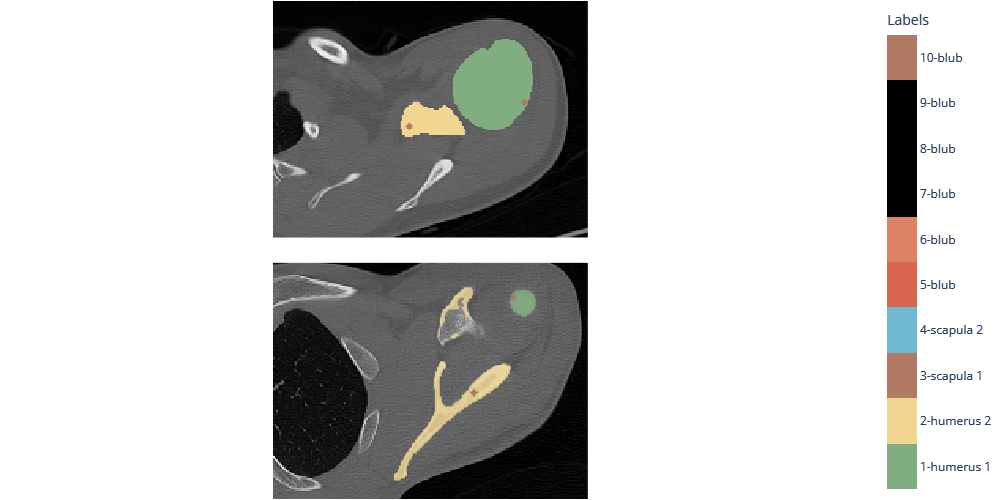}\\ 
    \hline
    & \cblacktriangledown[0.4]{grey} & \cwhitetriangledown[0.4]{grey} & \cblacktriangleleft[0.4]{grey} & \cwhitetriangleleft[0.4]{grey} \\
    & \includegraphics[width=0.15\textwidth, trim=280 0 440 0, clip]{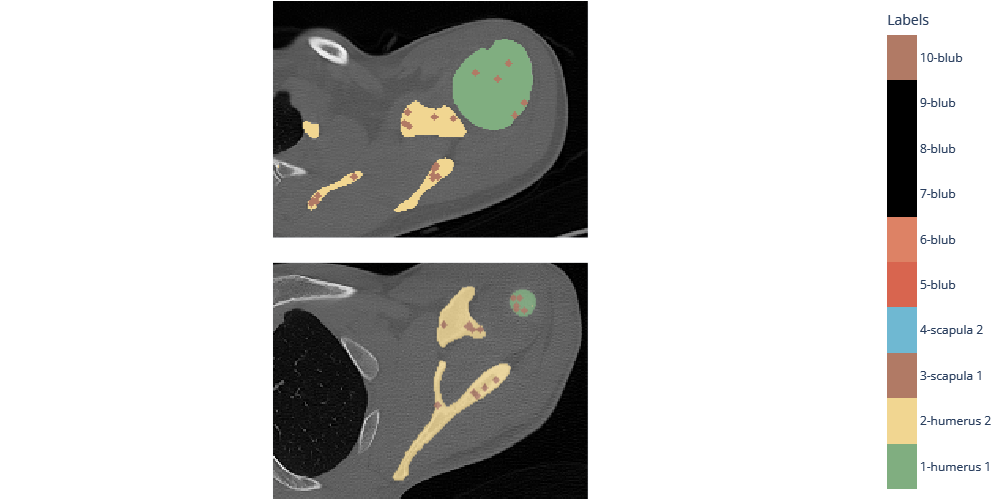} & \includegraphics[width=0.15\textwidth, trim=280 0 440 0, clip]{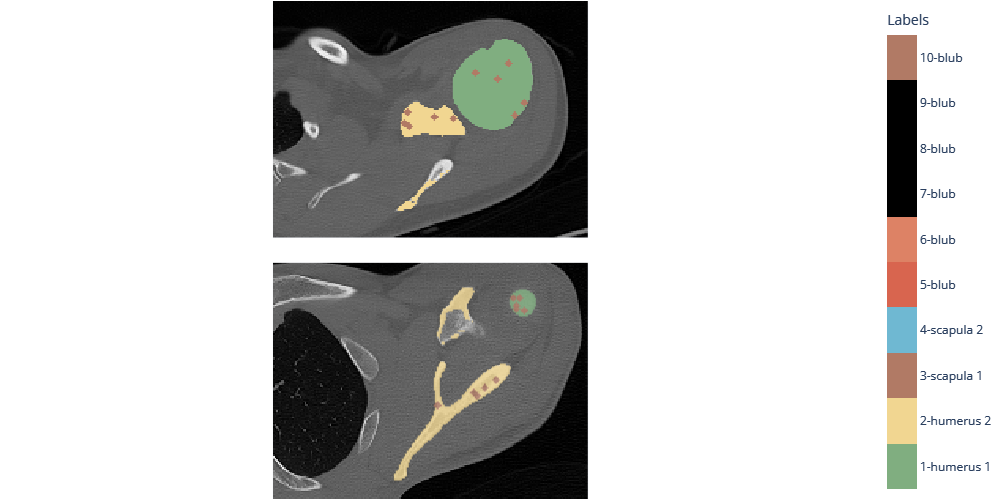} & \includegraphics[width=0.15\textwidth, trim=280 0 440 0, clip]{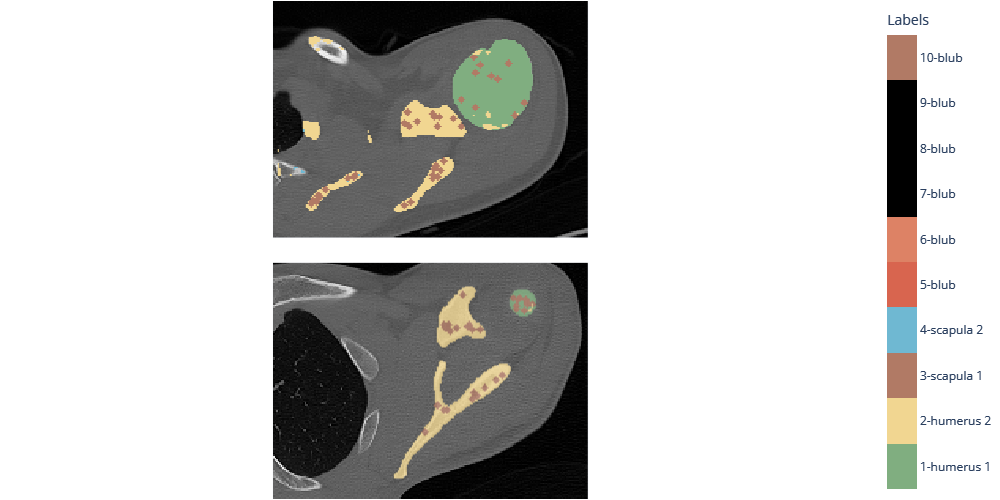} & \includegraphics[width=0.15\textwidth, trim=280 0 440 0, clip]{point_examples/random_ten_multiple_7_5.png}\\
    & \cblackstartriangledown[0.4]{grey} & \cwhitestartriangledown[0.4]{grey} & \cblackstartriangledowndot[0.4]{grey} & \cwhitestartriangledowndot[0.4]{grey} \\
     & \includegraphics[width=0.15\textwidth, trim=280 0 440 0, clip]{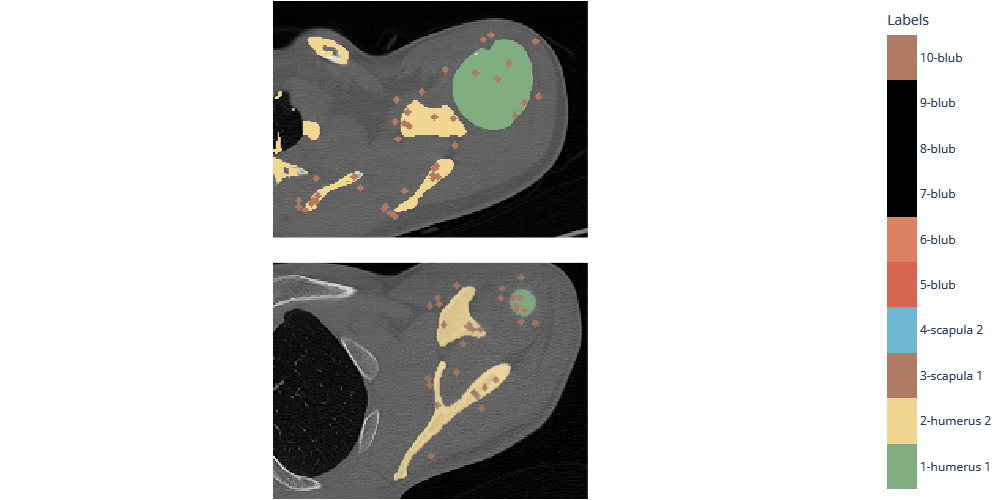} & \includegraphics[width=0.15\textwidth, trim=280 0 440 0, clip]{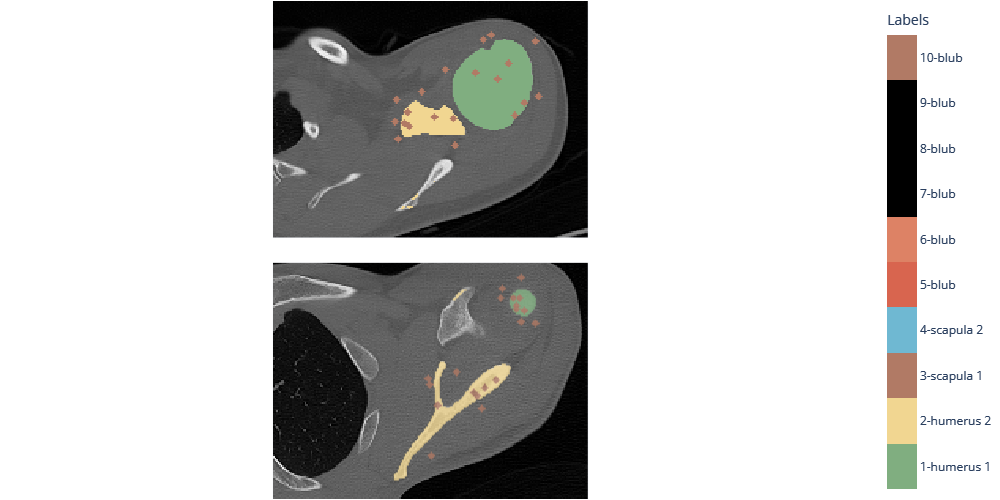} & \includegraphics[width=0.15\textwidth, trim=280 0 440 0, clip]{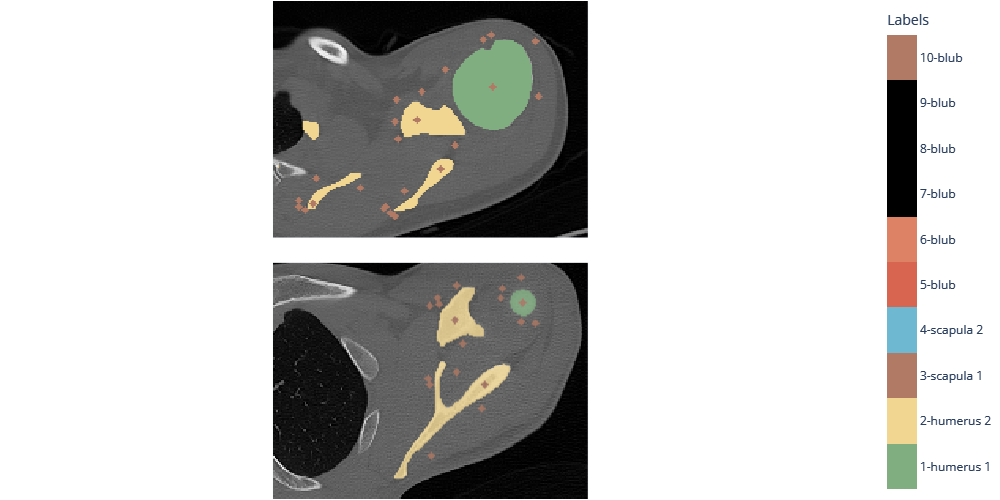} & \includegraphics[width=0.15\textwidth, trim=280 0 440 0, clip]{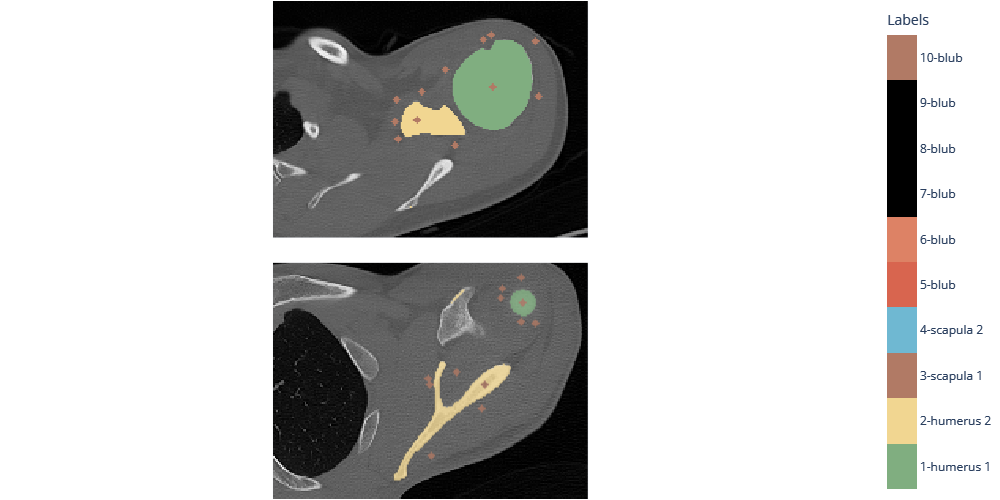}\\
    
\end{tabular}
\caption{Examples of selected point-based strategies in axial view for reference label and  \textit{\textsc{Sam} B} with different point-based prompting strategies. The labels/prompts are color-encoded: yellow - scapula; green - humerus; brown - point-based prompts for  \textit{\textsc{Sam} B} inference.}
\label{fig:examples_points}
\end{figure}

\newpage

\subsection{Per Model} \label{sec:appendx_example_model}

Selected samples of each dataset with low, medium and high associated DSC are displayed for the \textit{Med-\textsc{Sam}} with \texttt{bbox 1C} and \texttt{bbox 5C}, \textit{\textsc{Sam}-Med2d}, \textit{\textsc{Sam} B}, \textit{\textsc{Sam} L} and \textit{\textsc{Sam2} B+} and the settings \texttt{bbox 5C} (\cblacksquare[0.3]{grey}), \texttt{bbox+center 5C} (\cblacksquaredot[0.3]{grey}) and \texttt{5 pos.+neg.~5C} (\cblackstartriangledowndot[0.3]{grey}).

\begin{figure}[H]
\begin{tabular}{|l|cc|cc|cc|}
    \hline
    Setting & \multicolumn{2}{c|}{DSC $\downarrow$} & \multicolumn{2}{c|}{DSC median} & \multicolumn{2}{c|}{DSC $\uparrow$} \\ 
    \hline
    \multirow{2}{*}{\cblacksquare[0.6]{blue}} & \includegraphics[width=0.14\linewidth, trim=600 380 100 330, clip]{examples/med_sam_bad/bbox_multiple_shoulder_1.png} & \raisebox{-0.6\height}[0pt][0pt]{\includegraphics[width=0.11\linewidth, trim=250 150 300 50, clip]{examples/med_sam_bad/bbox_multiple_disjoint_cortical_20.png}} &
    \includegraphics[width=0.14\linewidth, trim=560 380 130 360, clip]{examples/med_sam_medium/bbox_multiple_shoulder_0.png} & \raisebox{-0.6\height}[0pt][0pt]{\includegraphics[width=0.11\linewidth, trim=380 200 400 100, clip]{examples/med_sam_medium/bbox_multiple_disjoint_cortical_1.png}} &
    \includegraphics[width=0.14\linewidth, trim=560 380 190 330, clip]{examples/med_sam_good/bbox_multiple_shoulder_7.png} & \raisebox{-0.6\height}[0pt][0pt]{\includegraphics[width=0.11\linewidth, trim=400 150 400 100, clip]{examples/med_sam_good/bbox_multiple_disjoint_cortical_19.png}} 
    \\
    & \includegraphics[width=0.14\linewidth, trim=370 290 370 280, clip]{examples/med_sam_bad/bbox_multiple_wrist_33.png} & &
    \includegraphics[width=0.14\linewidth, trim=370 290 370 280, clip]{examples/med_sam_medium/bbox_multiple_wrist_4.png} & &
    \includegraphics[width=0.14\linewidth, trim=370 290 370 280, clip]{examples/med_sam_good/bbox_multiple_wrist_0.png} & \\ 
    \hline
    \multirow{2}{*}{\cwhitesquare[0.6]{blue}} & \includegraphics[width=0.14\linewidth, trim=600 380 100 330, clip]{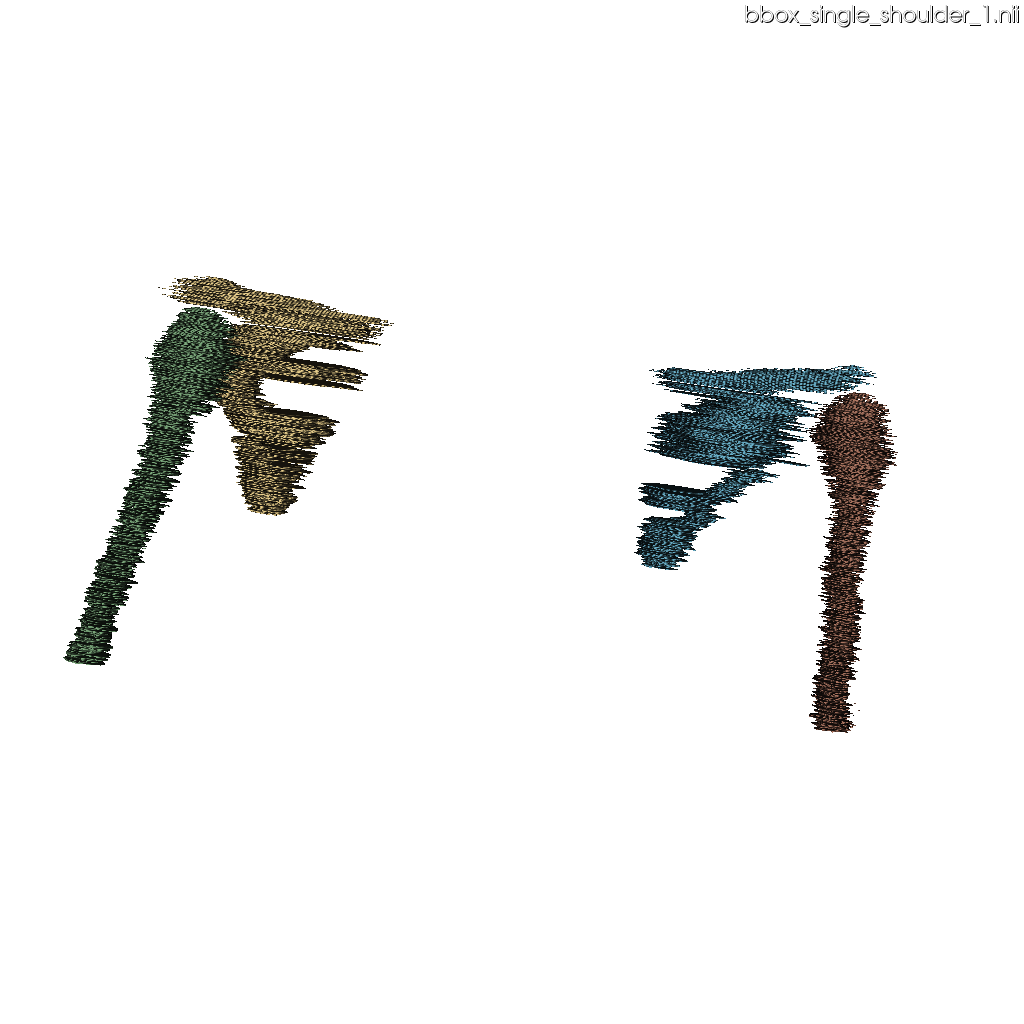} & \raisebox{-0.6\height}[0pt][0pt]{\includegraphics[width=0.11\linewidth, trim=250 150 300 50, clip]{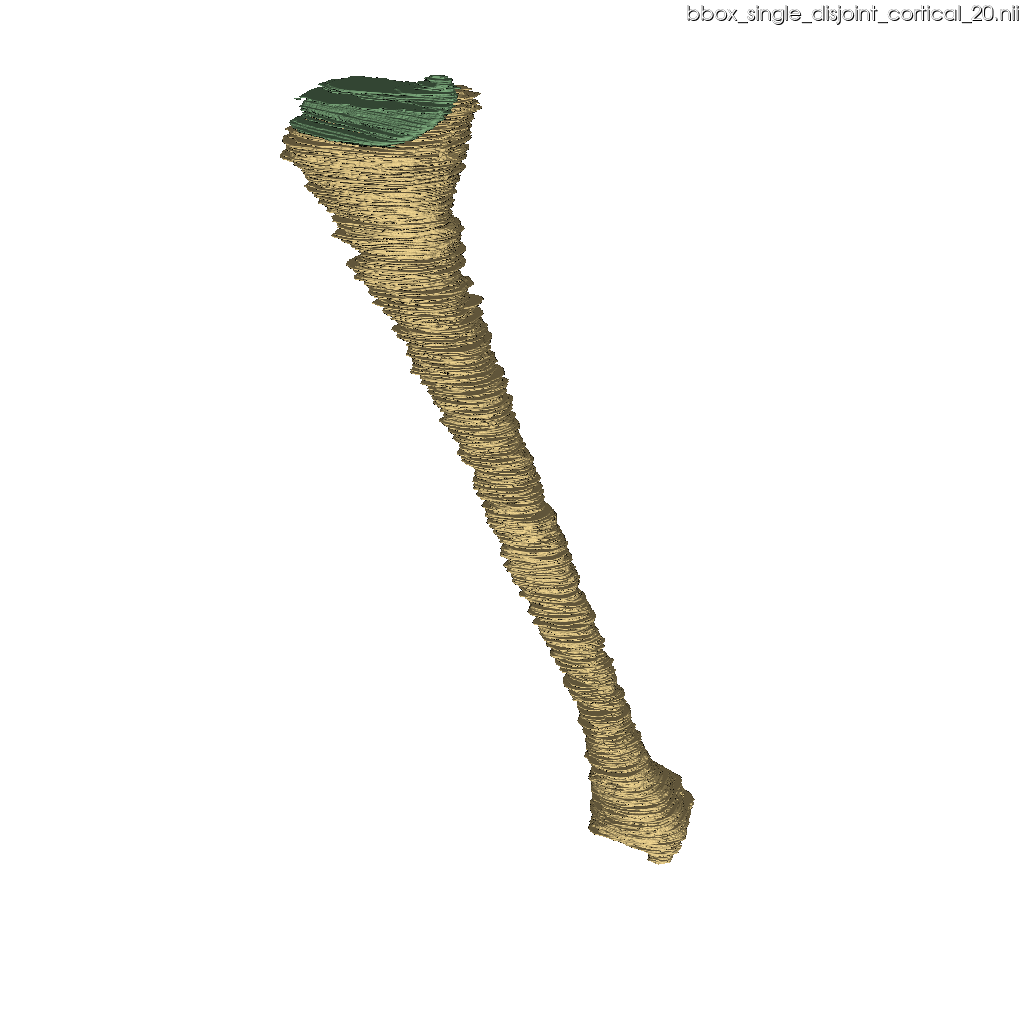}} &
    \includegraphics[width=0.14\linewidth, trim=560 380 130 380, clip]{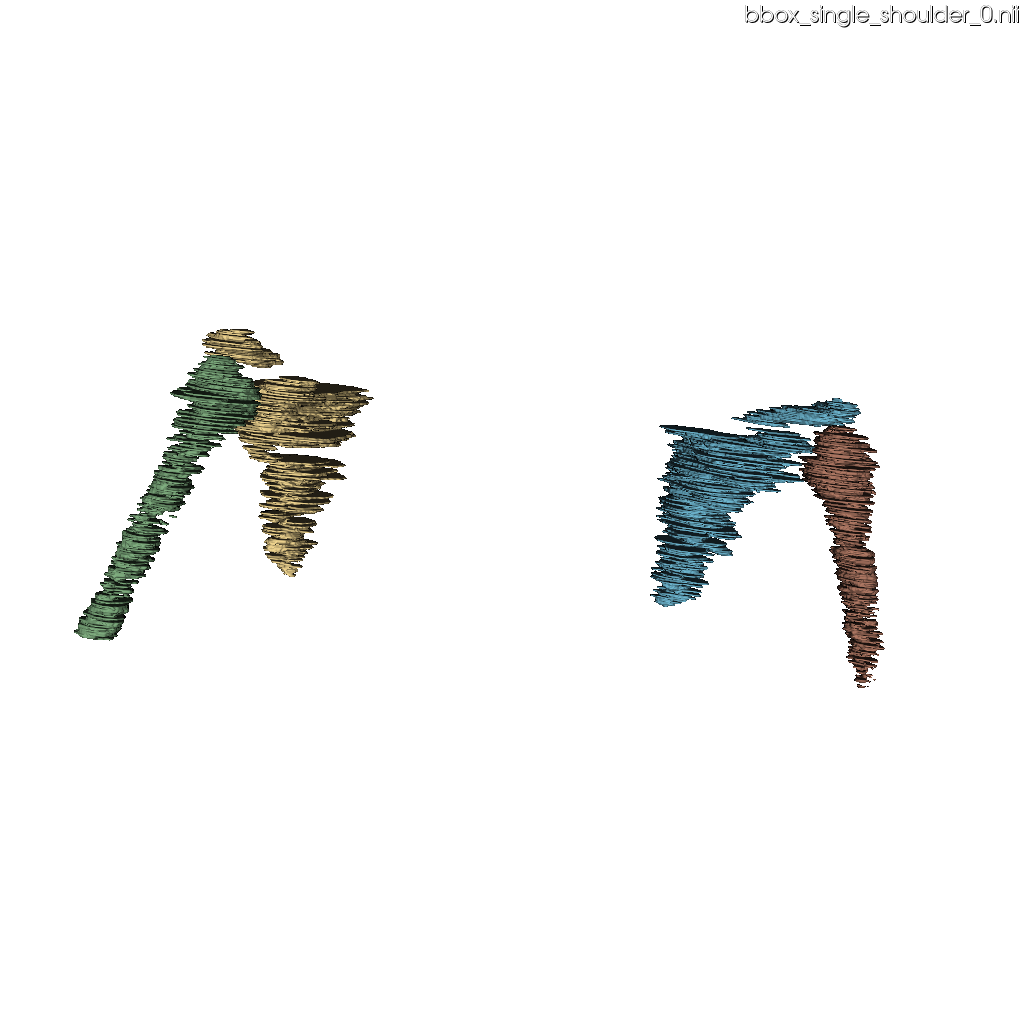} & \raisebox{-0.6\height}[0pt][0pt]{\includegraphics[width=0.11\linewidth, trim=400 150 400 100, clip]{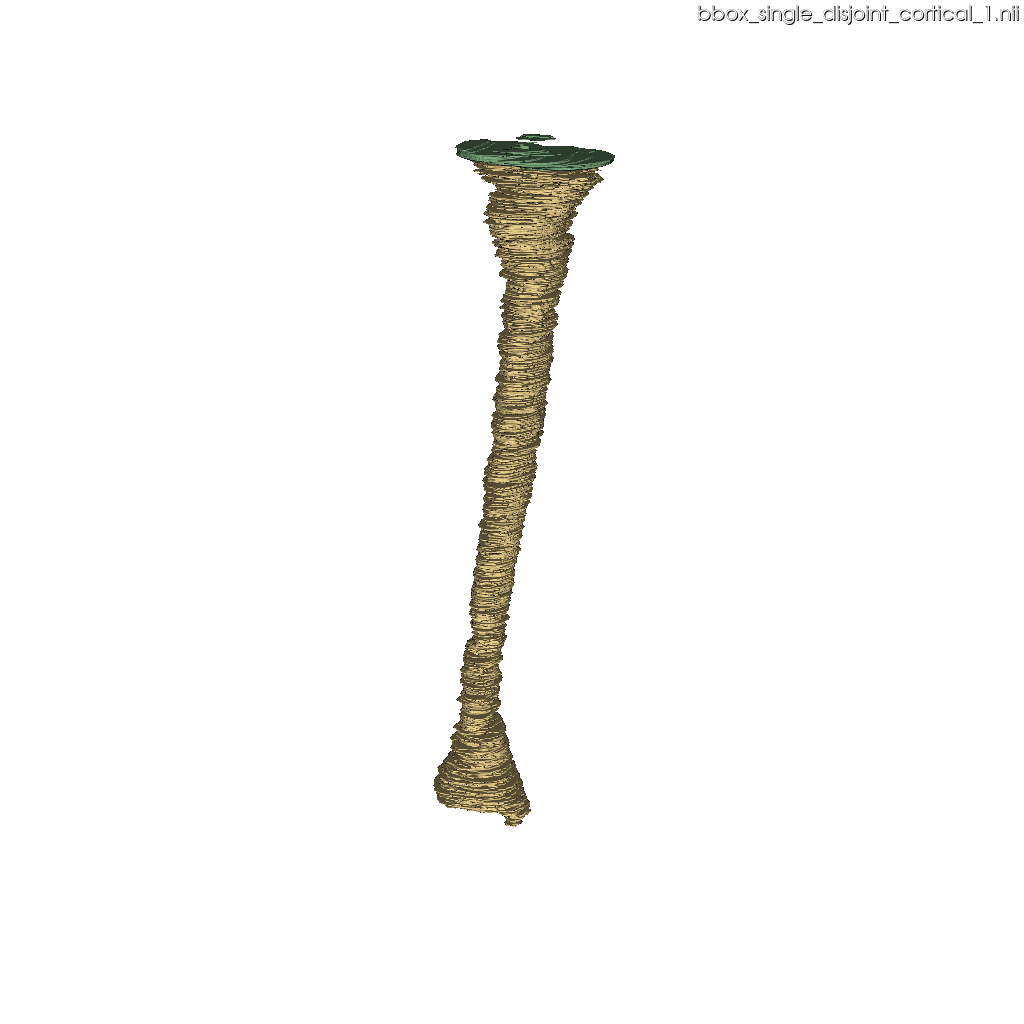}} &
    \includegraphics[width=0.14\linewidth, trim=560 380 190 330, clip]{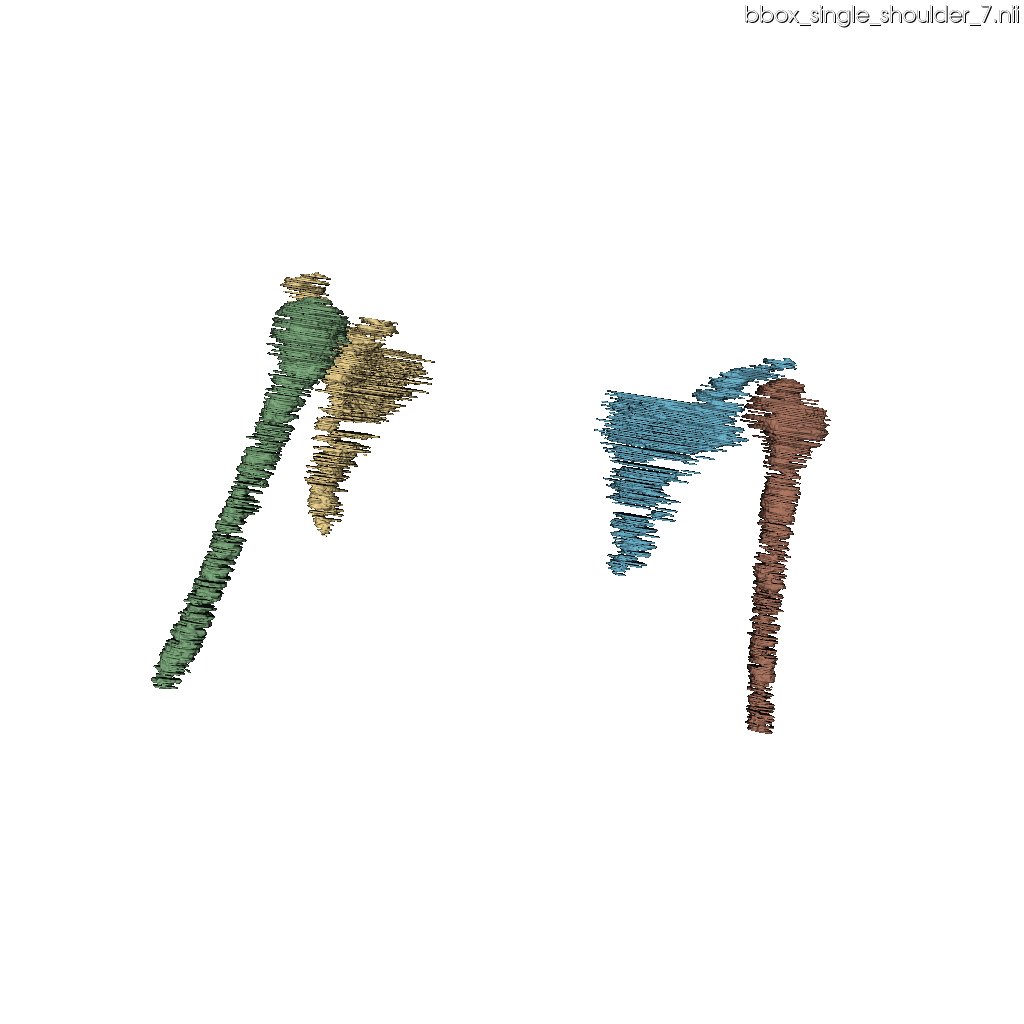} & \raisebox{-0.6\height}[0pt][0pt]{\includegraphics[width=0.11\linewidth, trim=400 150 400 100, clip]{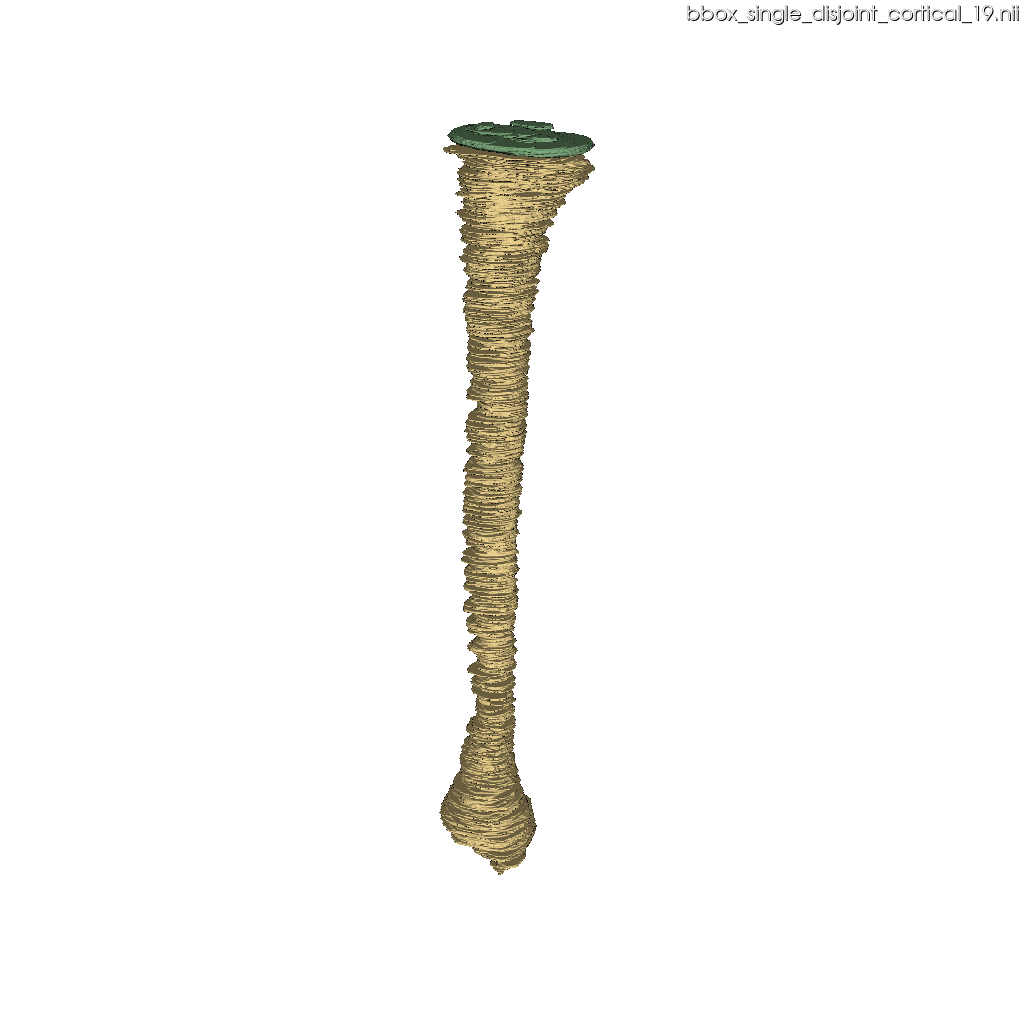}} 
    \\
    & \includegraphics[width=0.14\linewidth, trim=370 290 370 280, clip]{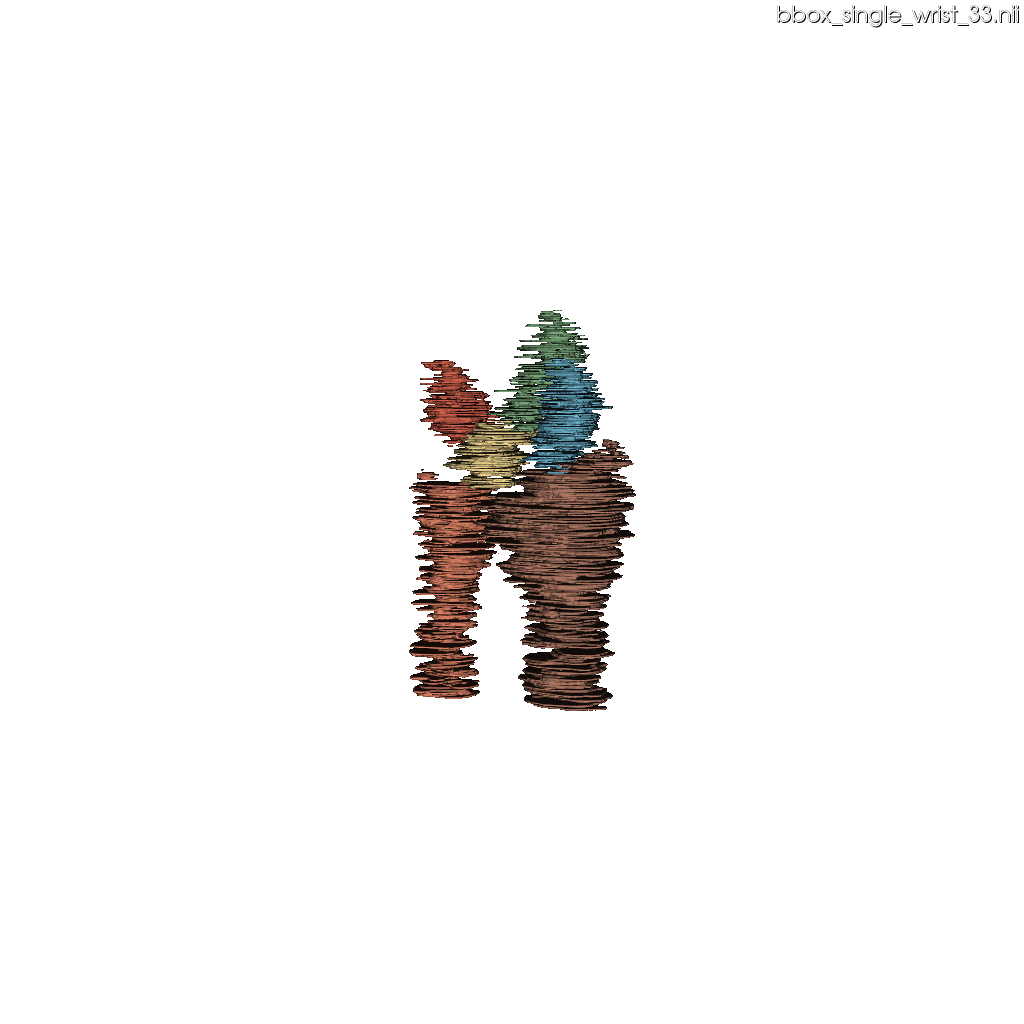} & &
    \includegraphics[width=0.14\linewidth, trim=370 290 370 280, clip]{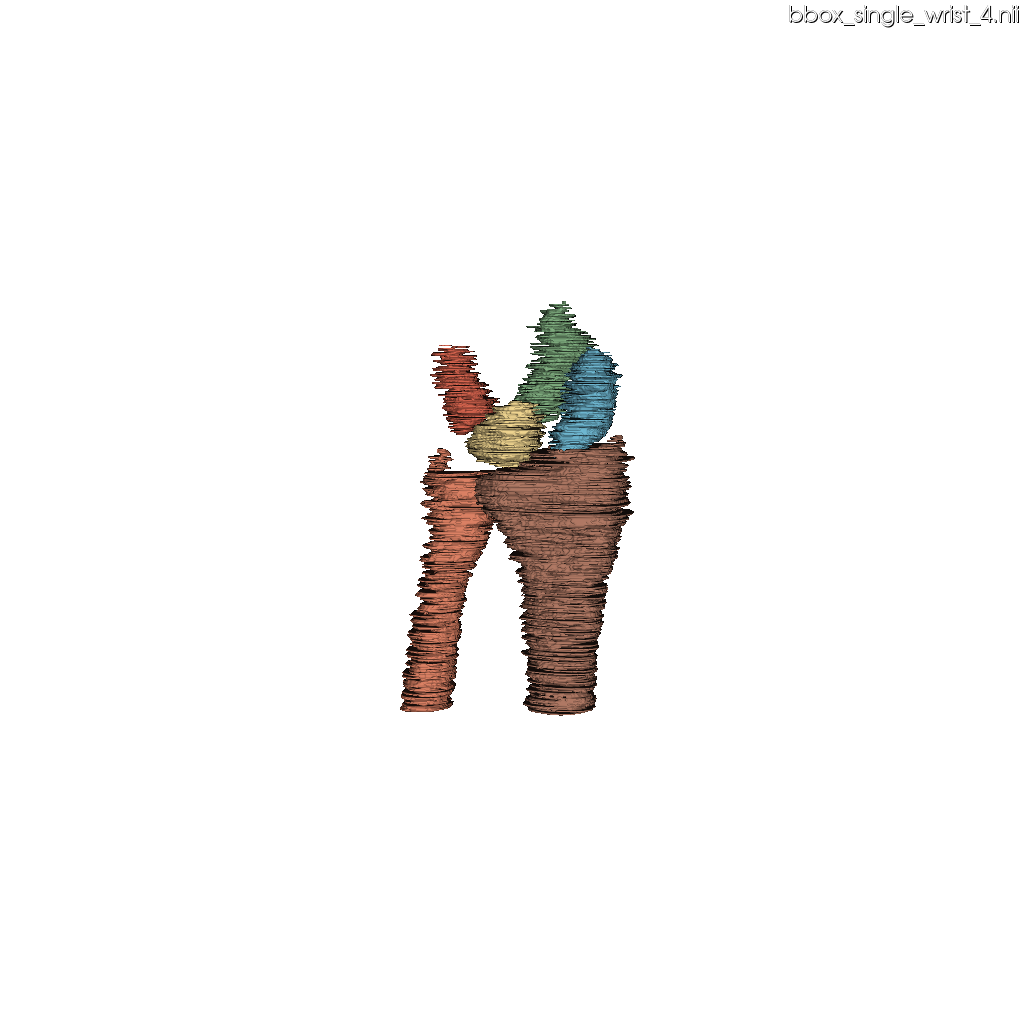} & &
    \includegraphics[width=0.14\linewidth, trim=370 290 370 280, clip]{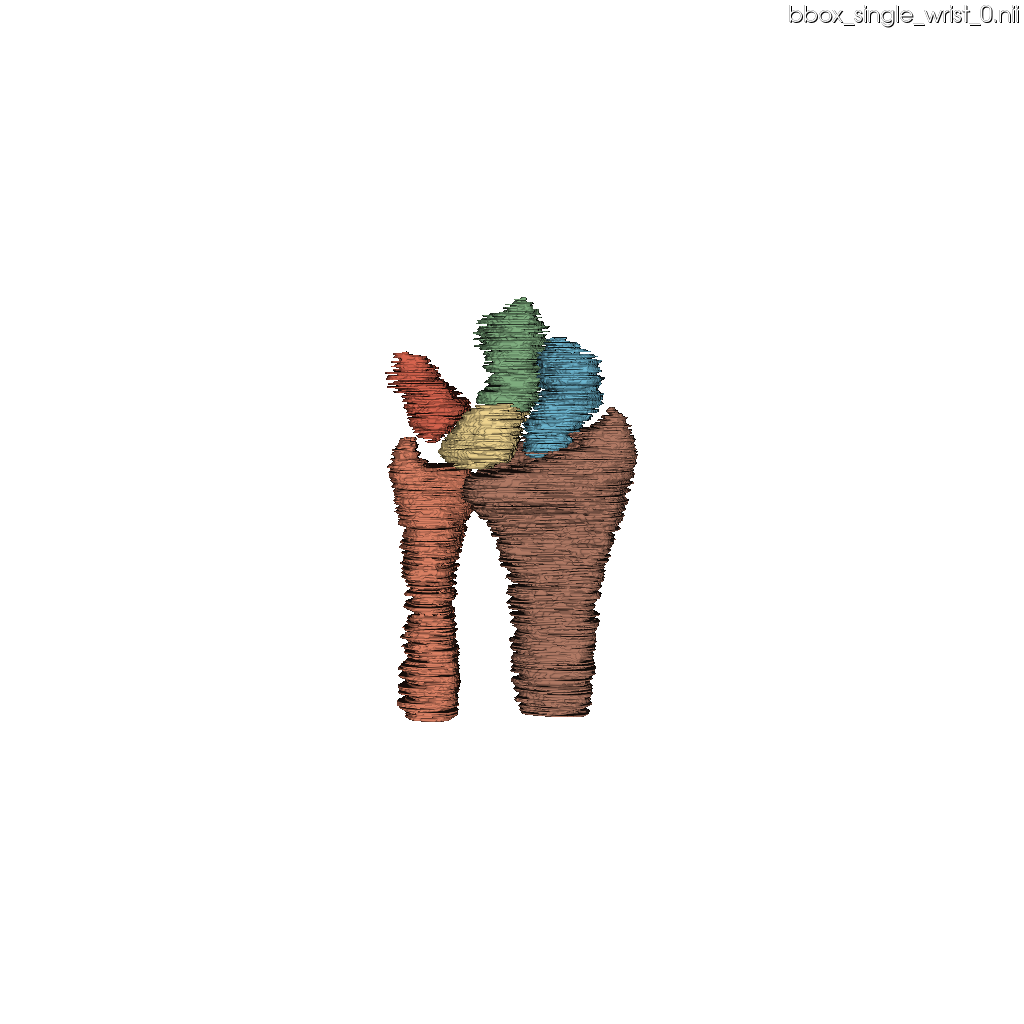} & \\
    \hline
\end{tabular}
\caption{Selected examples for \textit{Med-\textsc{Sam}} with low, medium and high DSC.}
\label{fig:example_medsam}
\end{figure}

\begin{figure}[H]
\begin{tabular}{|l|cc|cc|cc|}
    \hline
    Setting & \multicolumn{2}{c|}{DSC $\downarrow$} & \multicolumn{2}{c|}{DSC median} & \multicolumn{2}{c|}{DSC $\uparrow$} \\ 
    \hline
    \multirow{2}{*}{\cblacksquare[0.6]{red}} & \includegraphics[width=0.14\linewidth, trim=560 380 120 330, clip]{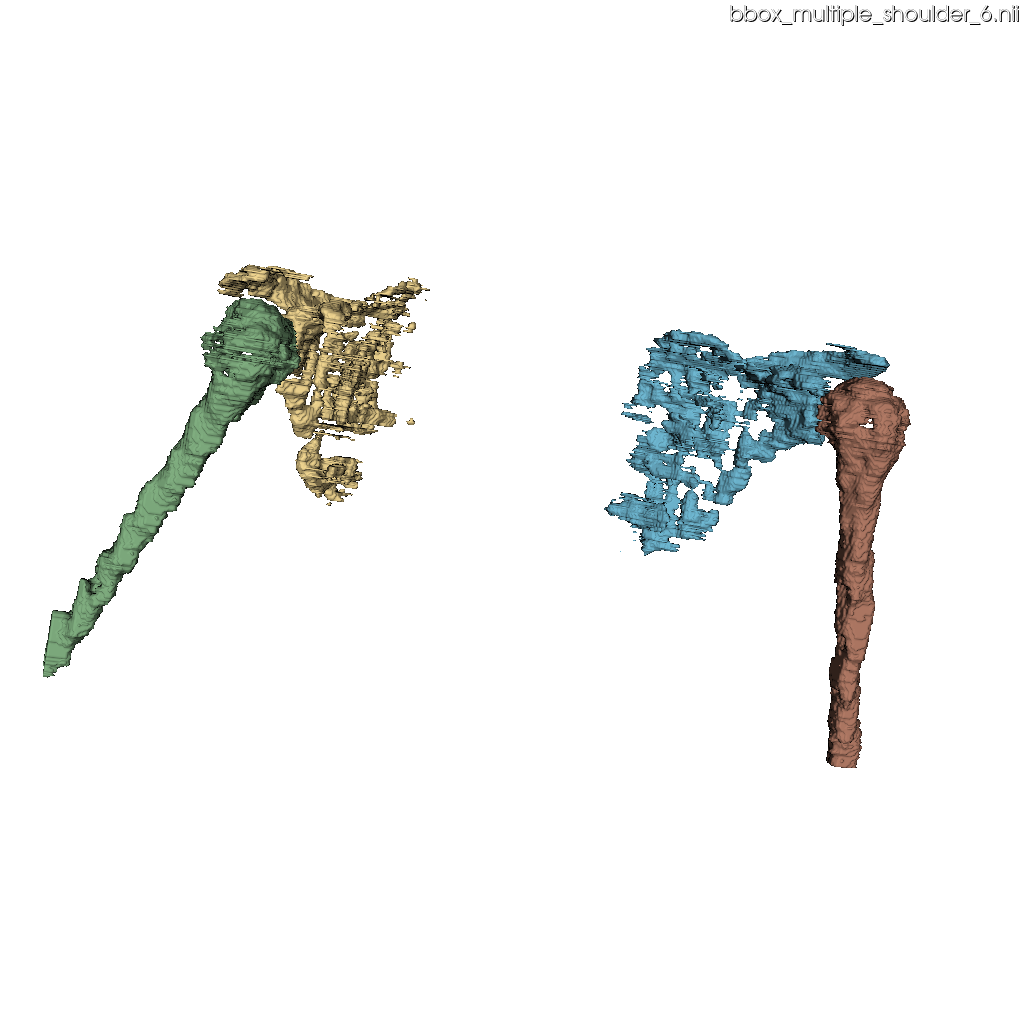} & \raisebox{-0.6\height}[0pt][0pt]{\includegraphics[width=0.11\linewidth, trim=400 50 370 30, clip]{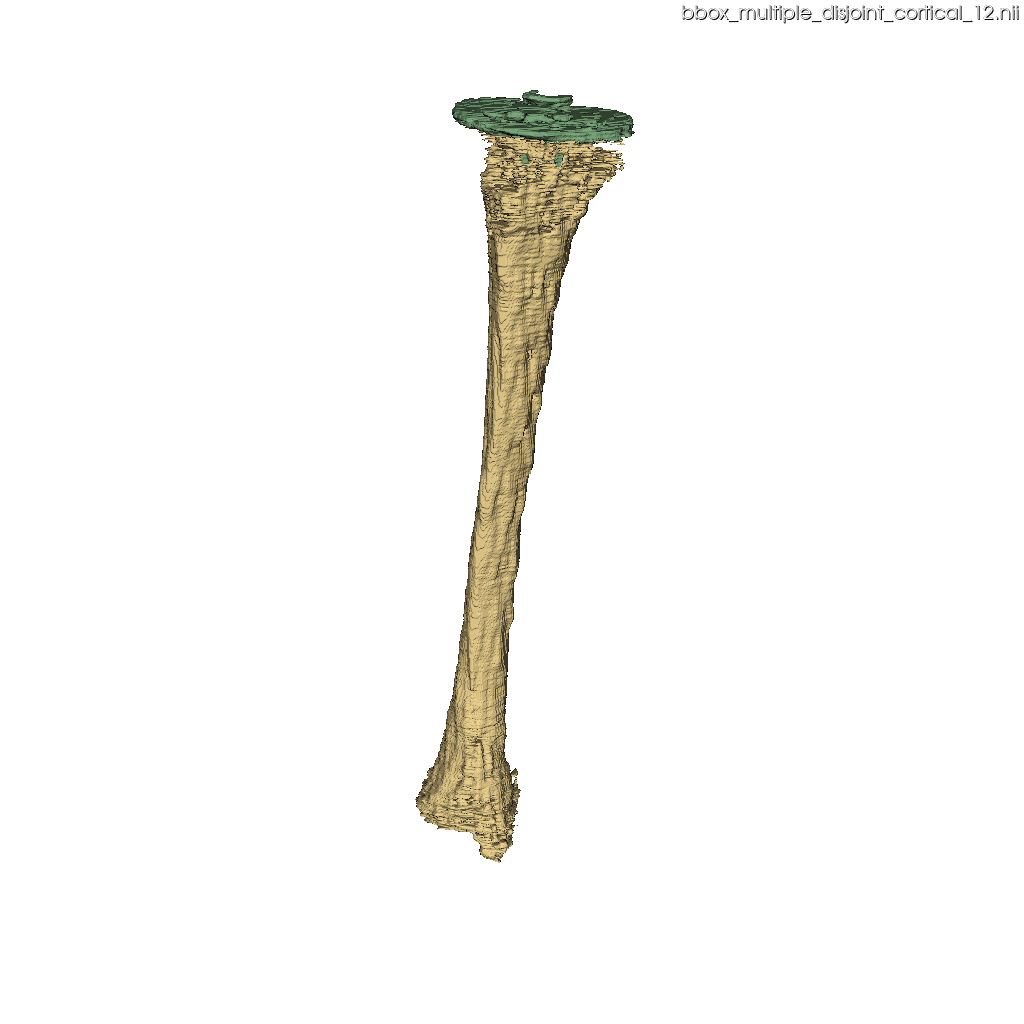}} &
    \includegraphics[width=0.14\linewidth, trim=560 380 120 330, clip]{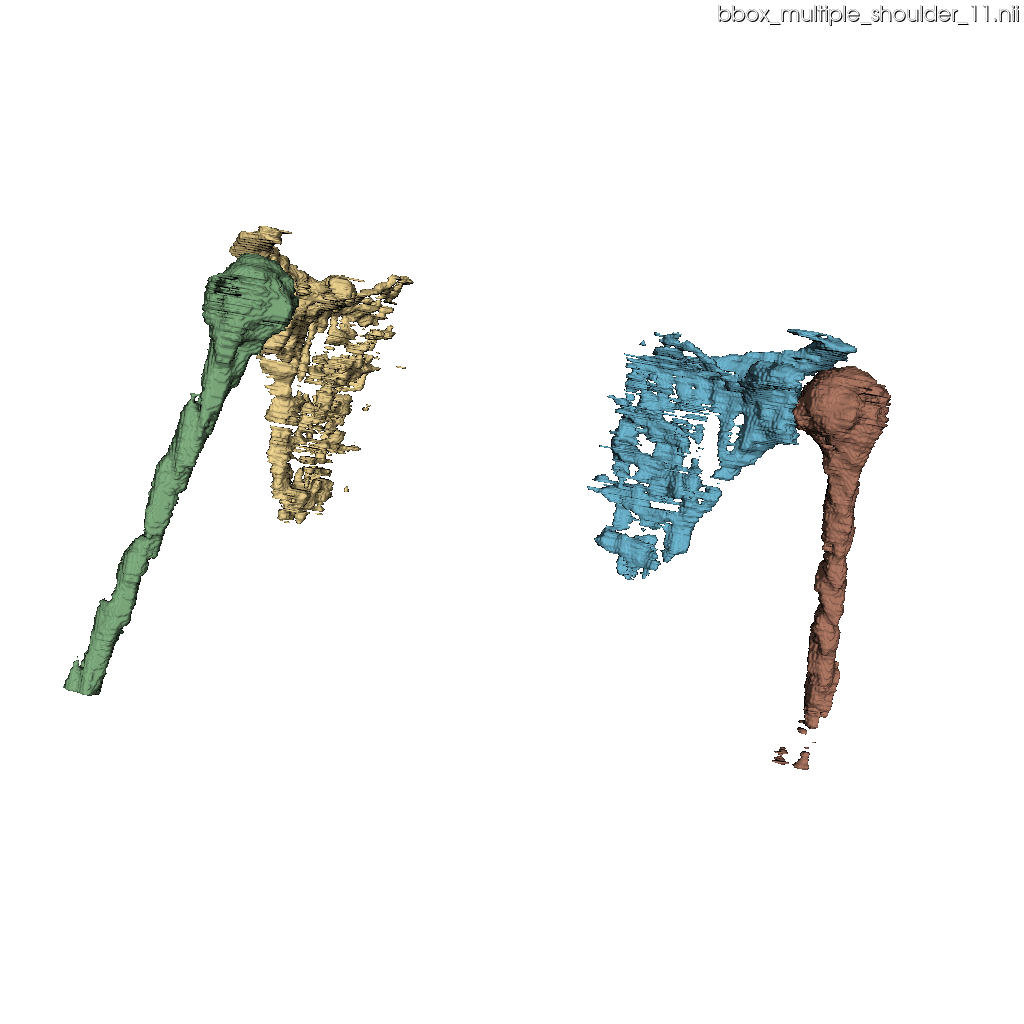} & \raisebox{-0.6\height}[0pt][0pt]{\includegraphics[width=0.11\linewidth, trim=400 200 400 100, clip]{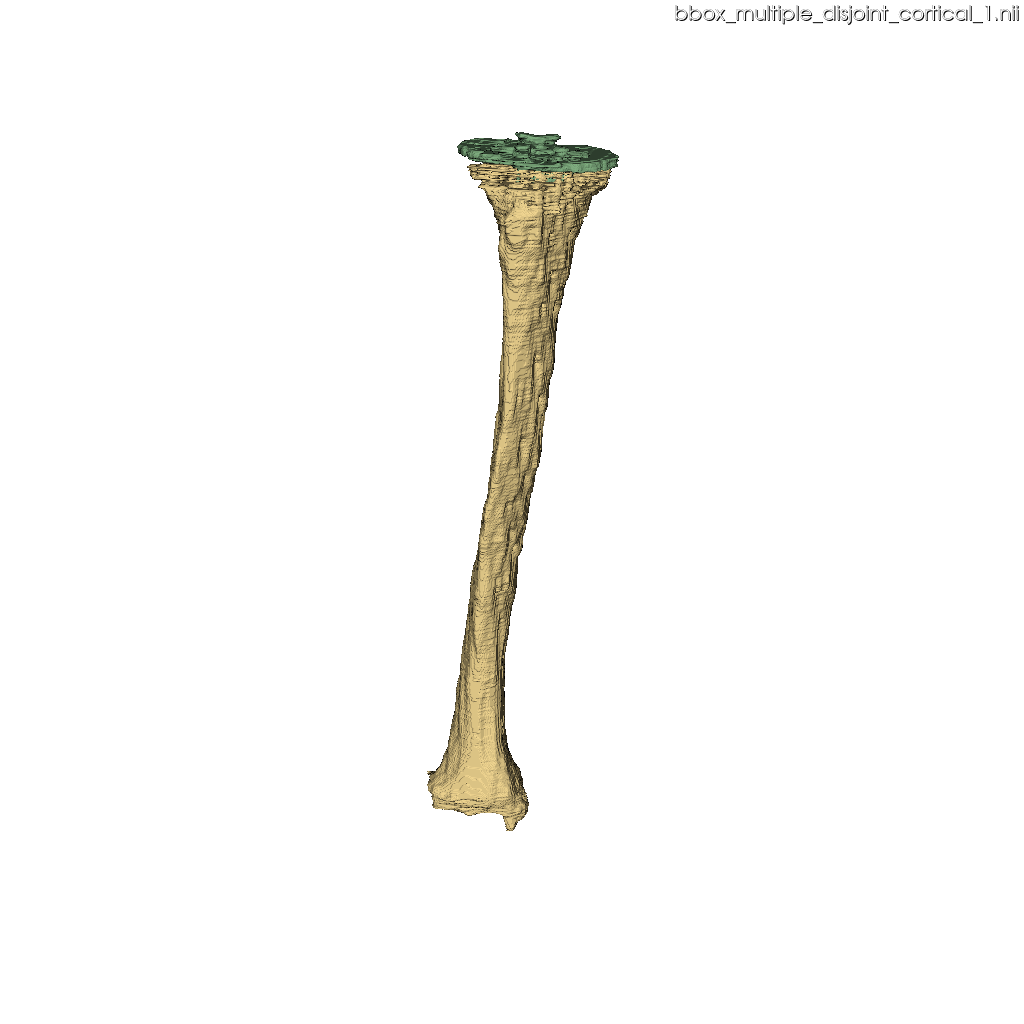}} &
    \includegraphics[width=0.14\linewidth, trim=560 380 190 330, clip]{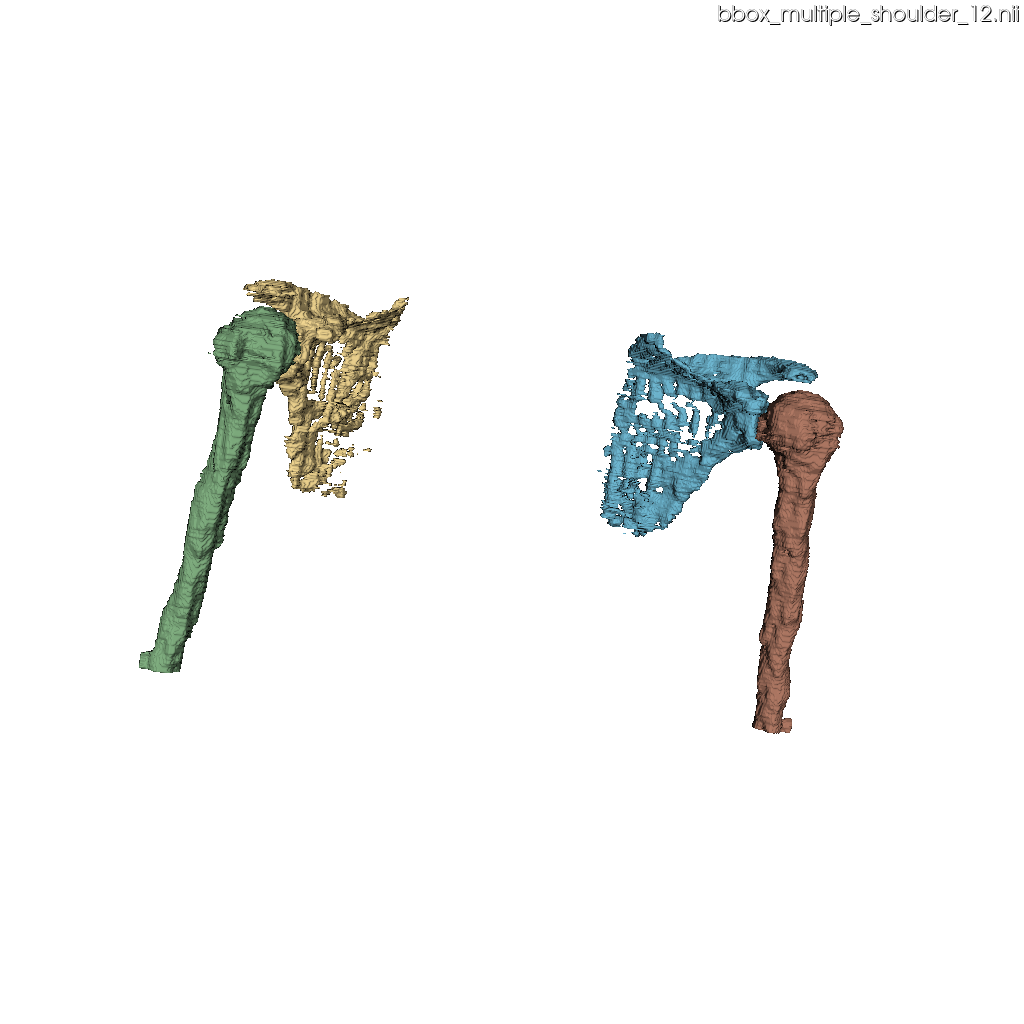} & \raisebox{-0.6\height}[0pt][0pt]{\includegraphics[width=0.11\linewidth, trim=400 150 400 100, clip]{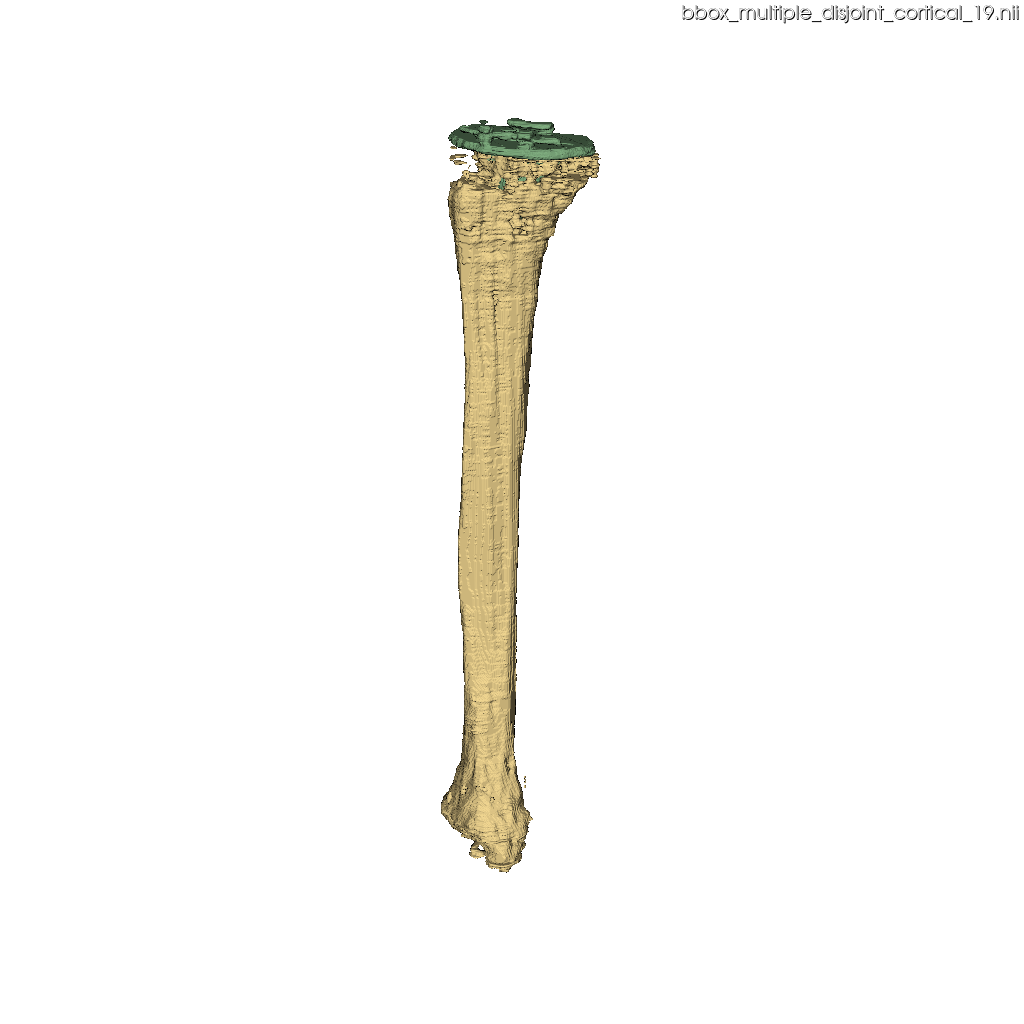}} 
    \\
    & \includegraphics[width=0.14\linewidth, trim=370 290 370 280, clip]{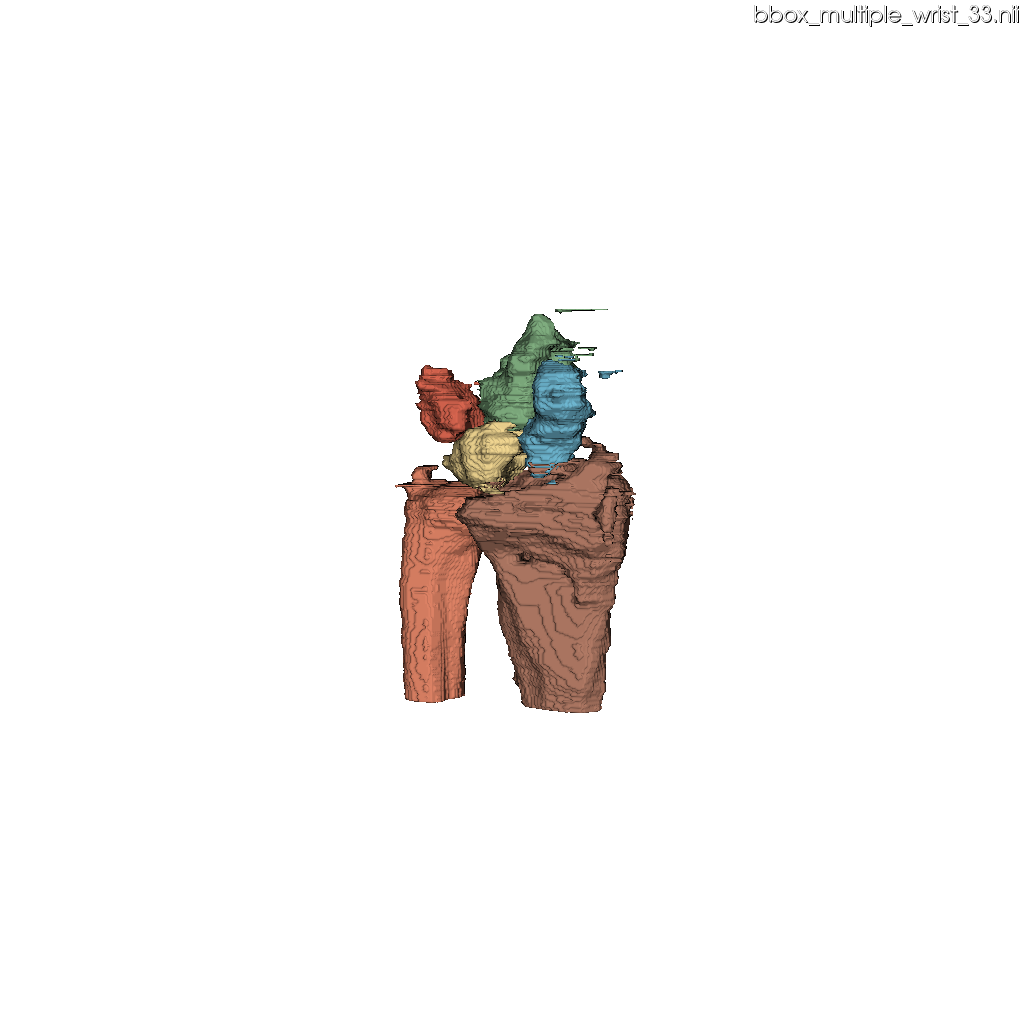} & &
    \includegraphics[width=0.14\linewidth, trim=370 290 370 280, clip]{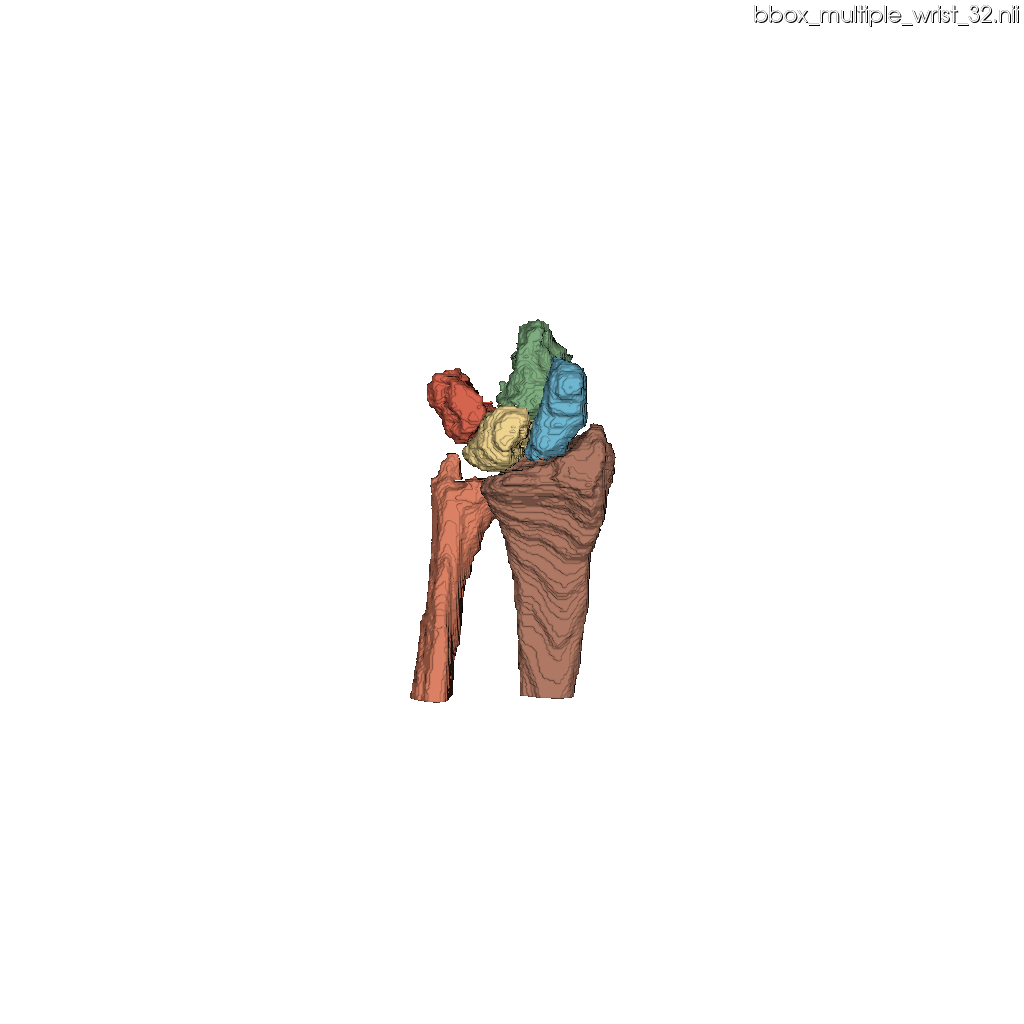} & &
    \includegraphics[width=0.14\linewidth, trim=370 290 370 280, clip]{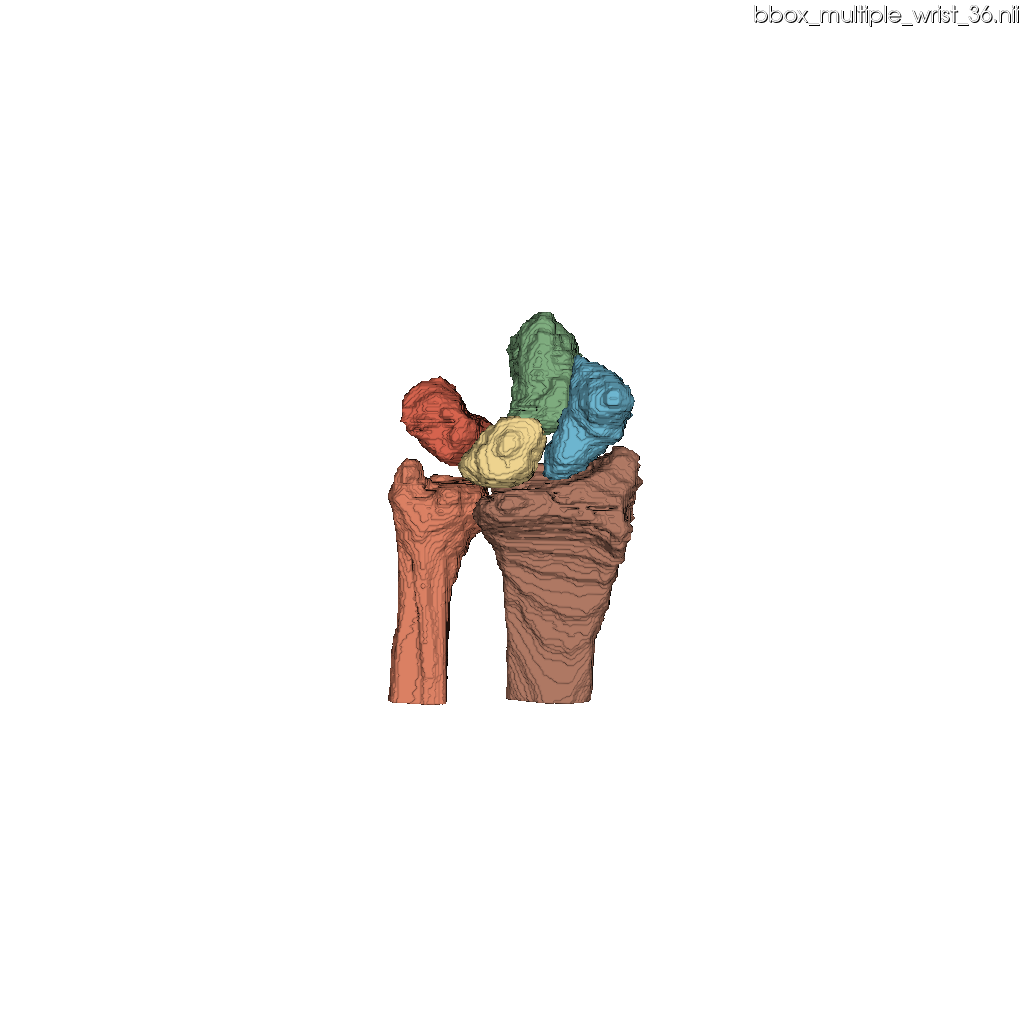} & \\ 
    \hline
    \multirow{2}{*}{\cblacksquaredot[0.6]{red}} & \includegraphics[width=0.14\linewidth, trim=560 380 190 330, clip]{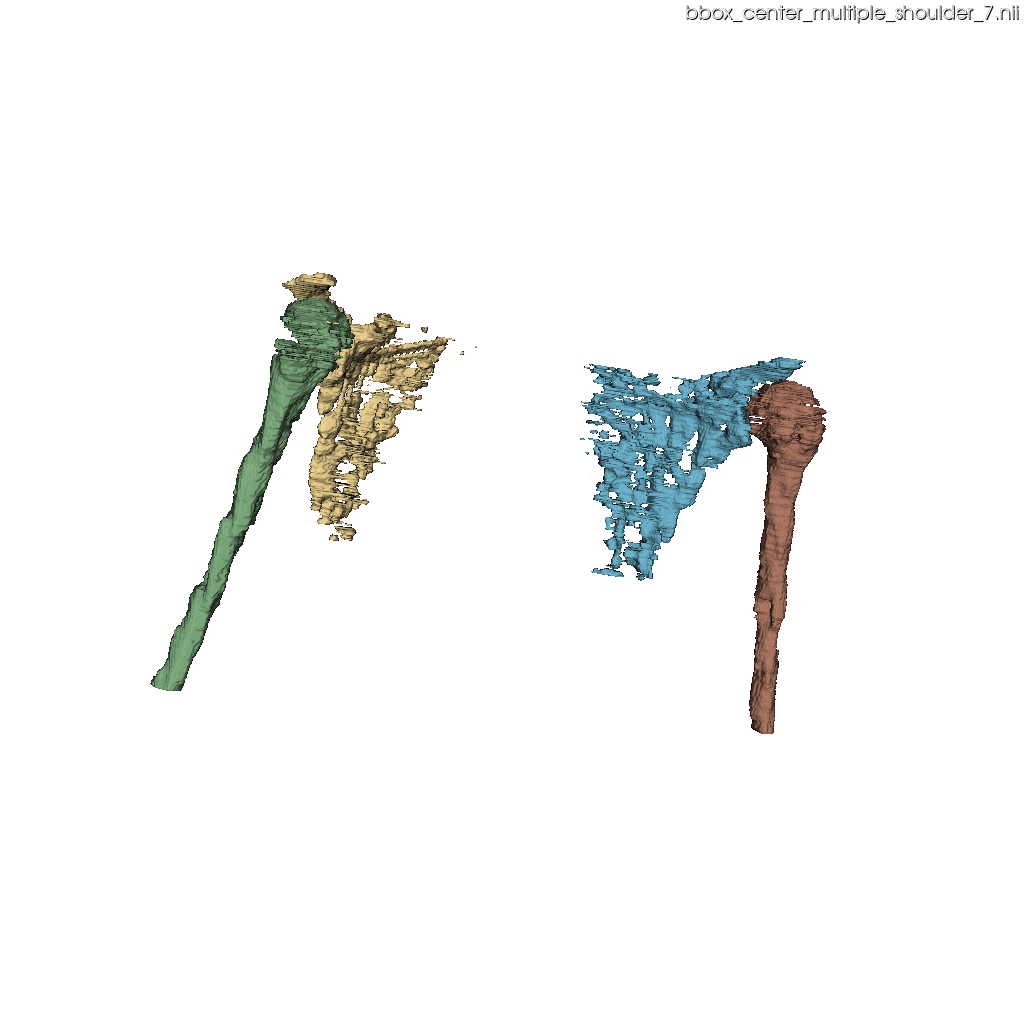} & \raisebox{-0.6\height}[0pt][0pt]{\includegraphics[width=0.11\linewidth, trim=400 130 400 120, clip]{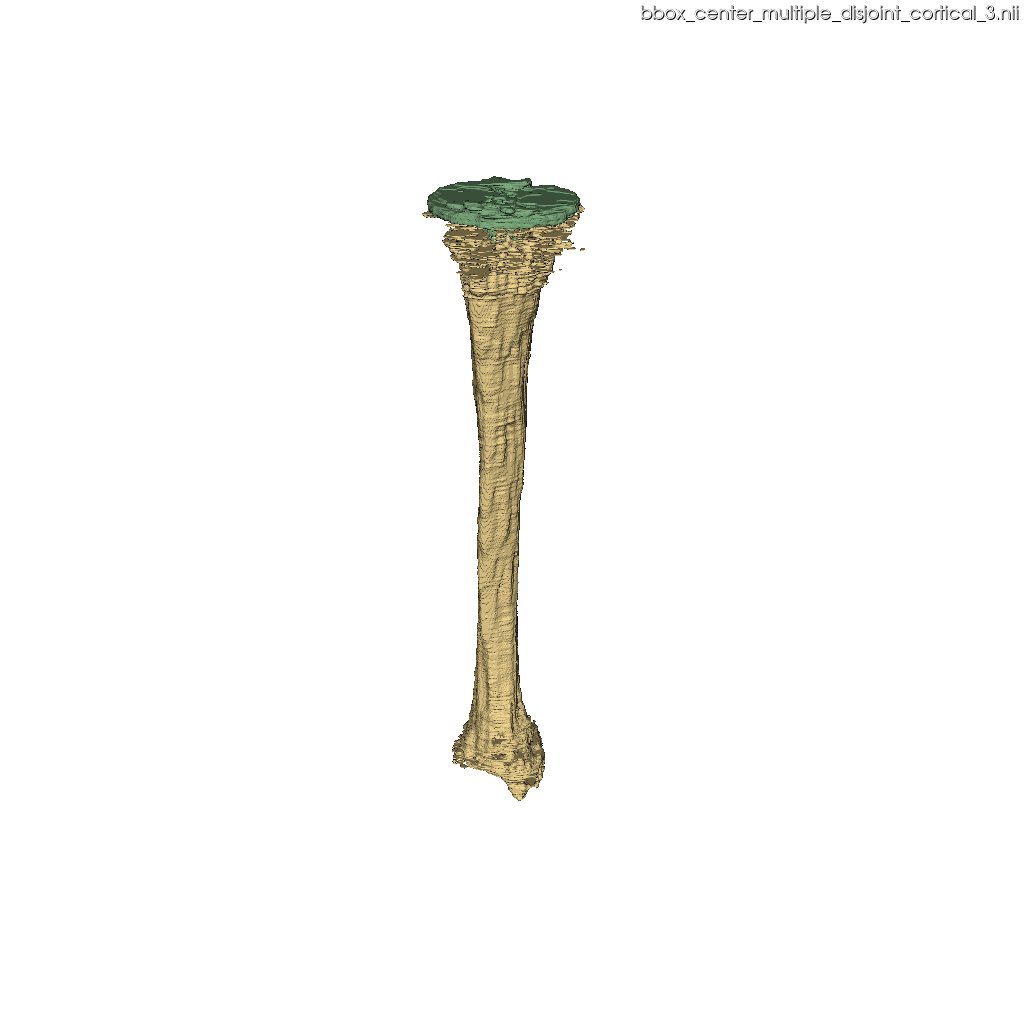}} &
    \includegraphics[width=0.14\linewidth, trim=560 350 100 300, clip]{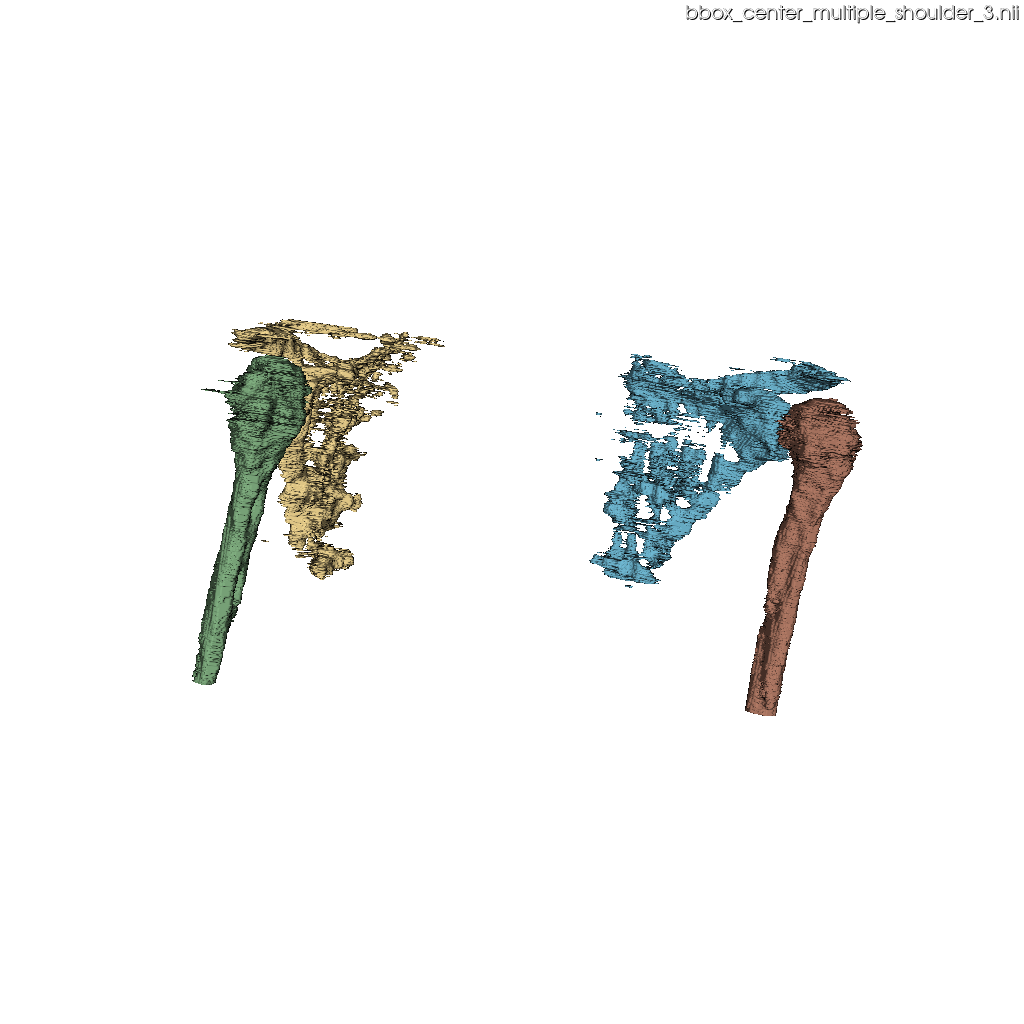} & \raisebox{-0.6\height}[0pt][0pt]{\includegraphics[width=0.11\linewidth, trim=400 150 400 50, clip]{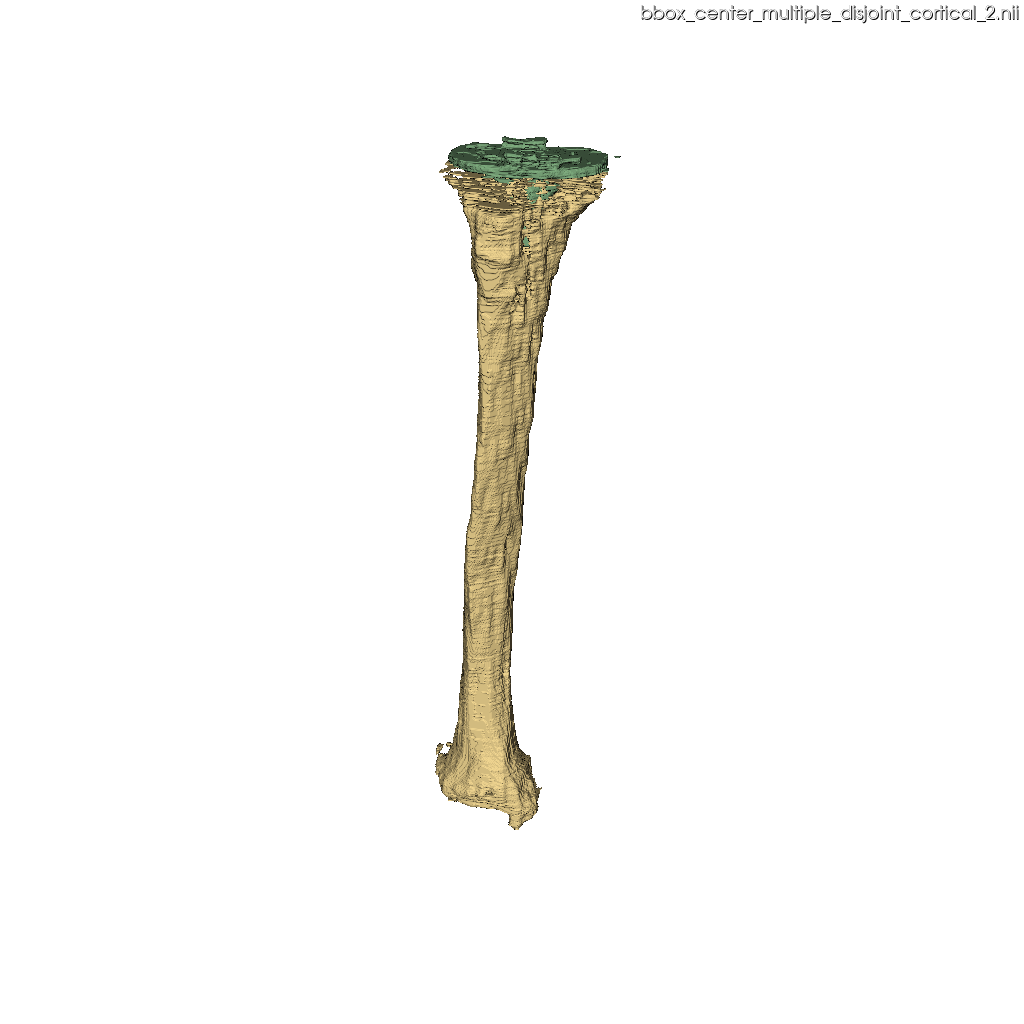}} &
    \includegraphics[width=0.14\linewidth, trim=560 380 190 330, clip]{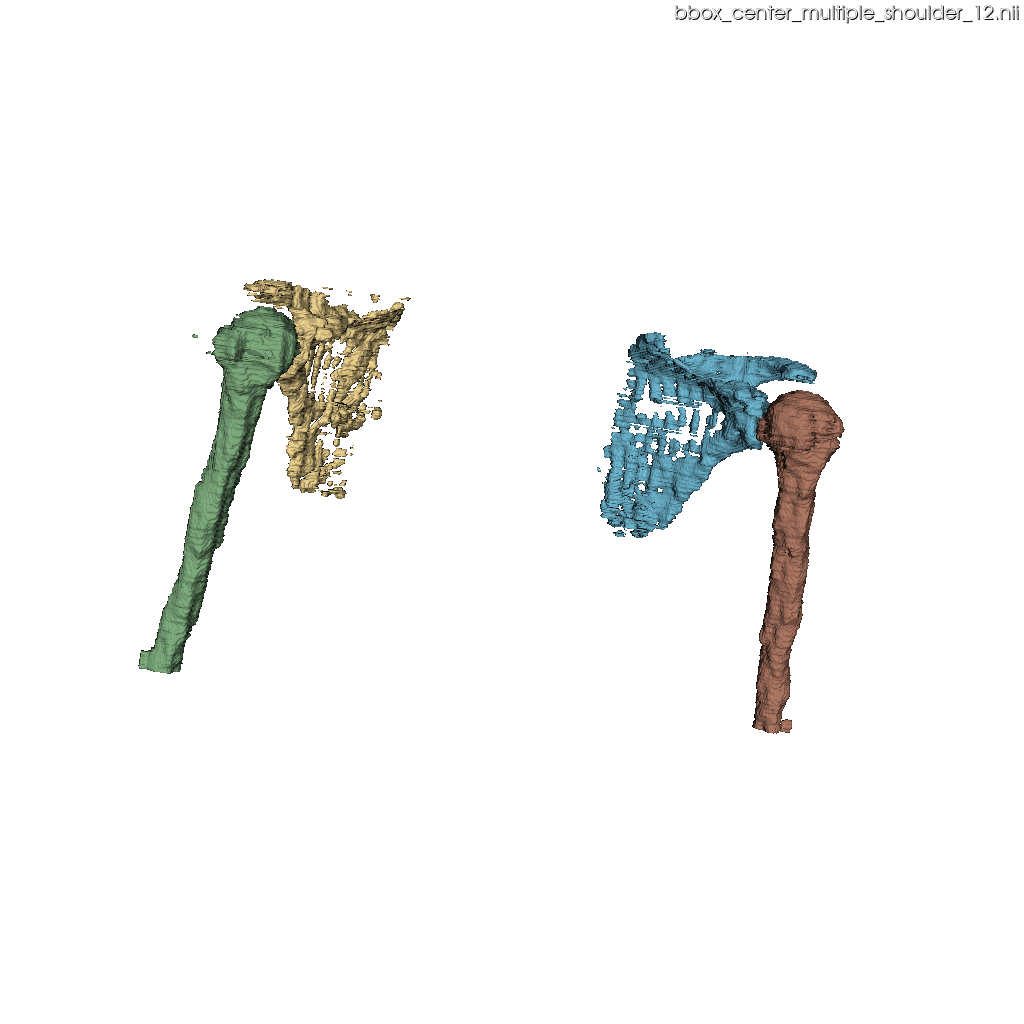} & \raisebox{-0.6\height}[0pt][0pt]{\includegraphics[width=0.11\linewidth, trim=400 150 400 100, clip]{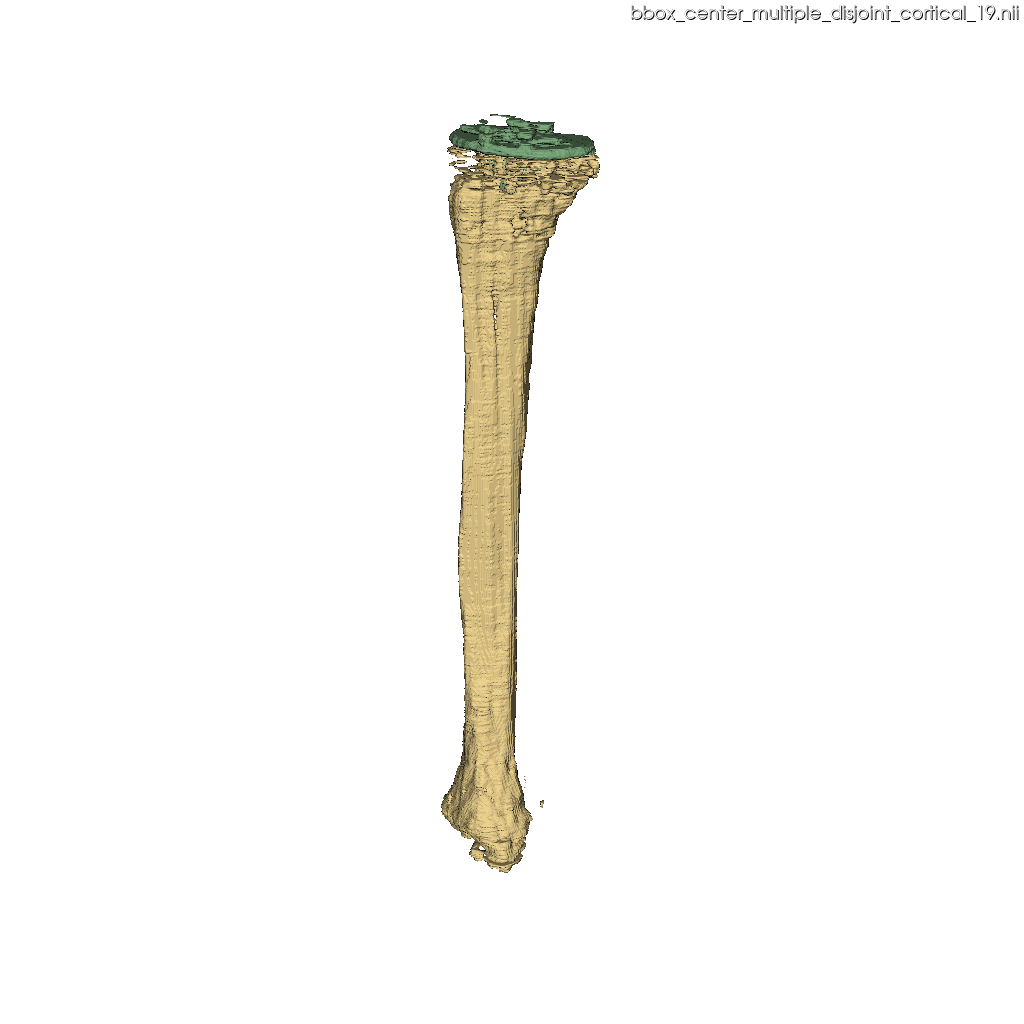}} 
    \\
    & \includegraphics[width=0.14\linewidth, trim=370 290 370 280, clip]{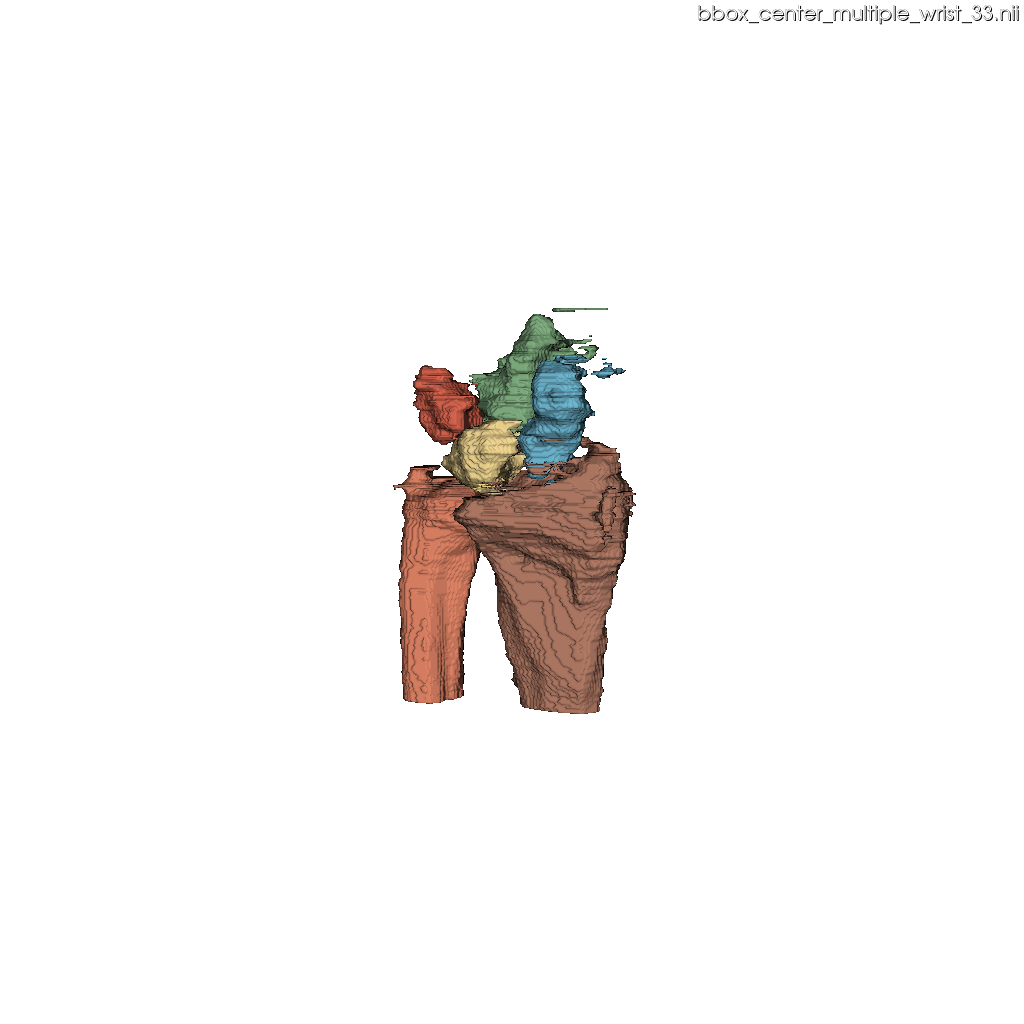} & &
    \includegraphics[width=0.14\linewidth, trim=370 290 370 280, clip]{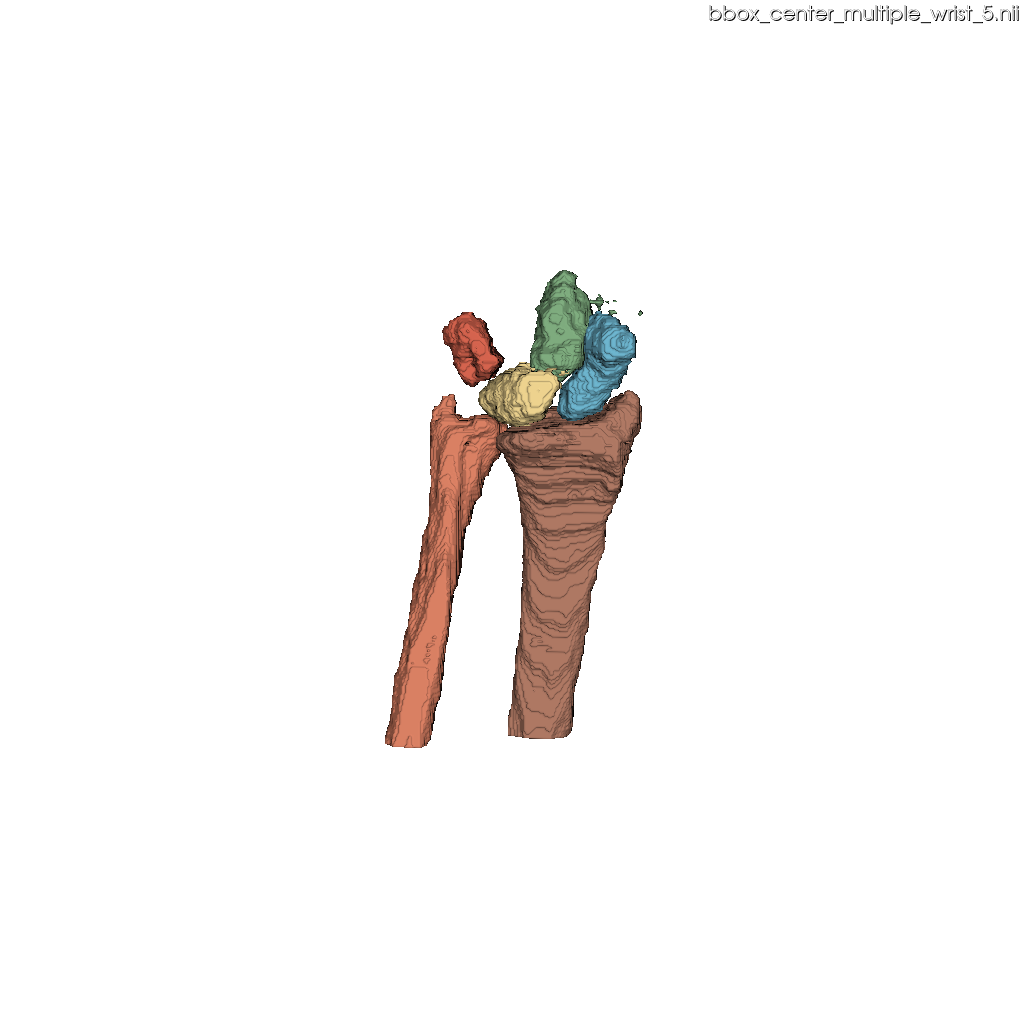} & &
    \includegraphics[width=0.14\linewidth, trim=370 290 370 280, clip]{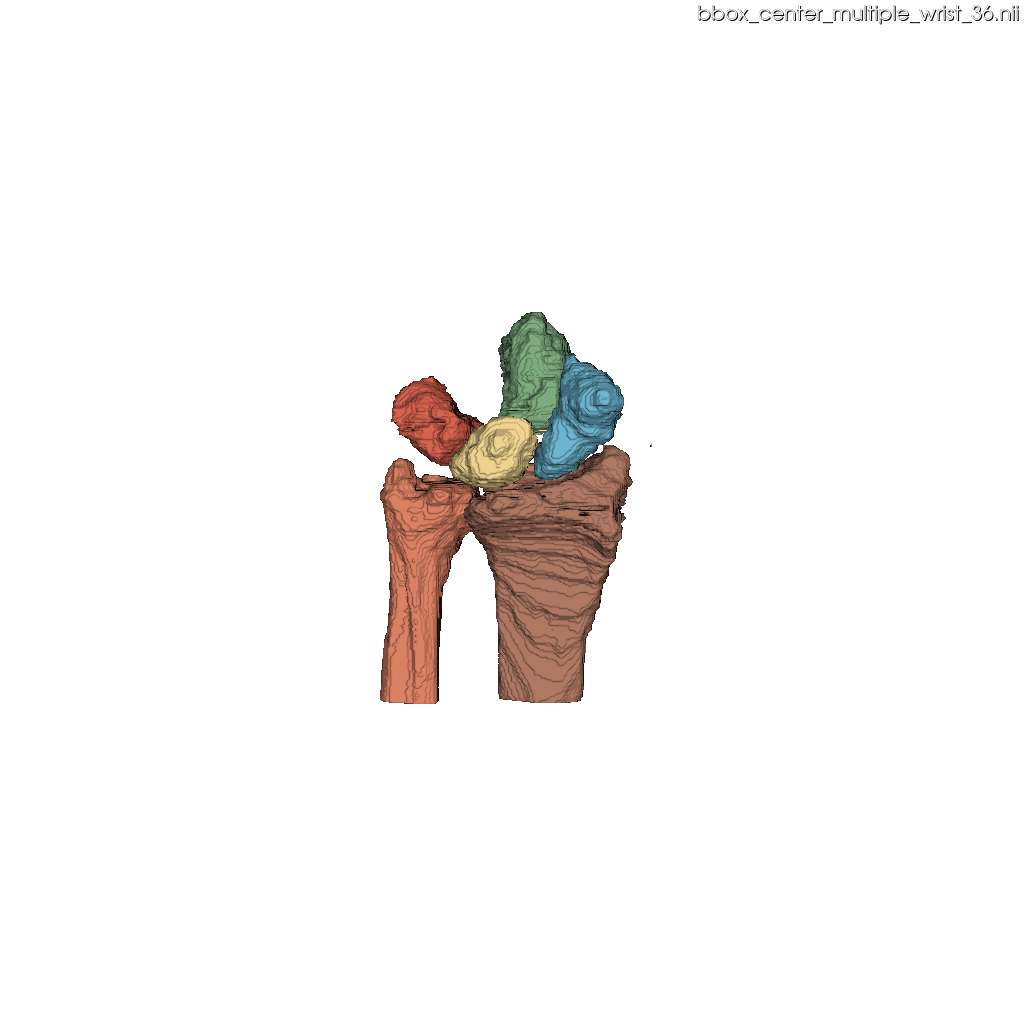} & \\ 
    \hline
    \multirow{2}{*}{\cblackstartriangledown[0.6]{red}} & \includegraphics[width=0.14\linewidth, trim=560 320 100 340, clip]{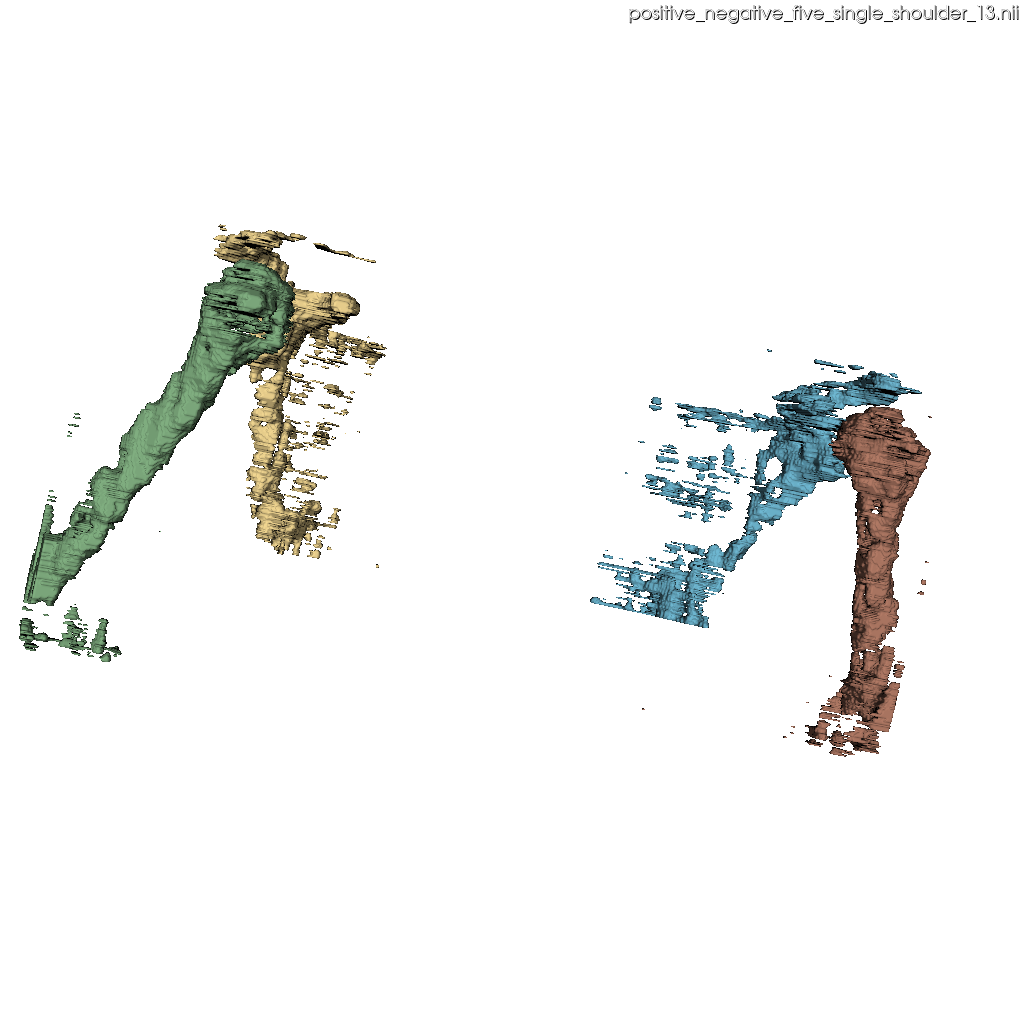} & \raisebox{-0.6\height}[0pt][0pt]{\includegraphics[width=0.11\linewidth, trim=400 100 380 100, clip]{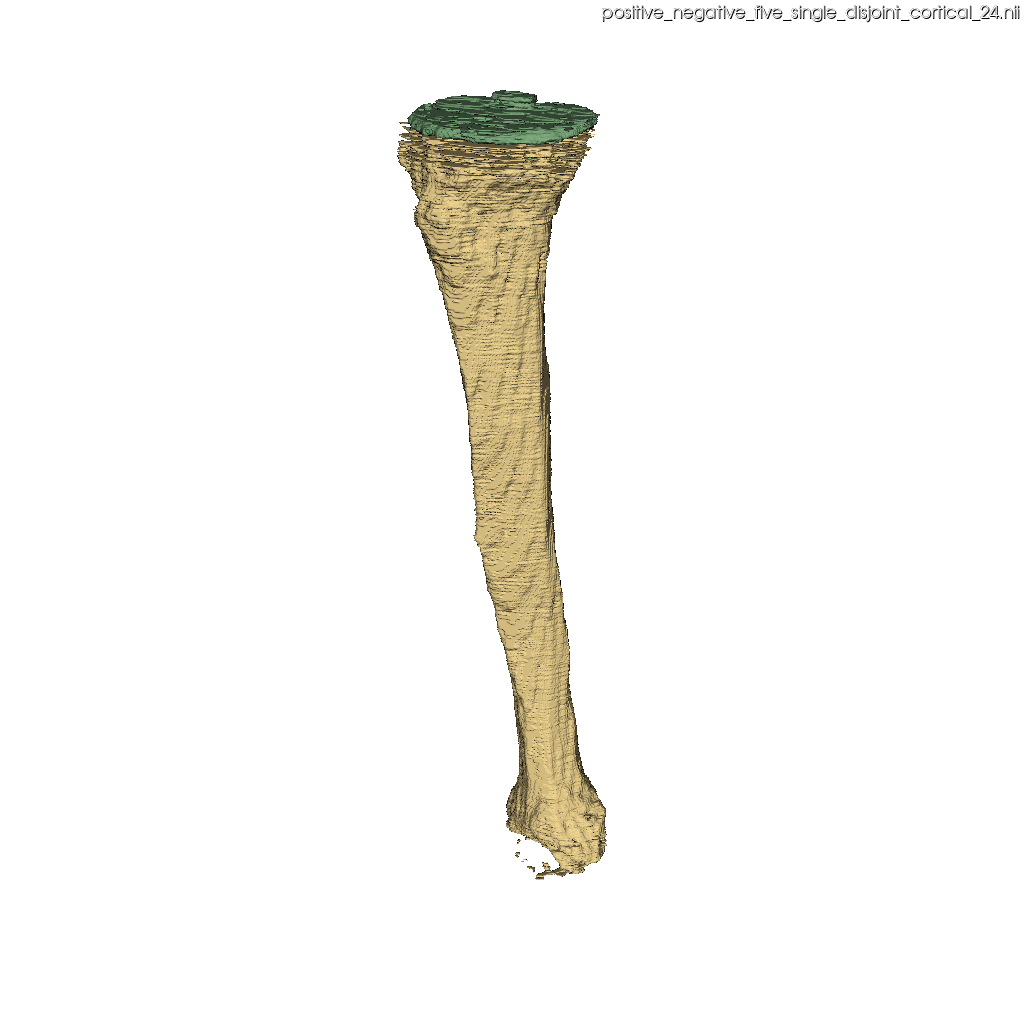}} &
    \includegraphics[width=0.14\linewidth, trim=560 380 190 330, clip]{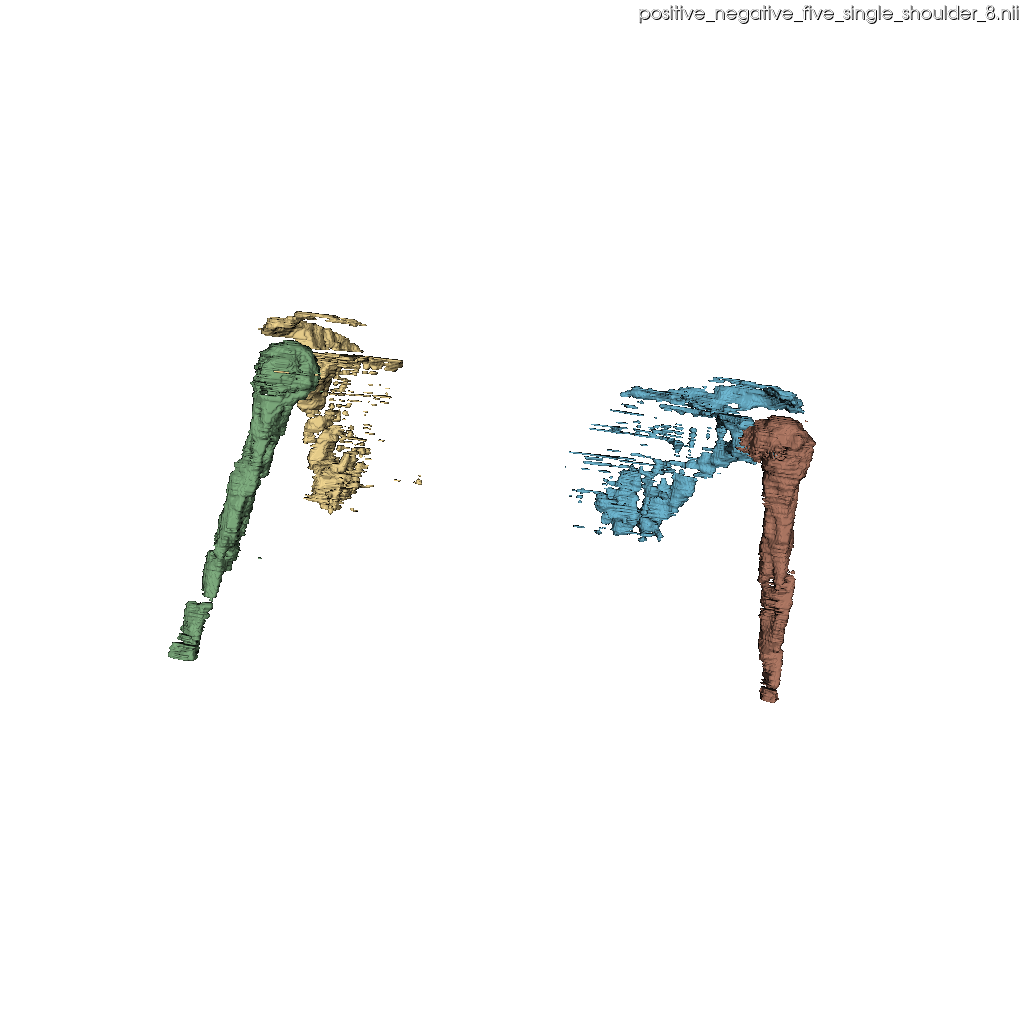} & \raisebox{-0.6\height}[0pt][0pt]{\includegraphics[width=0.11\linewidth, trim=400 100 370 50, clip]{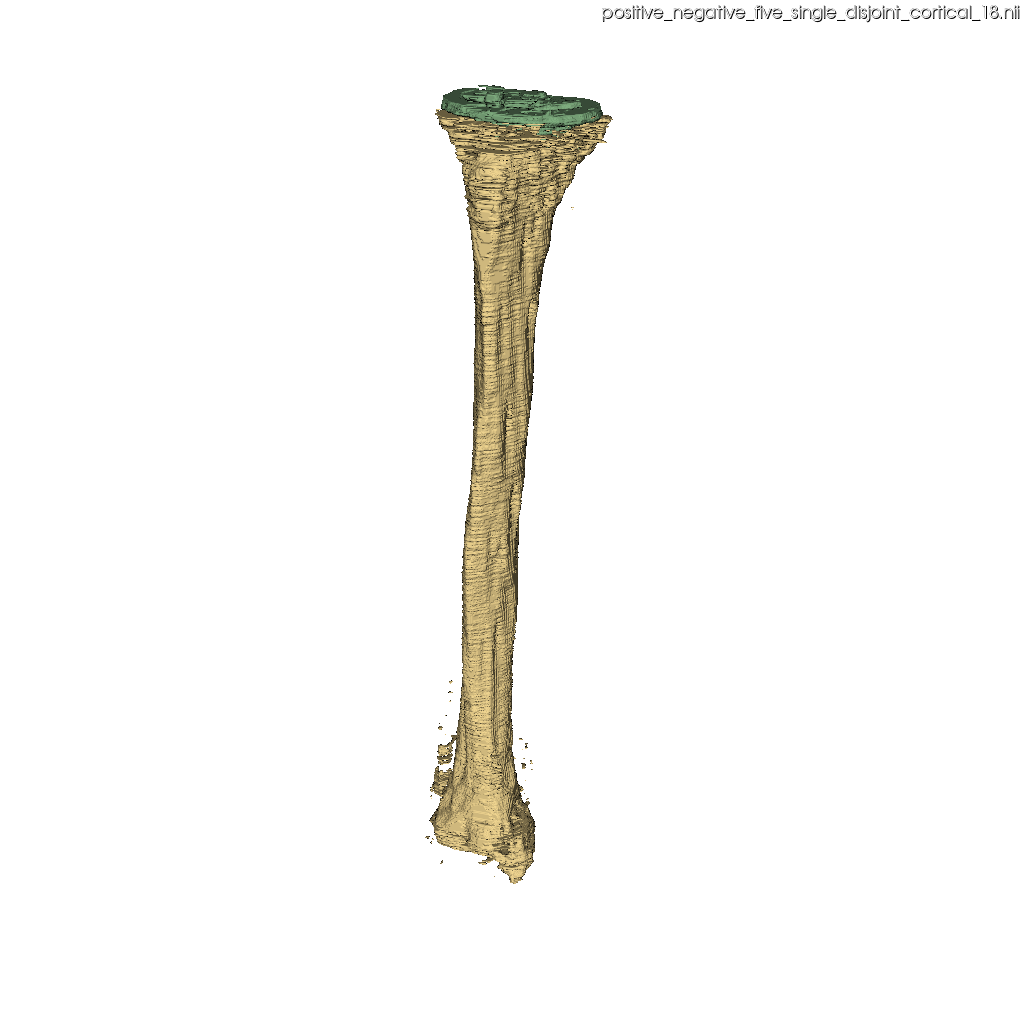}} &
    \includegraphics[width=0.14\linewidth, trim=560 380 190 300, clip]{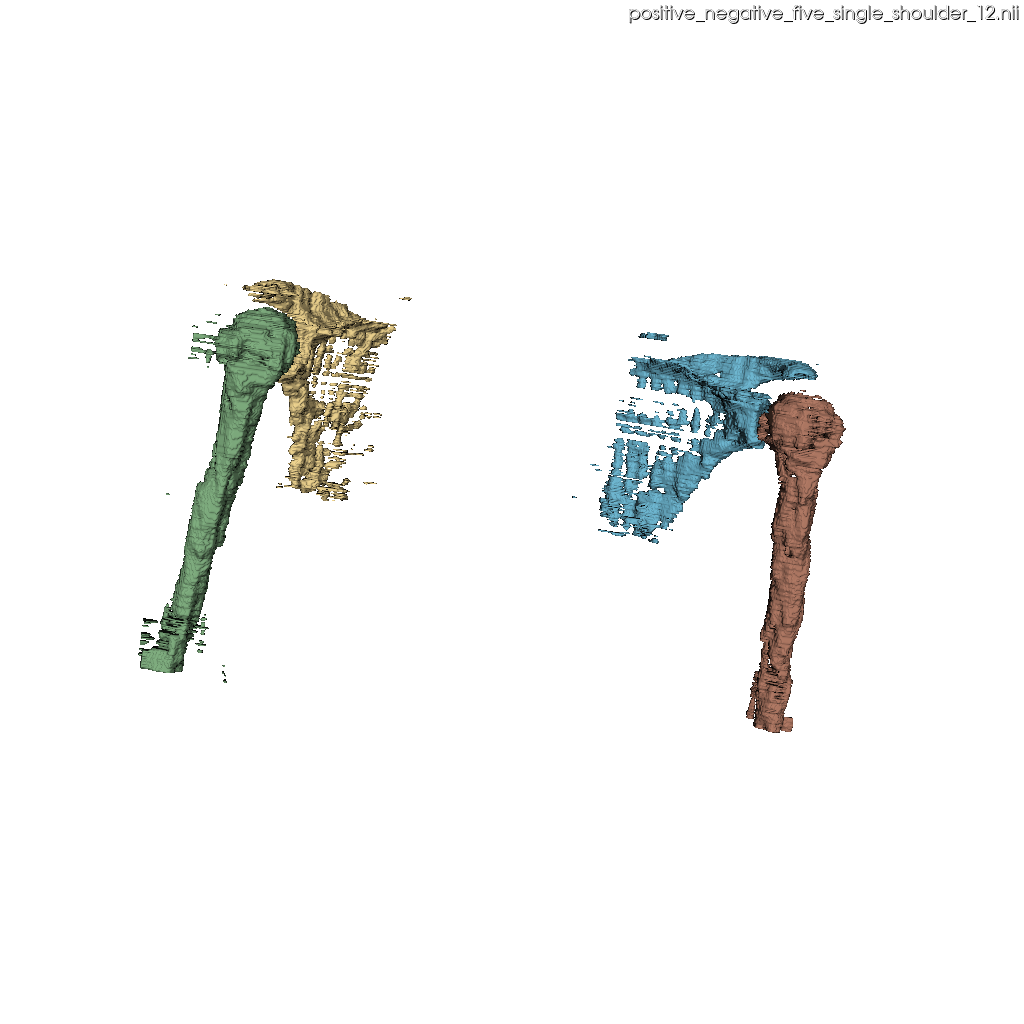} & \raisebox{-0.6\height}[0pt][0pt]{\includegraphics[width=0.11\linewidth, trim=400 150 400 100, clip]{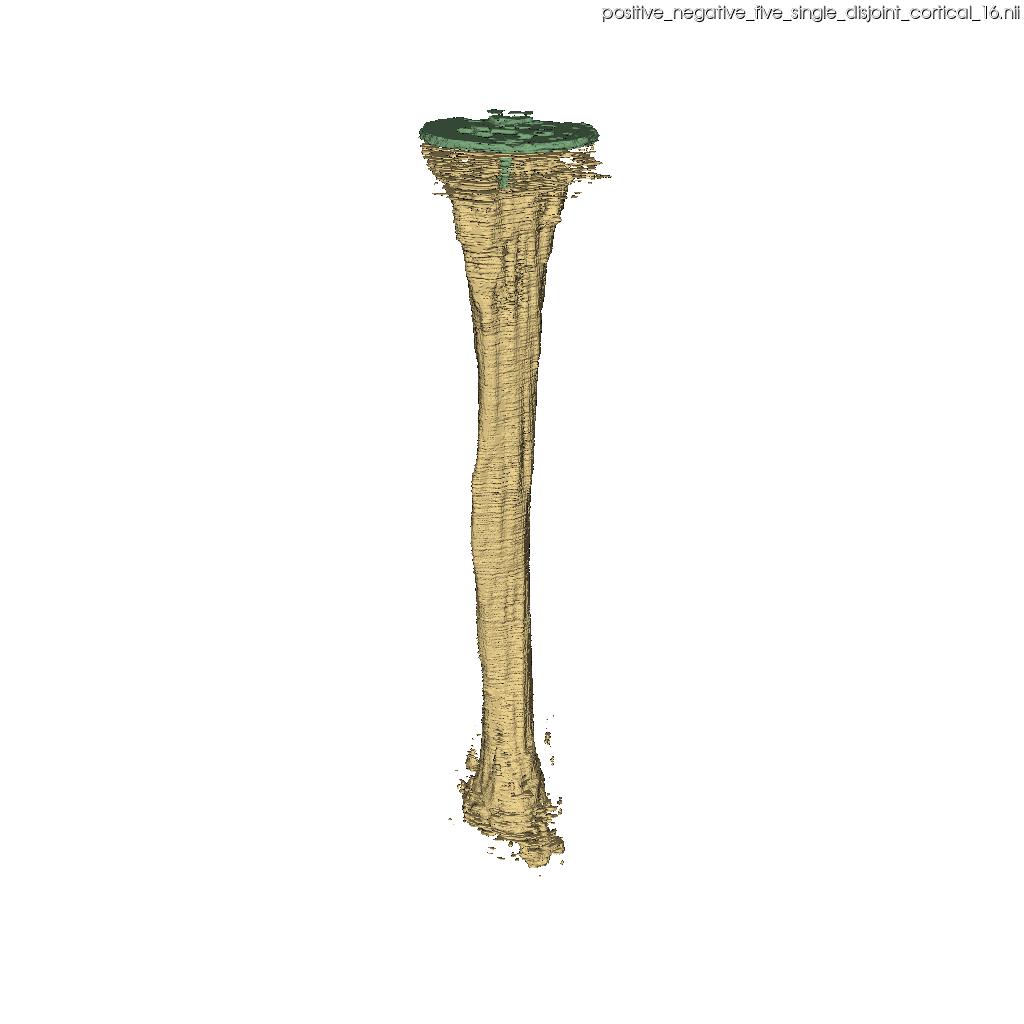}} 
    \\
    & \includegraphics[width=0.14\linewidth, trim=370 290 370 280, clip]{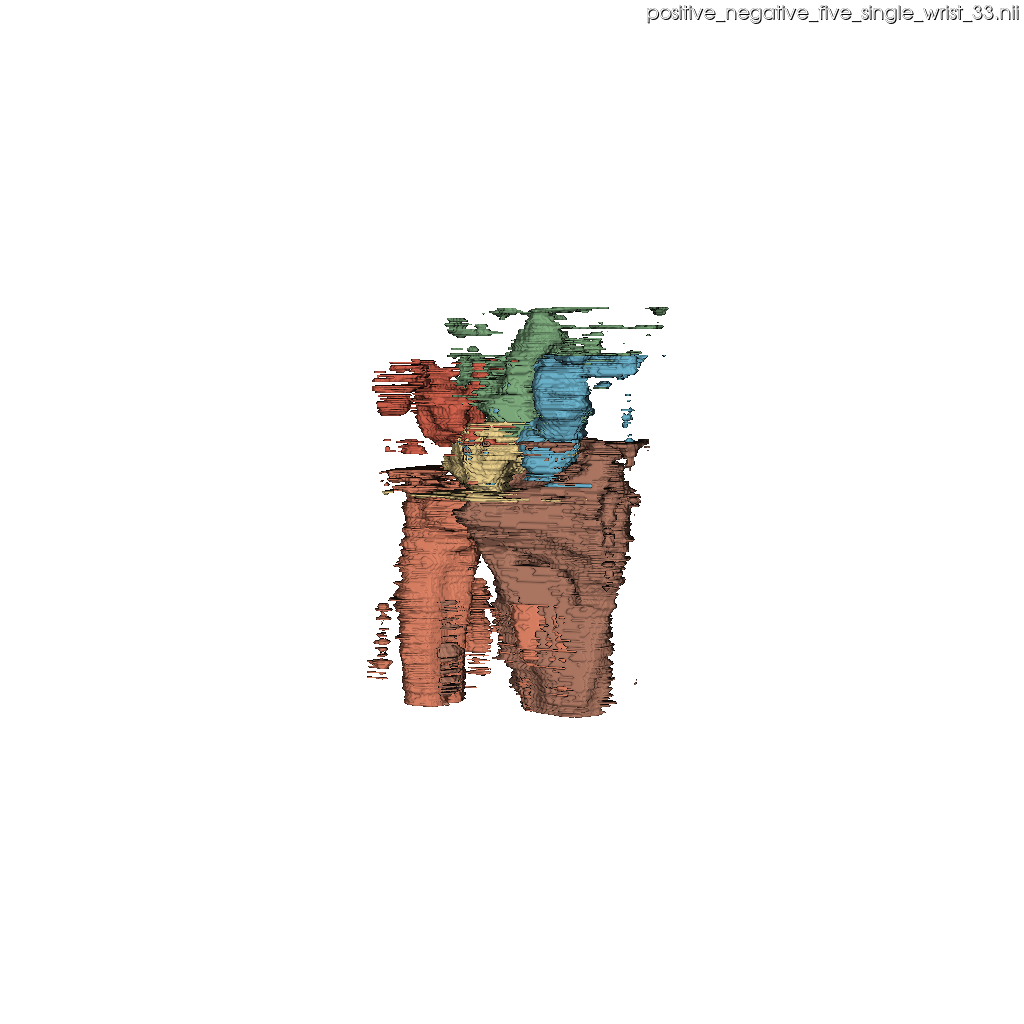} & &
    \includegraphics[width=0.14\linewidth, trim=370 290 370 280, clip]{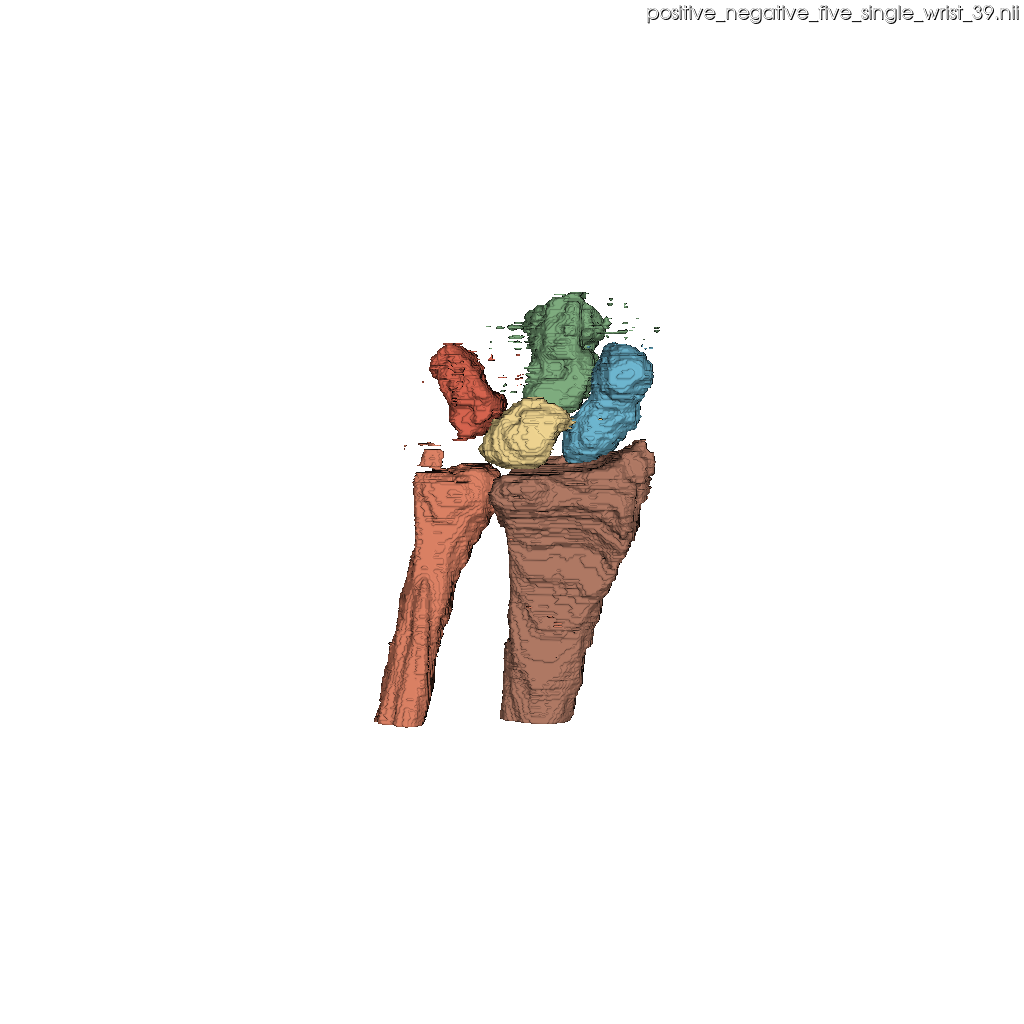} & &
    \includegraphics[width=0.14\linewidth, trim=370 290 370 280, clip]{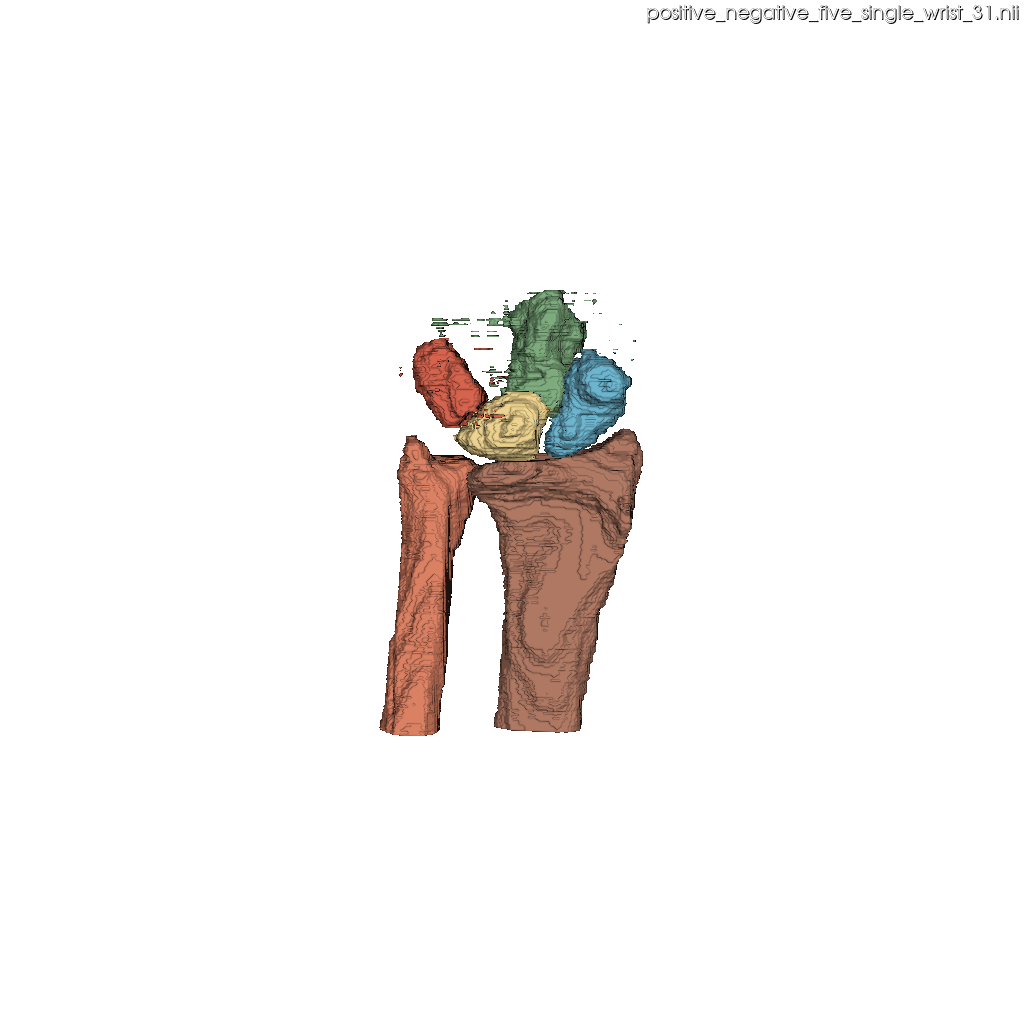} & \\ 
    \hline
\end{tabular}
\caption{Selected examples for \textit{\textsc{Sam}-Med2d} with low, medium and high DSC.}
\label{fig:example_sammed2d}
\end{figure}

\begin{figure}[H]
\begin{tabular}{|l|cc|cc|cc|}
    \hline
    Setting & \multicolumn{2}{c|}{DSC $\downarrow$} & \multicolumn{2}{c|}{DSC median} & \multicolumn{2}{c|}{DSC $\uparrow$} \\  
    \hline
    \multirow{2}{*}{\cblacksquare[0.6]{green}} & \includegraphics[width=0.14\linewidth, trim=580 380 100 320, clip]{examples/sam_b_bad/bbox_multiple_shoulder_13.png} & \raisebox{-0.6\height}[0pt][0pt]{\includegraphics[width=0.11\linewidth, trim=400 200 430 150, clip]{examples/sam_b_bad/bbox_multiple_disjoint_cortical_3.png}} &
    \includegraphics[width=0.14\linewidth, trim=560 380 190 330, clip]{examples/sam_b_medium/bbox_multiple_shoulder_8.png} & \raisebox{-0.6\height}[0pt][0pt]{\includegraphics[width=0.11\linewidth, trim=400 200 410 120, clip]{examples/sam_b_medium/bbox_multiple_disjoint_cortical_2.png}} &
    \includegraphics[width=0.14\linewidth, trim=560 380 130 320, clip]{examples/sam_b_good/bbox_multiple_shoulder_5.png} & \raisebox{-0.6\height}[0pt][0pt]{\includegraphics[width=0.11\linewidth, trim=400 170 410 120, clip]{examples/sam_b_good/bbox_multiple_disjoint_cortical_13.png}} 
    \\
    & \includegraphics[width=0.14\linewidth, trim=370 290 370 280, clip]{examples/sam_b_bad/bbox_multiple_wrist_13.png} & &
    \includegraphics[width=0.13\linewidth, trim=370 310 370 250, clip]{examples/sam_b_medium/bbox_multiple_wrist_17.png} & &
    \includegraphics[width=0.14\linewidth, trim=370 290 370 280, clip]{examples/sam_b_good/bbox_multiple_wrist_31.png} & \\ 
    \hline
    \multirow{2}{*}{\cblacksquaredot[0.6]{green}} & \includegraphics[width=0.14\linewidth, trim=560 380 190 330, clip]{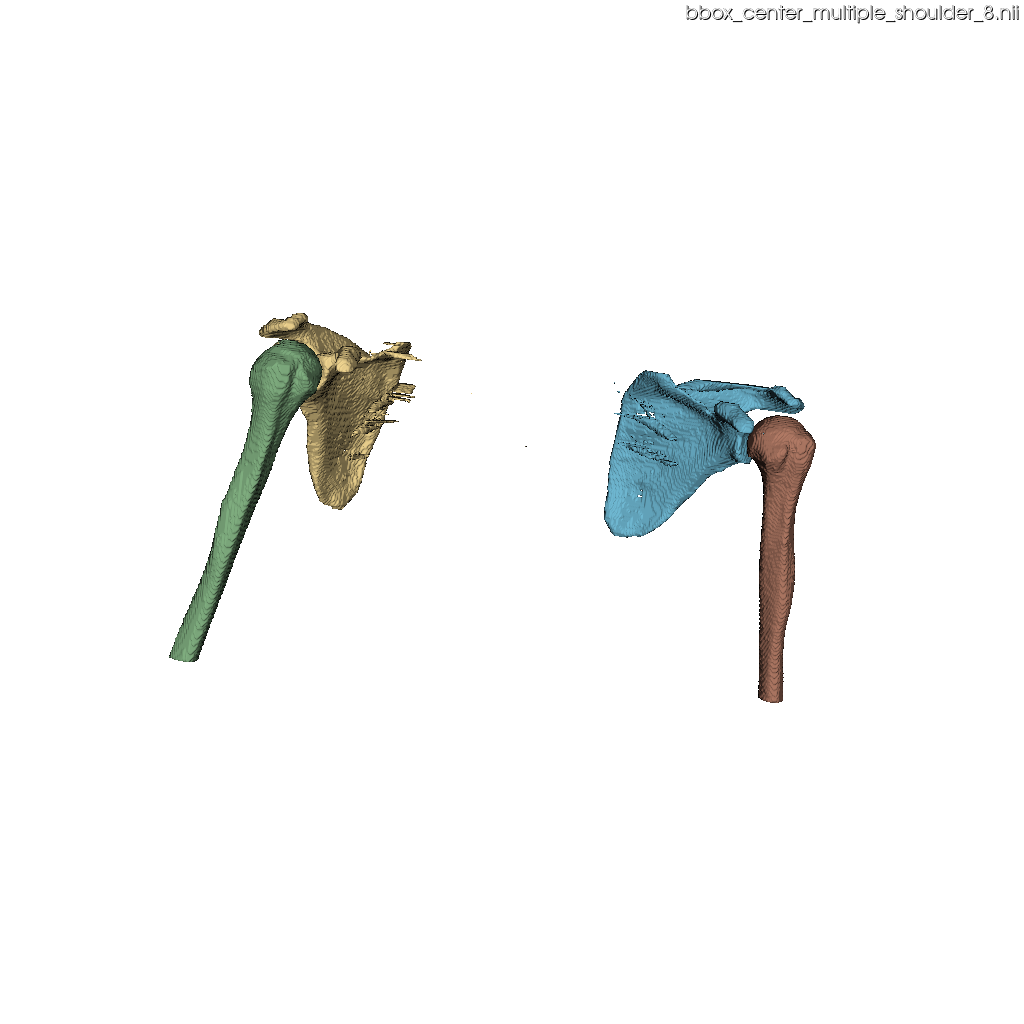} & \raisebox{-0.6\height}[0pt][0pt]{\includegraphics[width=0.11\linewidth, trim=400 200 430 150, clip]{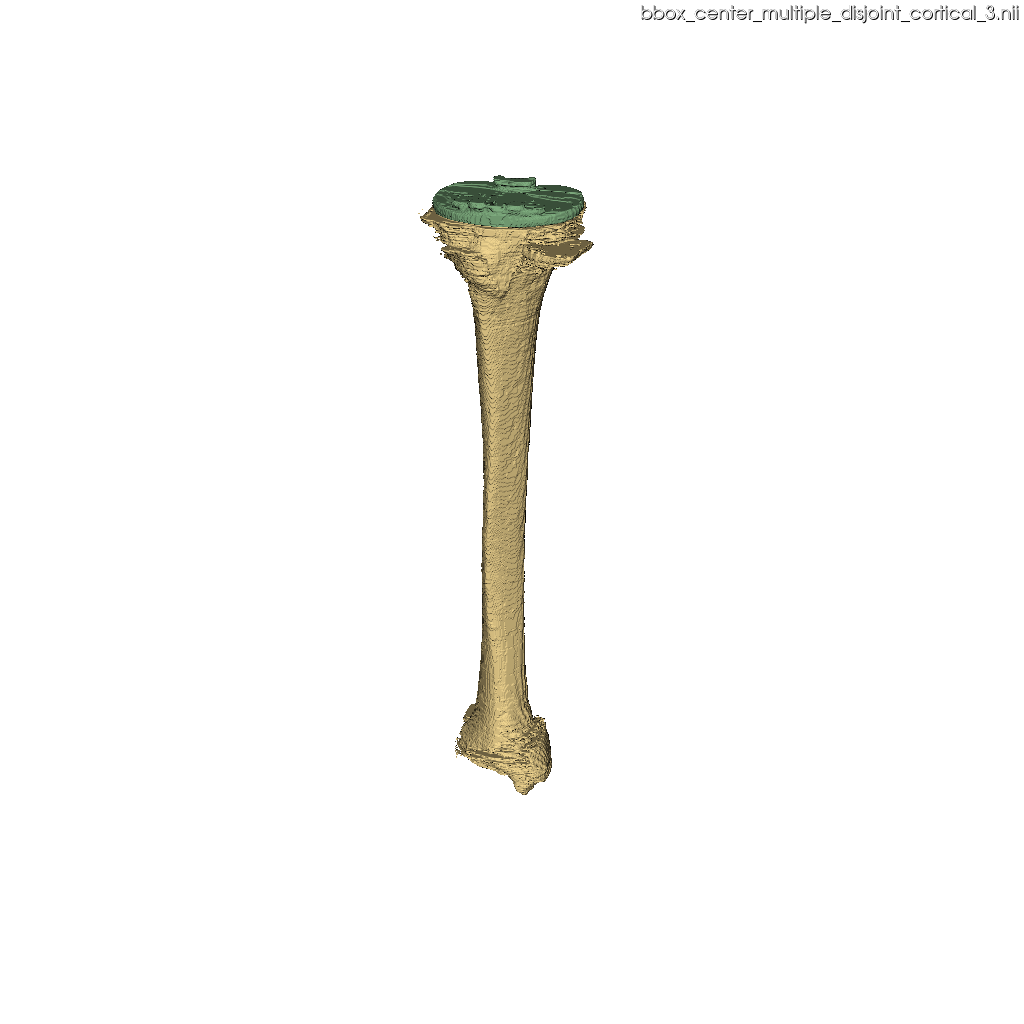}} &
    \includegraphics[width=0.14\linewidth, trim=570 380 110 310, clip]{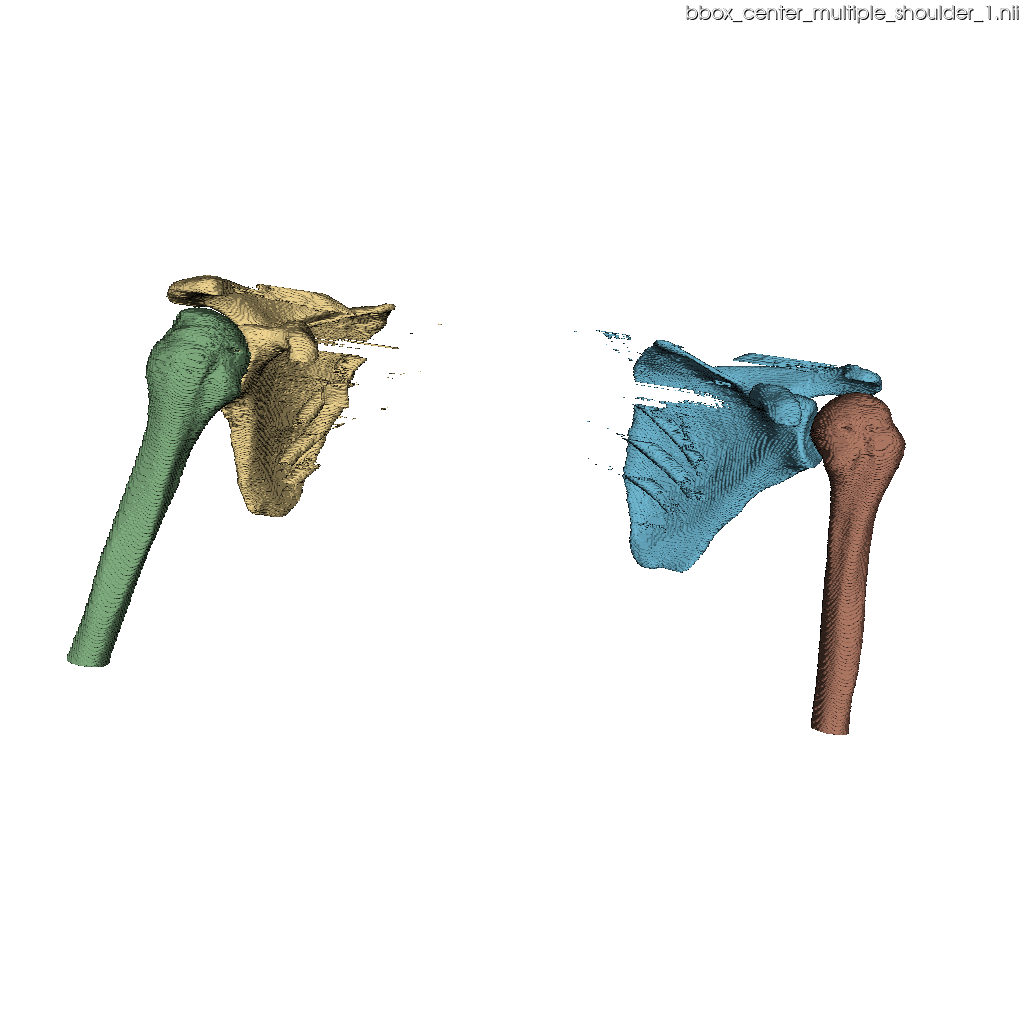} & \raisebox{-0.6\height}[0pt][0pt]{\includegraphics[width=0.11\linewidth, trim=400 200 400 120, clip]{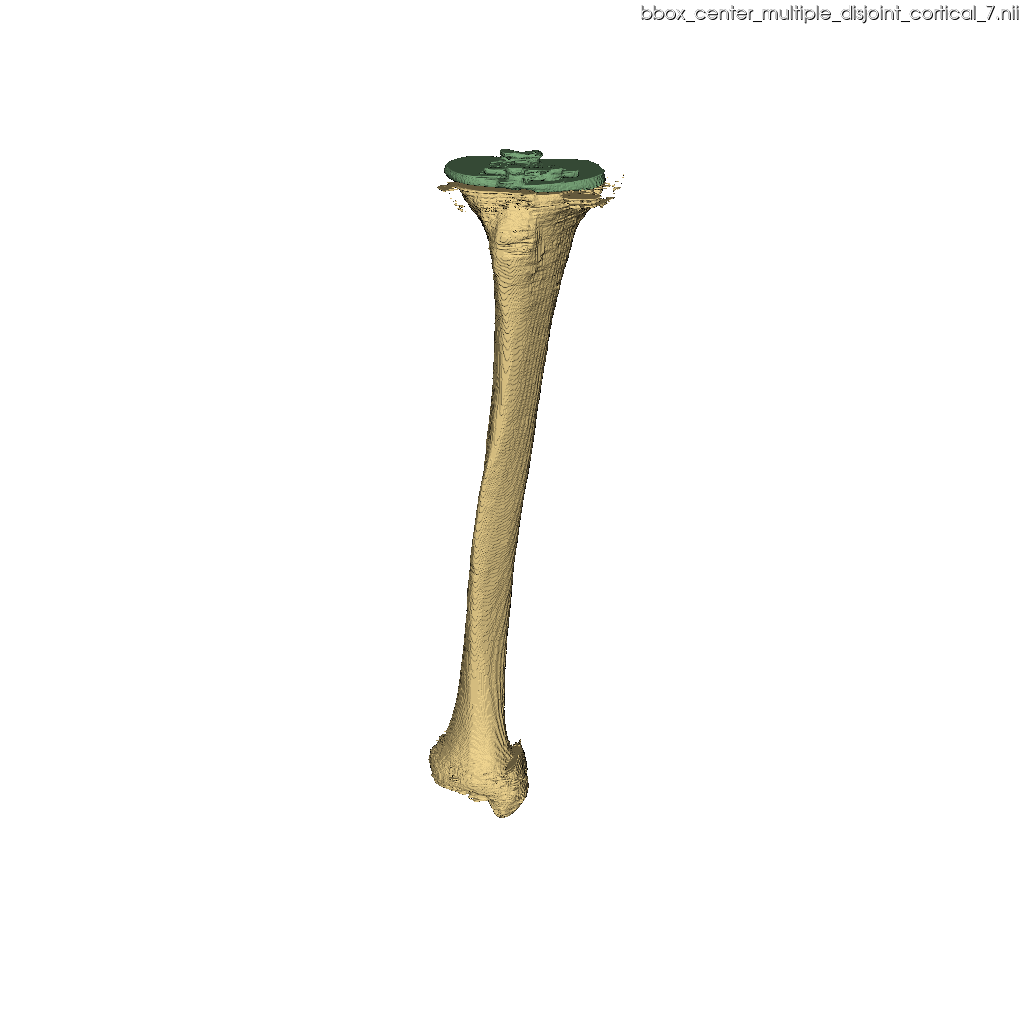}} &
    \includegraphics[width=0.14\linewidth, trim=570 380 170 330, clip]{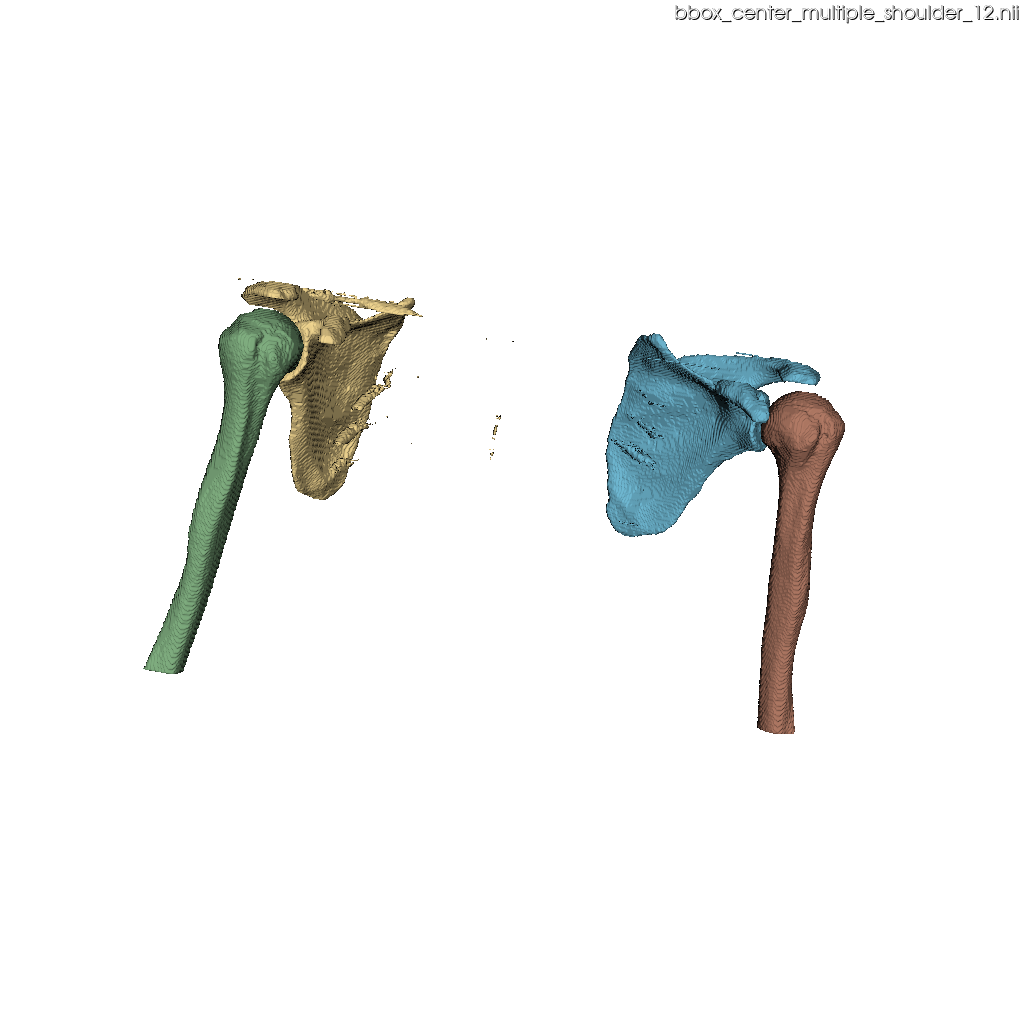} & \raisebox{-0.6\height}[0pt][0pt]{\includegraphics[width=0.11\linewidth, trim=400 170 410 120, clip]{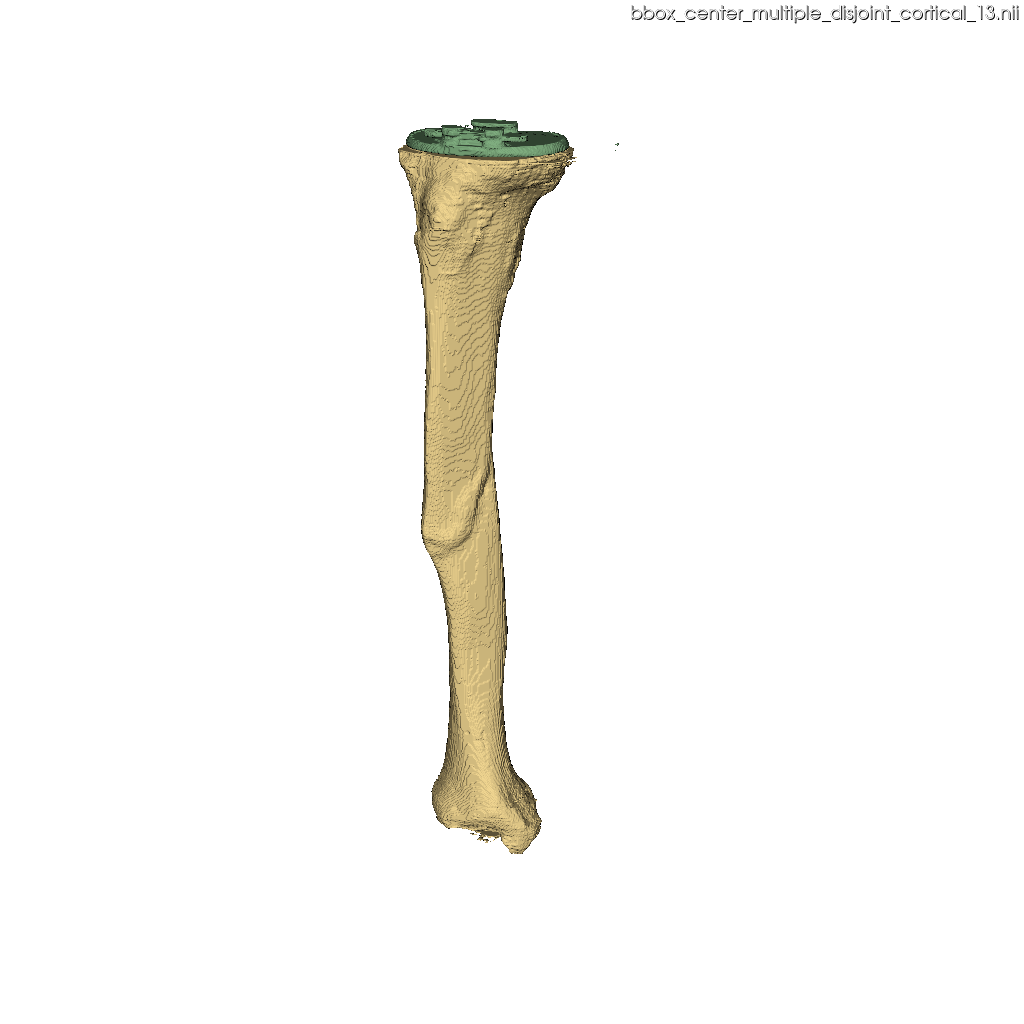}} 
    \\
    & \includegraphics[width=0.14\linewidth, trim=370 310 370 250, clip]{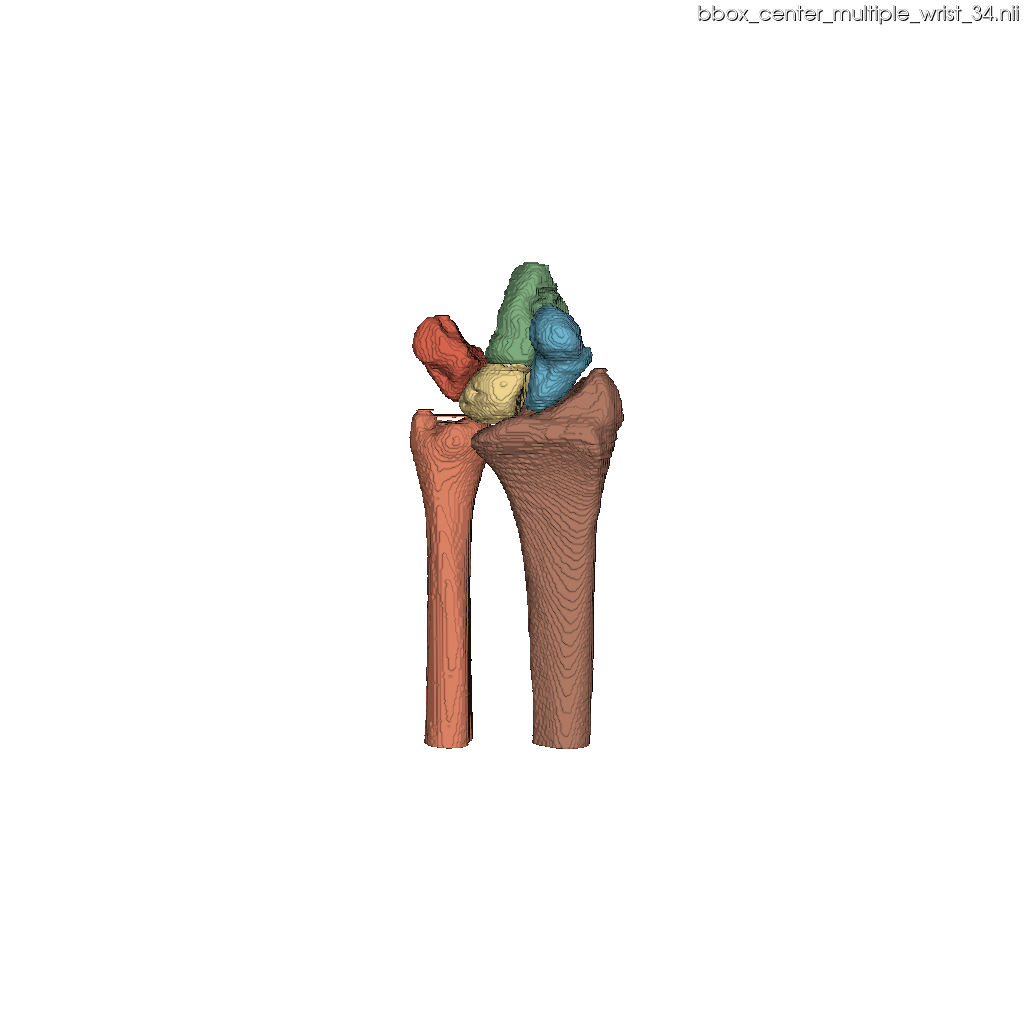} & &
    \includegraphics[width=0.14\linewidth, trim=370 290 370 280, clip]{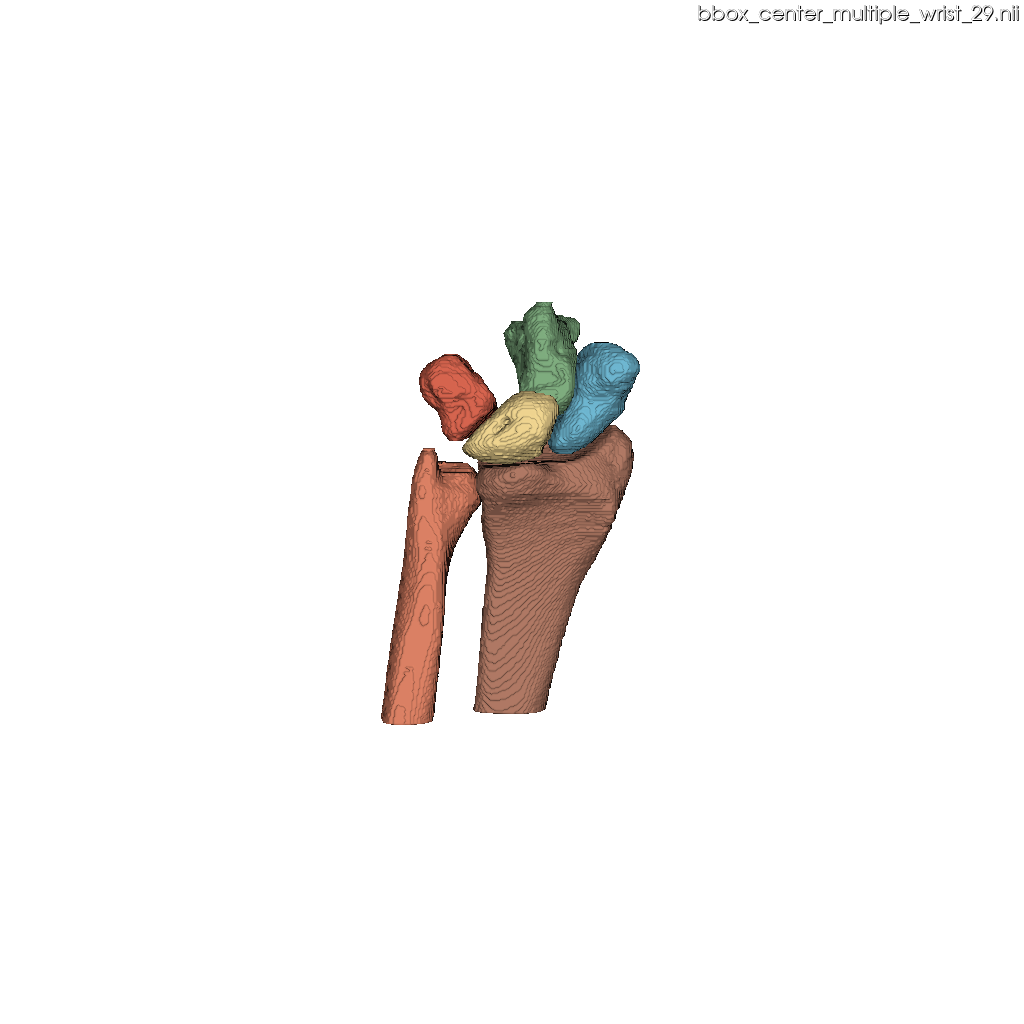} & &
    \includegraphics[width=0.14\linewidth, trim=370 290 370 280, clip]{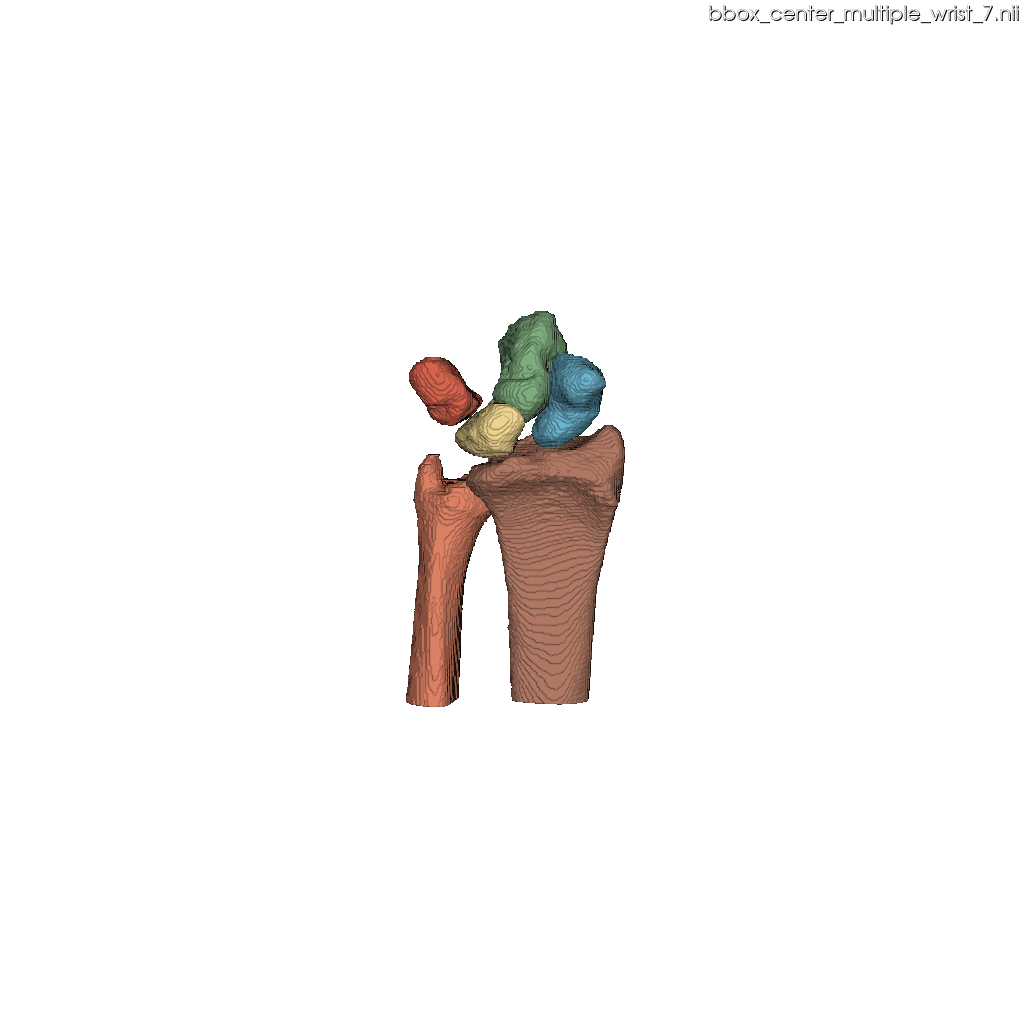} & \\
    \hline
    \multirow{2}{*}{\cblackstartriangledown[0.6]{green}} & \includegraphics[width=0.14\linewidth, trim=560 380 190 330, clip]{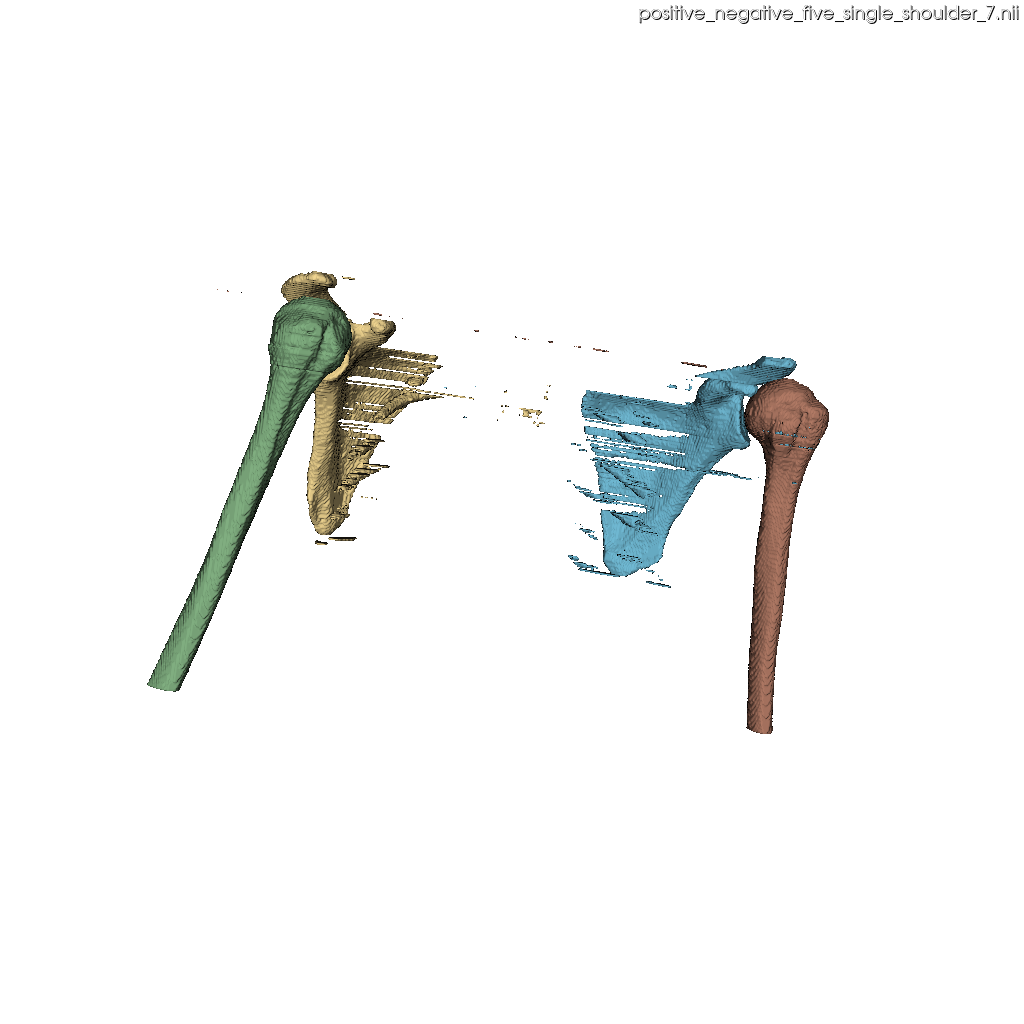} & \raisebox{-0.6\height}[0pt][0pt]{\includegraphics[width=0.11\linewidth, trim=310 150 400 60, clip]{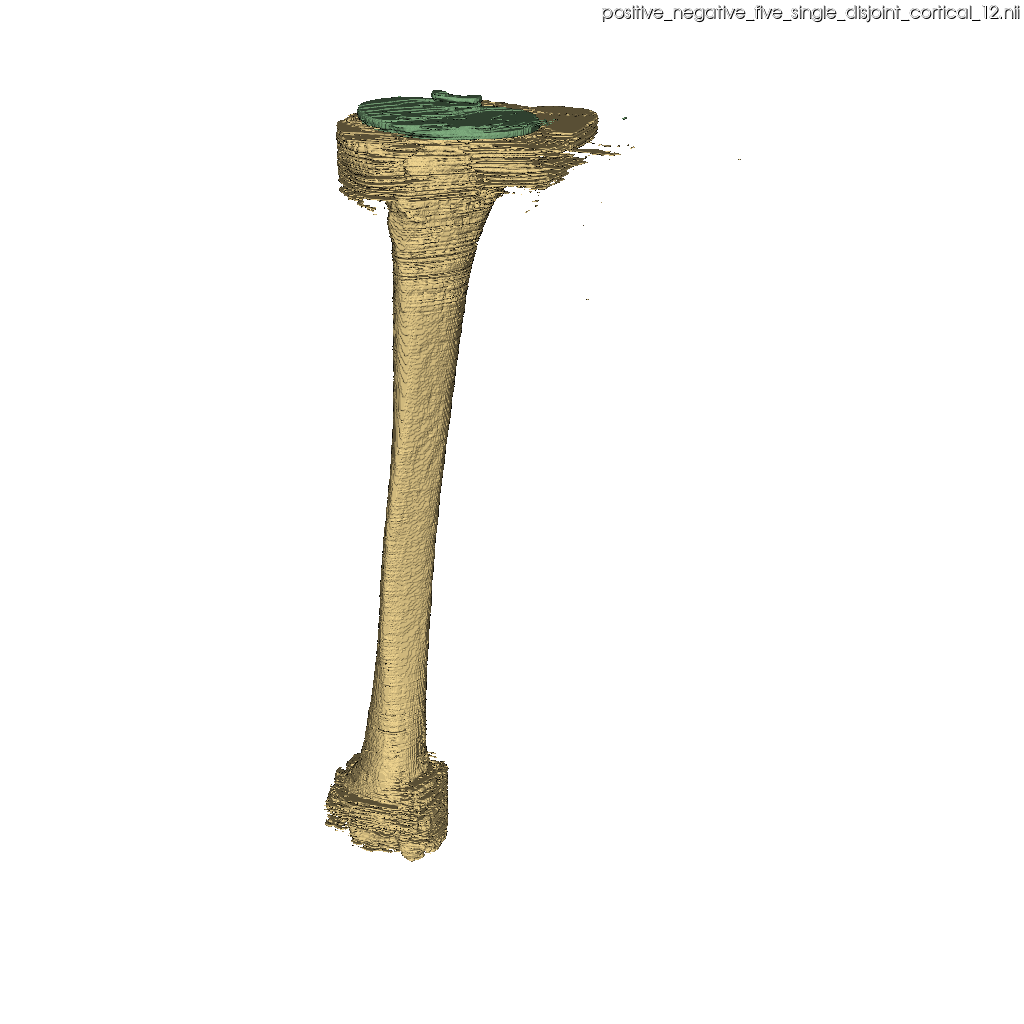}} &
    \includegraphics[width=0.14\linewidth, trim=580 380 100 350, clip]{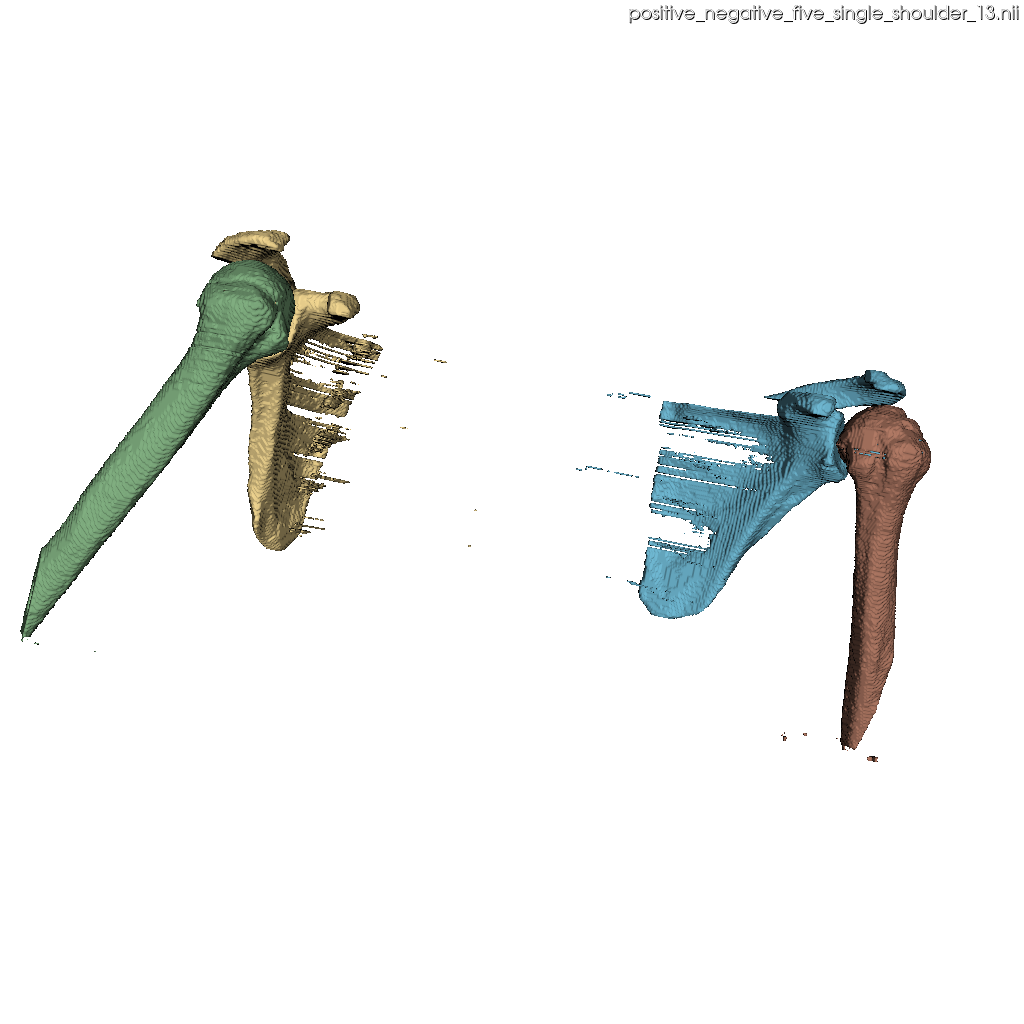} & \raisebox{-0.6\height}[0pt][0pt]{\includegraphics[width=0.11\linewidth, trim=420 170 380 100, clip]{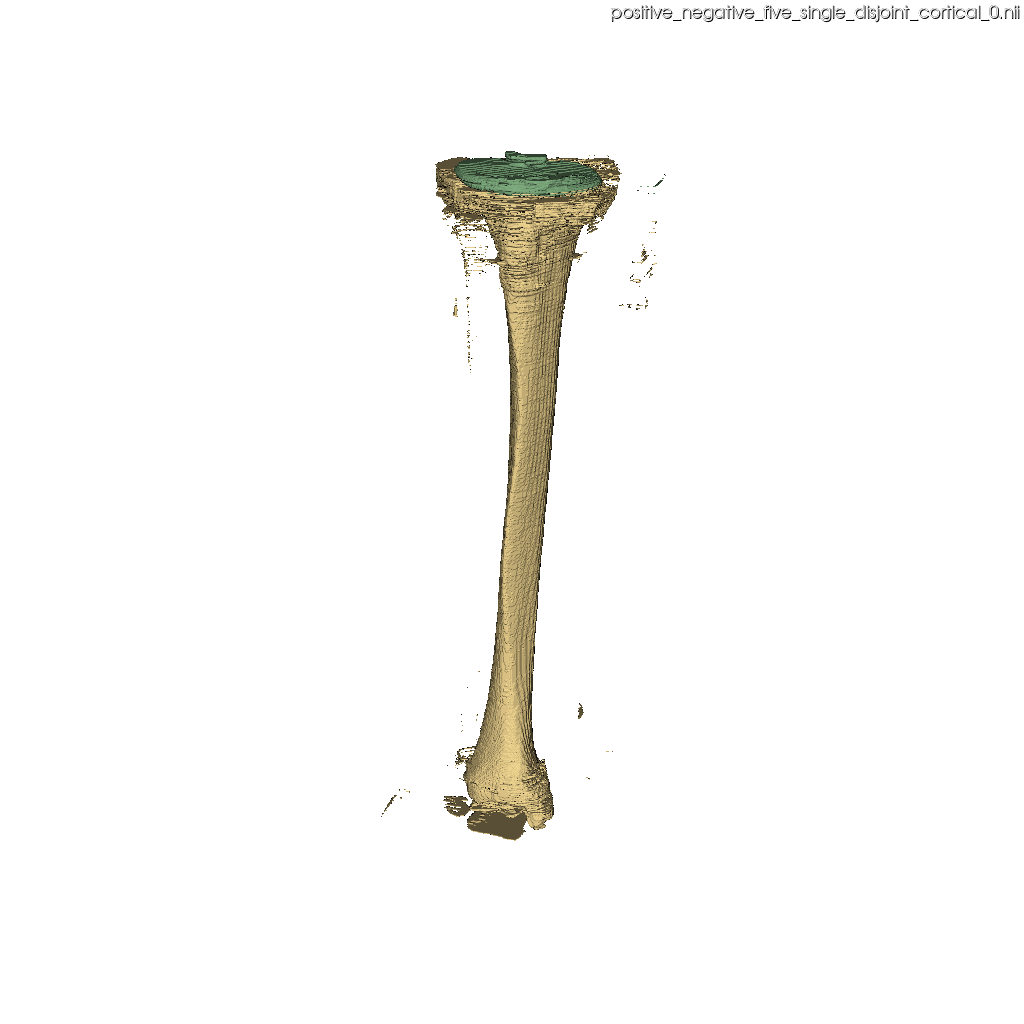}} &
    \includegraphics[width=0.14\linewidth, trim=560 380 120 310, clip]{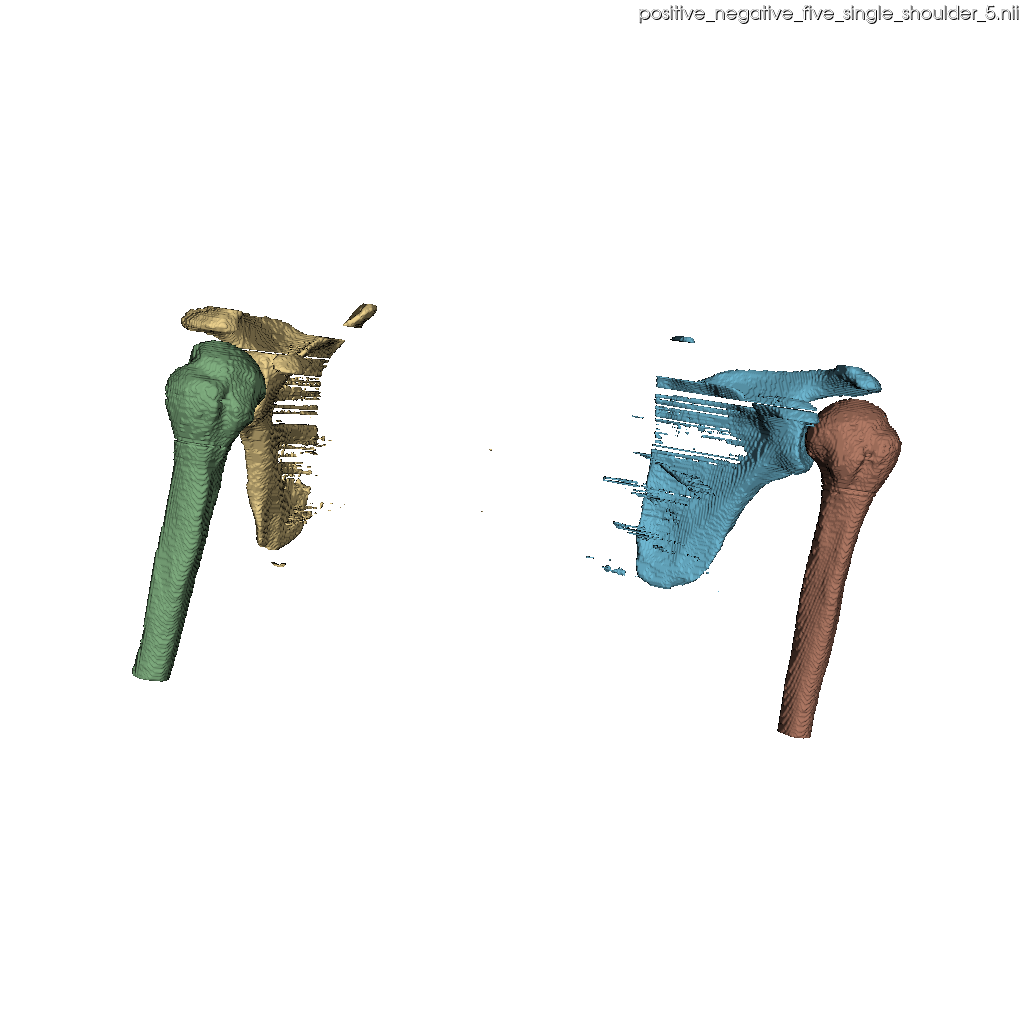} & \raisebox{-0.6\height}[0pt][0pt]{\includegraphics[width=0.11\linewidth, trim=330 190 400 120, clip]{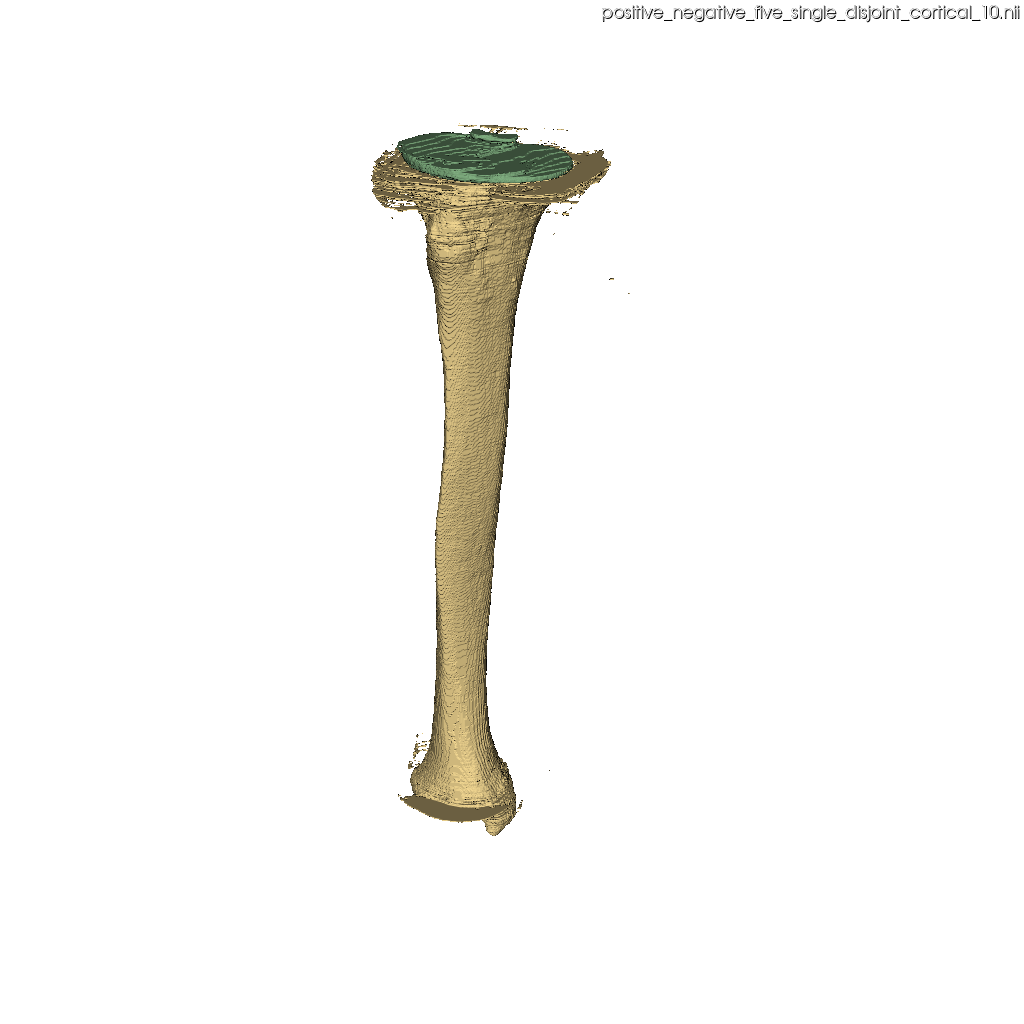}}
    \\
    & \includegraphics[width=0.14\linewidth, trim=370 290 370 280, clip]{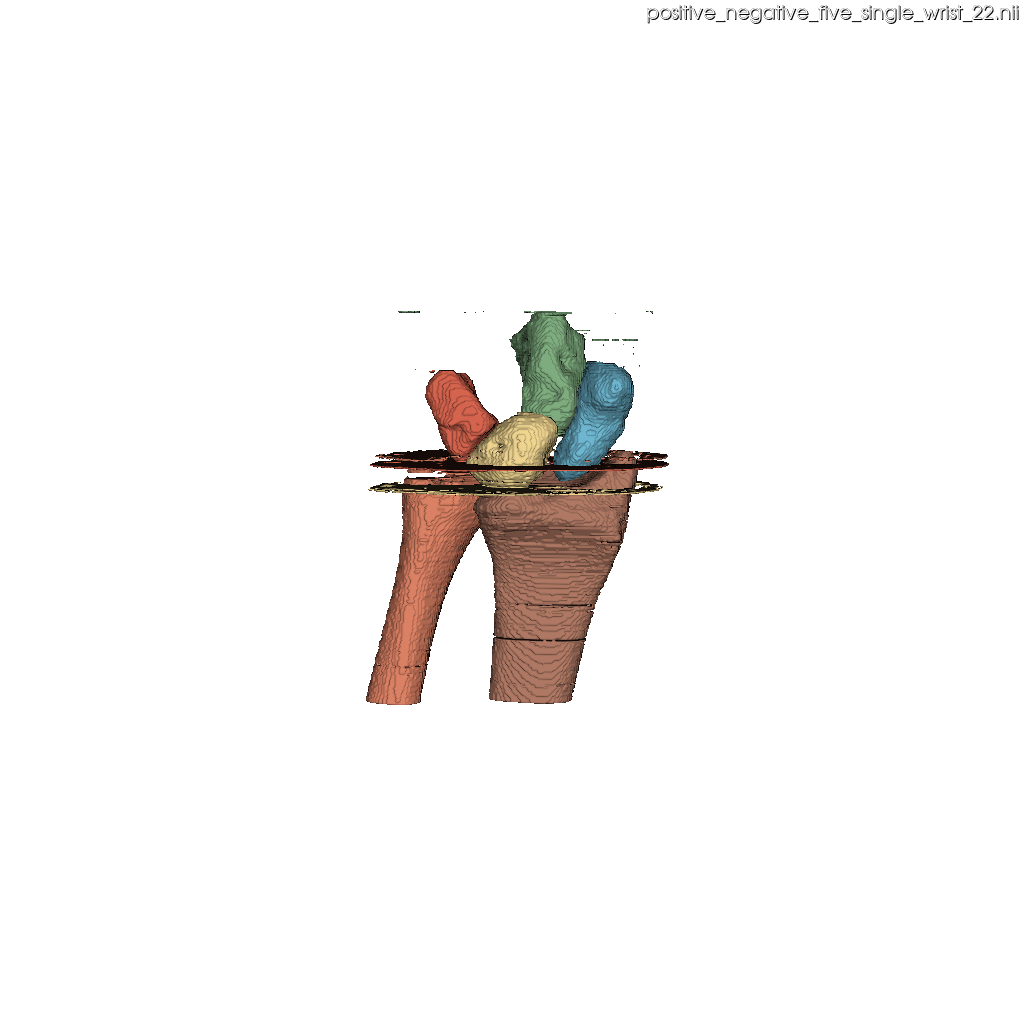} & &
    \includegraphics[width=0.14\linewidth, trim=370 290 370 280, clip]{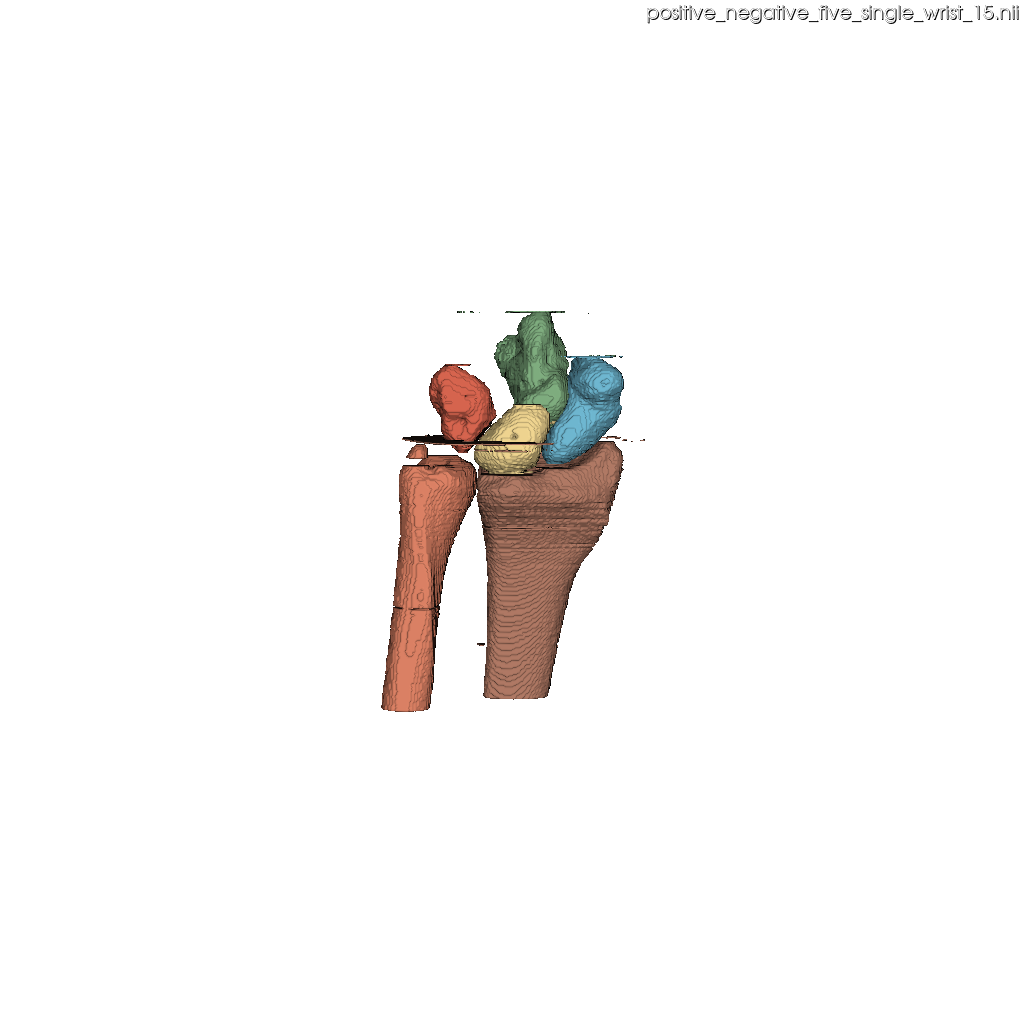} & &
    \includegraphics[width=0.14\linewidth, trim=370 290 380 280, clip]{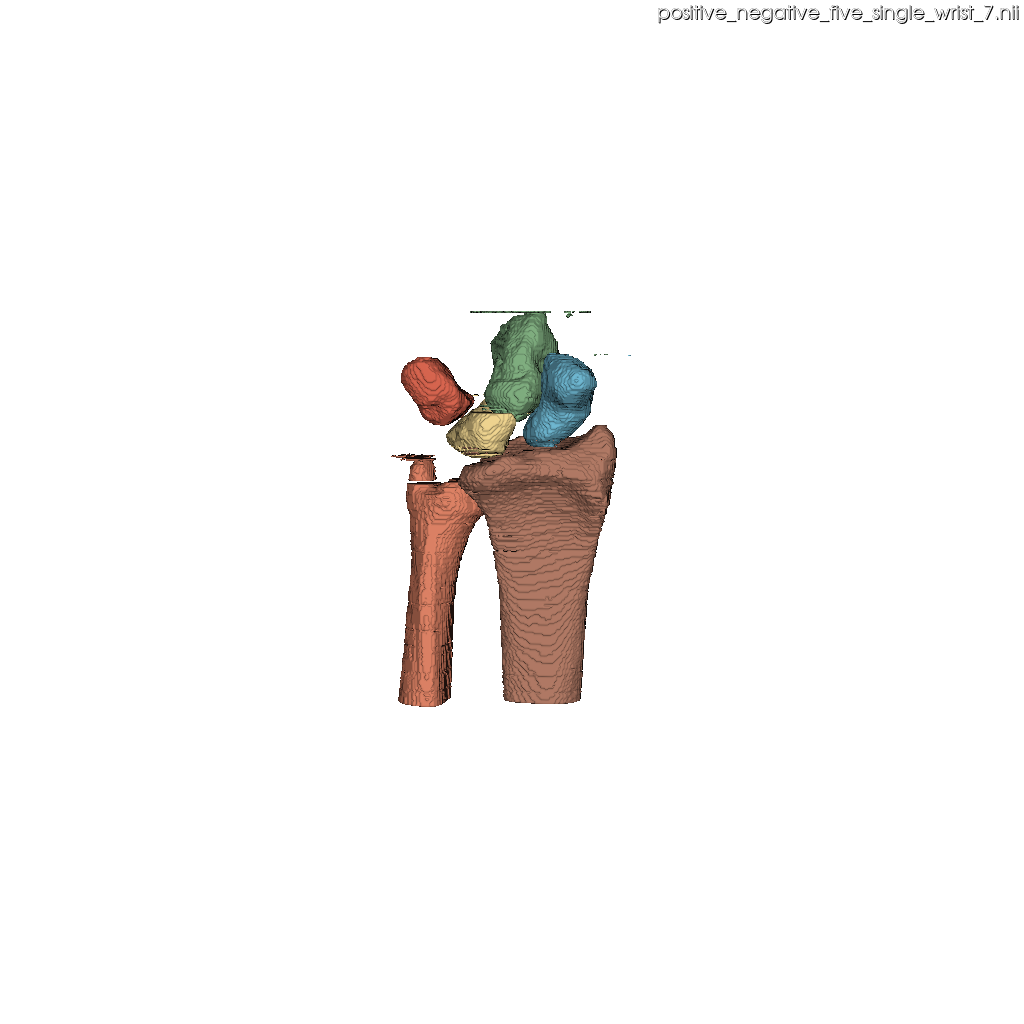} & \\  
    \hline
\end{tabular}
\caption{Selected examples for \textit{\textsc{Sam} B} with low, medium and high DSC.}
\label{fig:example_samb}
\end{figure}

\begin{figure}[H]
\begin{tabular}{|l|cc|cc|cc|}
    \hline
    Setting & \multicolumn{2}{c|}{DSC $\downarrow$} & \multicolumn{2}{c|}{DSC median} & \multicolumn{2}{c|}{DSC $\uparrow$} \\ 
    \hline
    \multirow{2}{*}{\cblacksquare[0.6]{oranje}} & \includegraphics[width=0.14\linewidth, trim=610 370 90 330, clip]{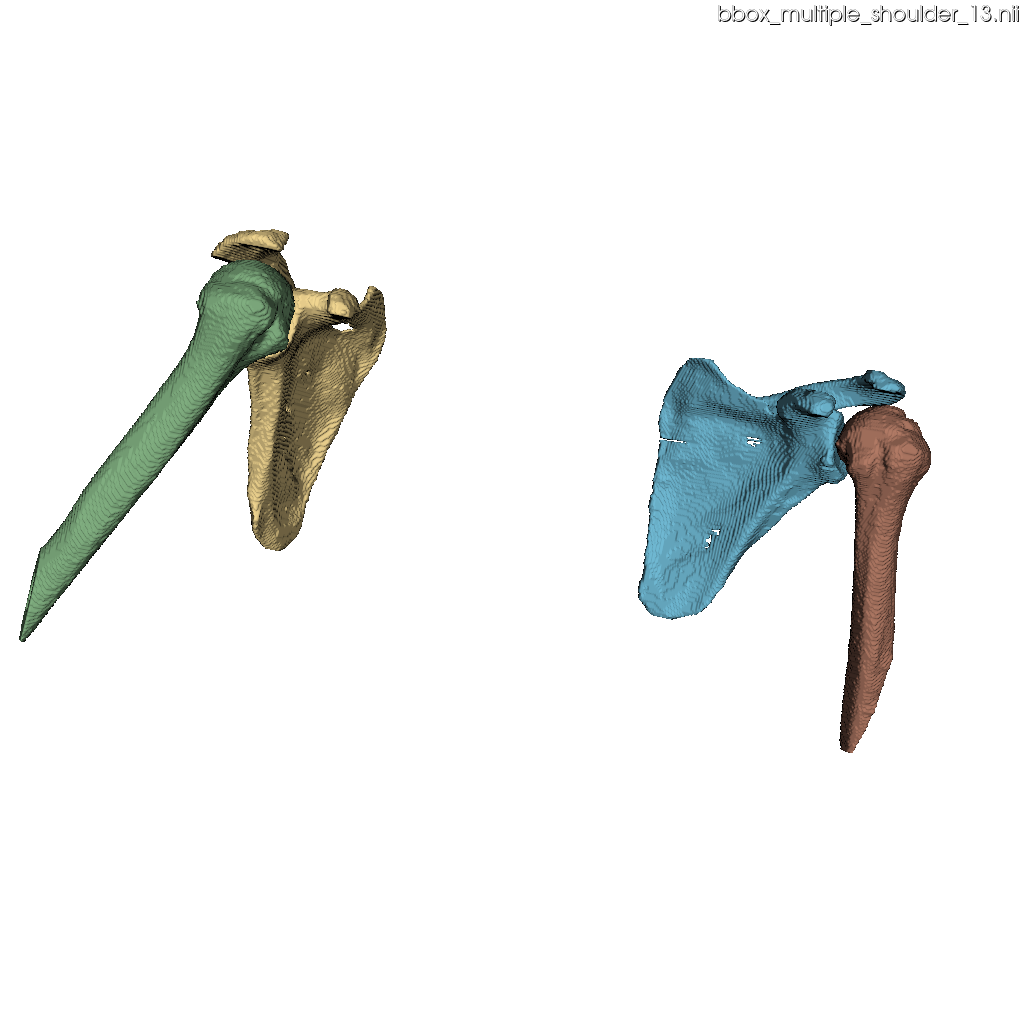} & \raisebox{-0.6\height}[0pt][0pt]{\includegraphics[width=0.11\linewidth, trim=420 200 430 150, clip]{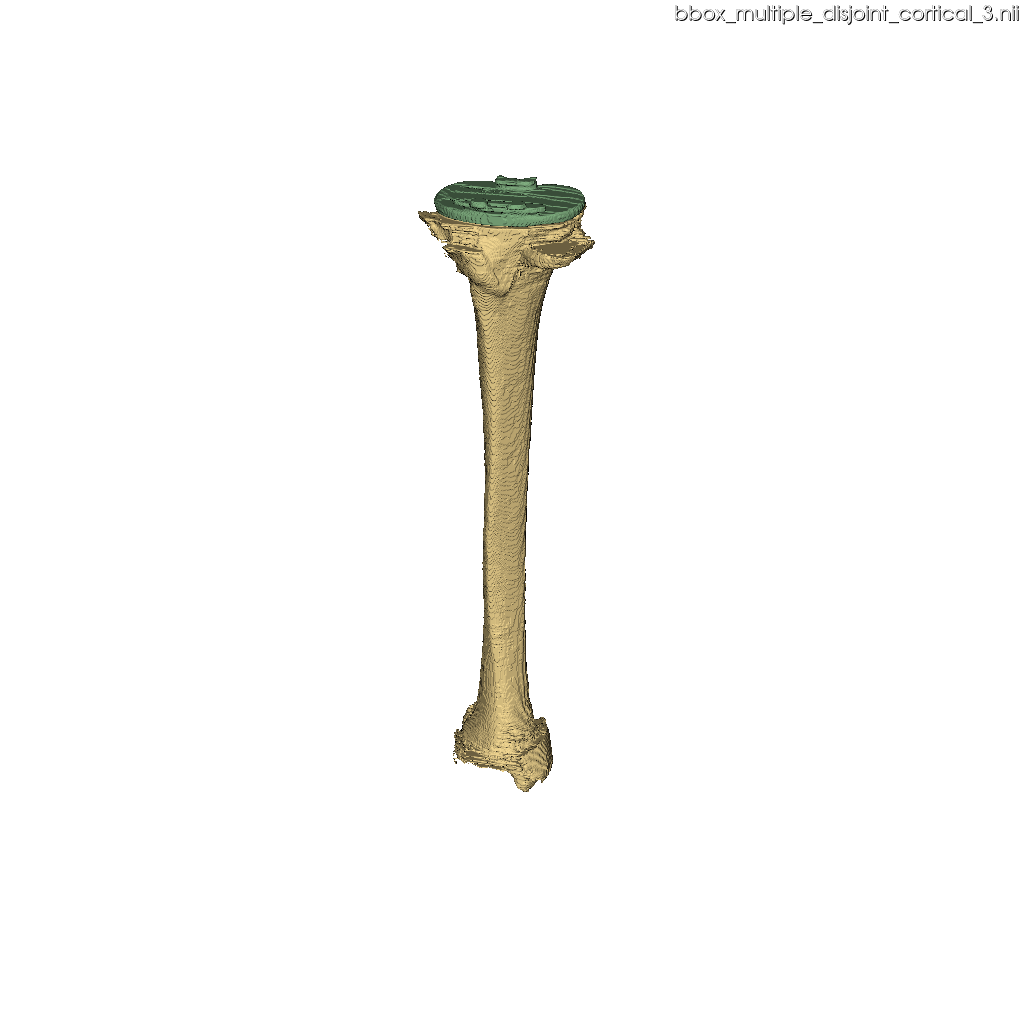}} &
    \includegraphics[width=0.14\linewidth, trim=560 380 190 320, clip]{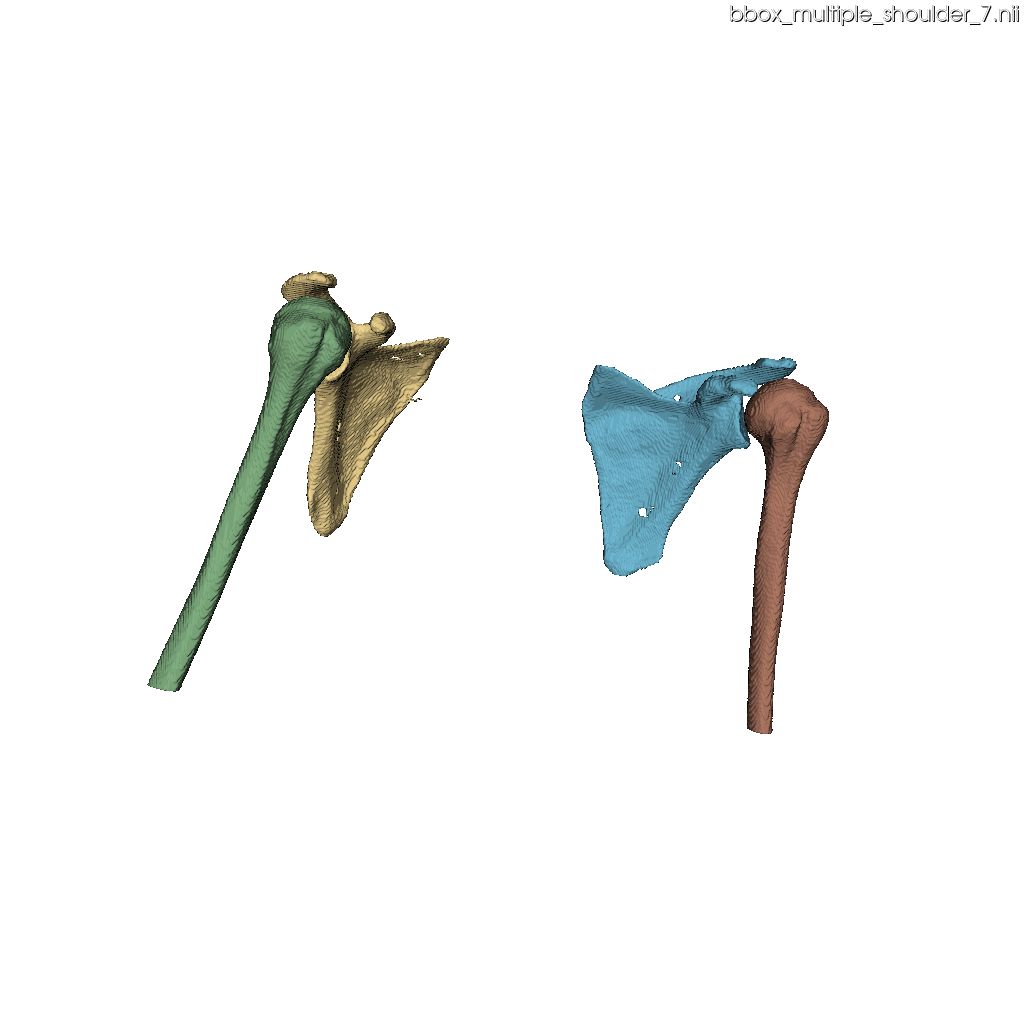} & \raisebox{-0.6\height}[0pt][0pt]{\includegraphics[width=0.11\linewidth, trim=420 200 410 150, clip]{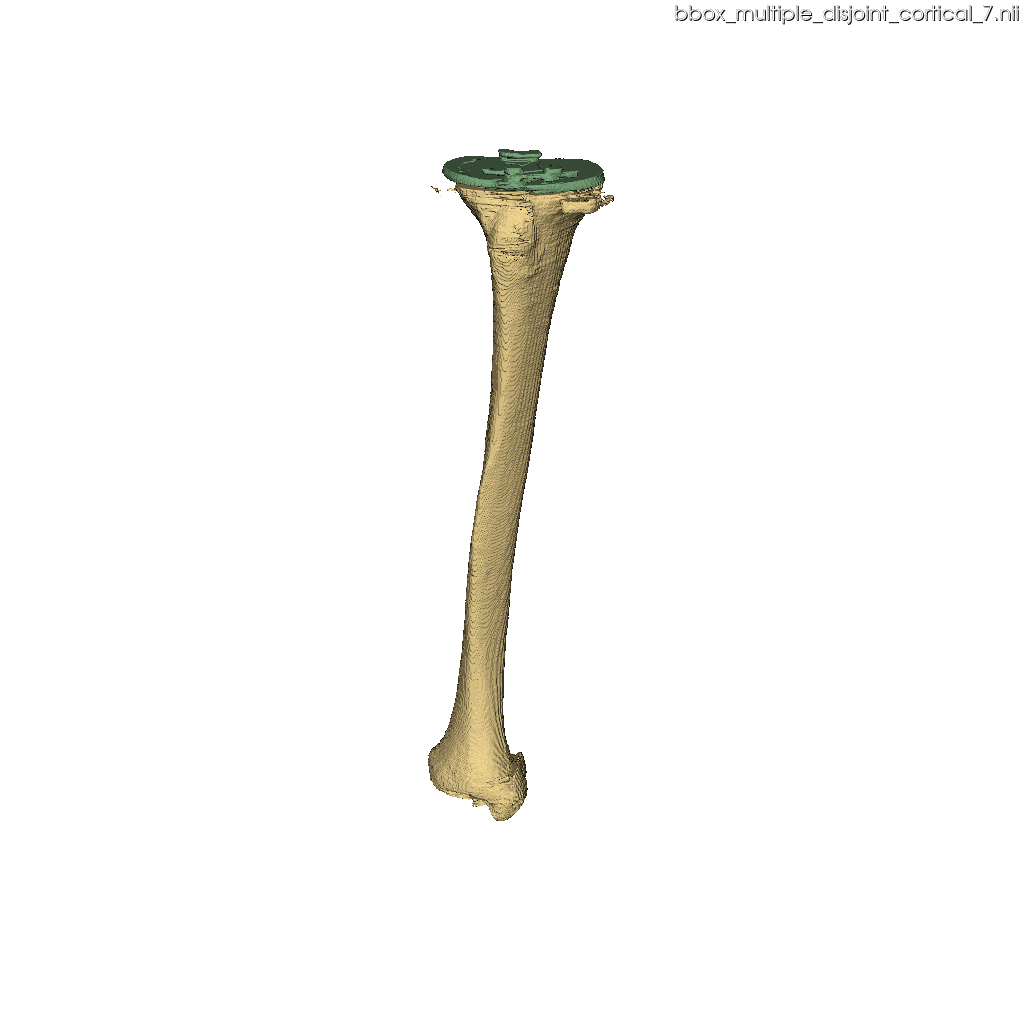}} &
    \includegraphics[width=0.14\linewidth, trim=610 380 140 370, clip]{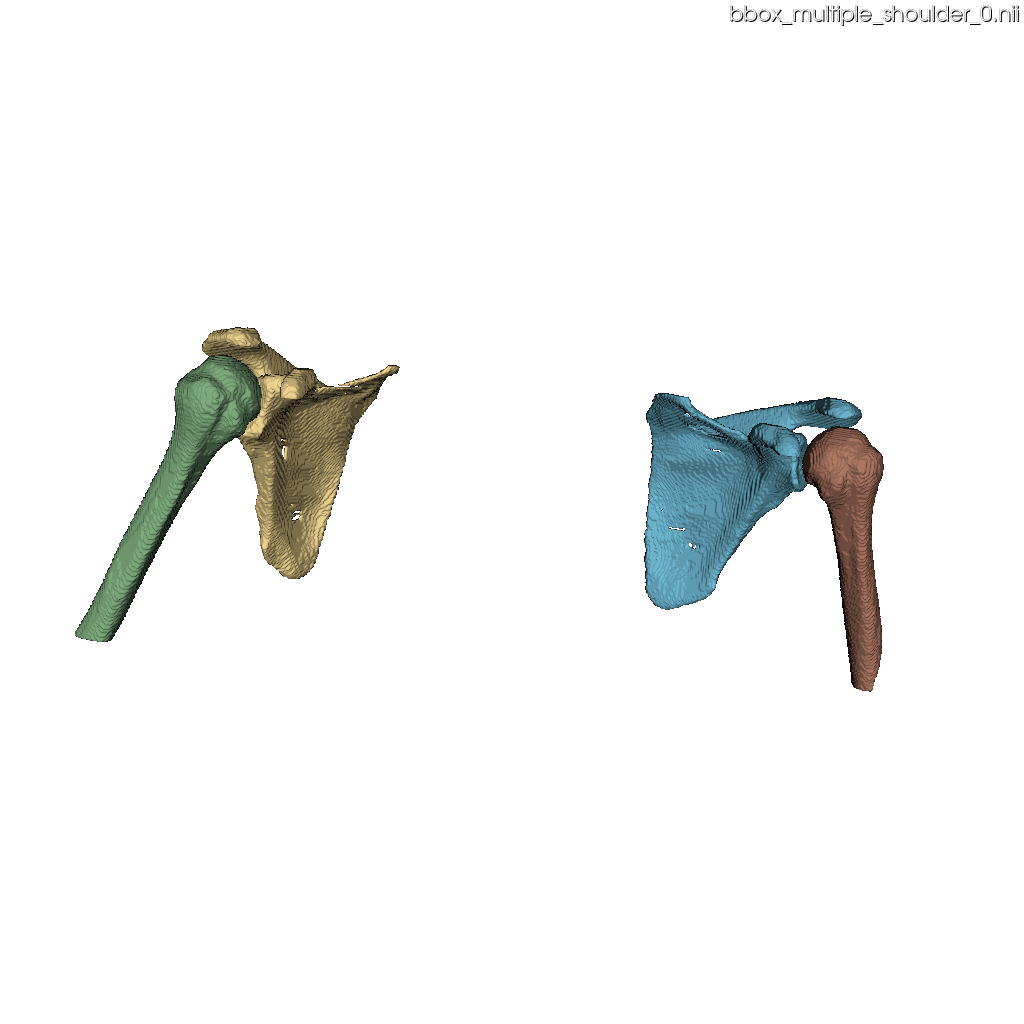} & \raisebox{-0.6\height}[0pt][0pt]{\includegraphics[width=0.1\linewidth, trim=430 100 410 60, clip]{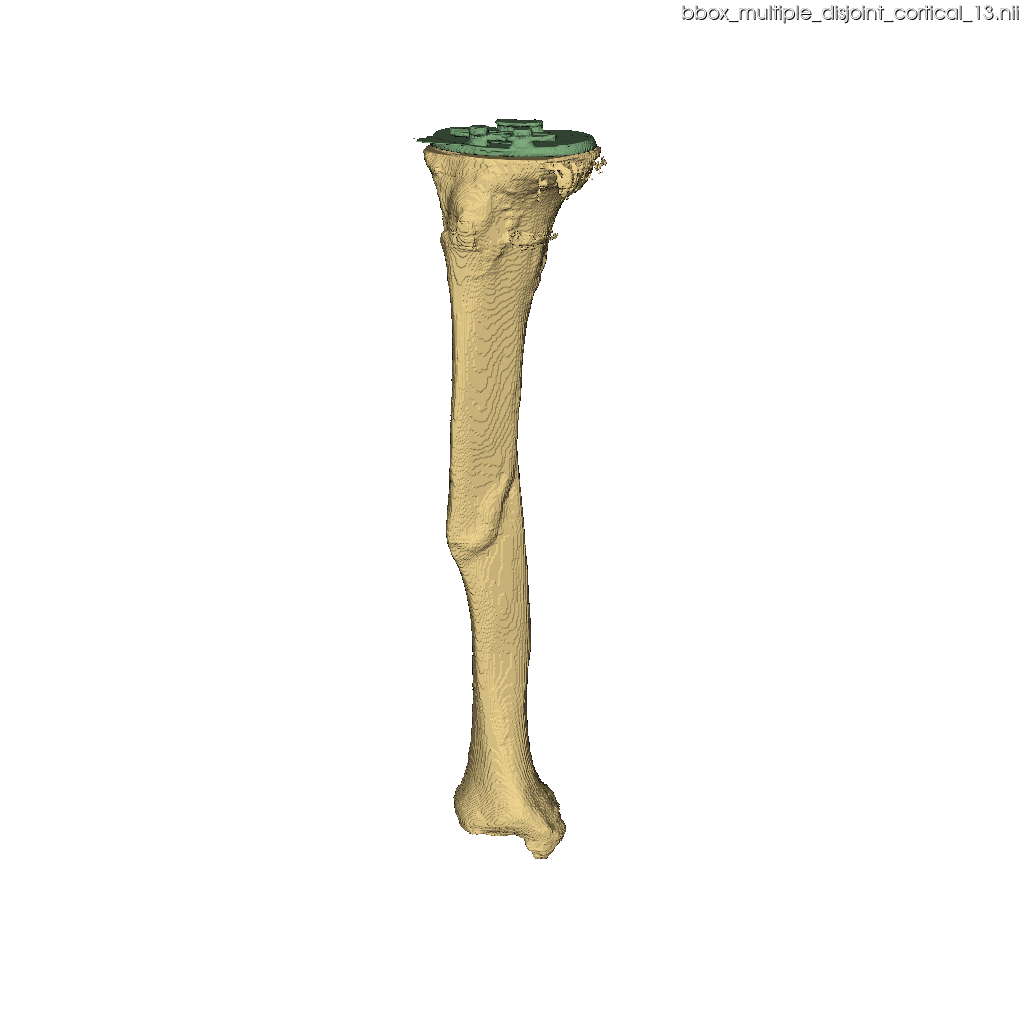}} 
    \\
    & \includegraphics[width=0.14\linewidth, trim=370 290 370 250, clip]{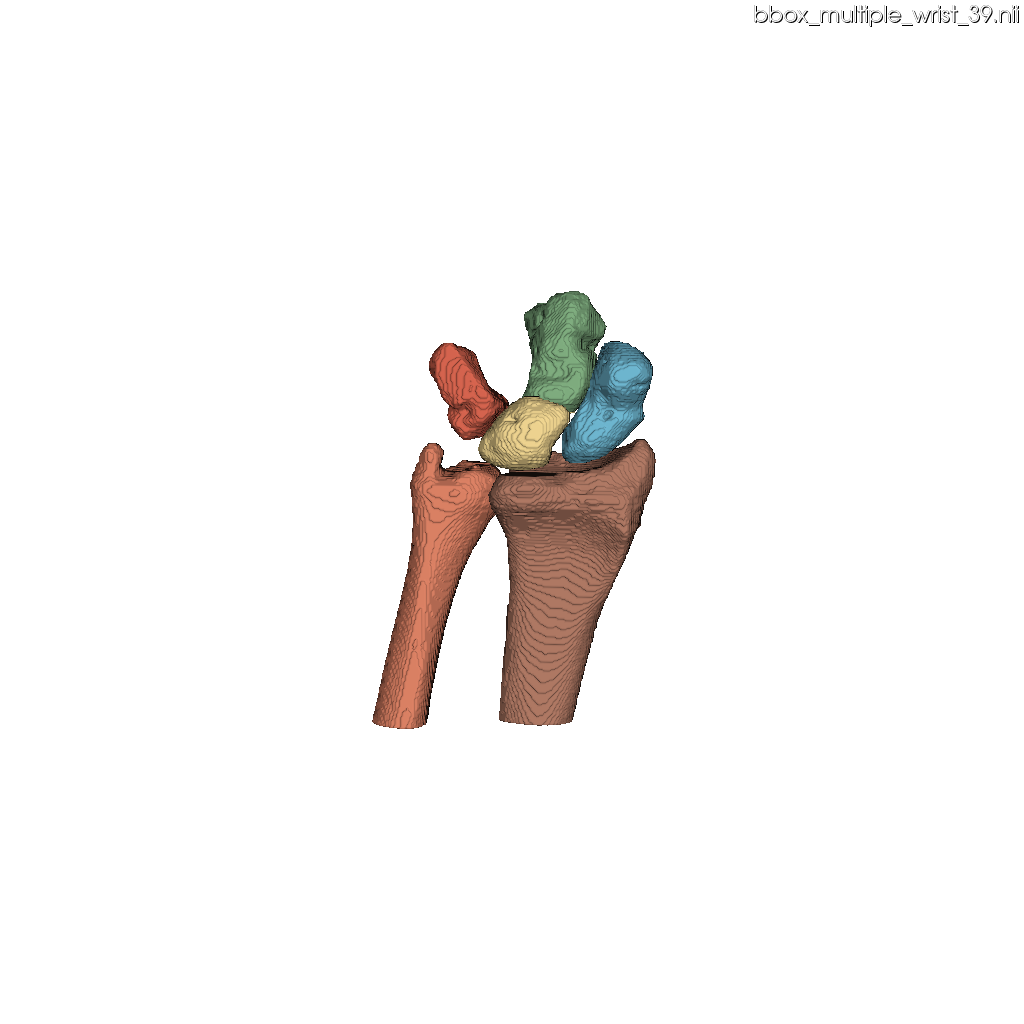} & &
    \includegraphics[width=0.13\linewidth, trim=370 290 370 260, clip]{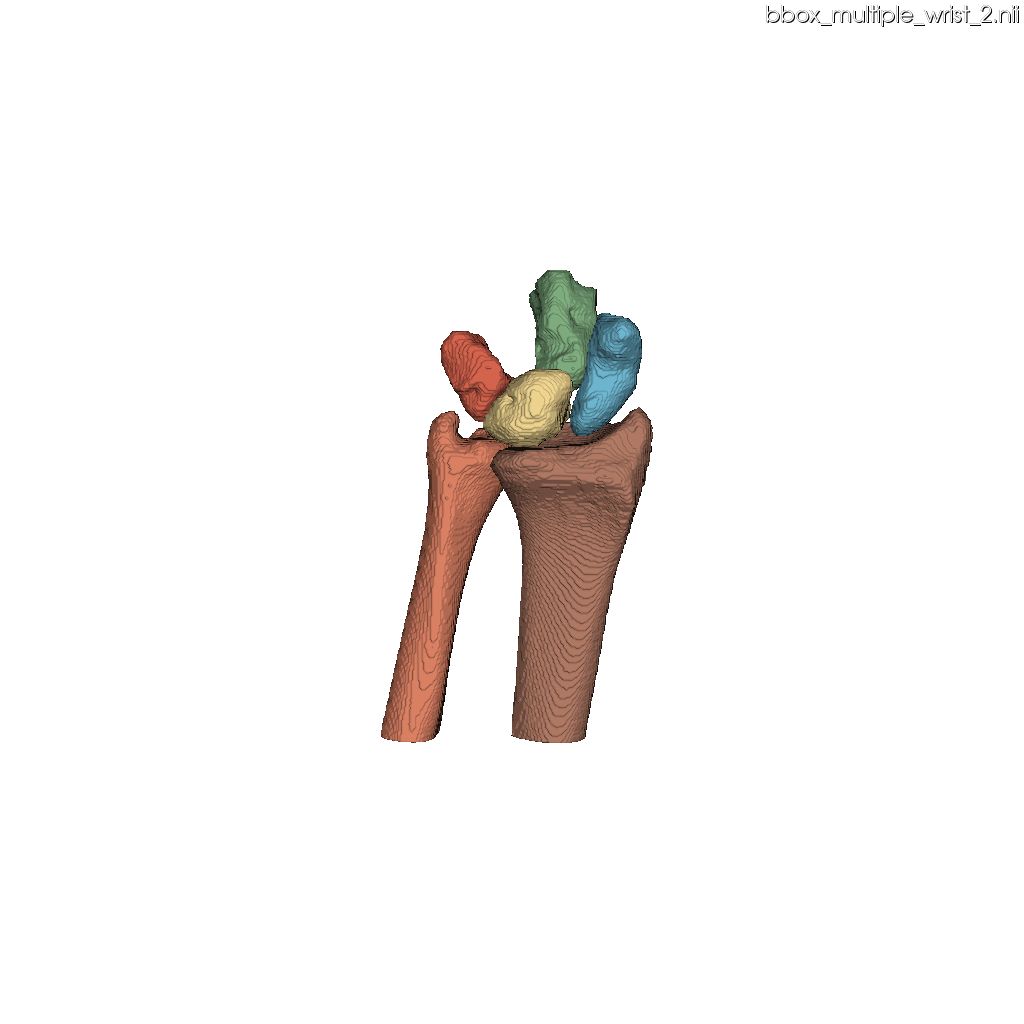} & &
    \includegraphics[width=0.14\linewidth, trim=390 290 370 280, clip]{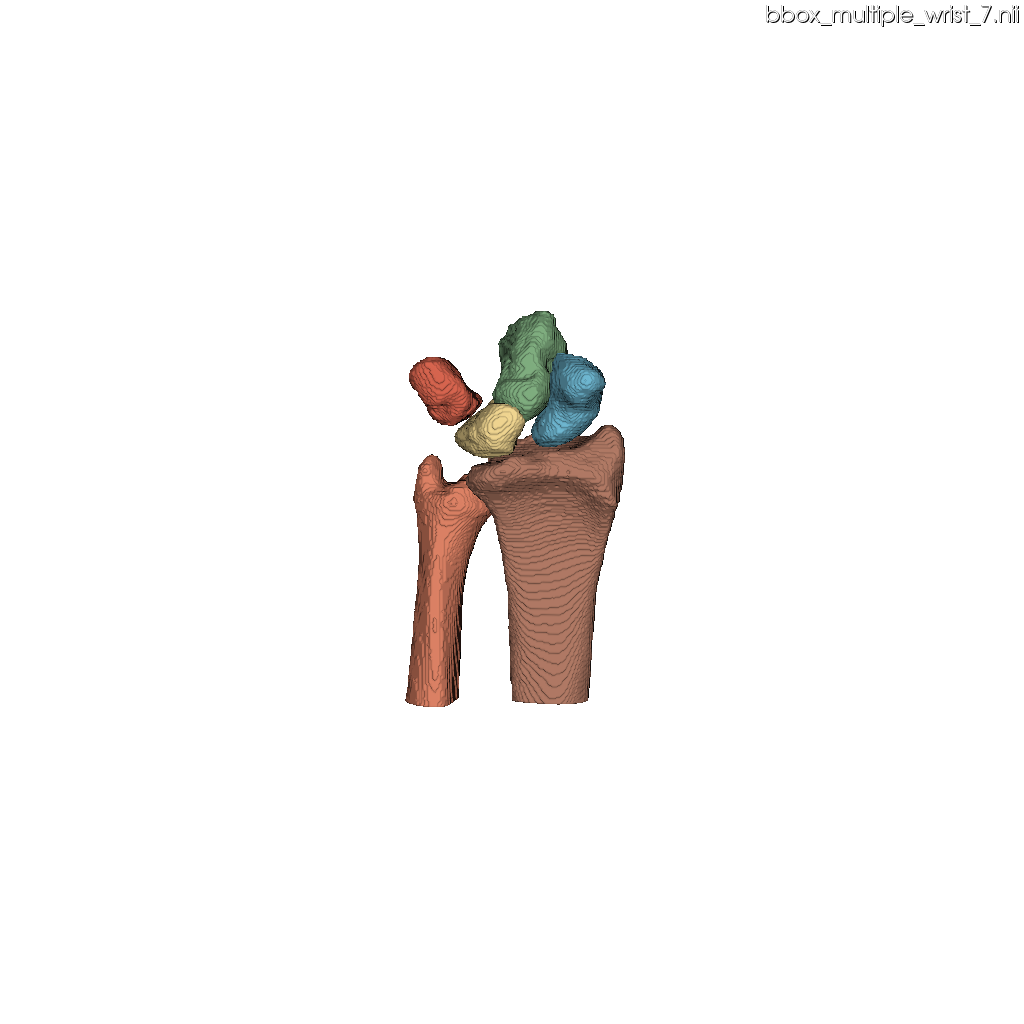} & \\ 
    \hline
    \multirow{2}{*}{\cblacksquaredot[0.6]{oranje}} & \includegraphics[width=0.14\linewidth, trim=560 380 190 330, clip]{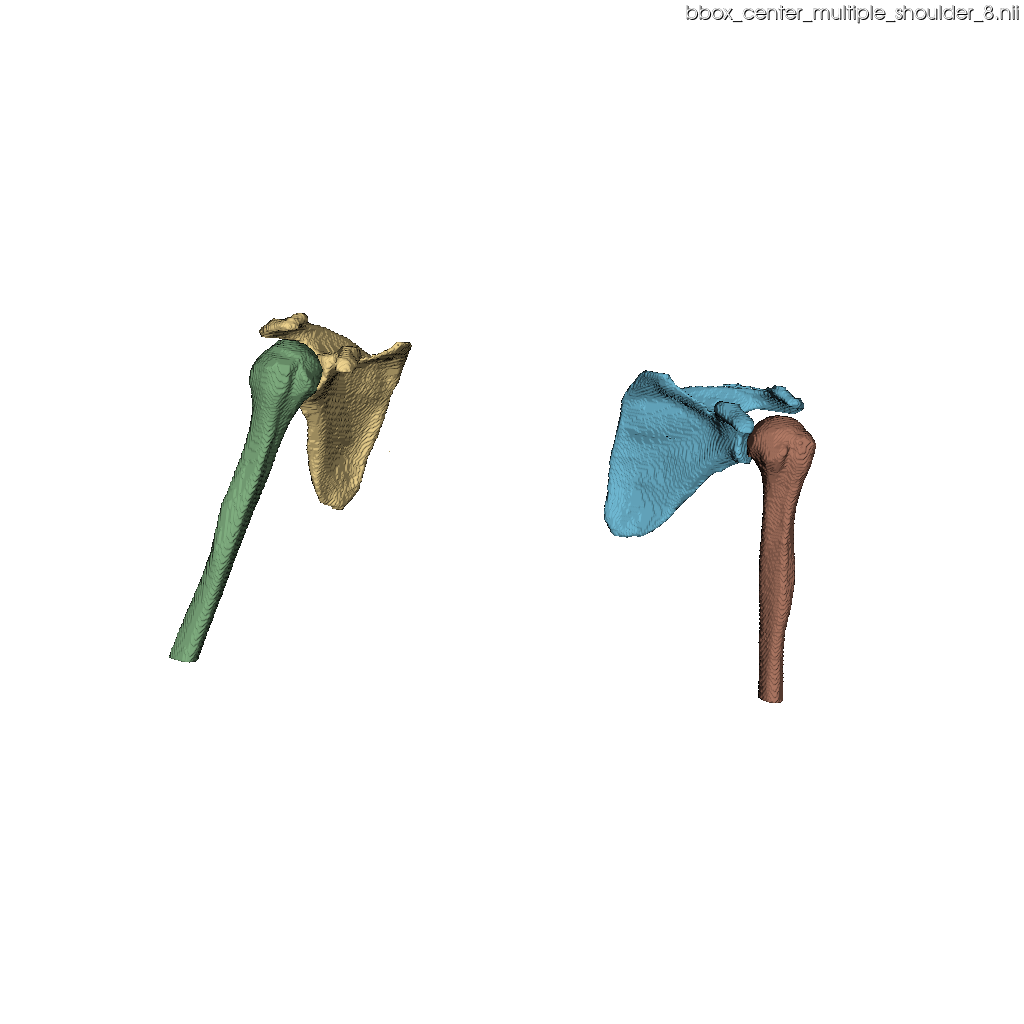} & \raisebox{-0.6\height}[0pt][0pt]{\includegraphics[width=0.11\linewidth, trim=400 200 430 150, clip]{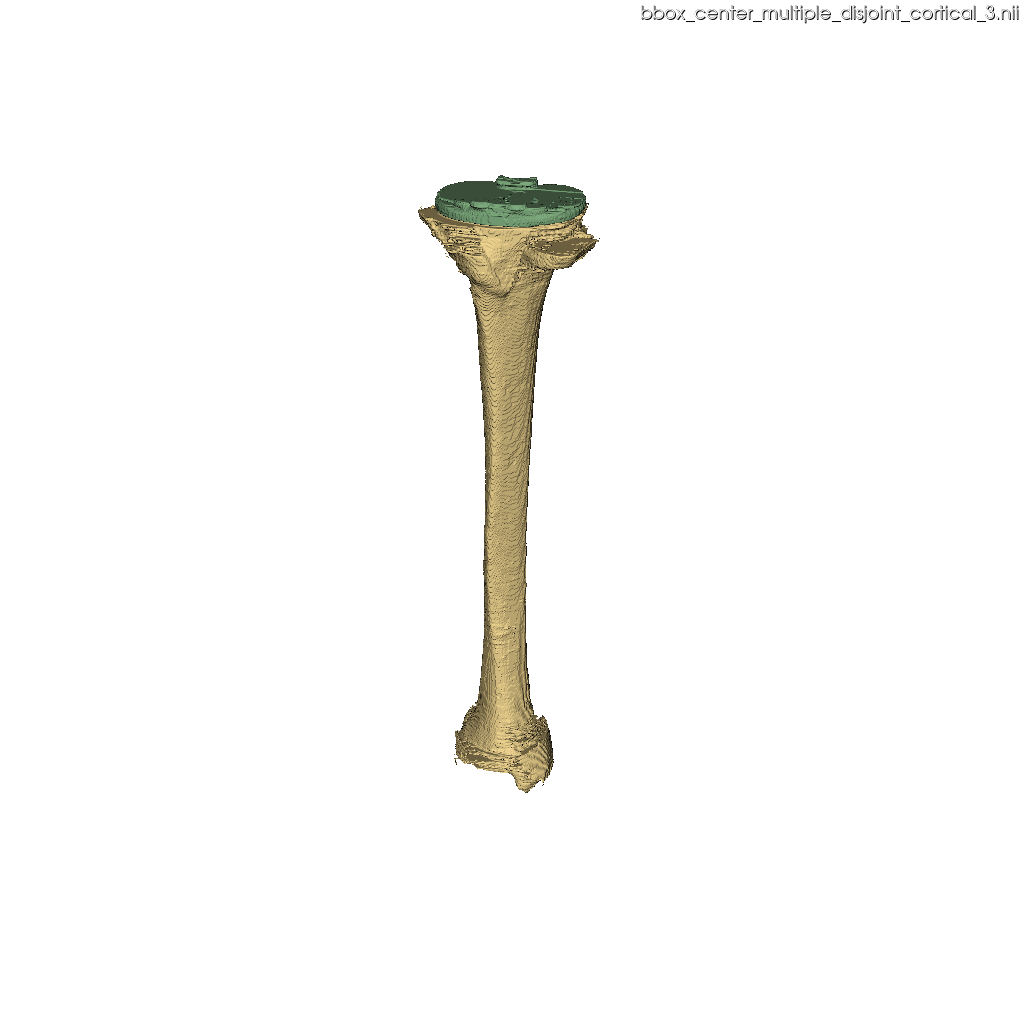}} &
    \includegraphics[width=0.14\linewidth, trim=590 360 120 370, clip]{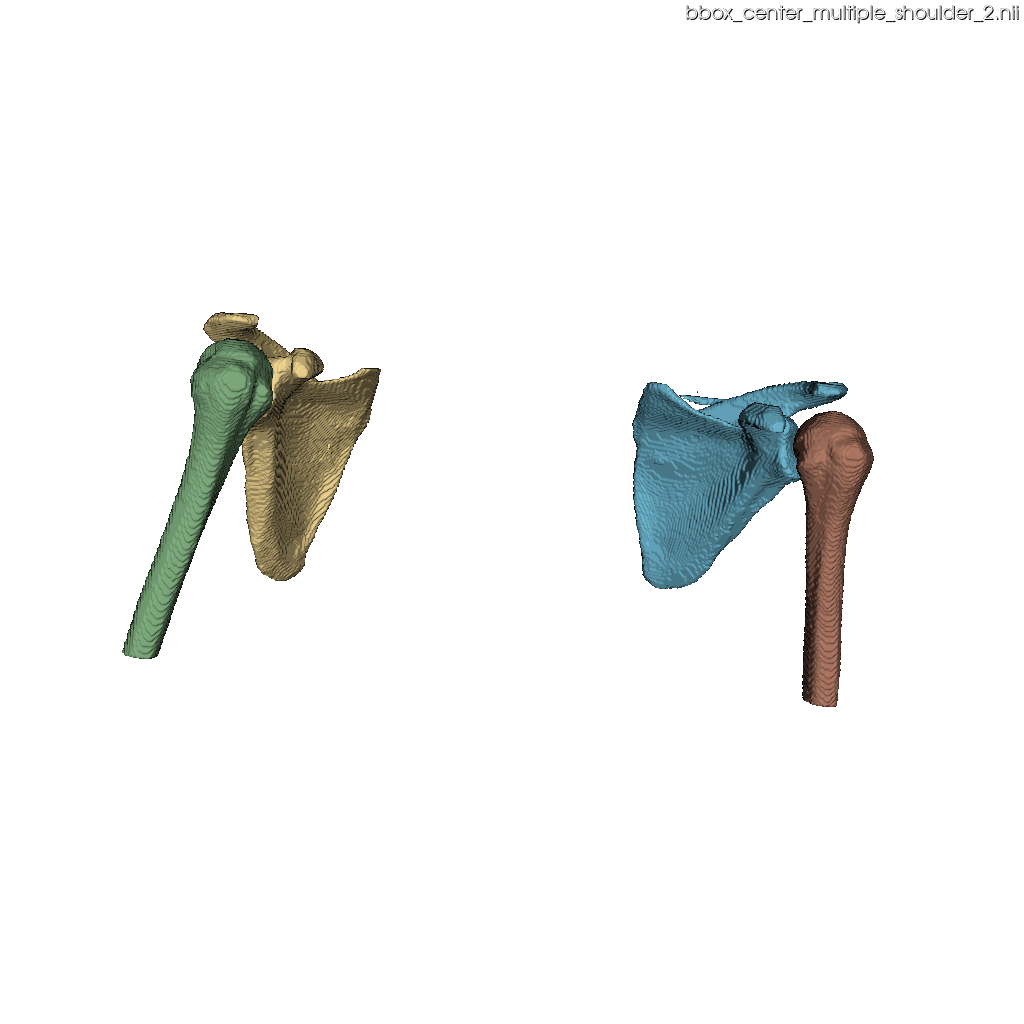} & \raisebox{-0.6\height}[0pt][0pt]{\includegraphics[width=0.11\linewidth, trim=400 200 400 150, clip]{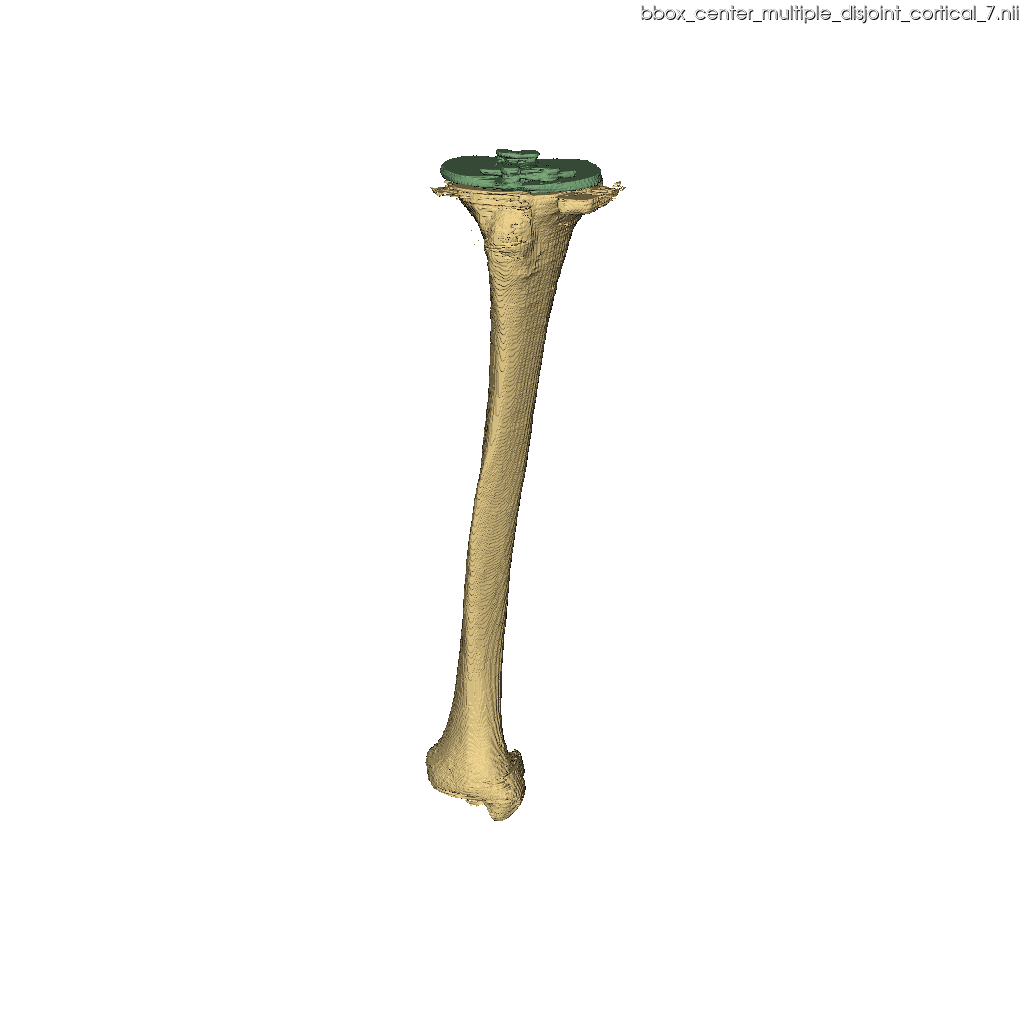}} &
    \includegraphics[width=0.14\linewidth, trim=560 380 170 330, clip]{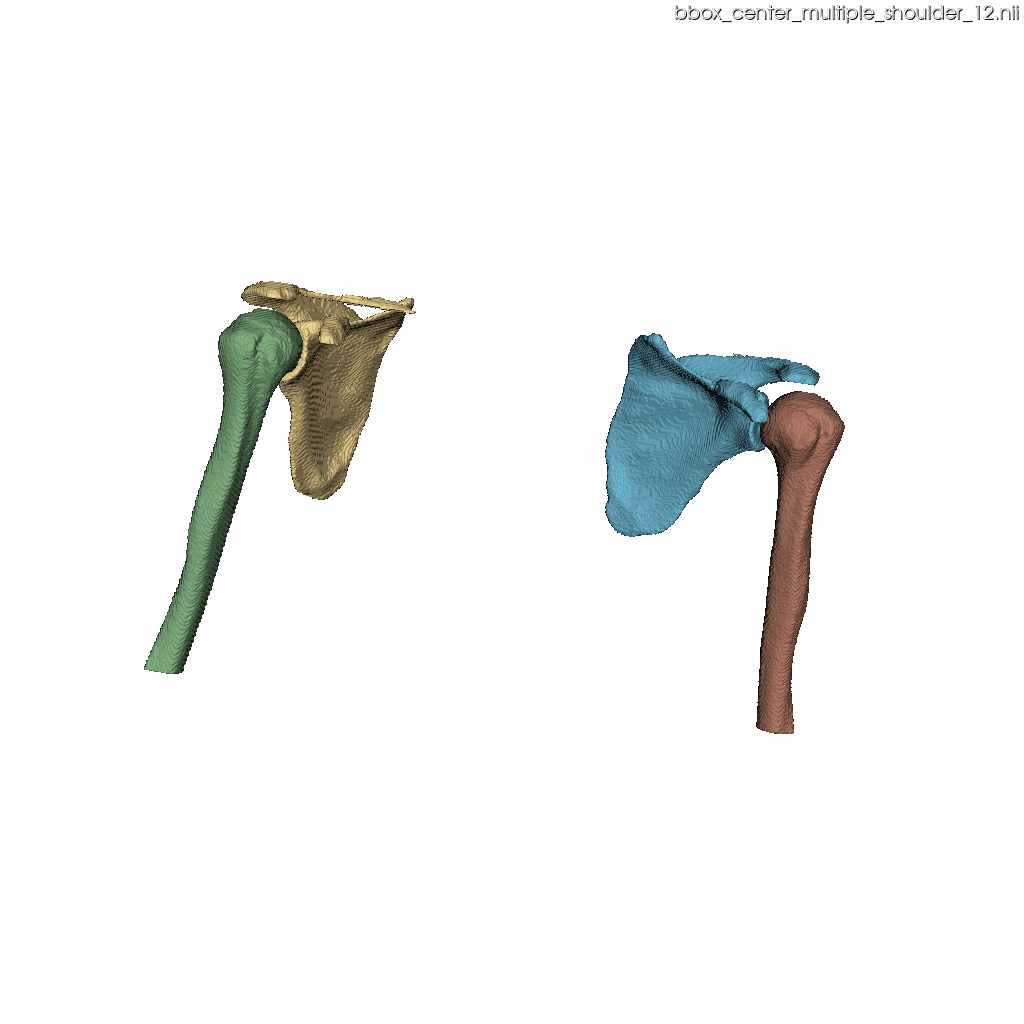} & \raisebox{-0.6\height}[0pt][0pt]{\includegraphics[width=0.11\linewidth, trim=400 150 400 100, clip]{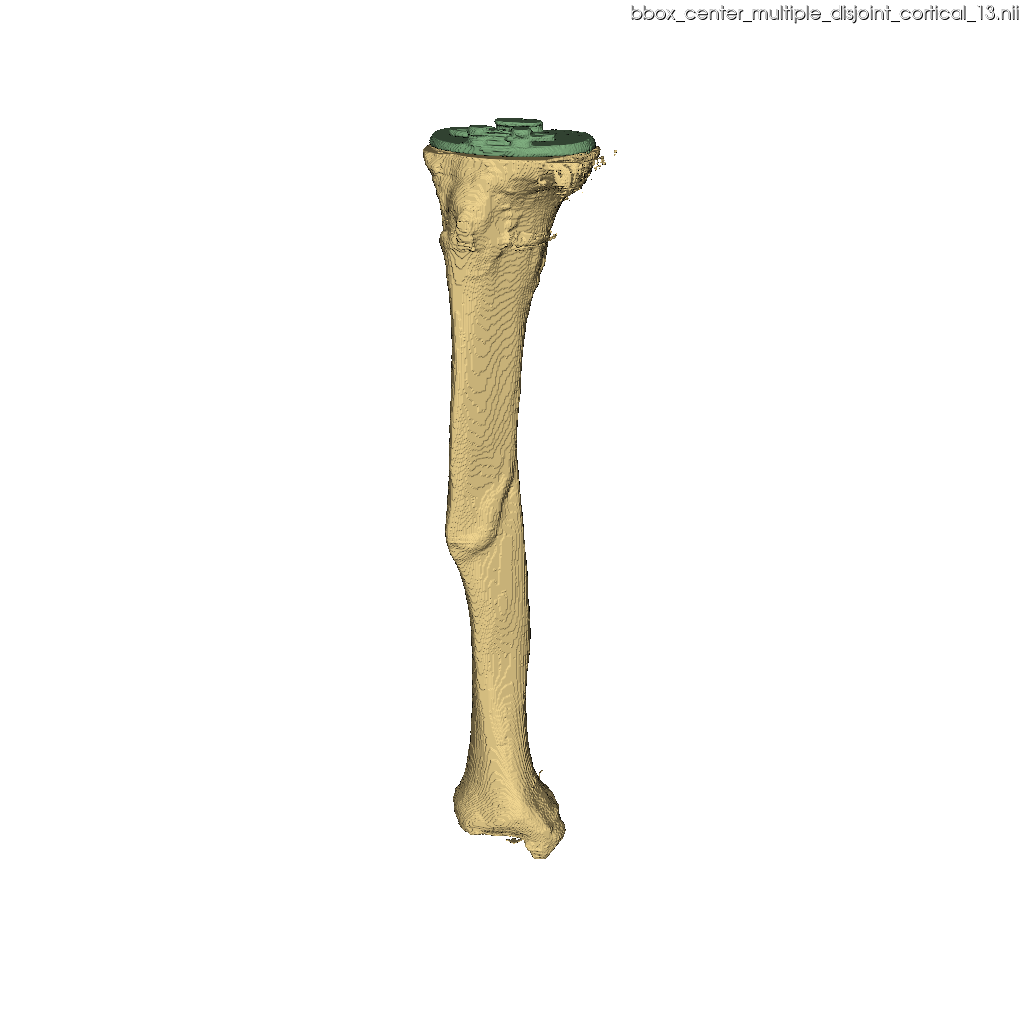}} 
    \\
    & \includegraphics[width=0.14\linewidth, trim=370 340 370 250, clip]{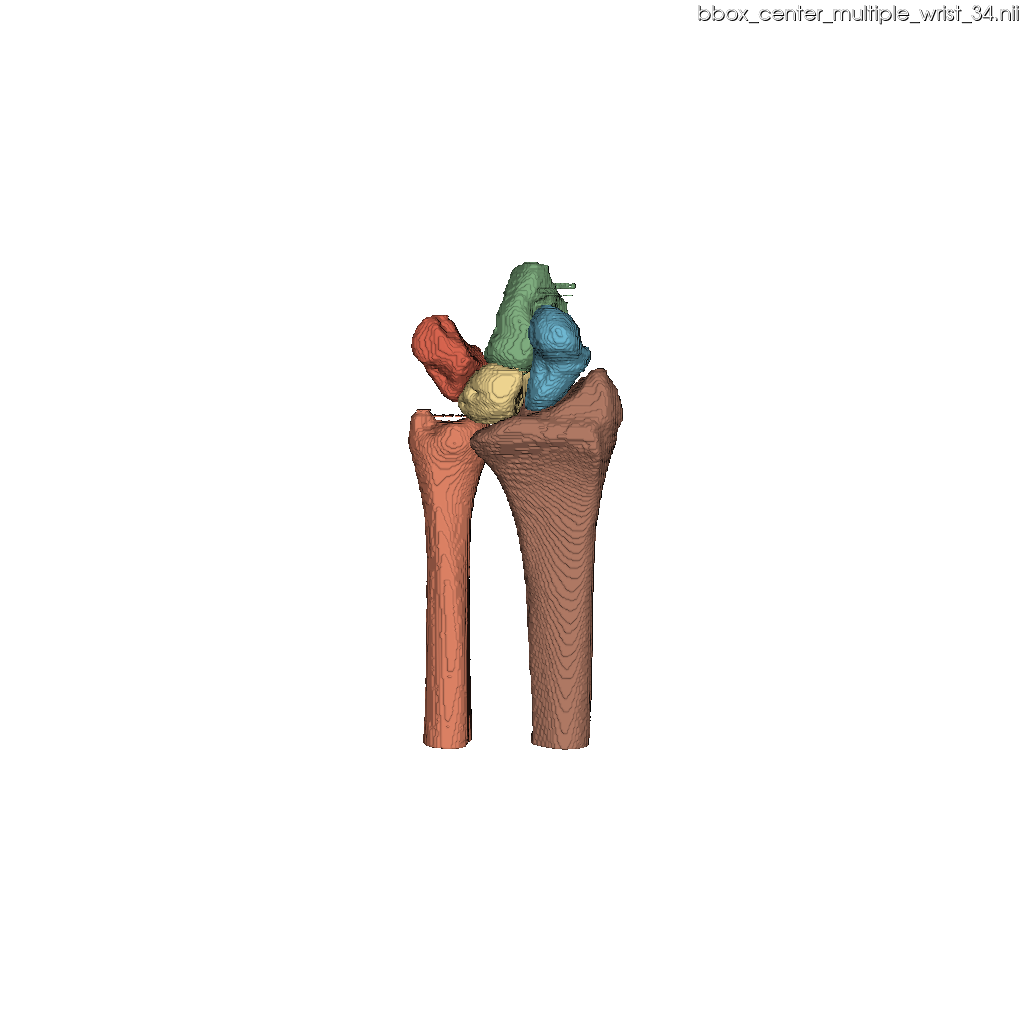} & &
    \includegraphics[width=0.14\linewidth, trim=370 290 370 280, clip]{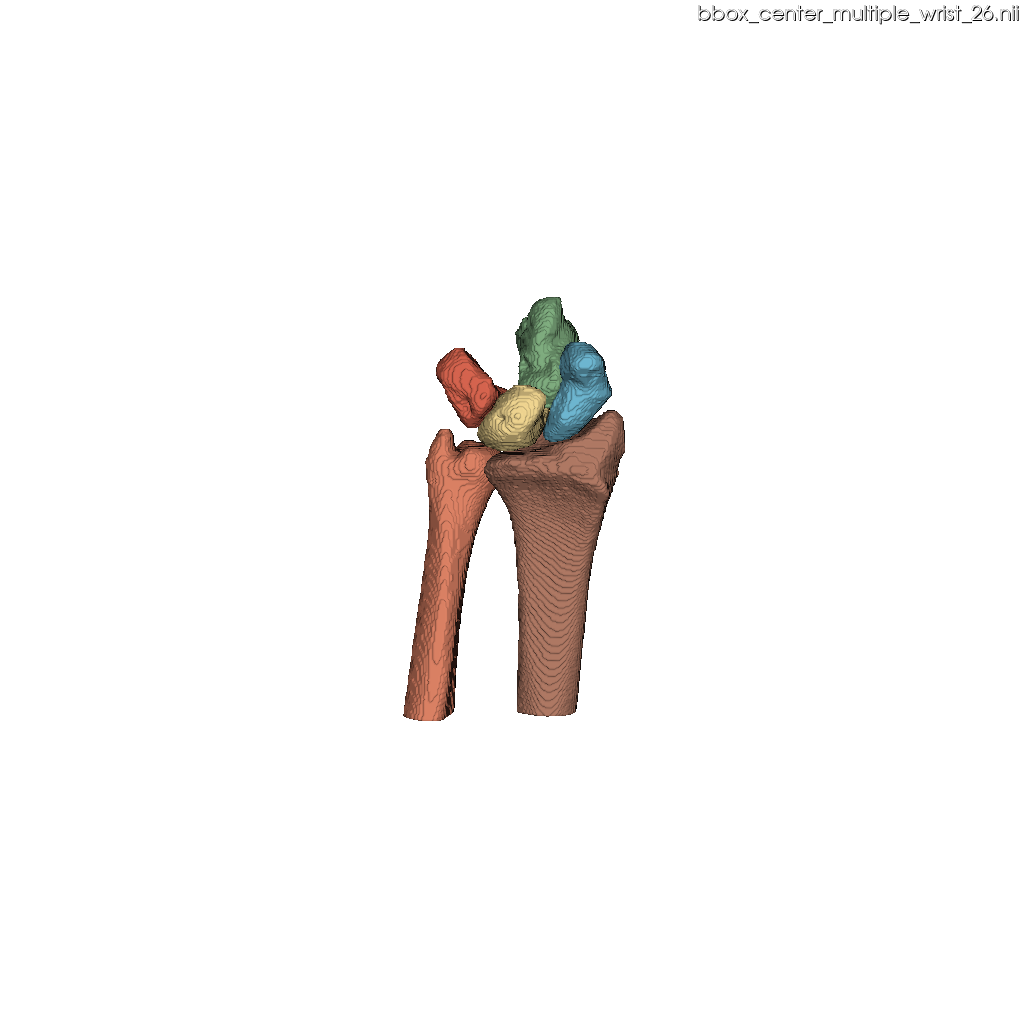} & &
    \includegraphics[width=0.14\linewidth, trim=370 290 370 280, clip]{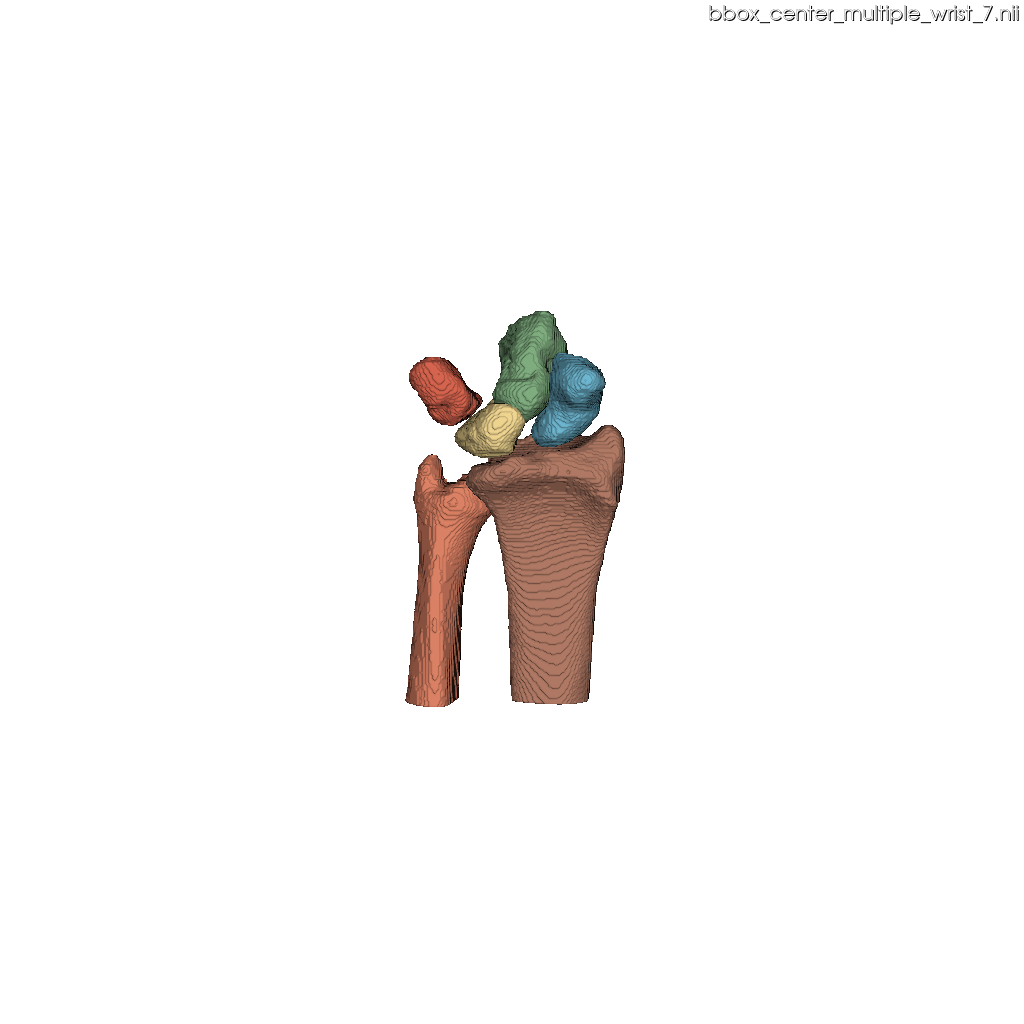} & \\ 
    \hline
    \hline
    \multirow{2}{*}{\cblackstartriangledown[0.6]{oranje}} & \includegraphics[width=0.14\linewidth, trim=580 380 110 370, clip]{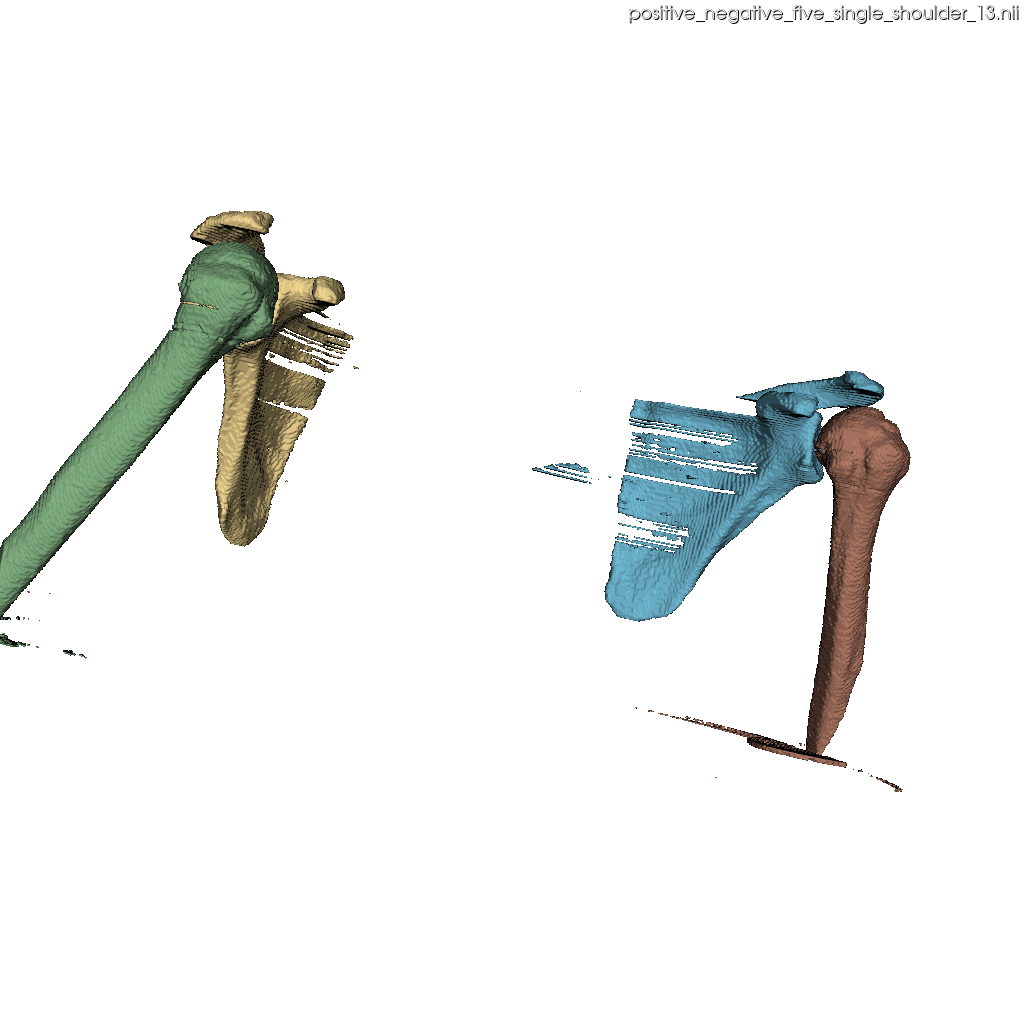} & \raisebox{-0.6\height}[0pt][0pt]{\includegraphics[width=0.11\linewidth, trim=380 150 340 40, clip]{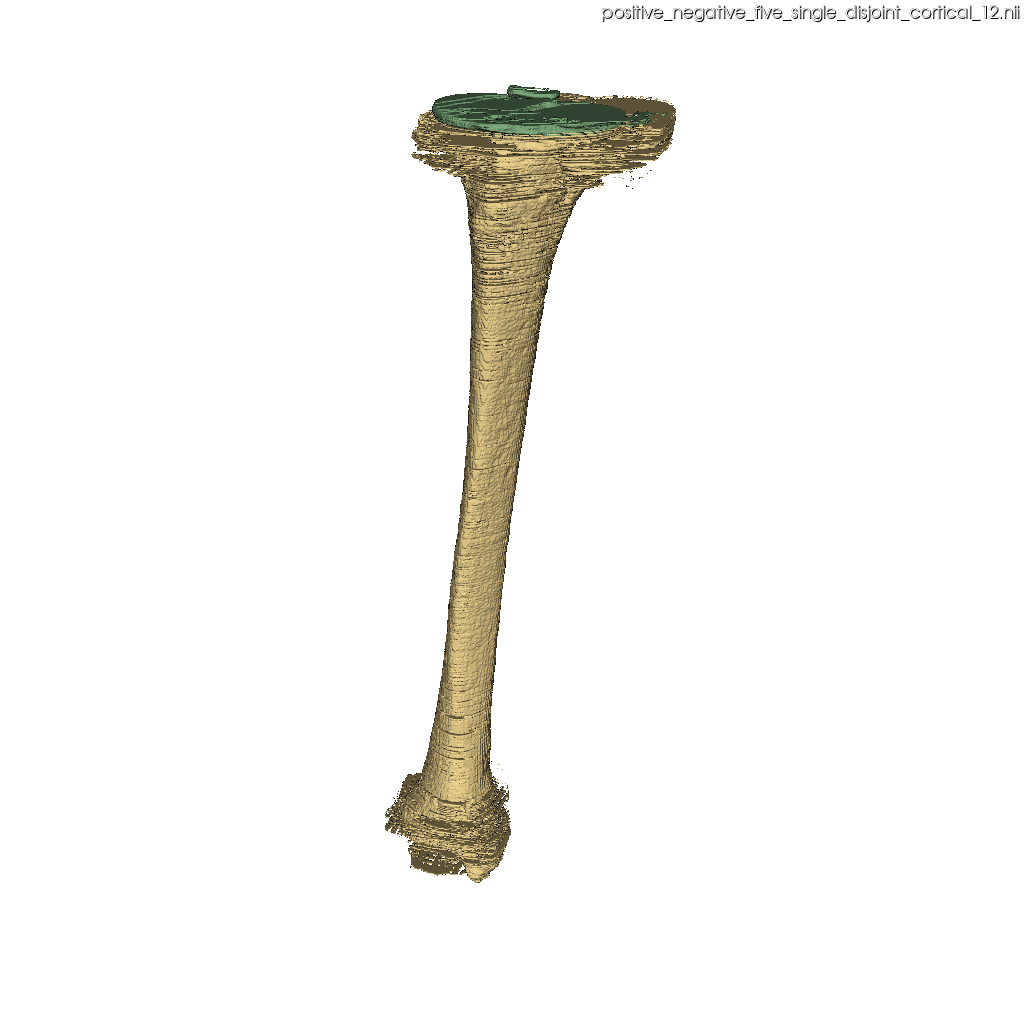}} &
    \includegraphics[width=0.14\linewidth, trim=560 380 170 330, clip]{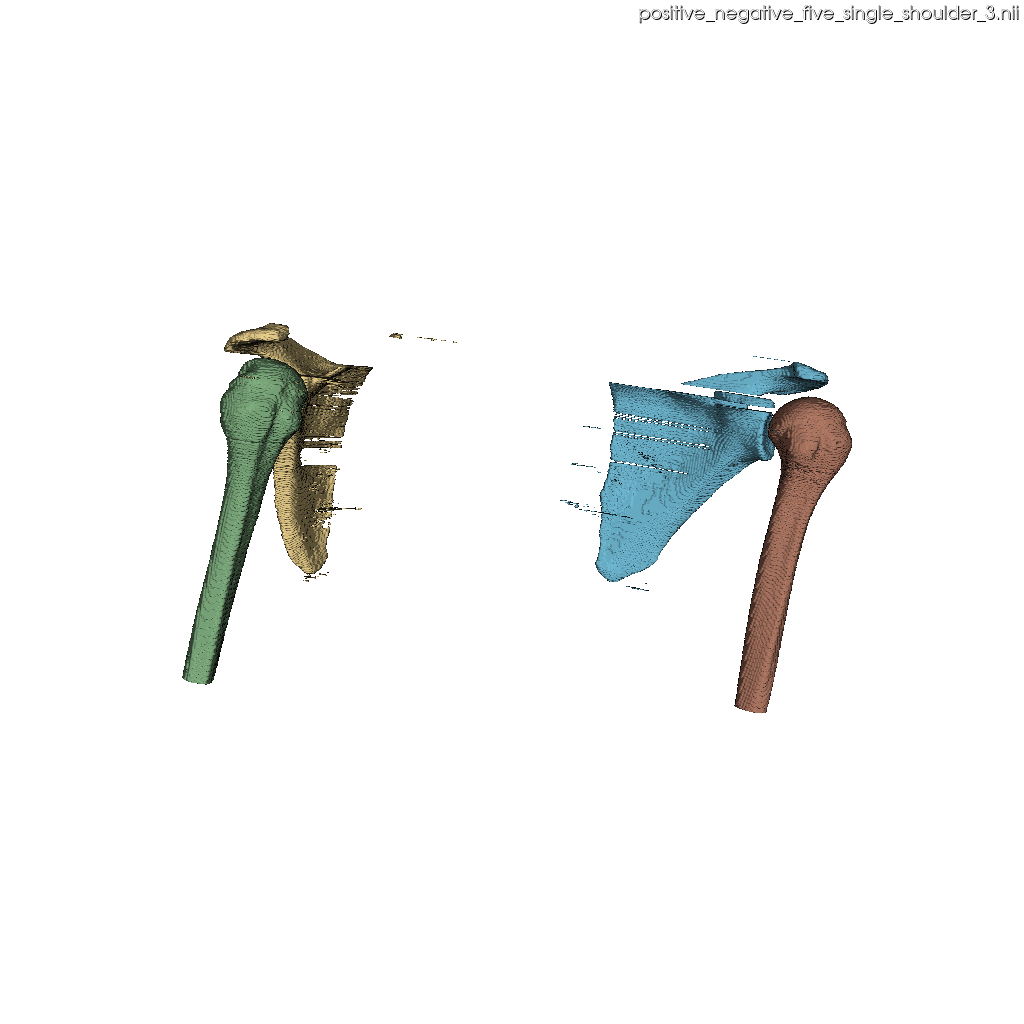} & \raisebox{-0.6\height}[0pt][0pt]{\includegraphics[width=0.11\linewidth, trim=380 100 350 60, clip]{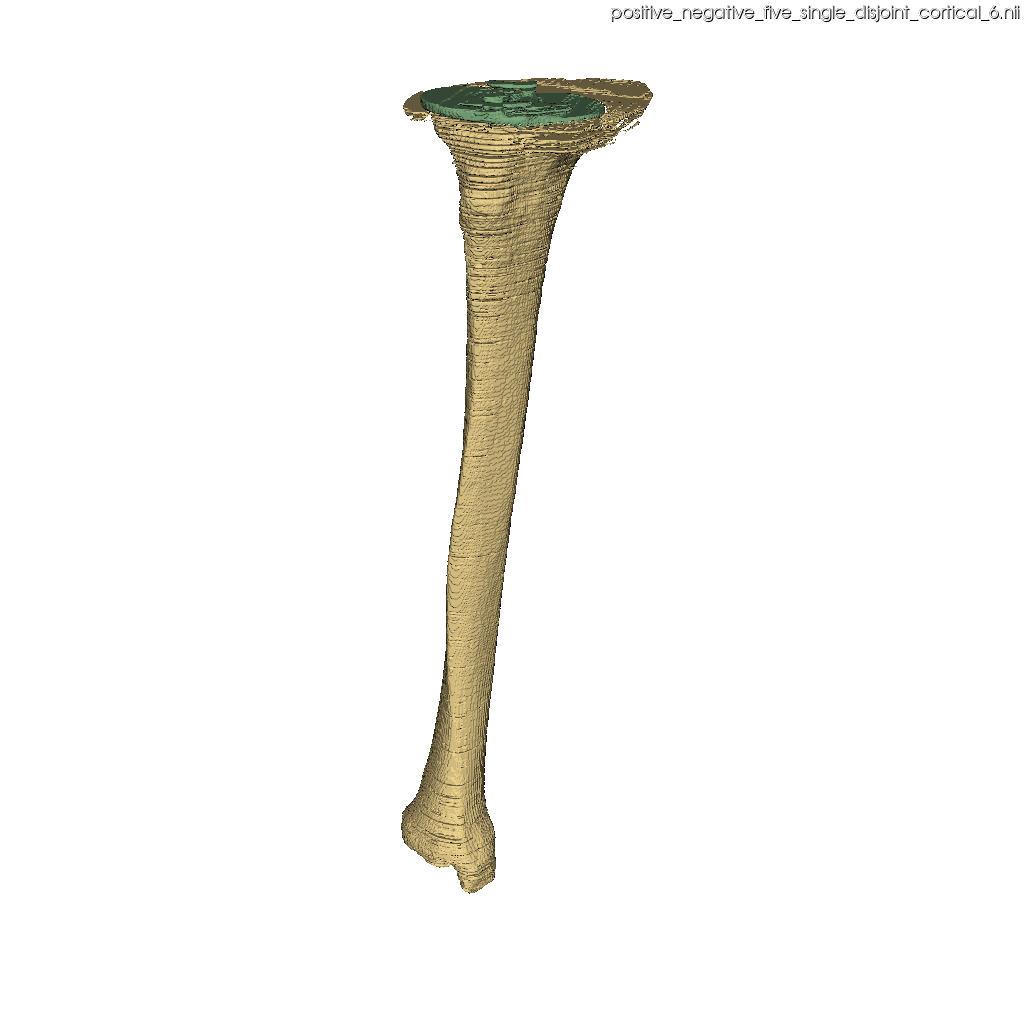}} &
    \includegraphics[width=0.14\linewidth, trim=580 380 100 350, clip]{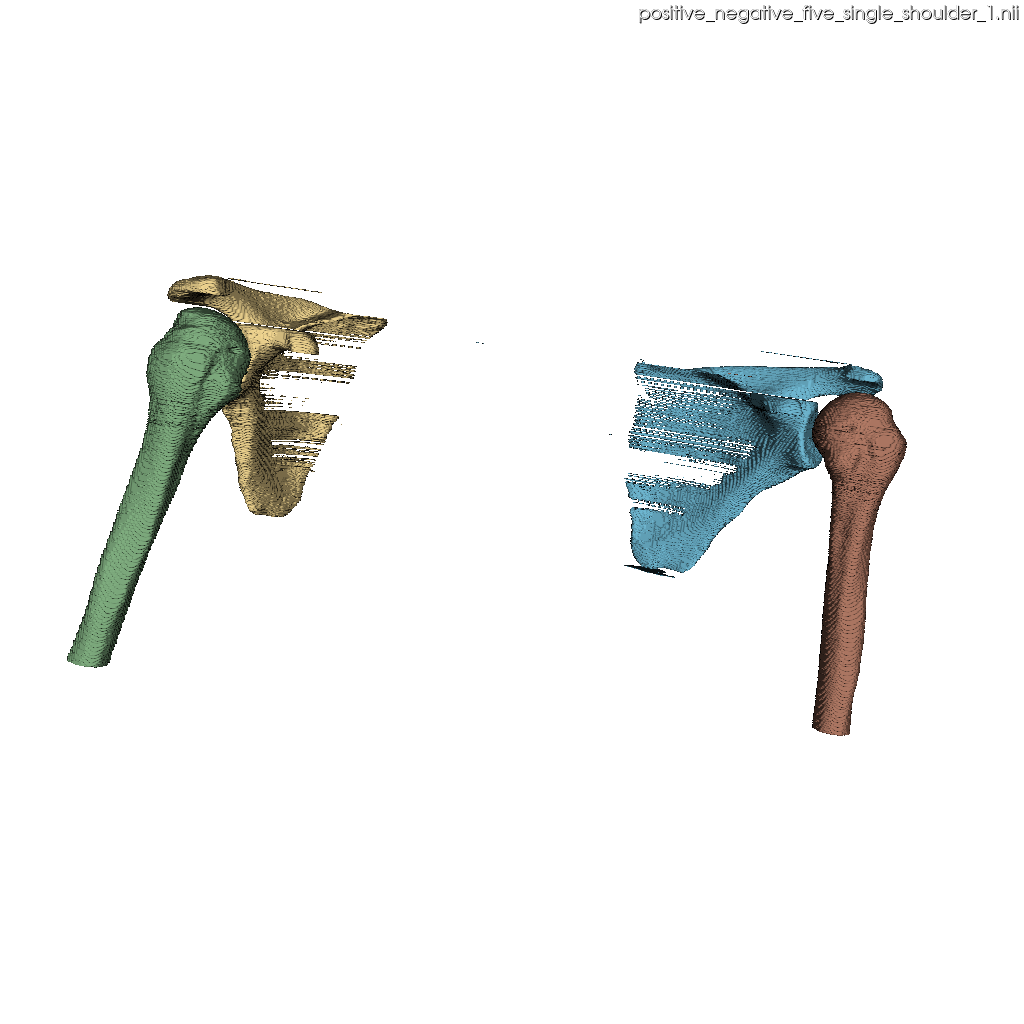} & \raisebox{-0.6\height}[0pt][0pt]{\includegraphics[width=0.11\linewidth, trim=380 130 390 80, clip]{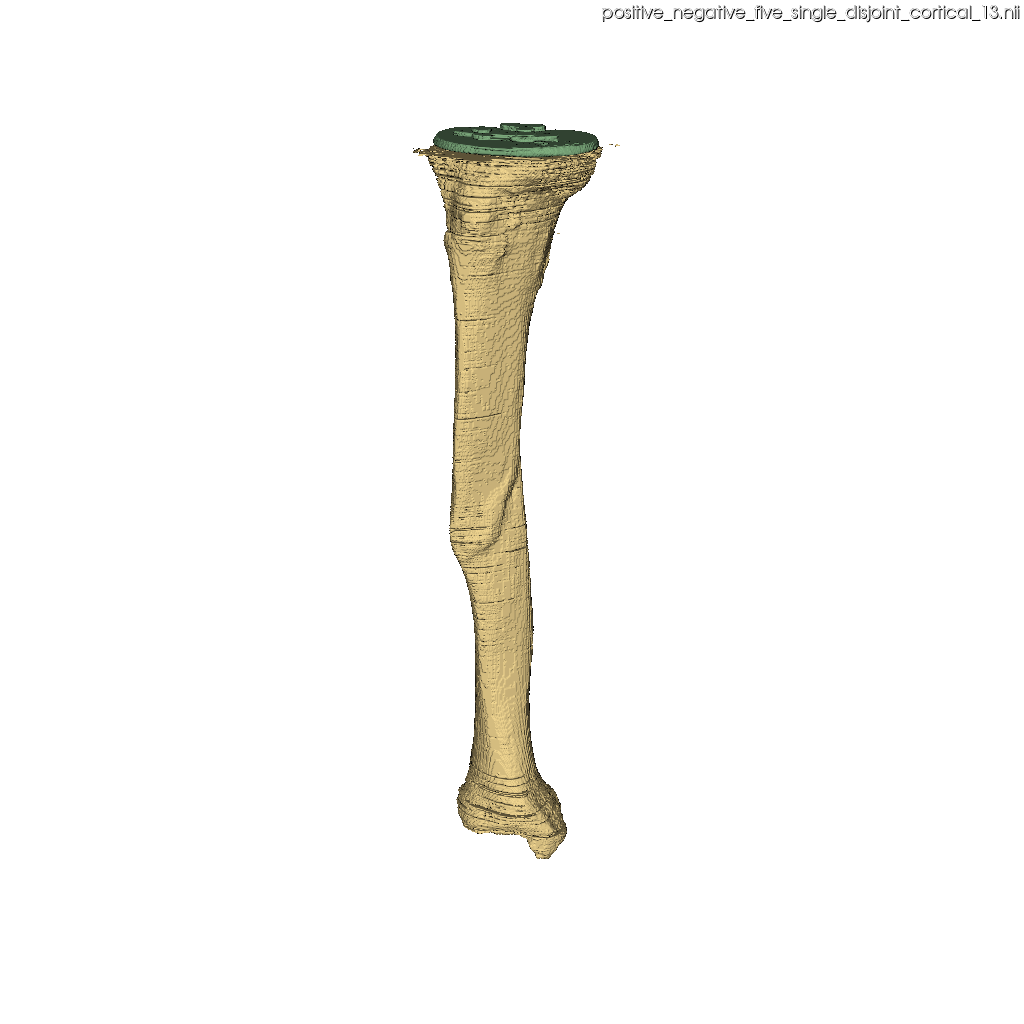}}
    \\
    & \includegraphics[width=0.14\linewidth, trim=370 290 370 280, clip]{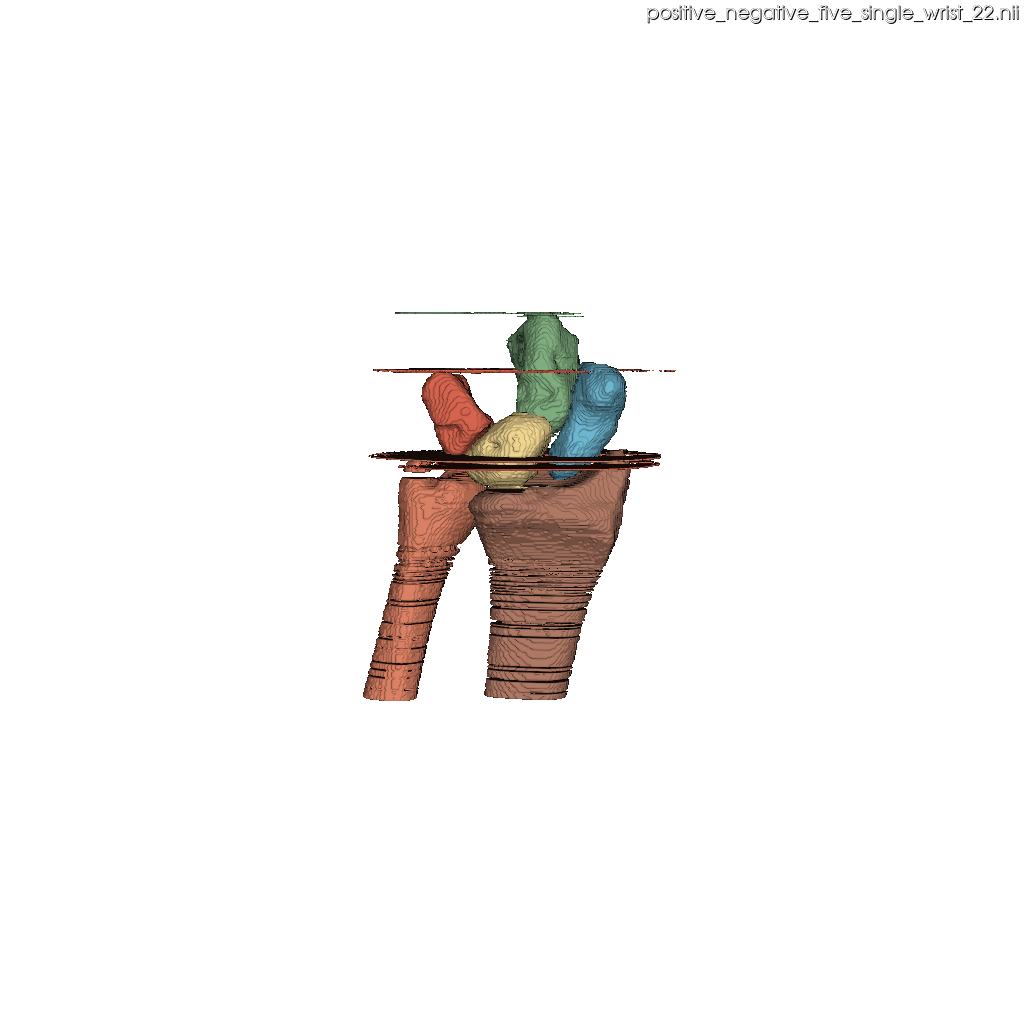} & &
    \includegraphics[width=0.13\linewidth, trim=370 290 370 280, clip]{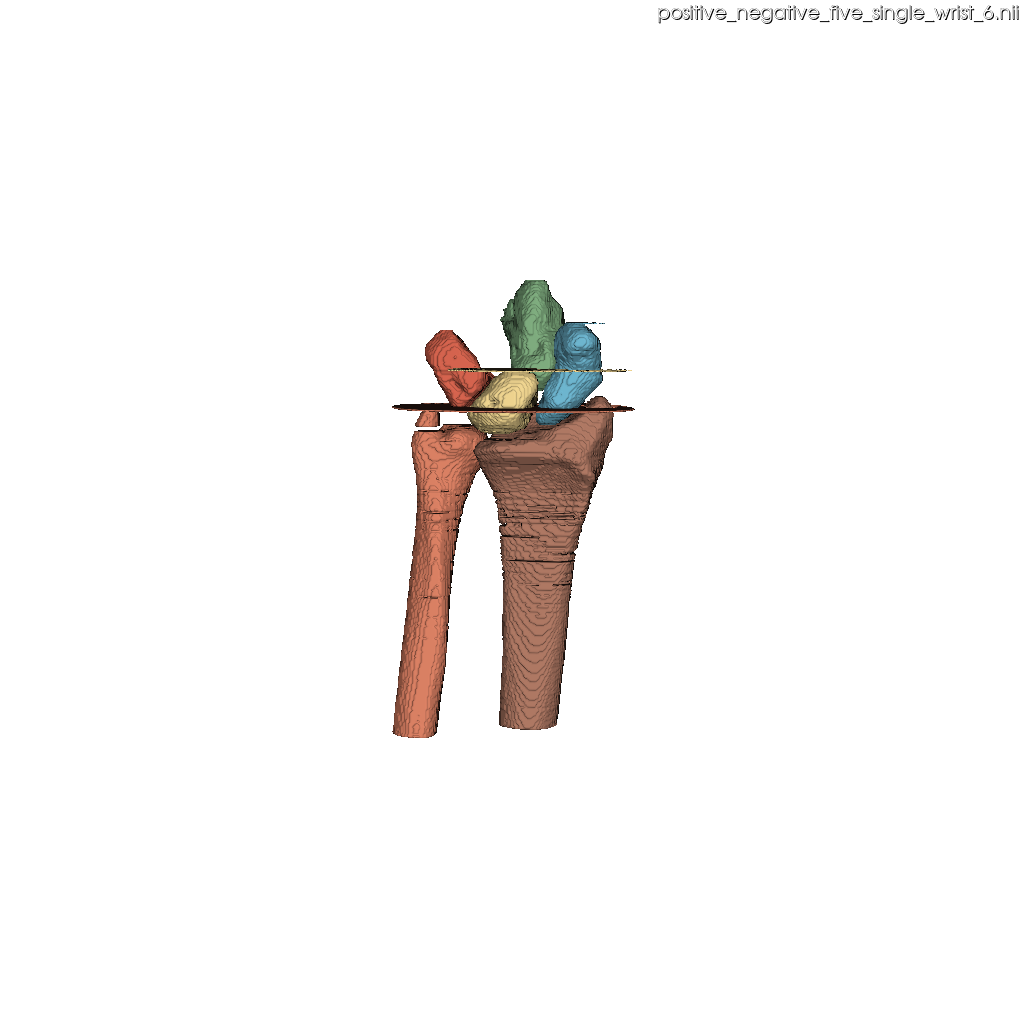} & &
    \includegraphics[width=0.14\linewidth, trim=370 290 370 280, clip]{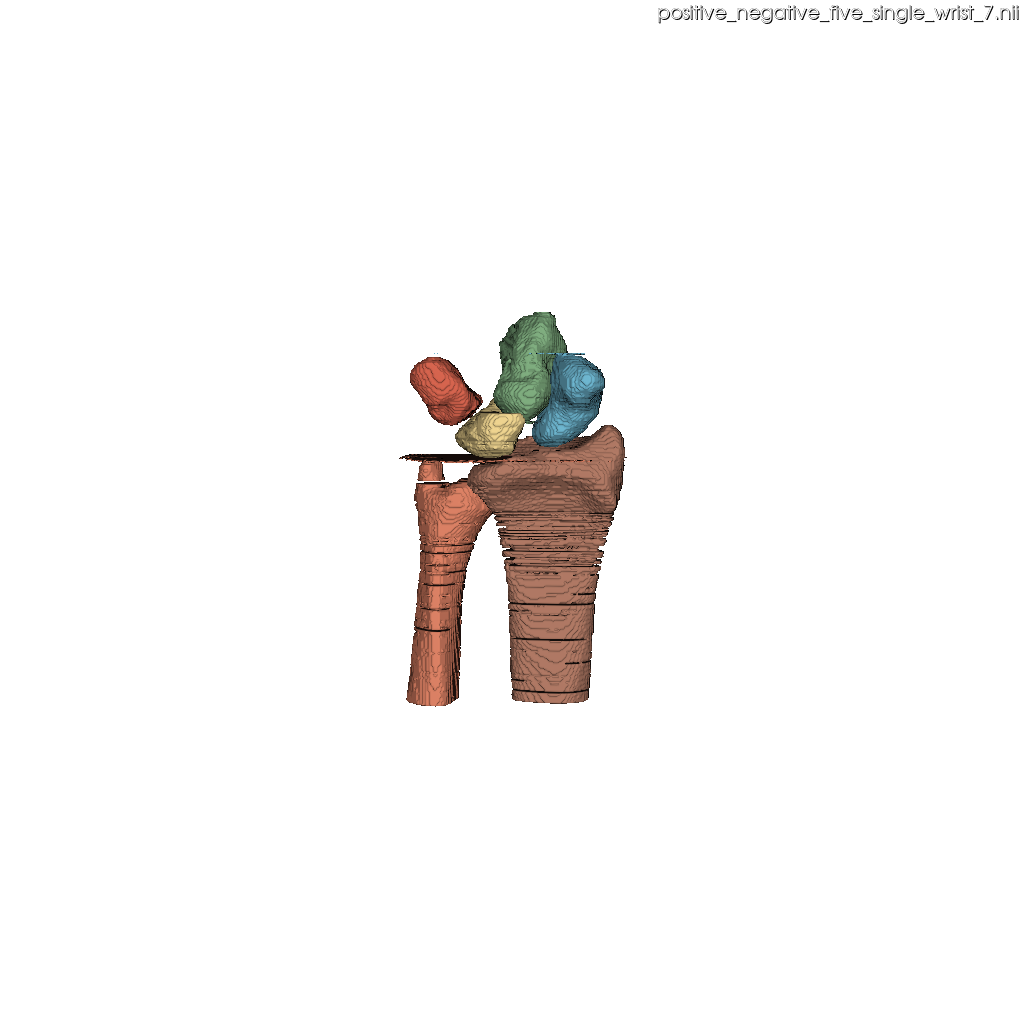} & \\  
    \hline
\end{tabular}
\caption{Selected examples for \textit{\textsc{Sam} H} with low, medium and high DSC.}
\label{fig:example_samh}
\end{figure}

\begin{figure}[H]
\begin{tabular}{|l|cc|cc|cc|}
    \hline
    Setting & \multicolumn{2}{c|}{DSC $\downarrow$} & \multicolumn{2}{c|}{DSC median} & \multicolumn{2}{c|}{DSC $\uparrow$} \\ 
    \hline
    \multirow{2}{*}{\cblacksquare[0.6]{pomme}} & \includegraphics[width=0.14\linewidth, trim=570 370 100 350, clip]{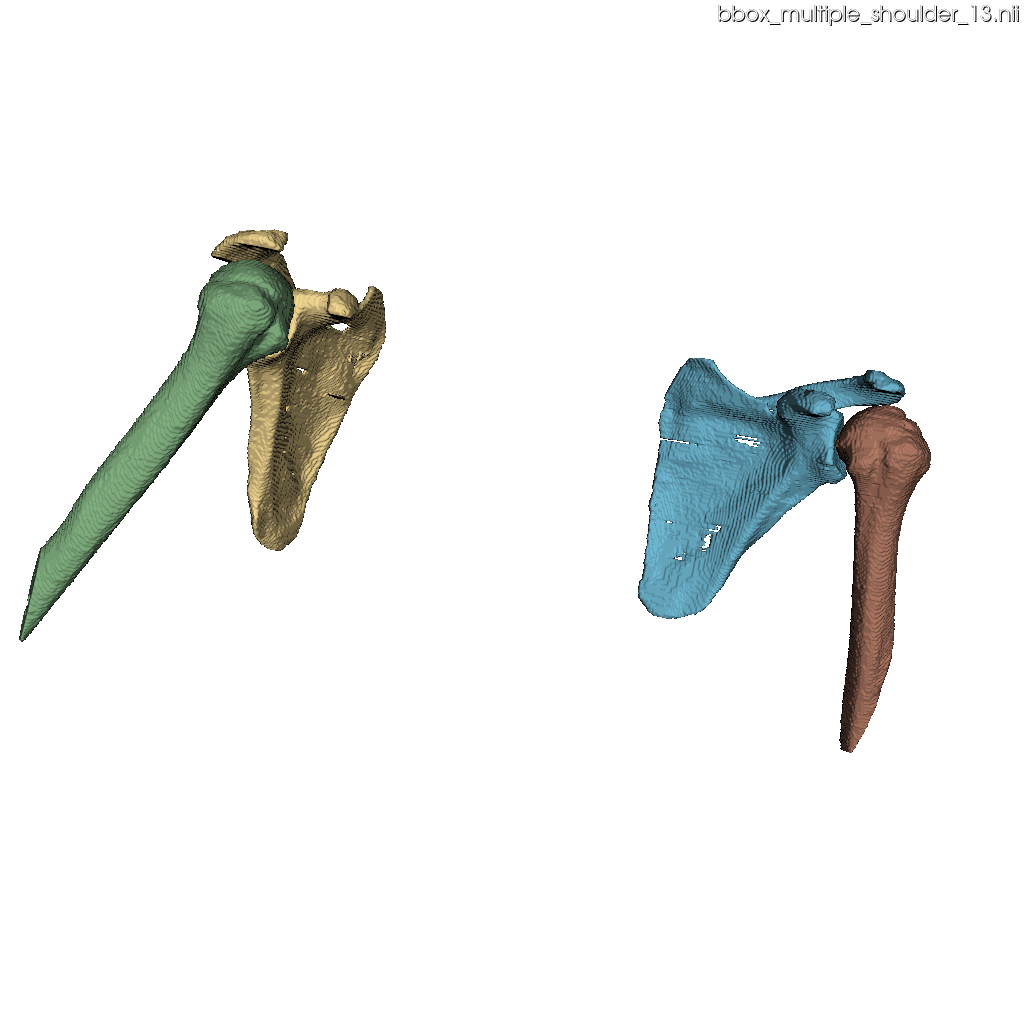} & \raisebox{-0.6\height}[0pt][0pt]{\includegraphics[width=0.11\linewidth, trim=400 200 420 150, clip]{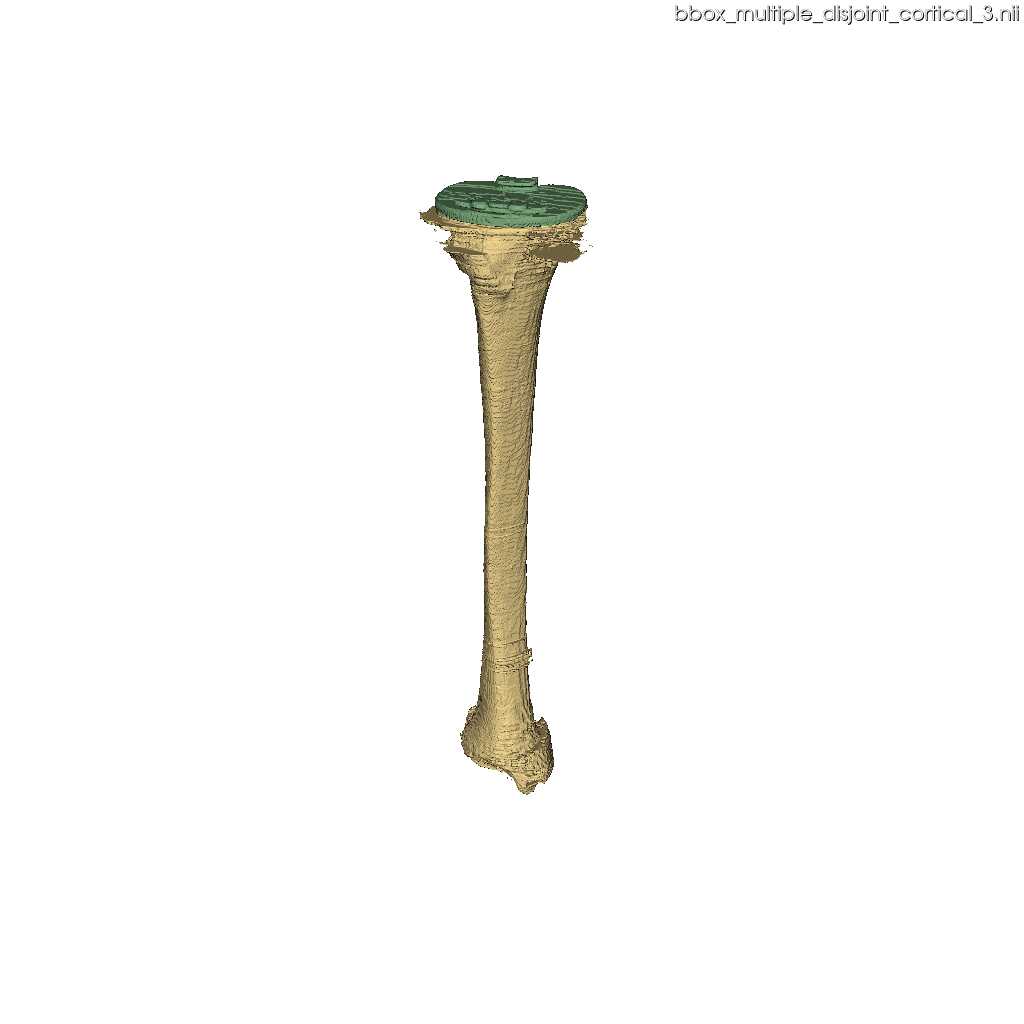}} &
    \includegraphics[width=0.14\linewidth, trim=560 370 140 330, clip]{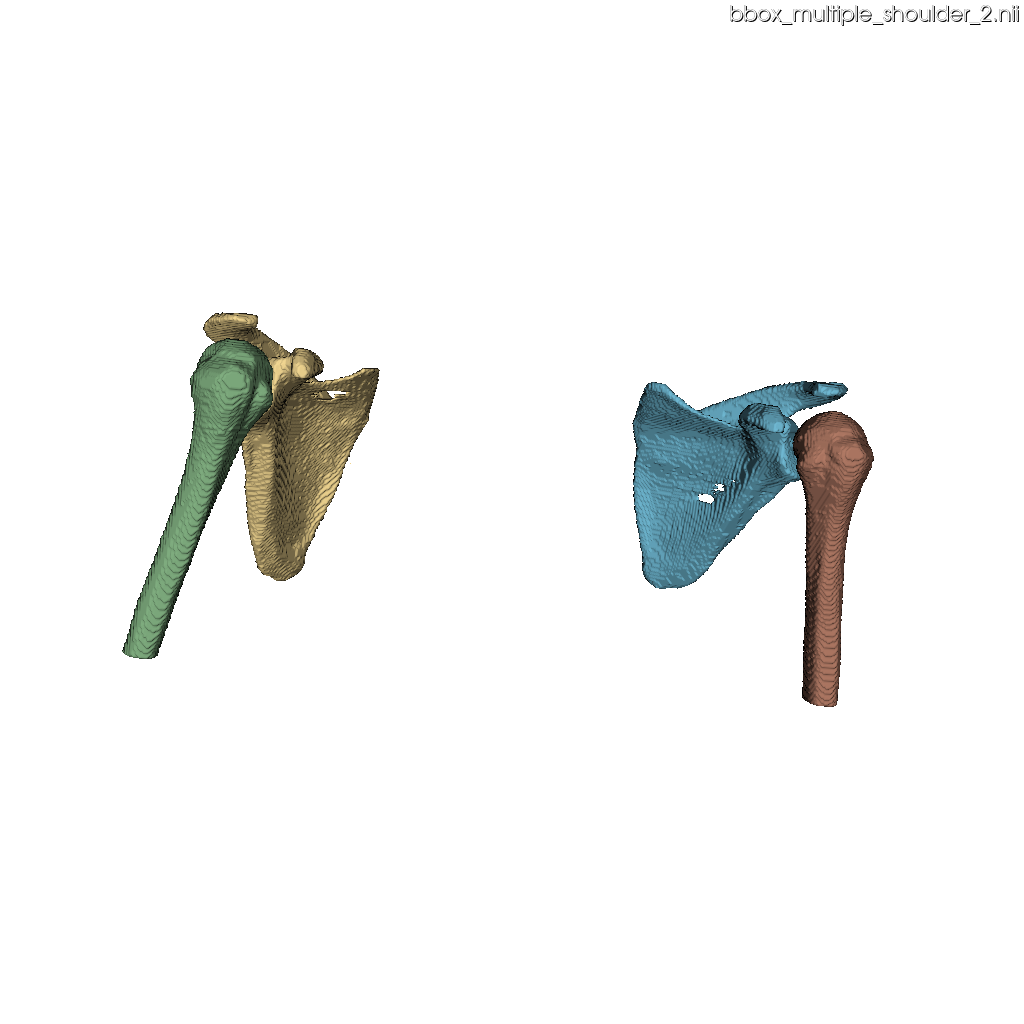} & \raisebox{-0.6\height}[0pt][0pt]{\includegraphics[width=0.11\linewidth, trim=400 200 400 120, clip]{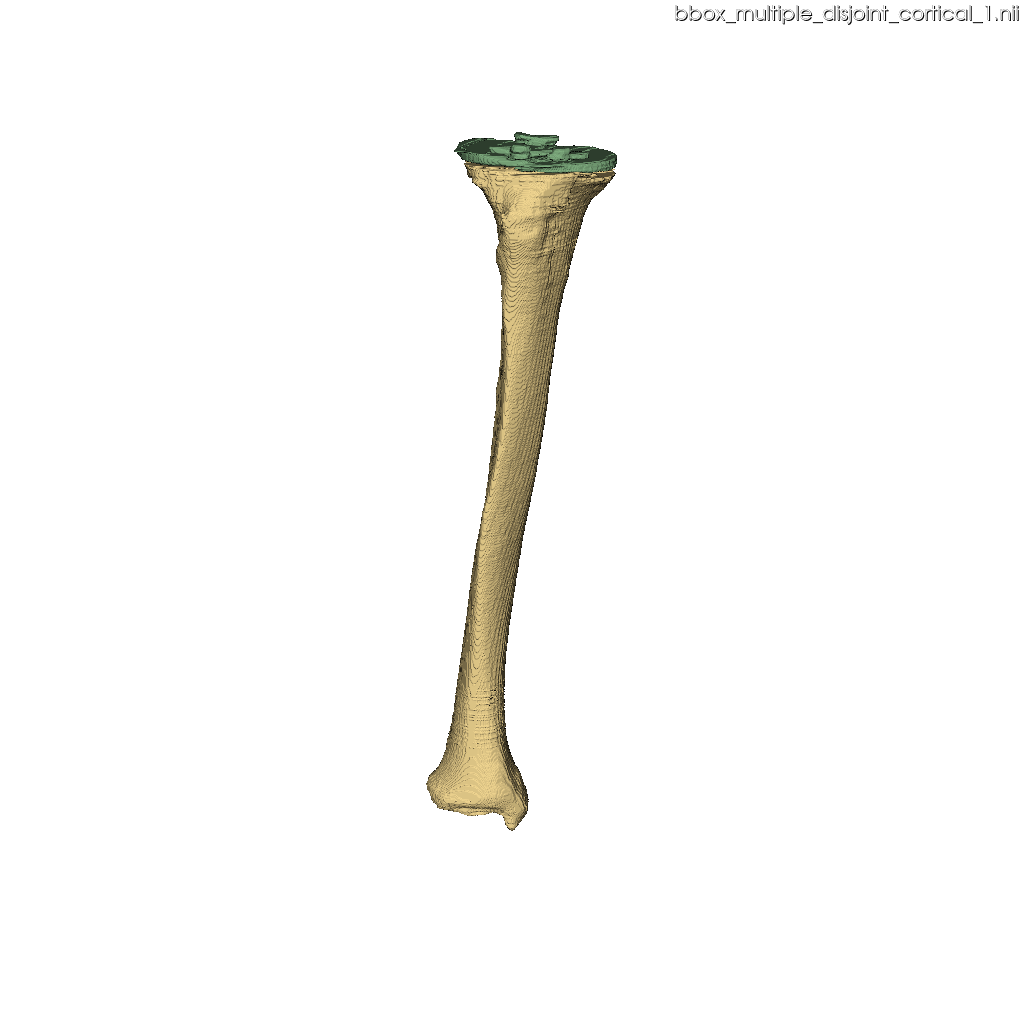}} &
    \includegraphics[width=0.14\linewidth, trim=560 380 120 310, clip]{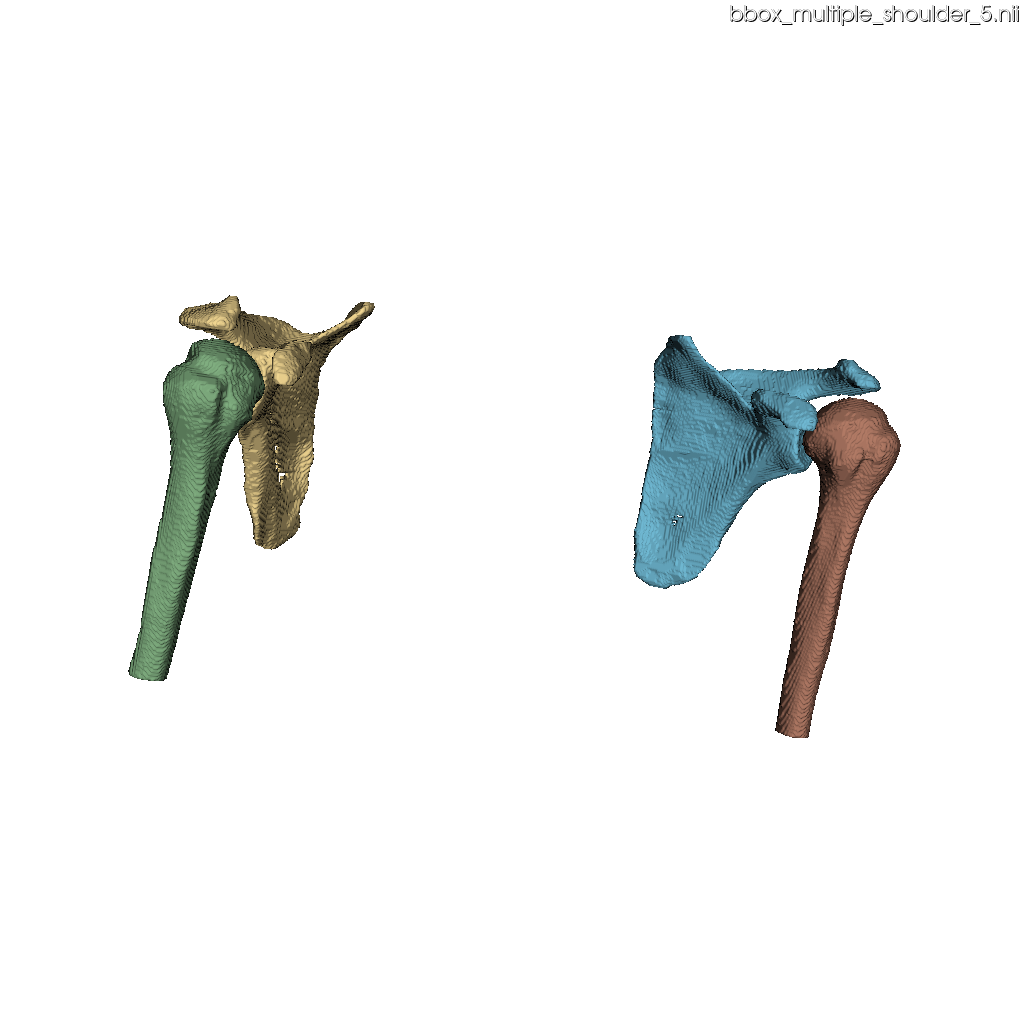} & \raisebox{-0.6\height}[0pt][0pt]{\includegraphics[width=0.1\linewidth, trim=400 170 410 90, clip]{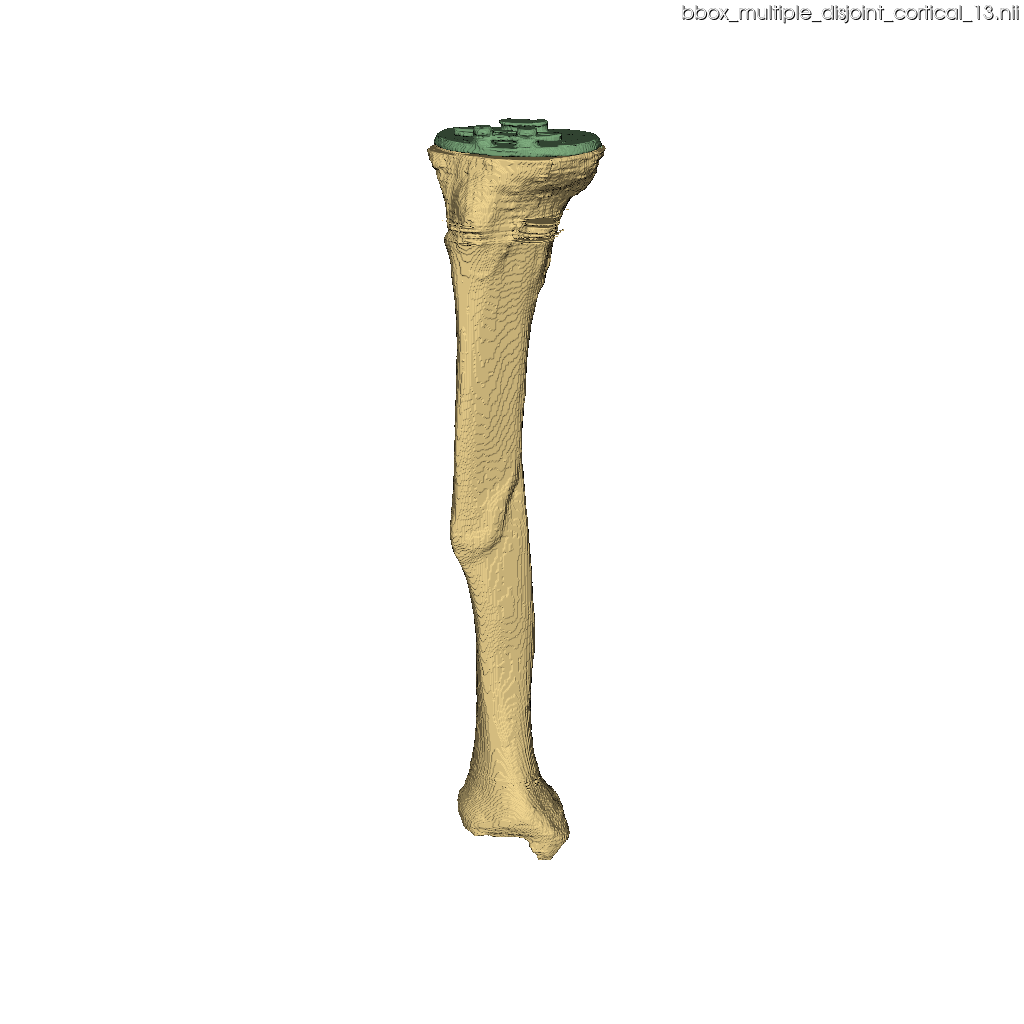}} 
    \\
    & \includegraphics[width=0.14\linewidth, trim=370 290 370 280, clip]{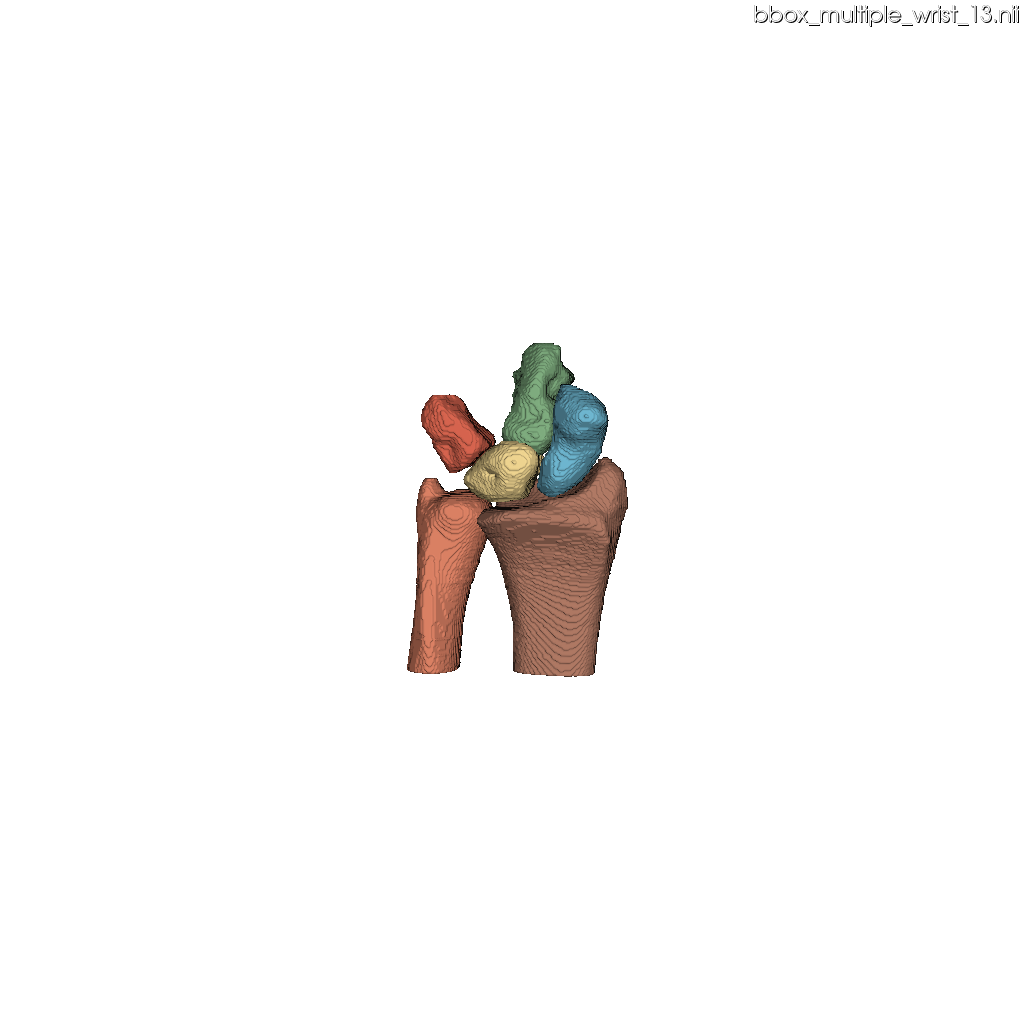} & &
    \includegraphics[width=0.14\linewidth, trim=370 290 370 280, clip]{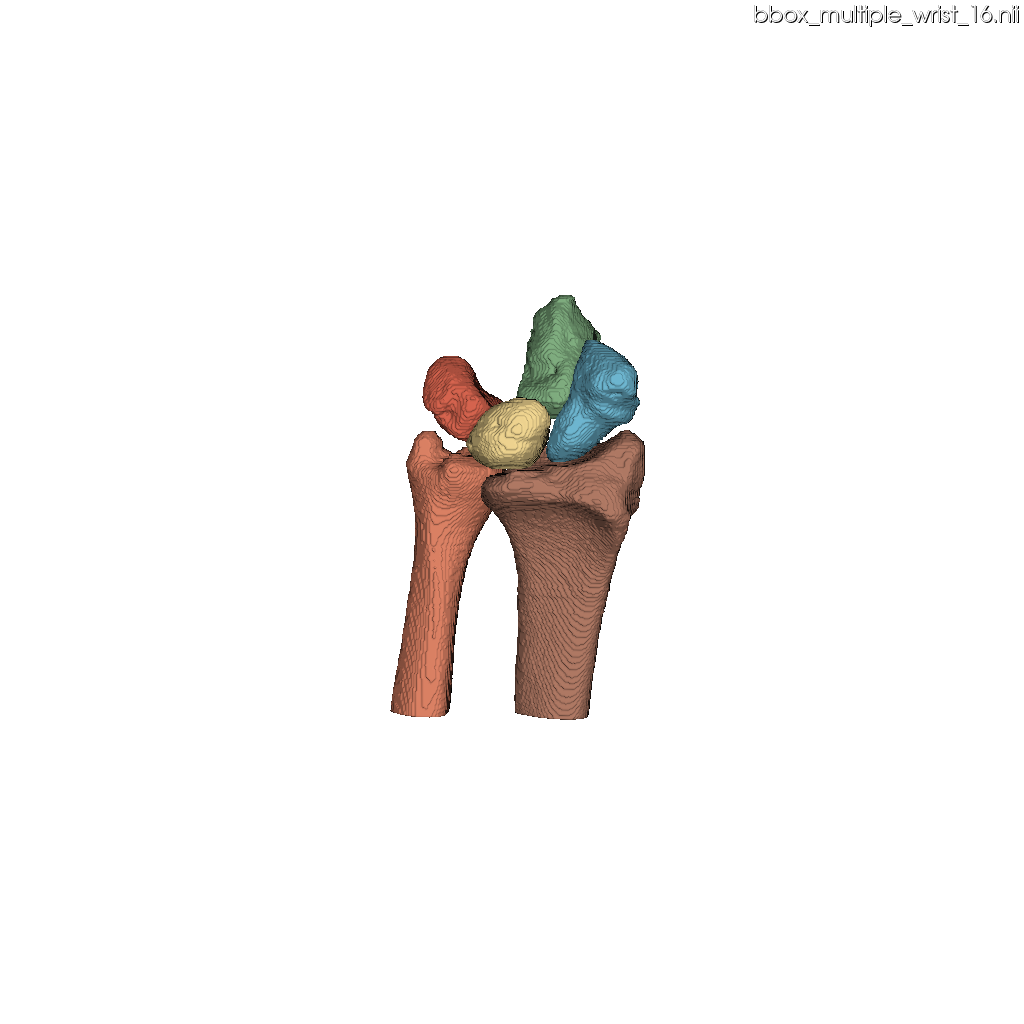} & &
    \includegraphics[width=0.14\linewidth, trim=370 290 370 280, clip]{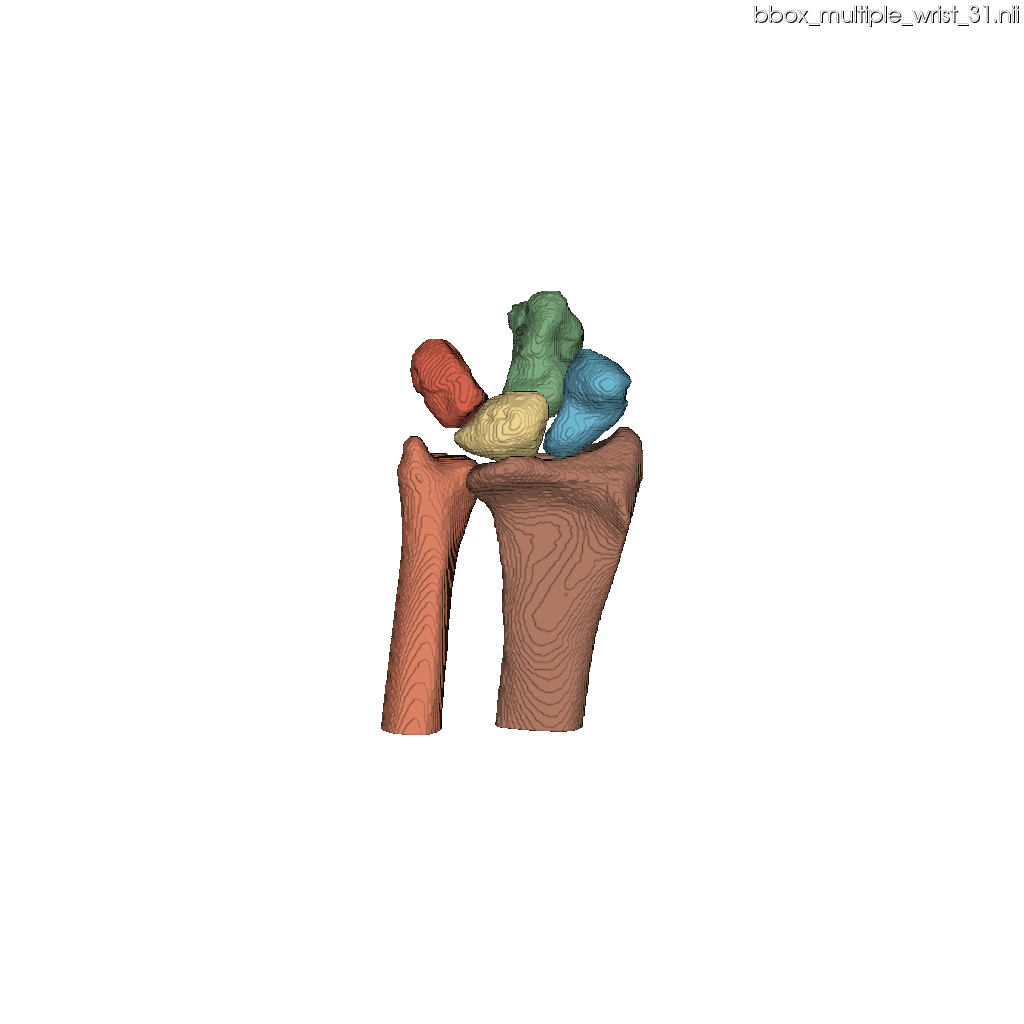} & \\ 
    \hline
    \multirow{2}{*}{\cblacksquaredot[0.6]{pomme}} & \includegraphics[width=0.14\linewidth, trim=580 380 100 340, clip]{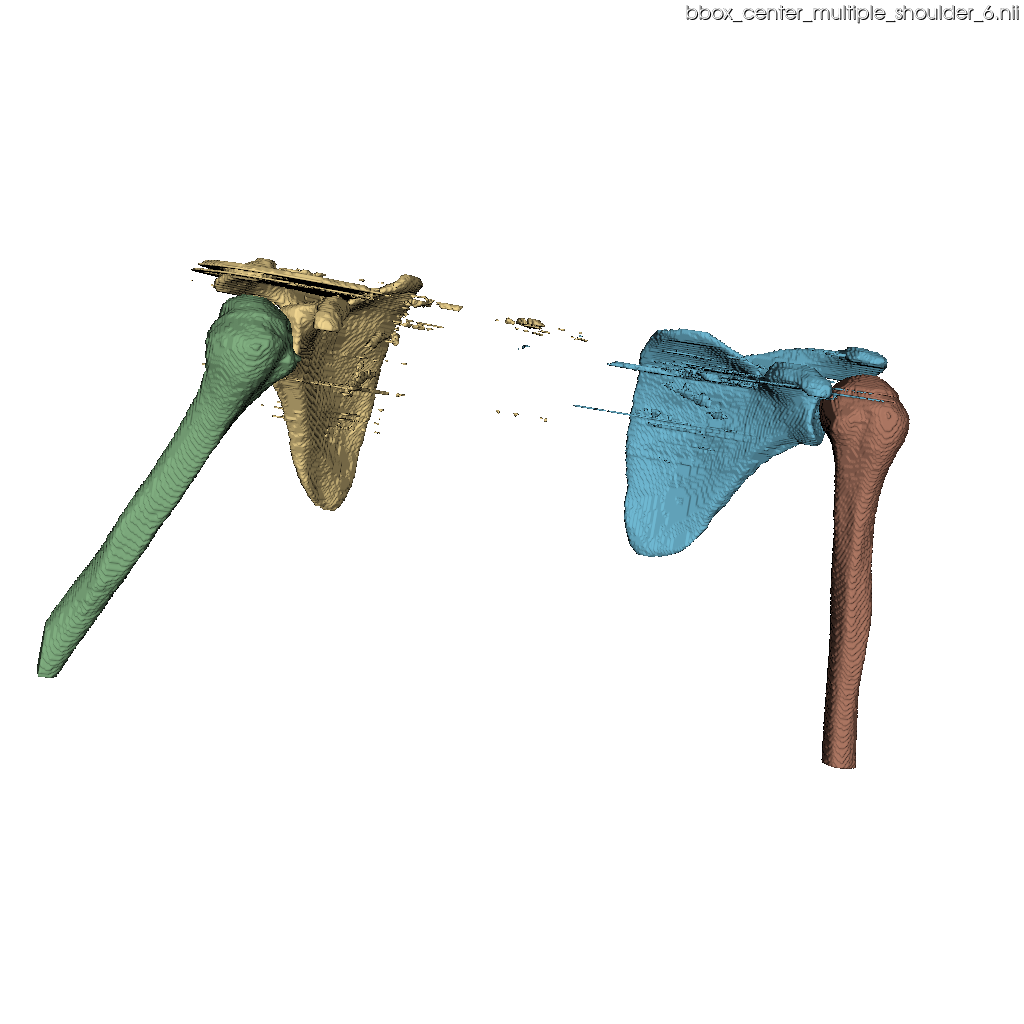} & \raisebox{-0.6\height}[0pt][0pt]{\includegraphics[width=0.11\linewidth, trim=400 200 430 150, clip]{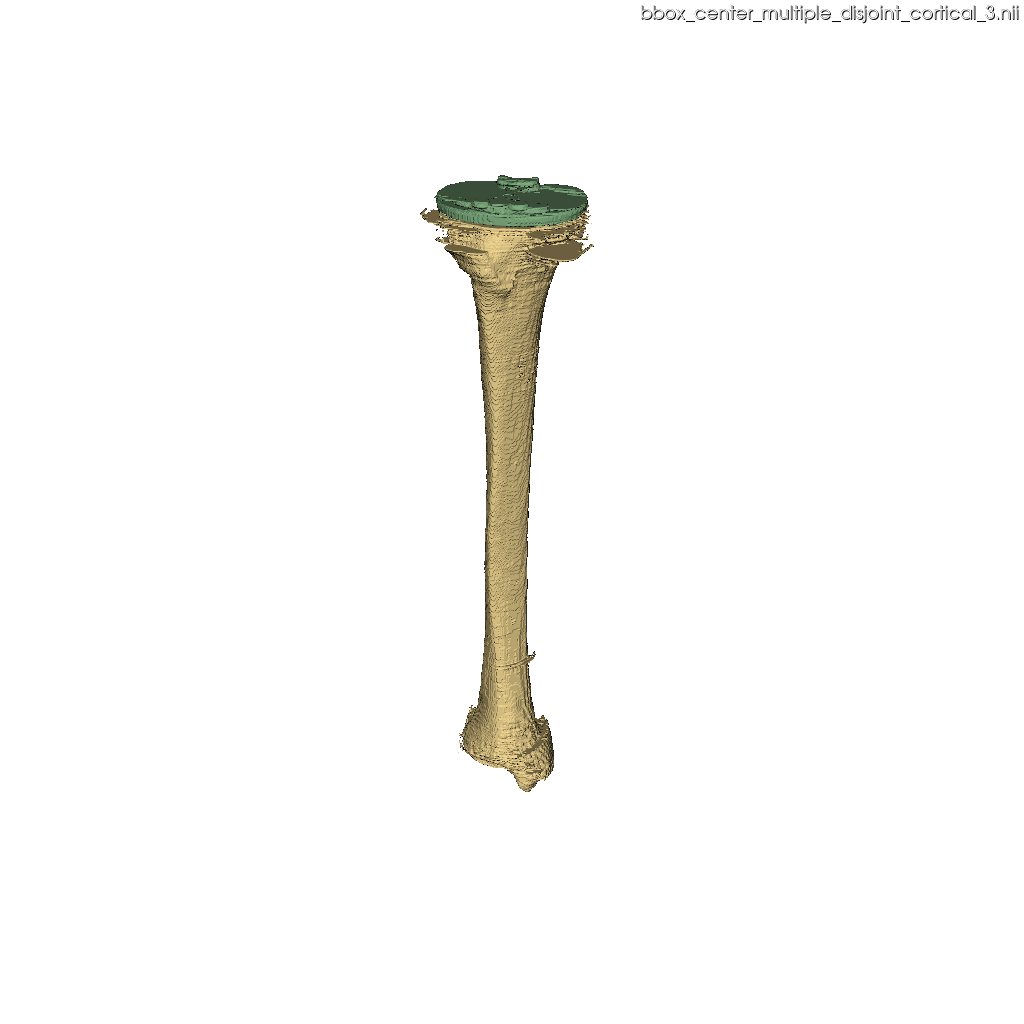}} &
    \includegraphics[width=0.1\linewidth, trim=560 380 190 330, clip]{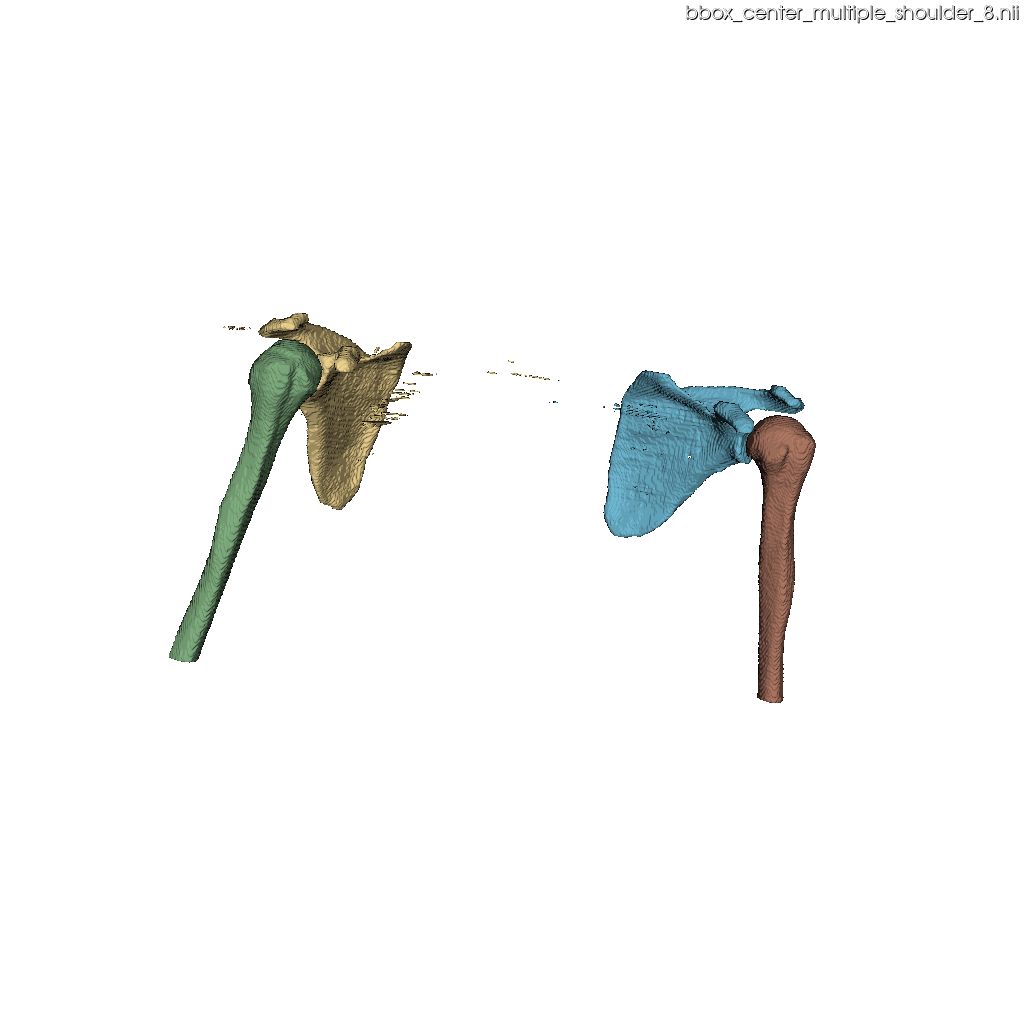} & \raisebox{-0.6\height}[0pt][0pt]{\includegraphics[width=0.11\linewidth, trim=400 200 400 110, clip]{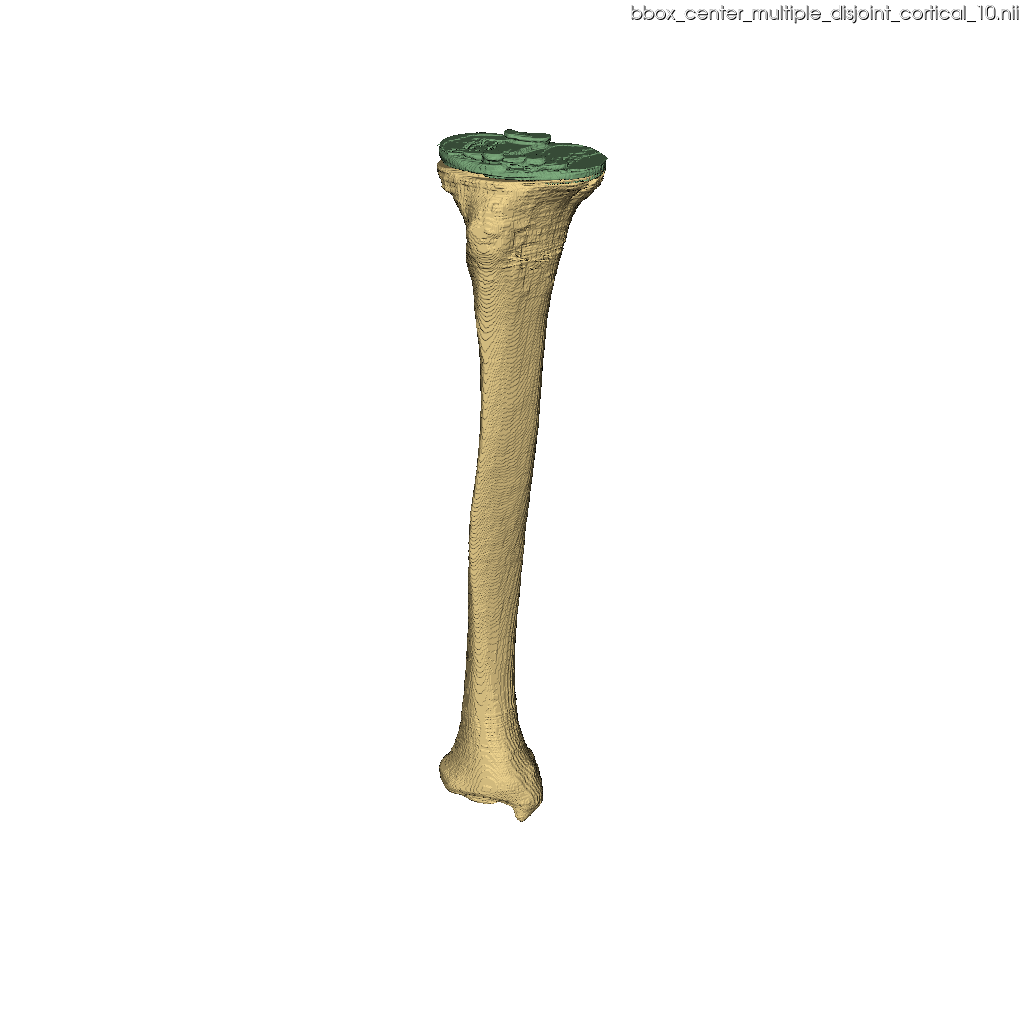}} &
    \includegraphics[width=0.14\linewidth, trim=560 380 170 330, clip]{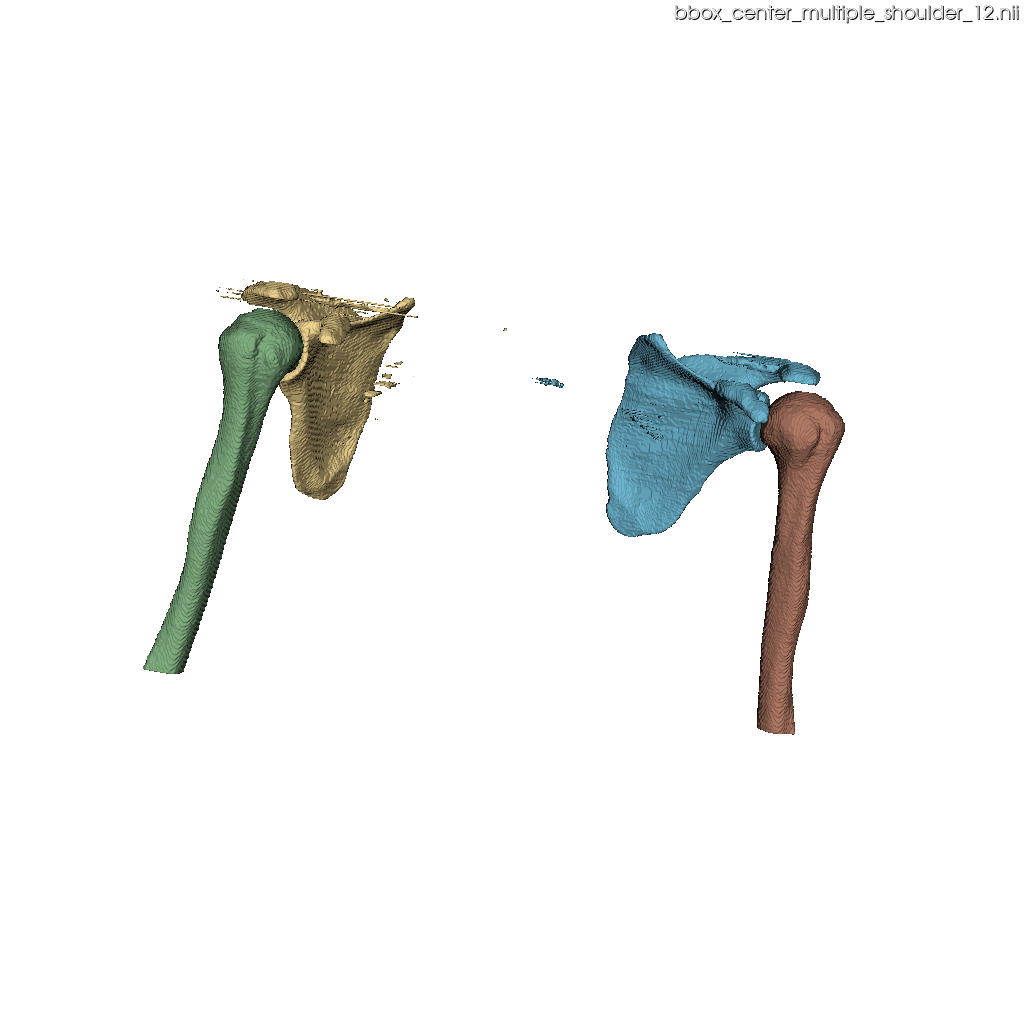} & \raisebox{-0.6\height}[0pt][0pt]{\includegraphics[width=0.11\linewidth, trim=400 150 400 110, clip]{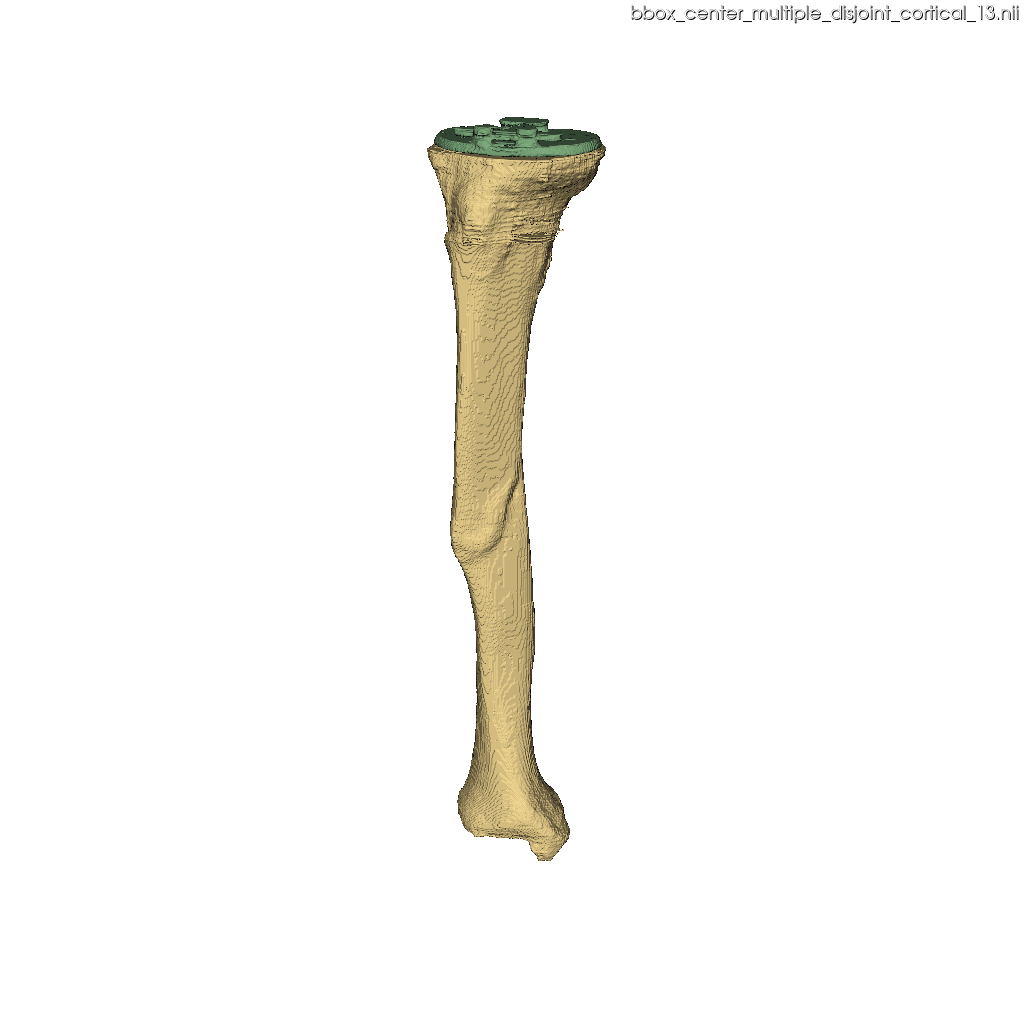}} 
    \\
    & \includegraphics[width=0.14\linewidth, trim=370 290 370 280, clip]{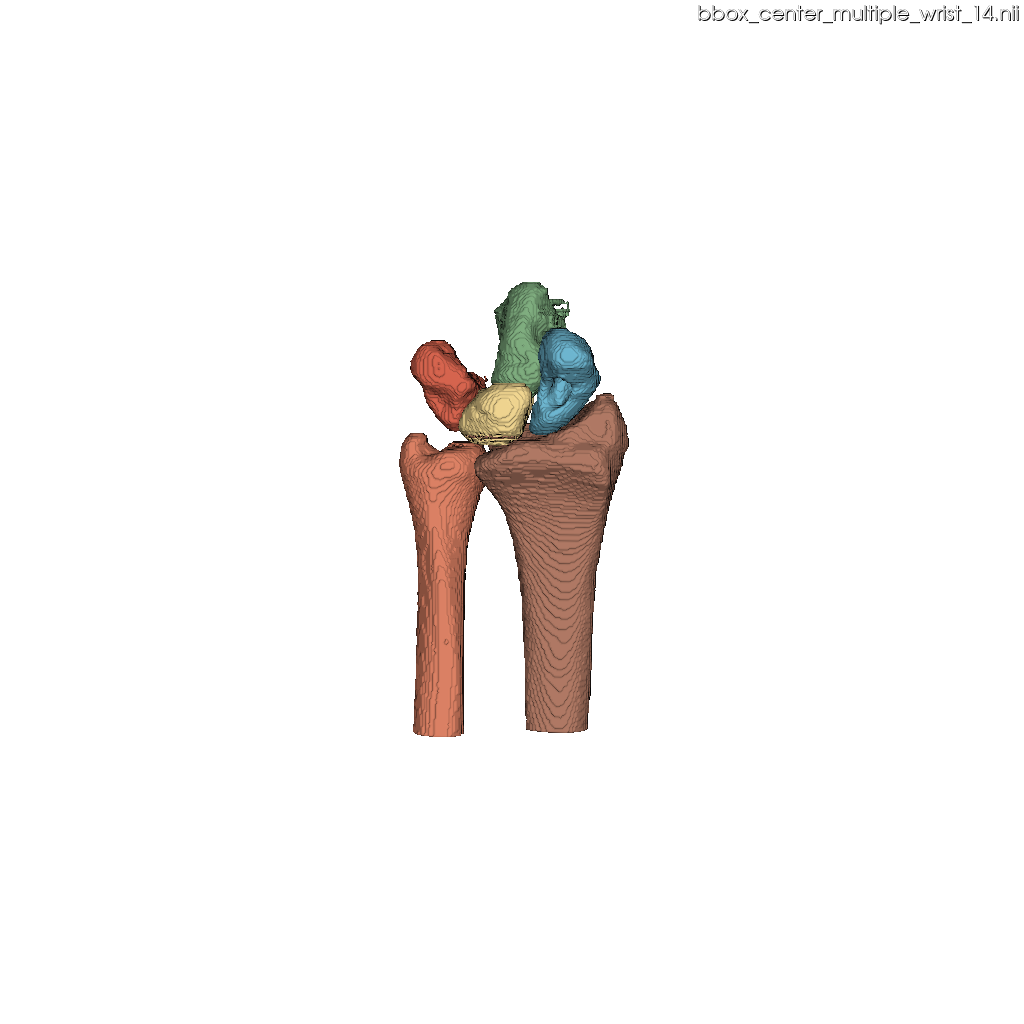} & &
    \includegraphics[width=0.14\linewidth, trim=370 290 370 280, clip]{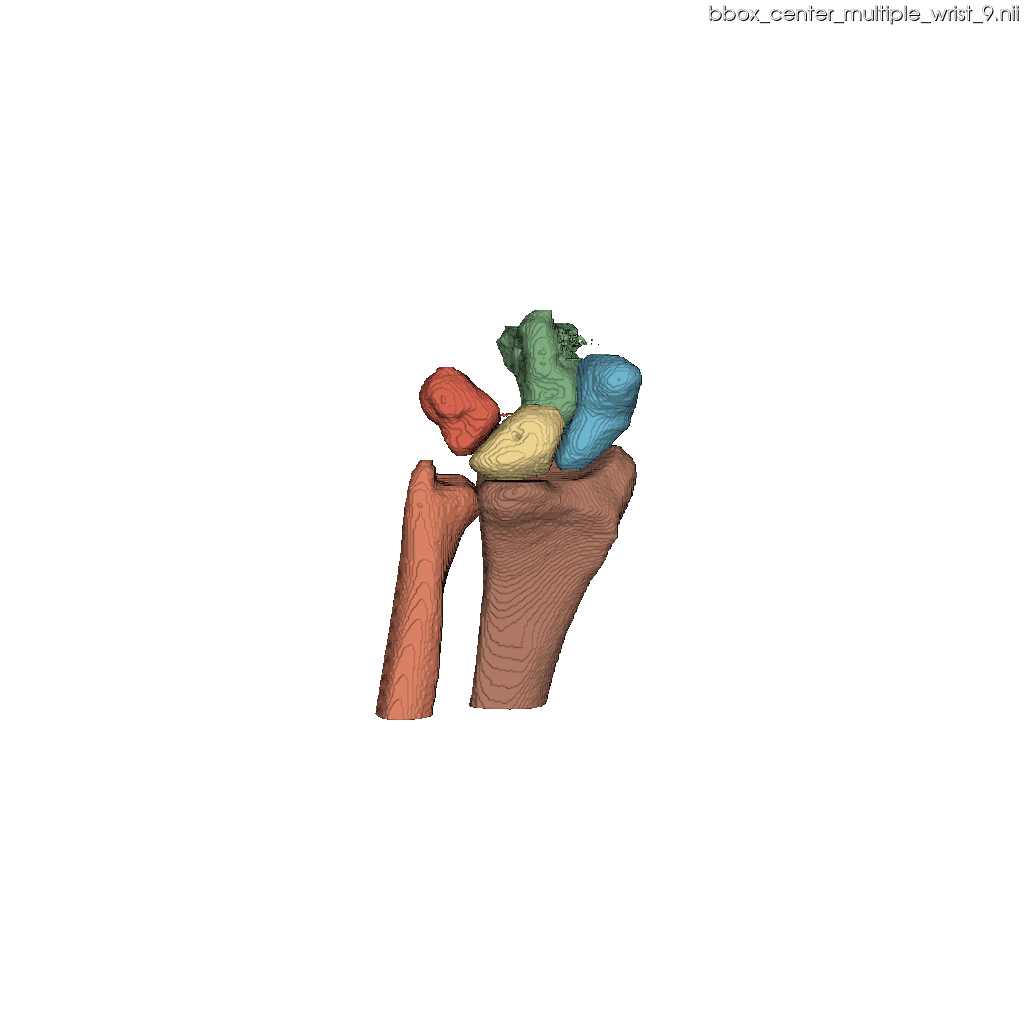} & &
    \includegraphics[width=0.14\linewidth, trim=370 290 370 280, clip]{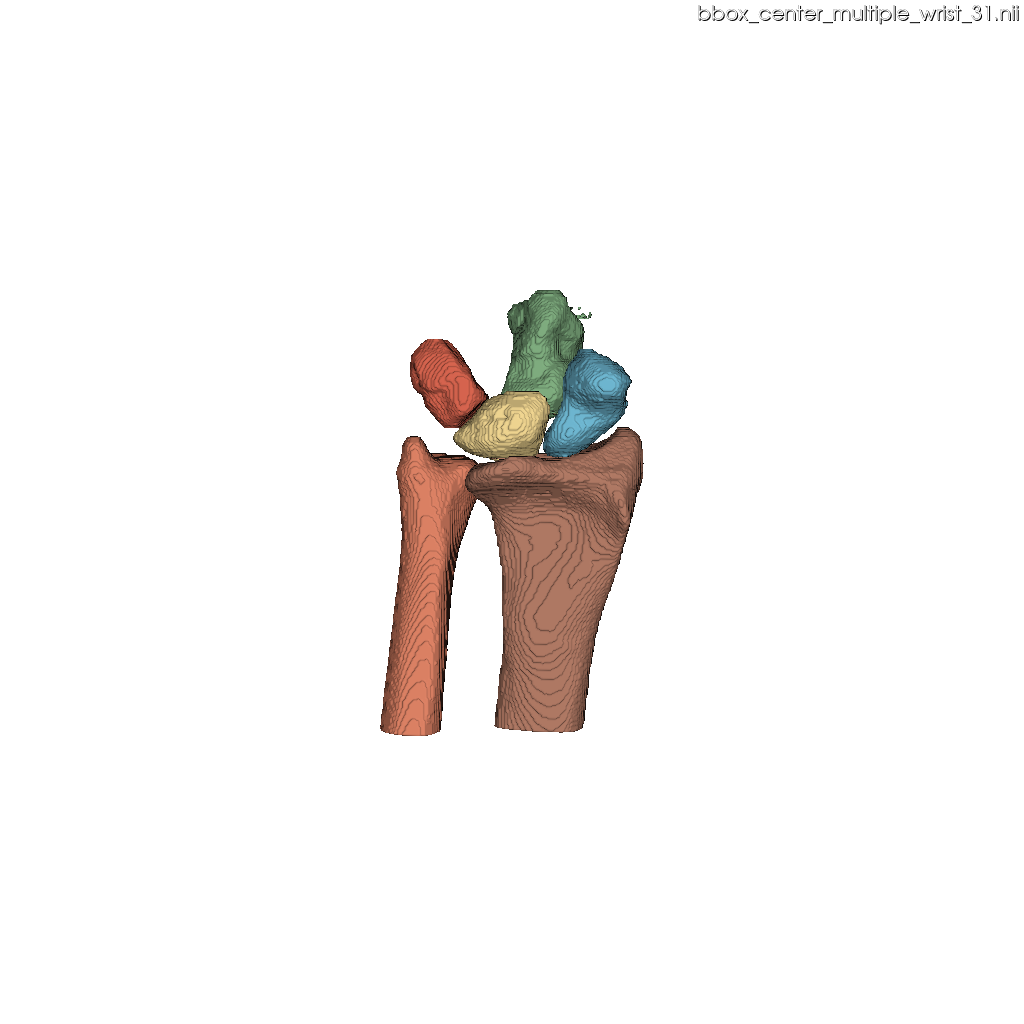} & \\ 
    \hline
    \multirow{2}{*}{\cblackstartriangledown[0.6]{pomme}} & \includegraphics[width=0.14\linewidth, trim=560 360 120 300, clip]{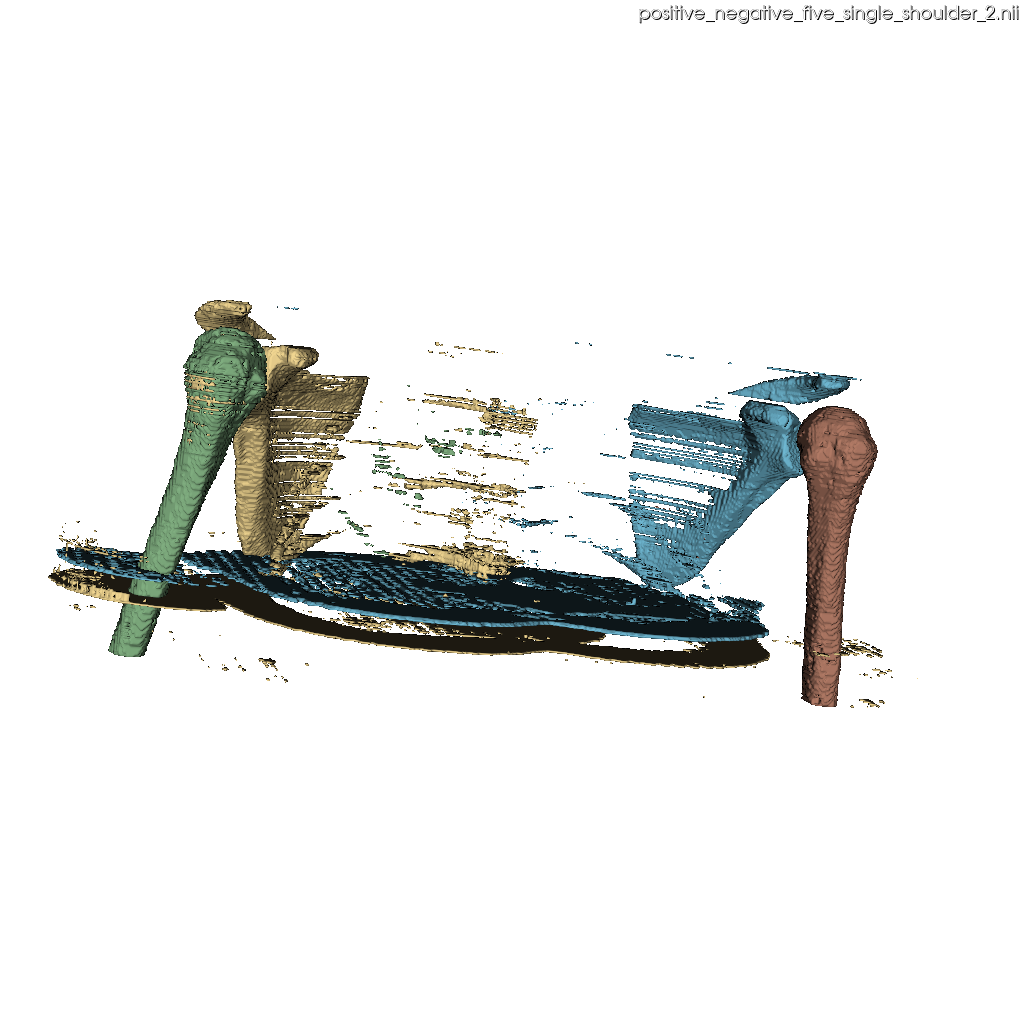} & \raisebox{-0.6\height}[0pt][0pt]{\includegraphics[width=0.11\linewidth, trim=300 100 370 30, clip]{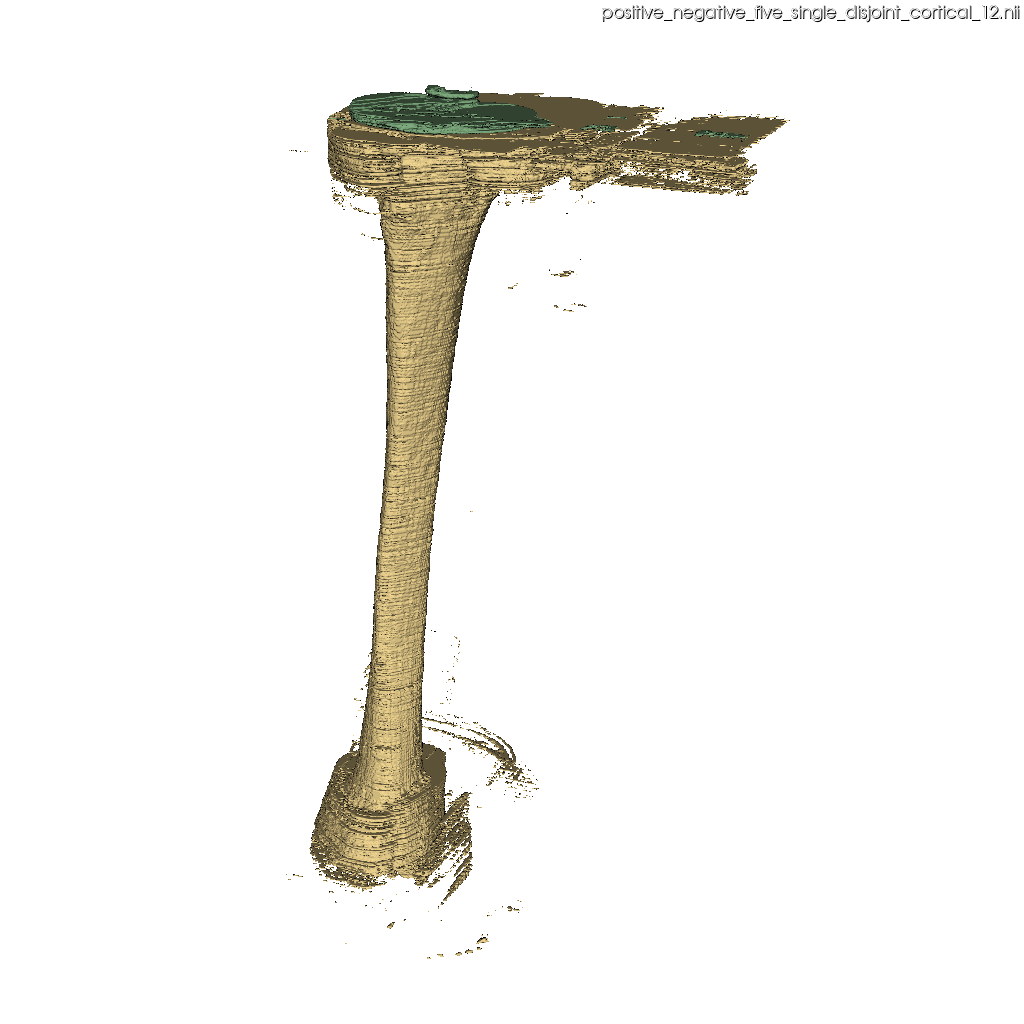}} &
    \includegraphics[width=0.14\linewidth, trim=560 360 110 300, clip]{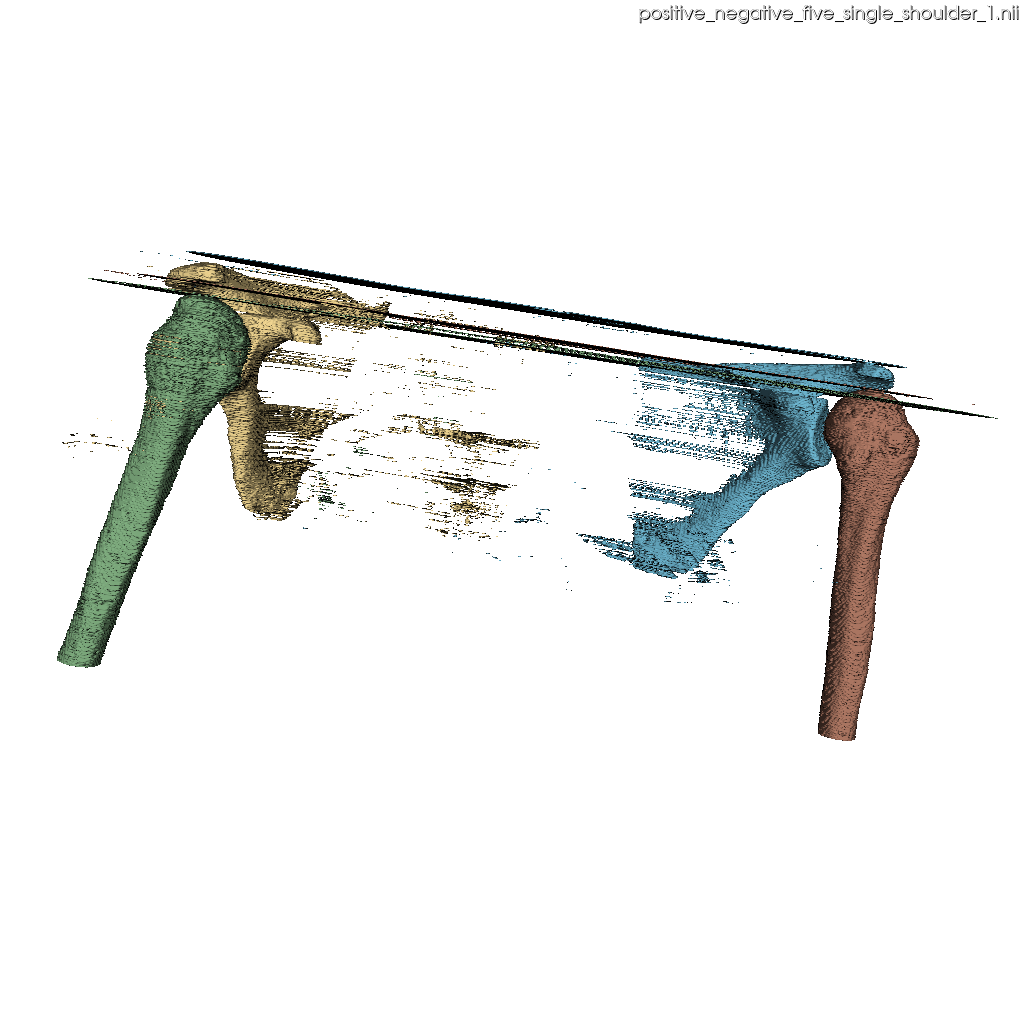} & \raisebox{-0.6\height}[0pt][0pt]{\includegraphics[width=0.11\linewidth, trim=320 100 370 40, clip]{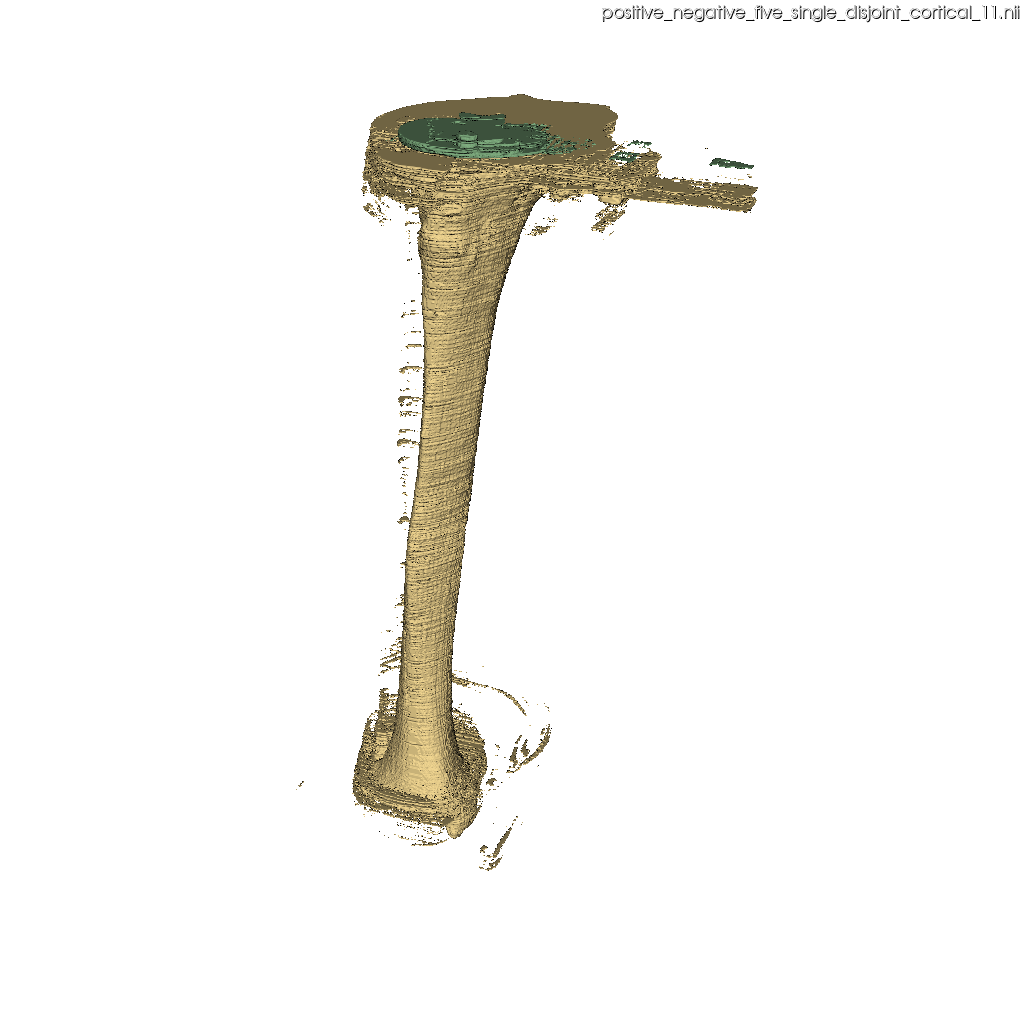}} &
    \includegraphics[width=0.14\linewidth, trim=570 340 70 300, clip]{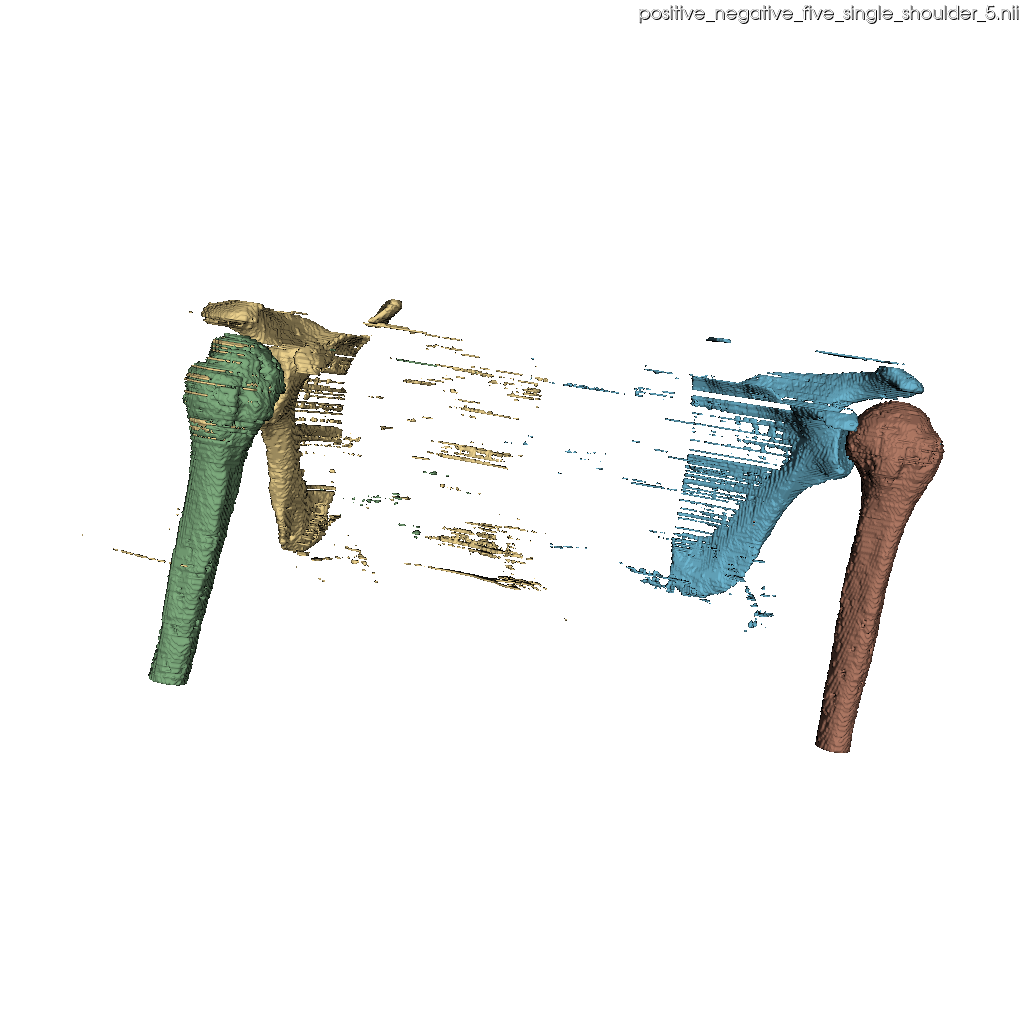} & \raisebox{-0.6\height}[0pt][0pt]{\includegraphics[width=0.11\linewidth, trim=420 150 370 70, clip]{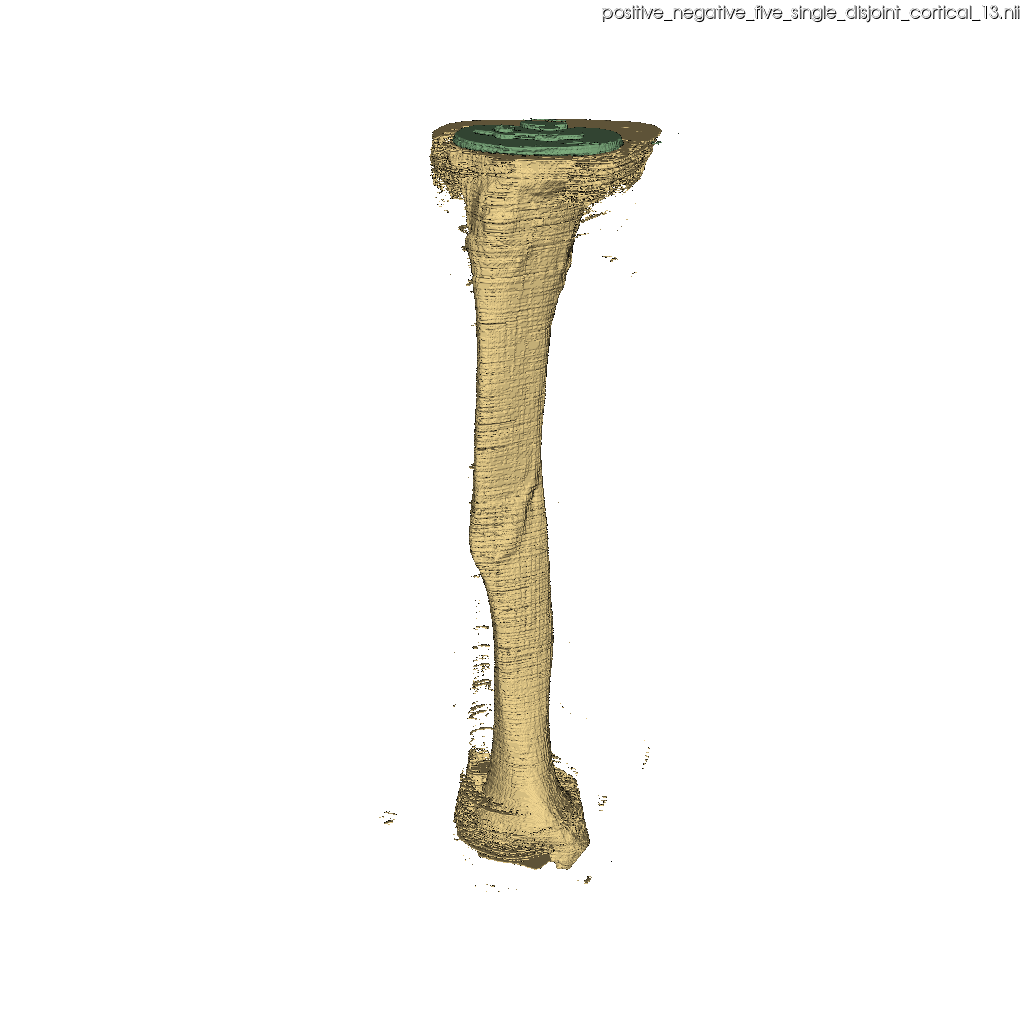}} 
    \\
    & \includegraphics[width=0.14\linewidth, trim=170 290 530 280, clip]{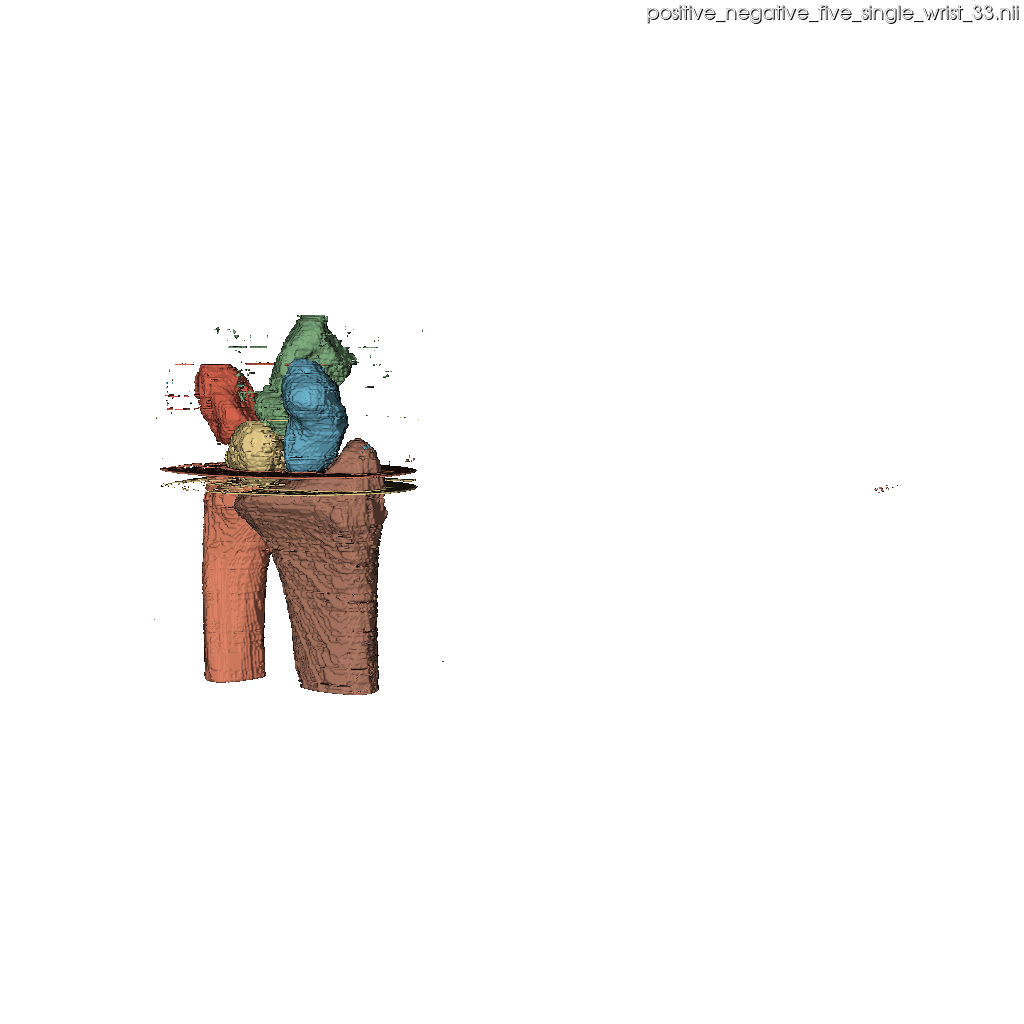} & &
    \includegraphics[width=0.14\linewidth, trim=380 290 380 280, clip]{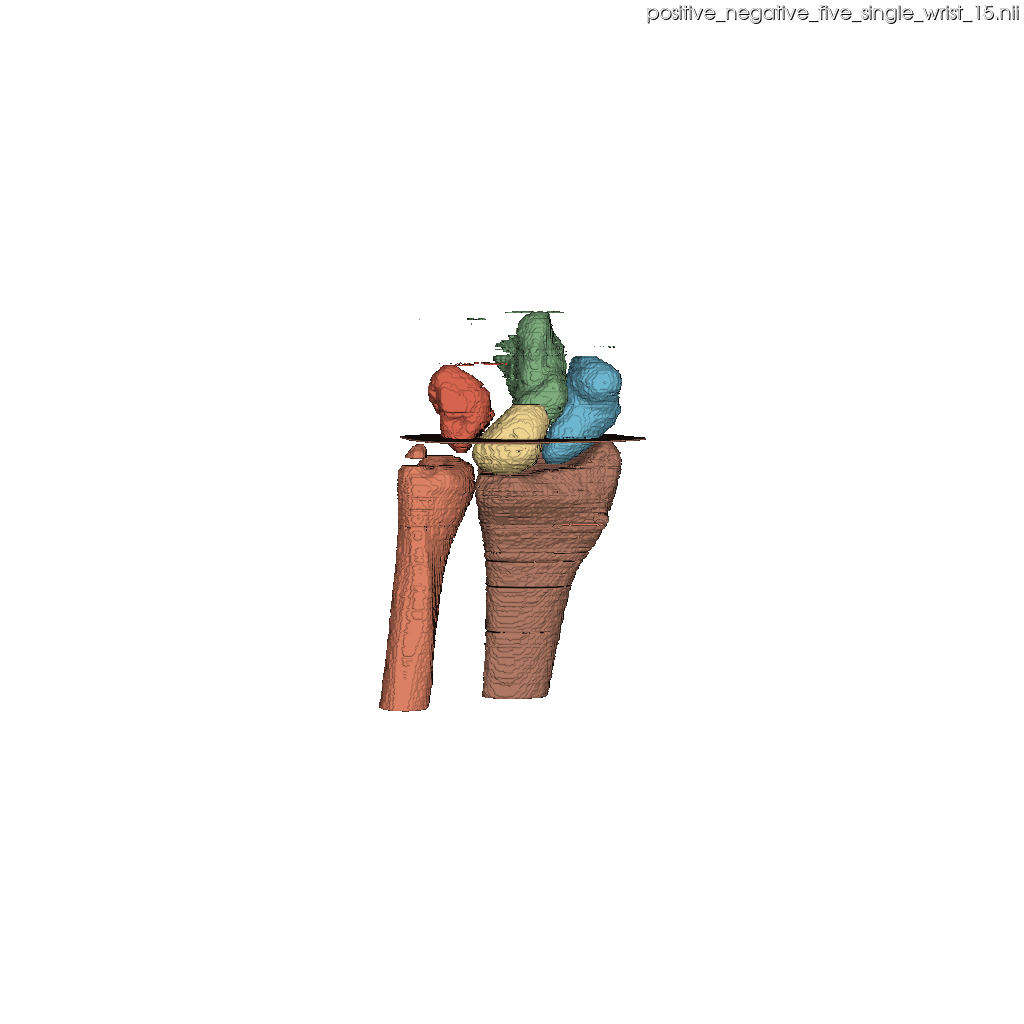} & &
    \includegraphics[width=0.14\linewidth, trim=370 290 370 280, clip]{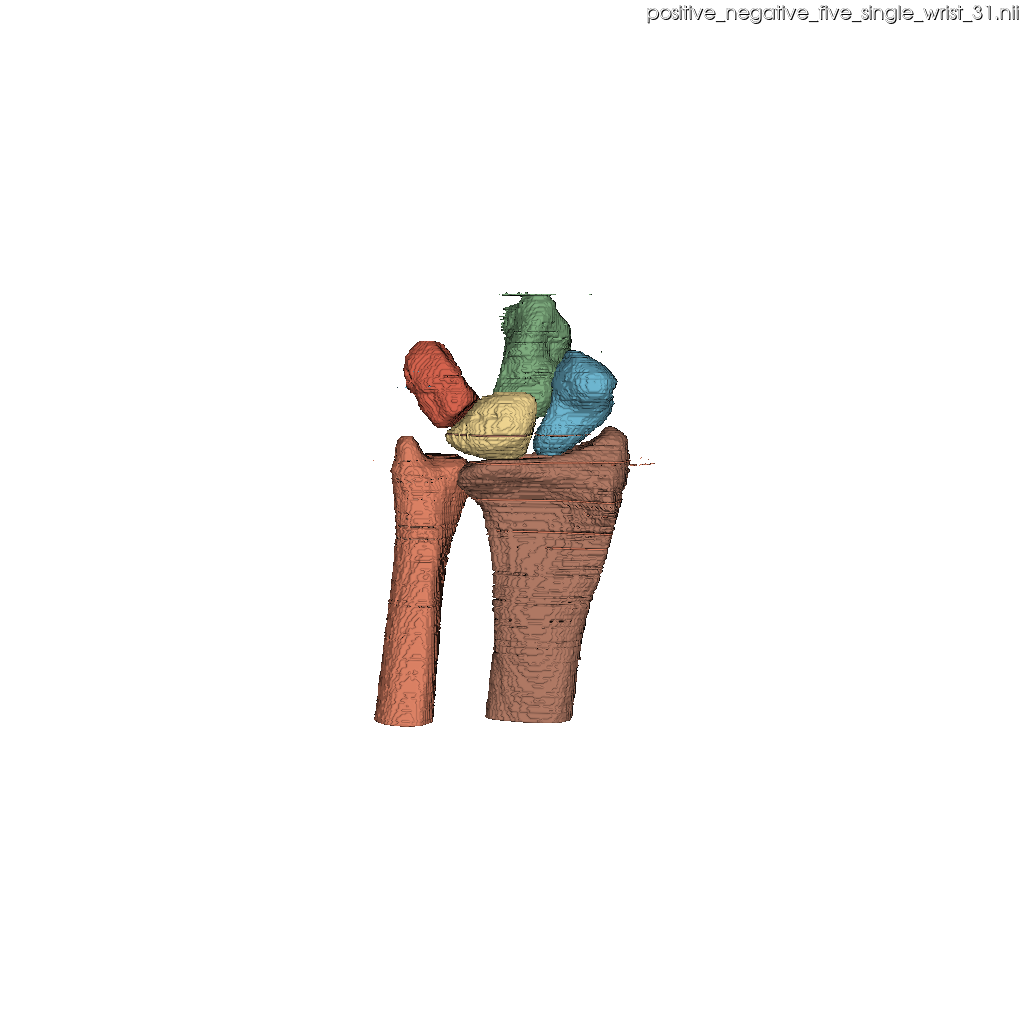} & \\ 
    \hline
\end{tabular}
\caption{Selected examples for \textit{\textsc{Sam2} B+} with low, medium and high DSC.}
\label{fig:example_sam2bplus}
\end{figure}

\newpage

\section{Segmentation performance}
\subsection{Performance for individual datasets}\label{sec:appendix_scatter_plots}
\begin{figure}[H]
    \begin{subfigure}[c]{1\textwidth}
        \centering
        \includegraphics[width=1.0\linewidth]{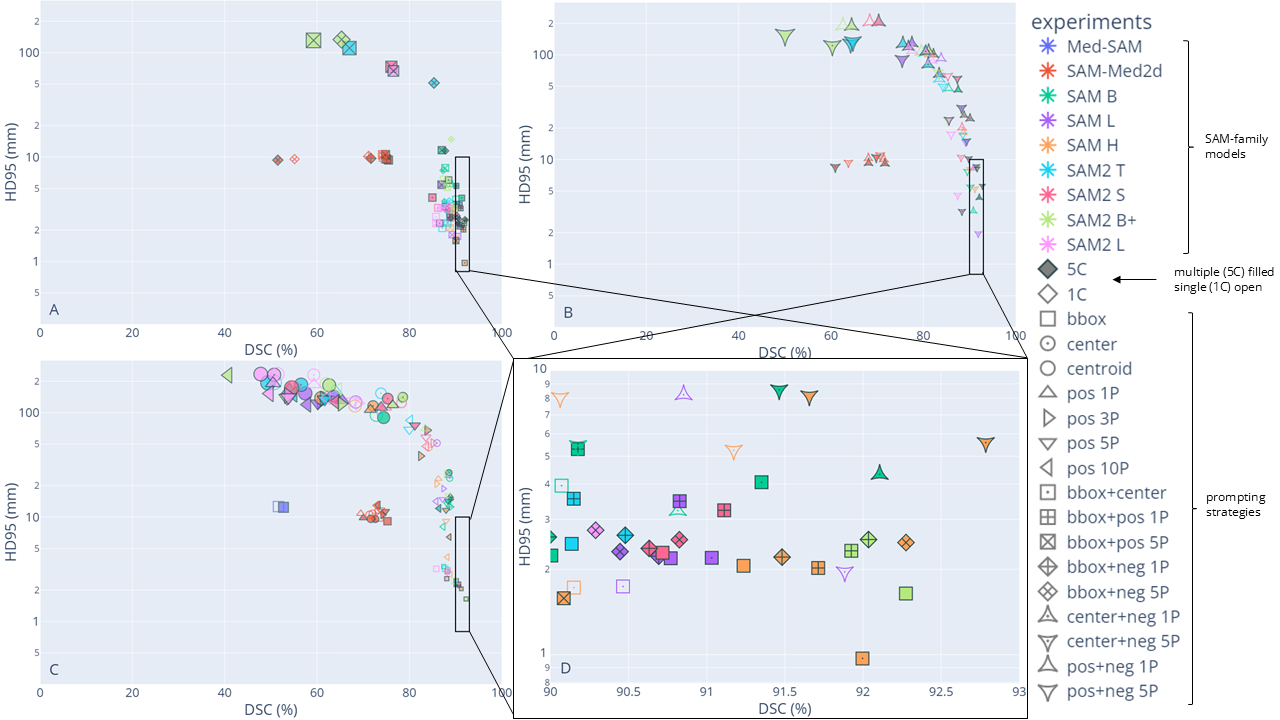}
        \caption{Knee dataset (D1)}
        \label{fig:scatter_plot_shoulder_datasets}
    \end{subfigure}
    \begin{subfigure}[c]{1\textwidth}
        \centering
        \includegraphics[width=1.0\linewidth]{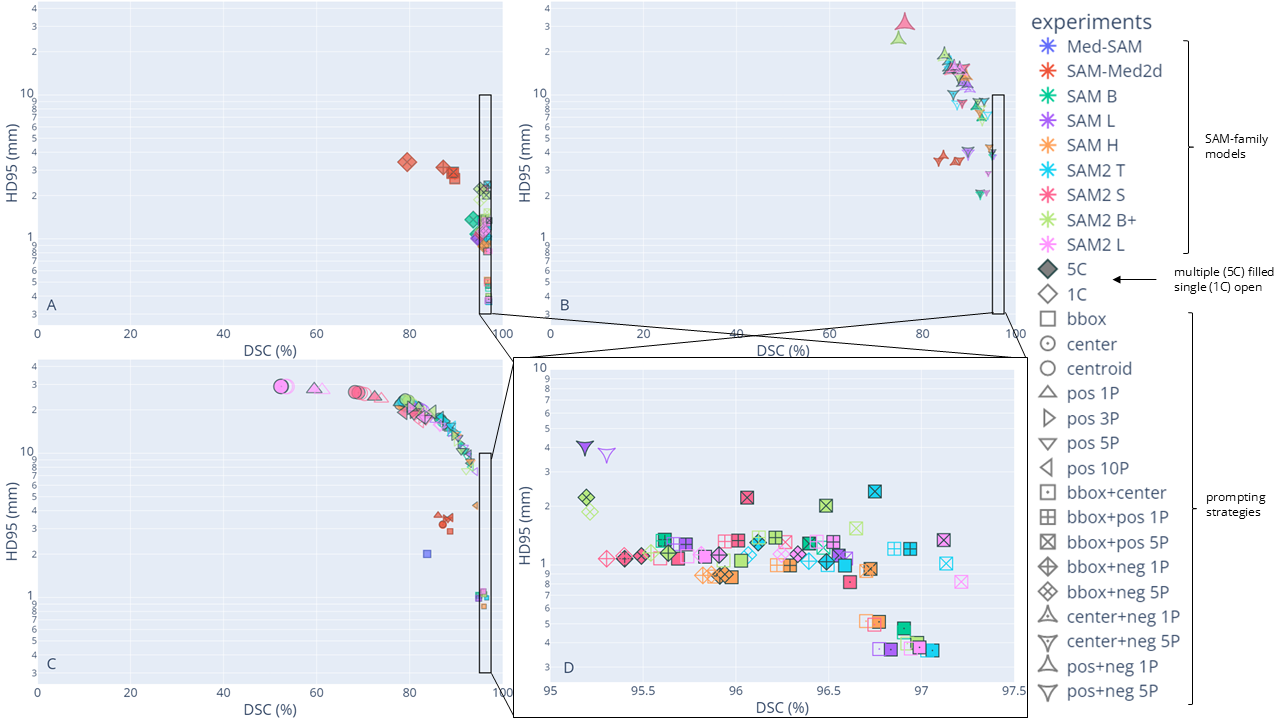}
        \caption{Wrist dataset (D2)}
    \end{subfigure}
\end{figure}%
\begin{figure}[H]\ContinuedFloat
    \begin{subfigure}[c]{1\textwidth}
        \centering
        \includegraphics[width=1.0\linewidth]{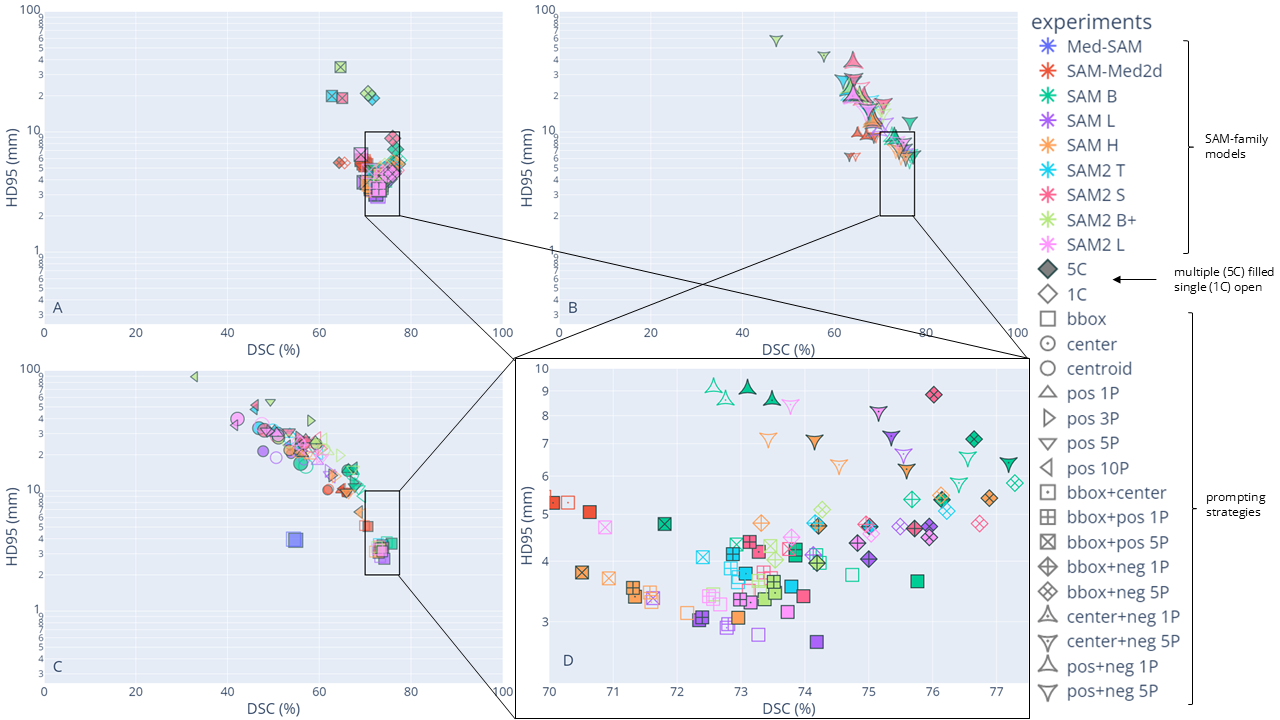}
        \caption{Knee dataset - cortical tibia bone (D3a)}
    \end{subfigure}
    \begin{subfigure}[c]{1\textwidth}
        \centering
        \includegraphics[width=1.0\linewidth]{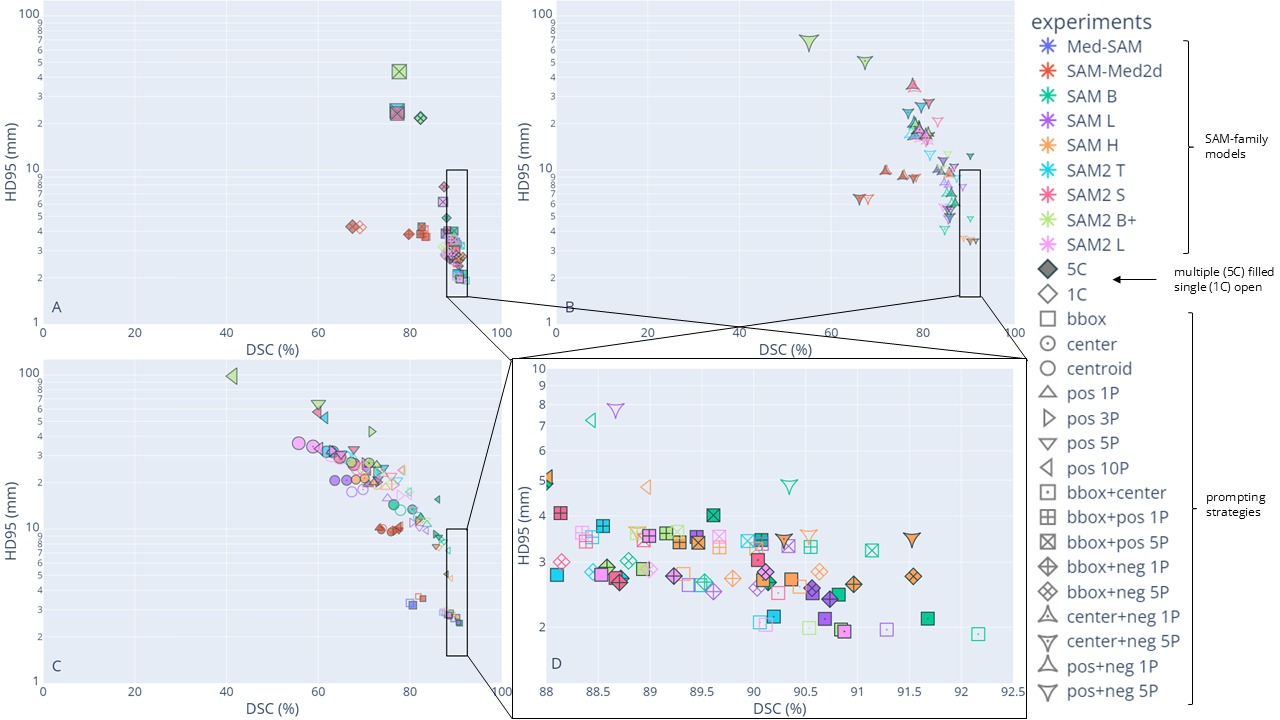}
        \caption{Knee dataset - full tibia bone (D3b)}
    \end{subfigure}
    \caption{Scatterplots for individual datasets of BPC prompts (A), PC prompts (B), OT prompts (C) and zoomed-in view to the lower right corner (D). The lower right corner, i.e., high DSC and low HD95 metrics, shows the performance strongest settings.}
    \label{fig:scatter_plot_details}
\end{figure}

\newpage
\subsection{Performance  with different number of points for individual datasets}\label{sec:appendix_barplot}
\begin{figure}[H]

    \begin{subfigure}[c]{1\textwidth}
    \centering
        \includegraphics[width=0.48\linewidth]{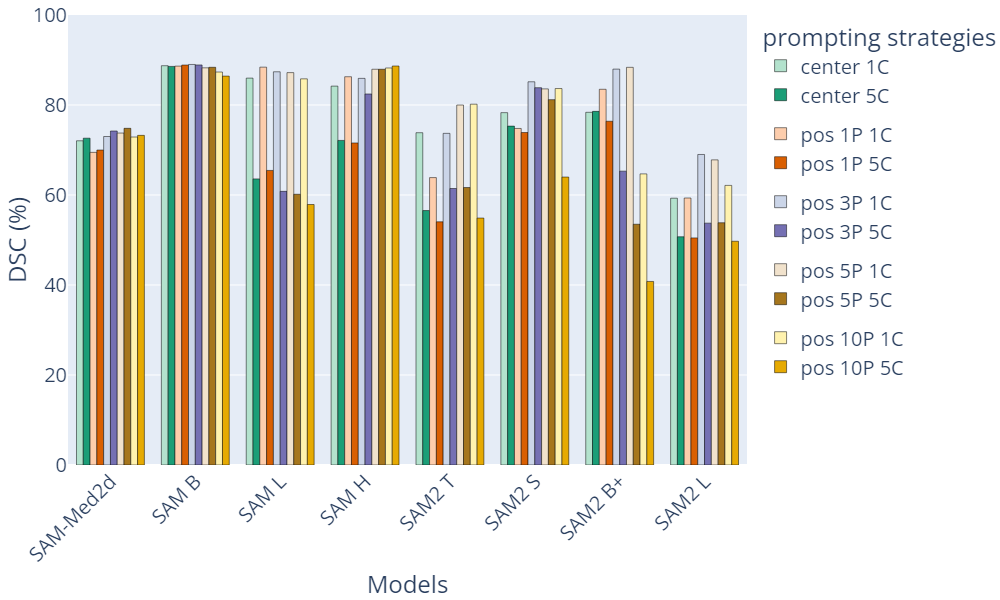}
        \hfill
        \includegraphics[width=0.48\linewidth]{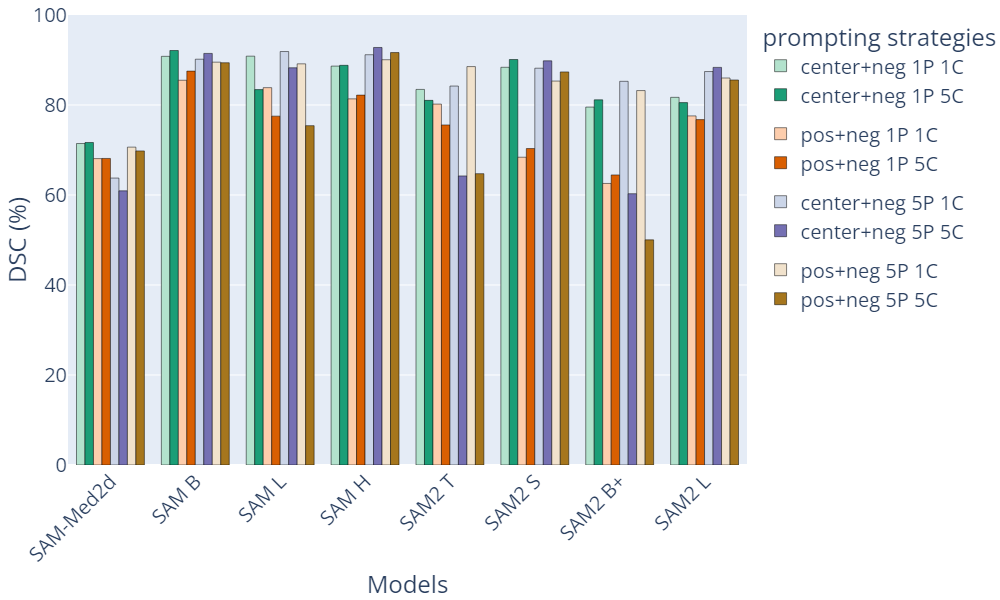}
        \vspace{-1em}
        \caption{Shoulder dataset (D1)}
        \label{fig:performance_points_shoulder}
    \end{subfigure}
\centering
    \begin{subfigure}[c]{1\textwidth}
    \centering
        \includegraphics[width=0.48\linewidth]{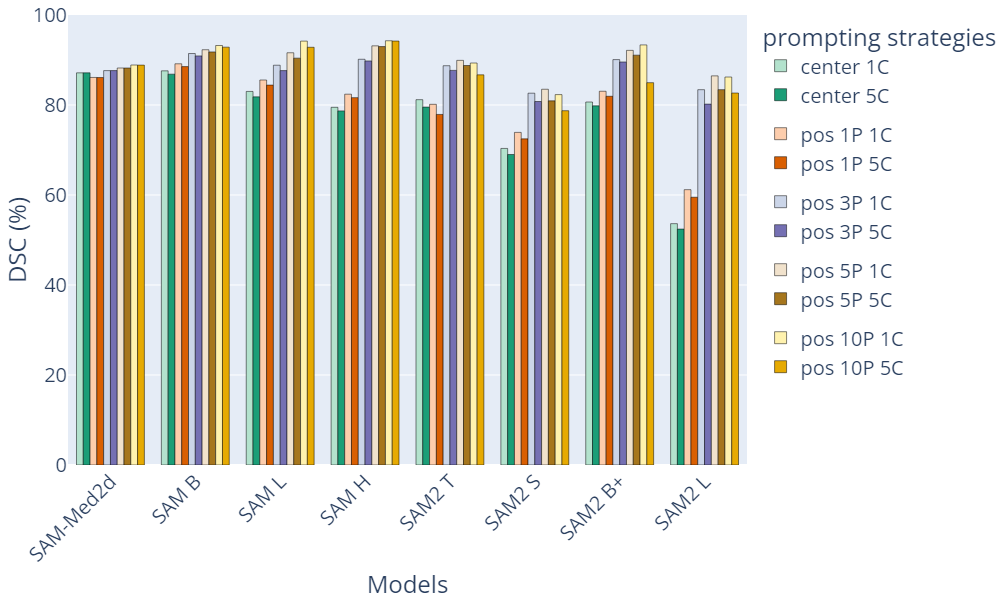}
        \hfill
        \includegraphics[width=0.48\linewidth]{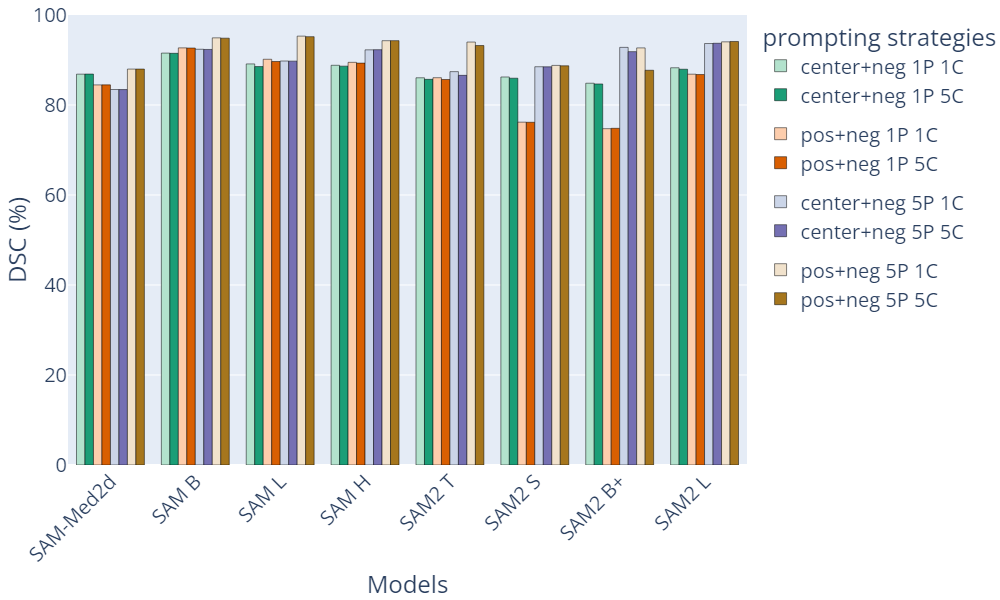}
        \vspace{-1em}
        \caption{Wrist dataset (D2)}
        \label{fig:performance_points_wrist}
    \end{subfigure}
\centering
    \begin{subfigure}[c]{1\textwidth}
    \centering
        \includegraphics[width=0.48\linewidth]{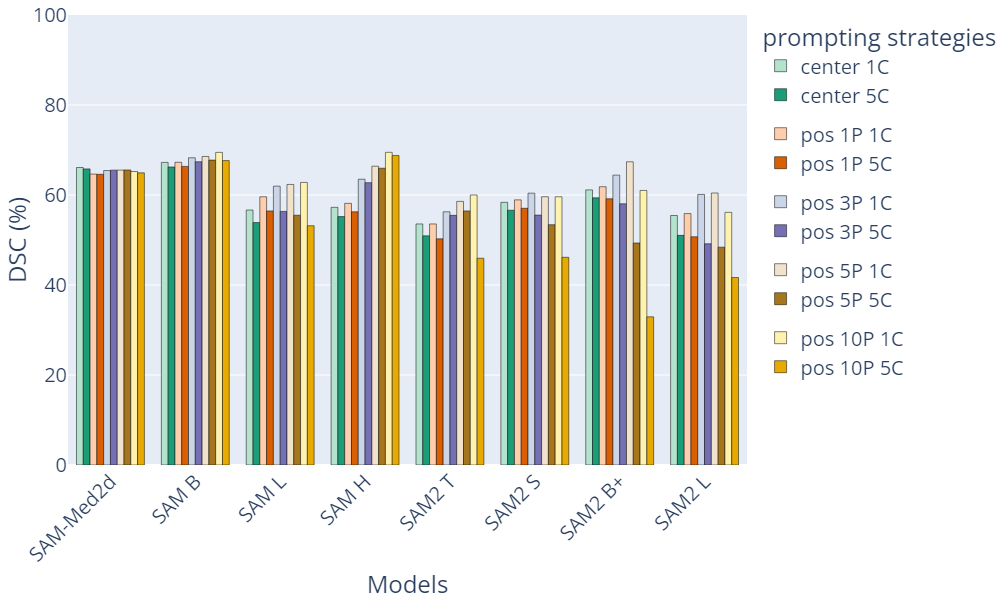}
        \hfill
        \includegraphics[width=0.48\linewidth]{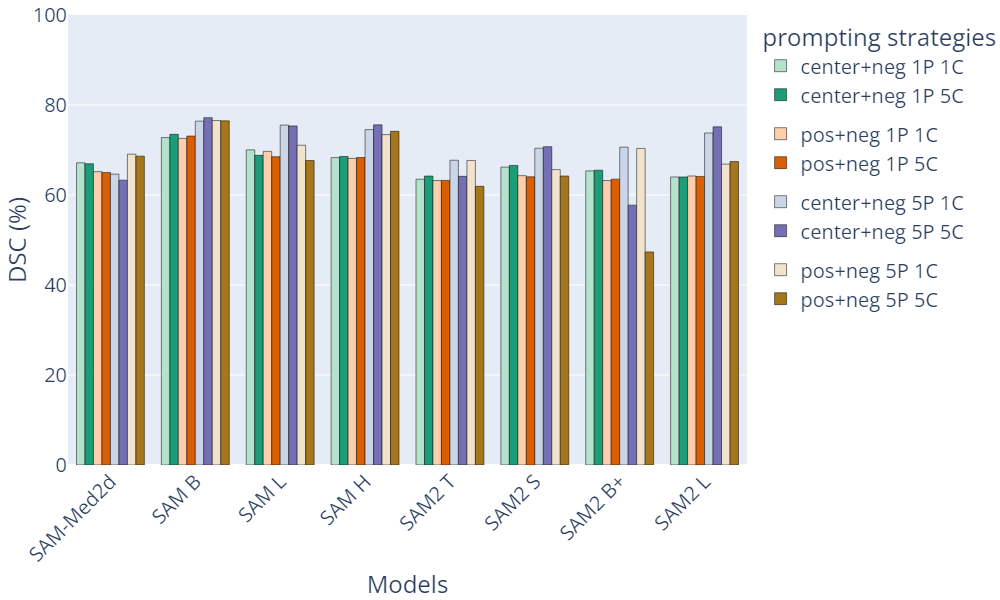}
        \vspace{-1em}
        \caption{Knee dataset - cortical bone (D3a)}
        \label{fig:performance_points_kneec}
    \end{subfigure}
\centering
    \begin{subfigure}[c]{1\textwidth}
    \centering
        \includegraphics[width=0.48\linewidth]{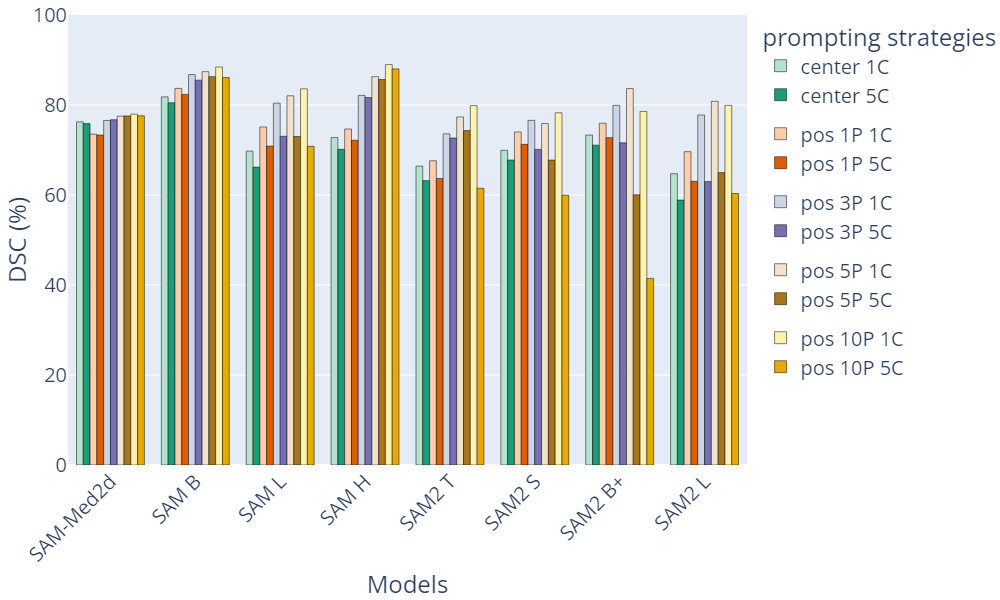}
        \hfill
        \includegraphics[width=0.48\linewidth]{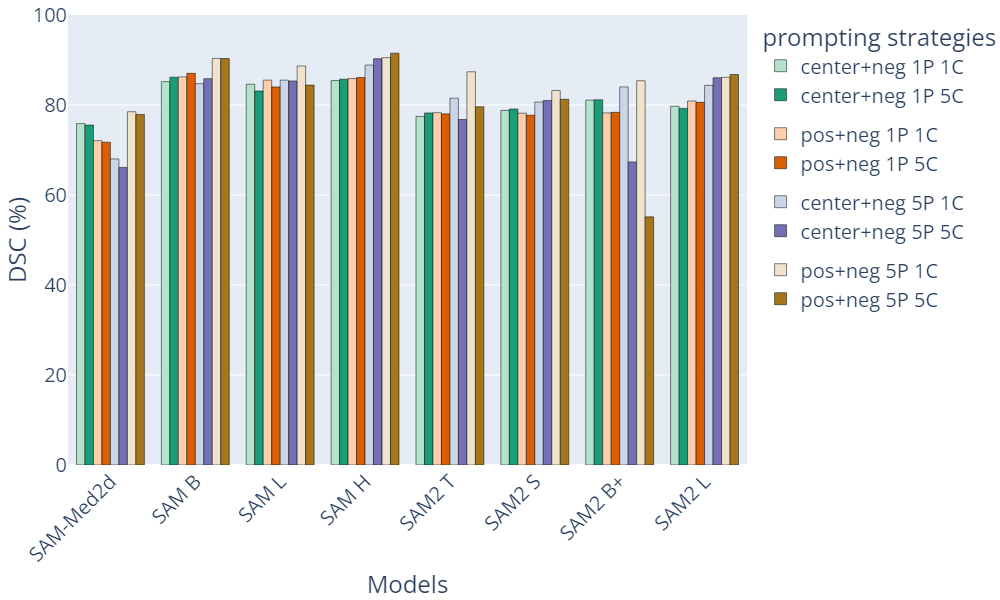}
        \vspace{-1em}
        \caption{Knee dataset - full bone (D3b)}
        \label{fig:performance_points_kneef}
    \end{subfigure}
    \caption{DSC (\%) performance for different number of points per model per datasets: center point and 1,3,5,10 random positive points (left); point combinations (right).}
    \label{fig:performance_points_detail}
\end{figure}

\newpage
\section{Performance of best performing prompts for all datasets}
\label{sec:appendix_results_detailed}
\footnotesize
\renewcommand{\arraystretch}{1.2}
\setlength{\tabcolsep}{2pt}
\begin{table}[H]
\centering
\caption{Best 2D prompting strategies for individual datasets and dedicated nnUNet model as comparison (\cblackstar[0.4]{yellow} for 2D, \cwhitestar[0.4]{yellow} for 3D full resolution). The ranking is averaged over all available models (i.e., 9 models for \cblacksquare[0.3]{black} and \cwhitesquare[0.3]{black}; 7 models for the others) and considers the position of the prompt ranking with respect to DSC (\%).}
\label{tab:best_settings_detail}
\begin{tabular}{|c|ccc|}
\hline
Prompt &  avg Ranking & \makecell{DSC (\%) \\ avg (std)} & \makecell{HD95 (\%) \\ avg (std)}\\
\hline
\multicolumn{4}{|l|}{D1 - Shoulder dataset} \\
\hline
\cblacksquarecross[0.3]{black}         &  3.25 &  88.72 (5.7) &    4.01 (3.4) \\
\cblacksquare[0.3]{black}              &  4.00 &  84.58 (5.6) &    4.11 (2.2) \\
\cblackdiamondcross[0.3]{black}        &  5.88 &  88.24 (3.6) &    3.36 (1.8) \\
\cblacksquaredot[0.3]{black}           &  6.75 &  87.22 (7.8) &    4.05 (4.6) \\
\cwhitesquarecross[0.3]{black}         &  8.38 &  87.19 (7.0) &    4.22 (1.9) \\
\cwhitesquaredot[0.3]{black}           &  9.88 &  86.51 (8.1) &    3.89 (2.8) \\
\cwhitestartriangledowndot[0.3]{black} & 11.00 &  85.26 (8.0) &  13.85 (23.2) \\
\cwhitesquarex[0.3]{black}             & 11.75 &  86.41 (8.4) &    4.34 (2.9) \\
\cwhitesquare[0.3]{black}              & 12.00 &  82.33 (7.6) &    4.86 (2.0) \\
\cblackdiamondx[0.3]{black}            & 12.38 &  81.73 (8.7) &  26.99 (27.9) \\
\hdashline
\cblackstar[0.4]{yellow} & - & 98.03 (1.1) & 0.87 (0.2) \\
\cwhitestar[0.4]{yellow} & - & 98.51 (0.8) & 0.91 (0.2) \\
\hline
\multicolumn{4}{|l|}{D2 - Wrist dataset} \\
\hline
\cblacksquaredot[0.3]{black}        &  1.62 &  95.98 (1.0) &  0.74 (0.6) \\
\cwhitesquaredot[0.3]{black}        &  2.38 &  95.97 (1.0) &  0.70 (0.4) \\
\cwhitesquarex[0.3]{black}          &  2.50 &  95.81 (1.0) &  1.35 (1.4) \\
\cblacksquarex[0.3]{black}          &  3.75 &  95.69 (1.3) &  1.78 (2.8) \\
\cblacksquarecross[0.3]{black}      &  5.00 &  95.30 (1.7) &  1.47 (1.3) \\
\cwhitesquarecross[0.3]{black}      &  5.75 &  95.24 (1.7) &  1.46 (1.3) \\
\cblacksquare[0.3]{black}           &  7.22 &  93.60 (2.7) &  1.33 (1.0) \\
\cwhitesquare[0.3]{black}           &  8.67 &  93.51 (2.7) &  1.34 (1.0) \\
\cblackdiamondcross[0.3]{black}     & 10.88 &  94.49 (2.6) &  1.31 (1.0) \\
\cwhitestartriangledown[0.3]{black} & 12.12 &  92.75 (3.8) &  6.11 (8.6) \\
\hdashline
\cblackstar[0.4]{yellow} & - & 98.68 (0.7) & 1.05 (7.9) \\
\cwhitestar[0.4]{yellow} & - & 98.83 (0.4) & 0.33 (0.0) \\
\hline
\end{tabular}
\quad
\begin{tabular}{|c|ccc|}
\hline
Prompt &  avg Ranking & \makecell{DSC (\%) \\ avg (std)} & \makecell{HD95 (\%) \\ avg (std)}\\
\hline
\multicolumn{4}{|l|}{D3a - Knee cortical datset} \\
\hline
\cwhitediamondx[0.3]{black}     & 4.00 &  74.59 (13.4) &   5.13 (3.8) \\
\cblackdiamondcross[0.3]{black} & 4.62 &  74.32 (15.7) &   4.66 (3.4) \\
\cblacksquare[0.3]{black}       & 5.33 &  71.47 (18.1) &   3.53 (2.4) \\
\cwhitediamondcross[0.3]{black} & 5.88 &  73.64 (15.1) &   4.73 (3.4) \\
\cblackdiamondx[0.3]{black}     & 7.38 &  73.50 (12.7) &  9.53 (10.8) \\
\cwhitesquare[0.3]{black}       & 8.11 &  70.64 (17.3) &   3.62 (2.3) \\
\cwhitesquaredot[0.3]{black}    & 8.25 &  72.65 (16.0) &   3.71 (2.4) \\
\cblacksquaredot[0.3]{black}    & 8.50 &  72.58 (15.9) &   3.81 (2.5) \\
\cwhitesquarecross[0.3]{black}  & 9.38 &  72.56 (15.9) &   3.83 (2.5) \\
\cblacksquarecross[0.3]{black}  & 9.62 &  72.46 (15.8) &   3.97 (2.6) \\
\hdashline
\cblackstar[0.4]{yellow} & - & 94.94 (3.0) & 14.23 (67.0) \\
\cwhitestar[0.4]{yellow} & - & 93.50 (4.5) & 6.59 (22.8) \\
\hline
\multicolumn{4}{|l|}{D3b - Knee full datset} \\
\hline
\cwhitesquaredot[0.3]{black}        &  2.00 &  89.81 (3.4) &  2.34 (1.4) \\
\cblacksquaredot[0.3]{black}        &  2.62 &  89.72 (3.7) &  2.46 (1.5) \\
\cwhitesquarex[0.3]{black}          &  4.38 &  89.06 (3.1) &  3.49 (2.7) \\
\cblacksquare[0.3]{black}           &  5.11 &  87.71 (3.6) &  2.83 (2.1) \\
\cblackdiamondcross[0.3]{black}     &  5.38 &  88.35 (3.3) &  2.81 (2.1) \\
\cwhitesquarecross[0.3]{black}      &  7.12 &  88.37 (2.5) &  3.48 (2.8) \\
\cblacksquarecross[0.3]{black}      &  8.00 &  88.24 (2.9) &  3.64 (2.9) \\
\cwhitediamondcross[0.3]{black}     &  9.62 &  87.45 (3.0) &  2.90 (2.1) \\
\cwhitesquare[0.3]{black}           &  9.78 &  86.62 (3.3) &  2.95 (2.0) \\
\cwhitestartriangledown[0.3]{black} & 10.12 &  86.27 (4.2) &  9.70 (9.8) \\
\hdashline
\cblackstar[0.4]{yellow} & - & 96.85 (3.0) & 12.51 (65.0) \\
\cwhitestar[0.4]{yellow} & - & 95.82 (4.9) & 4.53 (14.7) \\
\hline
\end{tabular}
\end{table}

\end{appendices}

\end{document}